\documentclass[svgnames]{article} 

\usepackage{microtype}
\usepackage{graphicx}
\usepackage{subcaption}
\usepackage{booktabs} 

\usepackage{hyperref}

\usepackage[accepted]{icml2026}

\usepackage{amsmath}
\usepackage{amsthm} 
\usepackage{amssymb}
\usepackage{mathtools}
\usepackage{graphicx}
\usepackage{booktabs}
\usepackage{caption}
\usepackage{subcaption}
\usepackage{array}
\usepackage{physics}
\usepackage{enumitem}
\usepackage{placeins}

\theoremstyle{plain}

\theoremstyle{definition}

\theoremstyle{remark}

\usepackage{amsthm}
\usepackage{thmtools}
\usepackage{thm-restate}

\usepackage{etoolbox} 
\newbool{includeapp}
\setbool{includeapp}{true} 

\usepackage{siunitx}

\definecolor{color1}{HTML}{0077BB}
\definecolor{color2}{HTML}{EE7733}
\definecolor{color3}{HTML}{33BBEE}
\definecolor{color4}{HTML}{EE3377}
\definecolor{color5}{HTML}{CC3311}
\definecolor{color6}{HTML}{009988}
\definecolor{color7}{HTML}{BBBBBB}

\input{icml_2026_content/feynman_setup}

\newcommand{\RR}{\mathbb{R}}
\newcommand{\EE}{\mathbb{E}}
\newcommand{\nin}{n_{\mathrm{in}}}
\newcommand{\nout}{n_{\mathrm{out}}}

\makeatletter
\newcommand{\scabbr}[1]{{\fontsize{0.7\dimexpr\f@size pt}{0}\selectfont\textnormal{#1}}}
\makeatother

\date{}

\icmltitlerunning{Finite-Width Neural Tangent Kernels from Feynman Diagrams}

\begin{document}

\twocolumn[
  \icmltitle{Finite-Width Neural Tangent Kernels from Feynman Diagrams}



  \icmlsetsymbol{equal}{*}

  \begin{icmlauthorlist}
    \icmlauthor{Max Guillen}{equal,chalmers}    
    \icmlauthor{Philipp Misof}{equal,chalmers}
    \icmlauthor{Jan E.\ Gerken}{chalmers}
  \end{icmlauthorlist}

  \icmlaffiliation{chalmers}{Department of Mathematical Sciences, Chalmers University of Technology and the University of Gothenburg, SE-412 96 Gothenburg, Sweden}
  \icmlcorrespondingauthor{Jan Gerken}{gerken@chalmers.se}

  \icmlkeywords{neural tangent kernel,feynman diagrams}

  \vskip 0.3in
]



\printAffiliationsAndNotice{\icmlEqualContribution}  


\begin{abstract}
  Neural tangent kernels (NTKs) are a powerful tool for analyzing deep, non-linear neural networks. In the infinite-width limit, NTKs can easily be computed for most common architectures, yielding full analytic control over the training dynamics. However, at infinite width, important properties of training such as NTK evolution or feature learning are absent. Nevertheless, finite width effects can be included by computing corrections to the Gaussian statistics at infinite width. We introduce Feynman diagrams for computing finite-width corrections to NTK statistics. These dramatically simplify the necessary algebraic manipulations and enable the computation of layer-wise recursion relations for arbitrary statistics involving preactivations, NTKs and certain higher-derivative tensors (dNTK and ddNTK) required to predict the training dynamics at leading order. We demonstrate the feasibility of our framework by extending stability results for deep networks from preactivations to NTKs and proving the absence of finite-width corrections for scale-invariant nonlinearities such as ReLU on the diagonal of the Gram matrix of the NTK. We numerically implement the complete set of equations necessary to compute the first-order corrections for arbitrary inputs and demonstrate that the results follow the statistics of sampled neural networks for widths $n\gtrsim 20$.
\end{abstract}


\section{Introduction}
\label{sec:intro}

\begin{figure*}[t]
  \centering
  \includegraphics[width=0.7\textwidth]{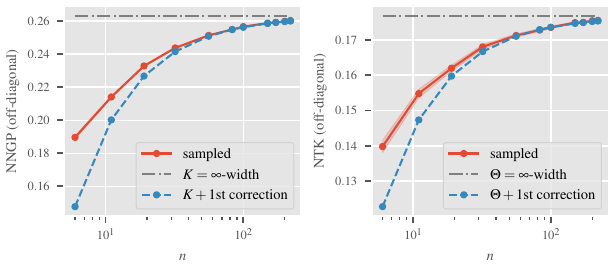}
  \vspace{-0.1cm}
  \caption{\emph{Finite-Width Corrected Kernels.} The Monte--Carlo (MC) estimated off-diagonal entry  NNGP $\overline{K}_{01}$ and NTK $\overline{\Theta}_{01}$ (red) at the fourth layer of a GeLU-MLP are shown at different hidden layer widths $n=n_\ell$ and compared to the first-order corrected finite-width solution $K^{(\ell)}_{01} +K^{\{1\}(\ell)}_{01} / n_\ell$ and $\Theta^{(\ell)}_{01} + \Theta^{\{1\}(\ell)}_{01} / n_\ell$ (blue), respectively, as well as to infinite-width results (gray). Sample sizes for the MC estimates of the NNGP and NTK are \num{e6} and \num{e5}, respectively. Error bars are included, but mostly covered by the mean line. For details, see Section \ref{sec:experiments}.}
  \label{fig:kernel_finite_width_corrected}
  \vspace{-0.5cm}
\end{figure*}

The neural tangent kernel (NTK) is defined as $\Theta(x,x')=J(x)J(x')^{\top}$, where $J(x)$ is the Jacobian of the neural network evaluated at $x$ with respect to the network parameters. Given a training dataset and loss function, the NTK completely specifies the evolution of the neural network to first order in the learning rate. This makes the NTK a powerful tool to approximate neural network training dynamics and many works have measured $\Theta$ during or after training for various applications~\cite{hayase2022, mok2022, engel2023a, zhou2024, wilson2025}. At the same time, the NTK can also be studied theoretically. However, the NTK is a random variable via its dependence on the initialization and it evolves non-linearly during training as the parameters are updated, making direct theoretical analyses of the NTK prohibitively difficult.

Nevertheless, it is well-known that neural networks simplify considerably in the infinite-width limit, in which the number of channels of all hidden layers are taken to infinity~\cite{neal1996, lee2018, matthews2018}. In particular, the NTK collapses onto its mean and does not evolve during training, it \emph{freezes}~\cite{jacot2018}. This deterministic NTK can be computed in closed form for deep, non-linear networks by applying layer-wise recursive relations which have been derived for all common neural network layers~\cite{novak2020a}. Therefore, both gradient descent~\cite{jacot2018} and Bayesian inference~\cite{he2020} can be solved in closed form at infinite width.



Although the study of the infinite-width limit has been very successful, the simplifications of the dynamics which enabled this progress also constitute an important limitation of this approach. In particular, at infinite width, the network is effectively linearized in its parameters~\cite{lee2019a} and becomes a Gaussian process. Furthermore, only the last layer evolves during training, a property termed \emph{no feature learning}. Correspondingly, some studies found considerable deviations between the properties observed for finite-width networks and those predicted by NTKs at infinite width~\cite{NEURIPS2020_ad086f59, wenger2024, liu2025}.

A solution for this problem is to consider how deviations from the strict infinite-width limit affect the NTK statistics. In particular, one can expand the training dynamics in a Taylor series in $1/n$, where $n$ is the width of the hidden layers~\cite{yaida2020, roberts2022}. In this expansion, the leading term, $1/n\rightarrow0$, corresponds to the infinite width limit and the first correction introduces finite-width effects. Including these finite-width effects in the analysis greatly extends the theoretical framework which now includes non-trivial NTK statistics, the evolution of the NTK and feature learning.

Since the infinite-width statistics are Gaussian, this constitutes an expansion of the parameter statistics around a Gaussian distribution. Such an expansion is very common in theoretical physics, in particular in quantum field theory (QFT), which describes the interactions of elementary particles in a statistical manner. In perturbative QFT, the \emph{action} (log-probability) of the theory is expanded around a leading Gaussian contribution which corresponds to non-interacting particles, while non-Gaussian corrections introduce interactions. Therefore, by employing methods from QFT, one can compute corrections to the infinite-width limit of neural network statistics and obtain results valid for networks of finite width, see Figure~\ref{fig:kernel_finite_width_corrected}. However, the algebraic manipulations involved in deriving these results are lengthy and usually phrased in a language unfamiliar to the machine-learning community. In contrast, when performing computations in particle physics, physicists mostly use \emph{Feynman diagrams} as an intuitive shorthand to greatly simplify the computations.

In this article, we introduce Feynman-like diagrams for the computation of finite-width corrections to NTKs of MLPs. We provide Feynman rules for straightforwardly computing all relevant neural network statistics involving preactivations, the NTK and related quantities. In this framework, computing expectation values reduces to drawing all diagrams compatible with the desired expectation value and translating them into algebraic expressions using the Feynman rules we provide.

In order to demonstrate the power of our framework, we investigate the stability of gradients as the depth of the network increases. At infinite width, a critical initialization variance can be derived, for which the preactivations and NTK do not grow exponentially in depth. It could be shown that the same critical initialization stabilizes the forward pass also at finite width using Feynman diagrams for preactivations~\cite{banta2024}. Using our NTK-Feynman diagrams, we can extend this result to the gradients at finite width. We also show that for scale-invariant activation functions such as ReLU, $\Theta(x,x)$ does not include finite-width corrections and the infinite-width statistics are exact.

We confirm our theoretical results with numerical experiments. We explicitly compute the first finite-width corrections for the NTK, the neural network Gaussian process kernel (NNGP) and related tensors and show that the corrected quantities align better with sample-averages of preactivations and Jacobians of GeLU-MLPs than the uncorrected quantities, as shown in Figure~\ref{fig:kernel_finite_width_corrected}. We furthermore demonstrate the stability of the critical initialization for statistics involving the NTK by sampling networks of increasing depth. We confirm our stability results for scale-invariant nonlinearities by sampling diagonal and off-diagonal elements of the kernels of ReLU-MLPs.

Our main contributions are as follows:
\begin{itemize}
\item We provide a set of rules for computing finite-width corrections to NTK statistics using Feynman diagrams, greatly simplifying the algebraic manipulations required. We verify that our rules result in the correct layer-wise recursion relations for the NTK and preactivation statistics at order $1/\text{width}$ and prove that the same rules can be used to compute corrections to all orders.
\item We demonstrate the power of our formalism by deriving the recursion relation of the leading order correction to the mean of the NTK using our Feynman diagrams. We furthermore extend previous results on preactivation stability to arbitrary statistics of preactivations and the NTK. Finally, by showing that all $1/\text{width}$ corrections of $\Theta(x,x)$ vanish for scale-invariant nonlinearities, we prove that its infinite-width limit is exact in this case.
\item We implement the recursion relations for the first-order finite-width corrections in the general multi-input case. To the best of our knowledge, no such implementation has been done before. We verify that the recursions predict the statistics of sampled neural networks with high accuracy for all tensors relevant for statistics involving the NNGP and NTK. Furthermore, we show that the initialization variance which stabilizes the infinite-width limit also stabilizes all statistics involving the NTK to all orders in $1/\text{width}$ by sampling tensors below, at and above criticality and measuring their dependence on depth. Finally, we verify by sampling that $\Theta(x,x)$ receives no finite-width corrections for the scale-invariant activation functions ReLU and LeakyReLU but corrections are present for GeLU (not scale-invariant). The code is available at \url{https://github.com/PhilippMisofCH/ntk-unlimited}.
\end{itemize}


\section{Related Work}
The analytical study of the NTK at infinite width has let to numerous insights about the training dynamics with results on trainability~\cite{xiao2020}, double-descent~\cite{geiger2020a}, generative adversarial networks~\cite{franceschi2022}, physics-informed neural networks~\cite{wang2022b}, data augmentation~\cite{gerken2024} and equivariant neural networks~\cite{misof2025}, among others. Infinite-width NTKs have also been used to optimize various aspects of the training process, such as architecture selection and bias detection~\cite{deoliveira2025}.

In the machine-learning literature, finite-width corrections in the formulation used in this article were studied in~\cite{yaida2020}. A book on the subject which develops the background step-by-step and contains many novel results is also available~\cite{roberts2022}. The analogy between QFT and neural networks has been explored from different perspectives in the physics literature. It was studied systematically for the first time in~\cite{halverson2021}. This work also established a connection to renormalization group flow which was studied in a non-perturbative setting in \cite{erbin2022}. In~\cite{grosvenor2022}, a connection to the $O(N)$ vector model was established. Different perturbations of the infinite-width limit related to the Edgeworth expansion have been studied as well~\cite{demirtas2024}. This line of work has led to results about symmetries~\cite{maiti2021}, initialization stability~\cite{banta2024}, orthogonal initializations~\cite{day2025} and scaling laws~\cite{maloney2022, zhang2025} among others.

Feynman diagrams appear in~\cite{banta2024, halverson2021, grosvenor2022, demirtas2024, maloney2022, zhang2025}, but are restricted to preactivation-statistics. In contrast, we use Feynman diagrams to compute NTK and joint prectivation-NTK statistics. This constitutes a substantial extension since preactivation statistics can only describe the initialization of the network, while NTK statistics are needed for training. Also \cite{dyer2019} uses Feynman diagrams to study corrections to the NTK, but their treatment is restricted to providing the scaling behavior of correlation functions, whereas our framework can evaluate the correlation functions explicitly.


\section{Finite-Width Corrections to the NTK}
\label{sec-finitewidth-corrections-NTK}

For an $L$-layer neural network $\mathcal{N}:\RR^{\nin}\rightarrow\RR^{\nout}$ with parameters $\theta$, the layer-$\ell$-\emph{empirical NTK} is defined by
\begin{align}
  \widehat\Theta_{ij}^{(\ell)}(x,x')=\sum_{\mu}\frac{\partial z^{(\ell)}_{i}(x)}{\partial\theta_{\mu}}\frac{\partial z^{(\ell)}_{j}(x')}{\partial\theta_{\mu}}\,,\label{eq:10}
\end{align}
where $z_{i}^{(\ell)}$ are the channel-$i$ preactivations of layer $\ell$ and we sum over all parameters in the layers $\ell'\leq \ell$. We will refer to channel indices $i,j,\dots$ as \emph{neural indices}. The NTK of the entire network is given by $\widehat\Theta=\widehat\Theta^{(L)}$.
This kernel is highly non-linear, evolves during training and is a stochastic variable due to its dependence on the initialization. To first order in the learning rate, the training dynamics of the neural network are completely specified by the NTK. Therefore, understanding the NTK is key to understanding the training dynamics of neural networks.

While the NTK captures the correlations of the gradients of the network, the \emph{neural network Gaussian process (NNGP) kernel}, defined by
\begin{align}
  \widehat K^{(\ell)}_{ij}(x,x')=z_{i}^{(\ell)}(x)z_{j}^{(\ell)}(x')\,,\label{eq:4}
\end{align}
captures the correlations of the preactivations. The NNGP also is a random variable due to initialization. Both kernels evolve during training, but we will consider them at initialization. Their functional forms depend on the architecture and we will restrict to MLPs for the remainder of this article.

To describe the statistics of the NTK, we introduce the \emph{NTK fluctuations} $\widehat{\Delta\Theta}_{ij}^{(\ell)}=\widehat\Theta^{(\ell)}_{ij}-\EE_{\theta}[\widehat\Theta^{(\ell)}_{ij}]$, where the expectation value is over initializations. In the infinite-width limit, in which the number $n$ of hidden channels in $\mathcal{N}$ goes to infinity, the NTK becomes constant in training time and collapses onto its mean~\cite{jacot2018}\footnote{This requires an appropriate parametrization of the neural network layers, for instance the so-called neural-tangent parametrization, see Appendix \ref{app:initialization}.}. Therefore, at infinite width, the NTK fluctuations vanish, $\widehat{\Delta\Theta}=0$. The constant, deterministic NTK at infinite width, the \emph{frozen} NTK, can be computed by layer-wise recursion relations for which efficient implementations exist~\cite{novak2020a}. 



To go beyond the infinite-width limit, one can consider a Taylor expansion of the neural network dynamics in $1/n$~\cite{yaida2020, roberts2022}. In this setting, the leading contribution corresponds to the Gaussian process behavior at $n\rightarrow\infty$ and higher-order corrections in $1/n$ introduce non-linear effects and feature evolution. The corrections to the NTK at higher orders are neither constant nor deterministic. The neural network statistics are then characterized by computing mixed moments of the NTK and preactivations in a given layer, for instance
\begin{align}
  \EE_{\theta}[z^{(\ell)}_{i_{1}}(x_{1})z^{(\ell)}_{i_{2}}(x_{2})\widehat{\Delta\Theta}^{(\ell)}_{i_{3}i_{4}}(x_{3},x_{4})]\,.\label{eq:2}
\end{align}
Since $\widehat{\Delta\Theta}=0$ at infinite width, \eqref{eq:2} is of order $1/n$. Because higher-order moments correspond to higher orders in $1/n$, by computing finitely many mixed moments, the statistics of the relevant quantities can be described completely to a certain order in $1/n$. Instead of computing mixed moments directly, we will instead work with so-called \emph{joint cumulants} (\emph{connected correlators} in physics), denoted by $\EE^{c}$. Intuitively, a joint cumulant is the corresponding mixed moment with all factorizations of the expectation value subtracted. A more detailed description is provided in Appendix~\ref{app:cumulants}.

In order to extract the building blocks of the cumulants at a certain order in $1/n$, we decompose the cumulants according to their dependence on the neural indices. For instance, we introduce the tensors $D$ and $F$ by decomposing the cumulant of~\eqref{eq:2} as~\cite{roberts2022}
\begin{align}
  &\EE^{c}_{\theta}[z^{(\ell+1)}_{i_{1}}(x_{1}),z^{(\ell+1)}_{i_{2}}(x_{2}),\widehat{\Delta\Theta}^{(\ell+1)}_{i_{3}i_{4}}(\textcolor{color1}{x_{3}},\textcolor{color1}{x_{4}})]\label{eq:3} \nonumber \\
  &\quad=\frac{1}{n_{\ell}}\left( D^{(\ell+1)}_{12\textcolor{color1}{34}}\delta_{i_{1}i_{2}}\delta_{i_{3}i_{4}}+F^{(\ell+1)}_{1\textcolor{color1}{3}2\textcolor{color1}{4}}\delta_{i_{1}i_{3}}\delta_{i_{2}i_{4}} \right. \nonumber \\
  &\, \qquad \left. + F^{(\ell+1)}_{1\textcolor{color1}{4}2\textcolor{color1}{3}}\delta_{i_{1}i_{4}}\delta_{i_{2}i_{3}} \right)\,.
\end{align}
Here, we have highlighted the inputs $x_{3}$, $x_{4}$ appearing in the NTK in blue and $D_{\alpha\beta\gamma\delta}$ is the Gram tensor of a function $D$ evaluated on the four inputs appearing in the cumulant, $D_{\alpha\beta\gamma\delta}=D(x_{\alpha},x_{\beta},x_{\gamma},x_{\delta})$ and similarly for $F$. We refer to $\alpha,\beta,\dots$ as \emph{sample indices}. We will decompose all cumulants into such tensors according to their neural indices. In particular, the rank-four tensors $A$, $B$, $D$ and $F$ together with the four-point cumulant of preactivations $V$ and the first mean NNGP and NTK corrections $K^{\{1\}}$ and $\Theta^{\{1\}}$ completely specify the statistics of preactivations and the NTK at order $1/n$~\cite{roberts2022}. We provide definitions of these objects in Appendix~\ref{app:defin-all-tens}. Intuitively, the tensors $D$ and $F$ capture joint preactivation-NTK statistics, while $A$ and $B$ describe NTK variations. In order to compute $1/n$ corrections to the training dynamics, the higher-derivative tensors \emph{dNTK} and \emph{ddNTK} are necessary. These will be decomposed into the tensors $P$, $Q$, $R$, $S$, $T$ and $U$, see Appendix~\ref{app:defin-all-tens}.





For a given neural network, these tensors can be computed from layer-wise recursive relations, for instance $F$ satisfies the relation~\cite{roberts2022}
\begingroup
\begin{align}
  F^{(\ell+1)}_{1\textcolor{color1}{3}2\textcolor{color1}{4}} &=(C_{W}^{(\ell+1)})^{2}\langle \sigma^{(\ell)}_{1}\sigma^{(\ell)}_{2}\sigma'^{(\ell)}_{\textcolor{color1}{3}}\sigma'^{(\ell)}_{\textcolor{color1}{4}} \rangle_{K^{(\ell)}}\Theta^{(\ell)}_{\textcolor{color1}{3}\textcolor{color1}{4}}\nonumber\\
  &\hspace{-0.5cm}+\frac{n_{\ell}}{n_{\ell-1}}(C_{W}^{(\ell+1)})^{2}\!\!\!\sum_{\alpha,\beta,\gamma,\delta=1}^{4}\langle \sigma^{(\ell)}_{1}\sigma'^{(\ell)}_{\textcolor{color1}{3}}z^{(\ell)}_{\alpha} \rangle_{K^{(\ell)}}\nonumber\\
  &\hspace{-0.5cm}\times\langle \sigma_{2}^{(\ell)}\sigma'^{(\ell)}_{\textcolor{color1}{4}}z^{(\ell)}_{\beta} \rangle_{K^{(\ell)}} K^{\alpha\gamma}_{(\ell)}K^{\beta\delta}_{(\ell)}F^{(\ell)}_{\gamma\textcolor{color1}{3}\delta\textcolor{color1}{4}}+\mathcal{O}\left(1/n\right)\,, \label{eq:F}
\end{align}
\endgroup
where the right-hand side only involves expectation values at layer $\ell$. Here,  $C_{W}^{(\ell)}$ is the variance of the initialization distribution of the weights in layer $\ell$ and $K^{(\ell)}_{\alpha\beta}=K^{(\ell)}(x_{\alpha},x_{\beta})=\EE_{\theta}[\widehat K^{(\ell)}(x_{\alpha},x_{\beta})]$ is the Gram matrix of the mean of the NNGP (similarly for $\Theta$). $K$ with upper indices denotes the inverse of the matrix $K_{\alpha\beta}$, i.e.\ $\sum_{\beta}K^{\alpha\beta}_{(\ell)}K^{(\ell)}_{\beta\gamma}=\delta_{\alpha\gamma}$. By $\langle \cdot \rangle_{K^{(\ell)}}$ we denote the Gaussian expectation value with mean zero and covariance given by the matrix $K^{(\ell)}_{\alpha\beta}$. We used the shorthands $z_{\alpha}=z(x_{\alpha})$ and $\sigma_{\alpha}=\sigma(z(x_{\alpha}))$. The prime designates the derivative of the activation function.

Unfortunately, arriving at such recursions is algebraically very laborious, hindering adoption of these expansions for machine-learning research. In this work, we introduce a framework for substantially simplifying these computations by using a graphical approach based on Feynman diagrams, allowing for the straightforward derivation of known and novel recursions such as the one discussed in Section~\ref{sec:recurs-relat-ntk}.


\section{Feynman Diagrams for NTKs}
\label{feynman-diagrams-NTKs}

Feynman diagrams are a standard tool to compute cumulants of quantum fields in theoretical particle physics. The probability distributions of these fields are Gaussians (corresponding to non-interacting particles) plus small, non-Gaussian corrections (corresponding to interactions). Feynman diagrams are graphs which symbolize these interactions. The set of \emph{Feynman rules} defines the vertices and edges which can appear in the graphs, what order in the expansion parameter (here: $1/n$) they carry and explains how to translate a given diagram into a mathematical term. The \emph{external vertices} correspond to the variables over which we take the expectation value and the computation proceeds by drawing all diagrams compatible with the given external vertices and the desired order according to the Feynman rules. The final result is obtained by summing up all the terms corresponding to the admissible Feynman diagrams.






\subsection{The Feynman Rules for NTK Statistics}\label{sec:feynman-rules}
The set of Feynman rules for computing cumulants of preactivations was explicitly derived in \cite{banta2024} from first principles. This approach relied heavily on the Gaussian nature of the conditional probability distribution $P(z^{(\ell+1)}|z^{(\ell)})$. Since the NTK is quadratic in the weights, their method is no longer admissible for evaluating cumulants involving the NTK. For this reason, we will adopt an alternative strategy based on the decomposition of the cumulants into rank-four tensors as described below~\eqref{eq:3}. In the following, we give an overview of the most important Feynman rules to provide a general idea of how Feynman diagrams are constructed in this context and translated into algebraic expressions. The full set of rules is provided in Appendix~\ref{app:feynman_rules}.


%
%
\begin{enumerate}[label=(\roman*),leftmargin=*,widest=iii]
  \item External vertices (which correspond to the quantities which appear in the expectation value) are represented by filled dots. \emph{Solid} and \emph{dotted external lines} represent preactivations and NTK fluctuations, respectively
  \vspace{-0.2cm}
  \begin{align}
    &z_{\alpha} \equiv
    \begin{tikzpicture}[baseline=-0.1cm]
      \begin{feynman}
        \vertex (l) {};
        \vertex[right = 0pt of l, dot, minimum size=3pt, label = {left: {\footnotesize $\alpha^{}$}}] (x1) {};
        \vertex[right = 30pt of x1, dot, minimum size=0pt] (b) {};
        \diagram*{
          (x1),
          (x1) -- [inner sep = 4pt] (b) 
        };
      \end{feynman}
    \end{tikzpicture}
    \qquad\widehat{\Delta \Theta}_{\textcolor{color1}{\alpha\beta}} \equiv
    \begin{tikzpicture}[baseline=0.05cm]
      \begin{feynman}
        \vertex (l) {};
        \vertex[right = 0pt of l, color1, dot, minimum size=3pt, label = {left: {\footnotesize $\textcolor{color1}{\alpha^{}}$}}] (x1) {};
        \vertex[above = 8pt of l, color1, dot, minimum size=3pt, label = {left: {\footnotesize $\textcolor{color1}{\beta^{}}$}}] (x2) {};
        \vertex[right = 30pt of x1, dot, minimum size=0pt] (b) {};
        \vertex[right = 30pt of x2, dot, minimum size=0pt] (bb) {};
        \diagram*{
          (x1), (x2),
          (x1) -- [color1, ghost, inner sep = 4pt] (b),
          (x2) -- [color1, ghost, inner sep = 4pt] (bb)
        };
      \end{feynman}
    \end{tikzpicture}\label{eq:6}
  \end{align}
  We will use different colors for external dotted lines corresponding to different NTKs. As detailed in Appendix~\ref{app:feynman_rules}, additional lines represent the higher derivative terms inside the dNTK and ddNTK.

\item The external lines from \eqref{eq:6} are attached to \emph{cubic interactions}, which are vertices connected to two external lines and one \emph{internal line}. For cubic interactions involving external NTK lines, we introduce the following Feynman rules
  \begin{align}
    \begin{tikzpicture}[baseline=(b)]
      \begin{feynman}
        \vertex (l) {};
        \vertex[below = 15pt of l, color1, dot, minimum size=3pt, label = {left: {\footnotesize $\textcolor{color1}{\alpha^{}}$}}] (x1) {};
        \vertex[above = 15pt of l, color1, dot, minimum size=3pt, label = {left: {\footnotesize $\textcolor{color1}{\beta^{}}$}}] (x2) {};
        \vertex[right = 10pt of l, dot, minimum size=0pt] (v12) {};
        \vertex[right = 30pt of v12, dot, minimum size=0pt] (b) {};
        \diagram*{
          (x1) --  [color1, ghost] (v12) --  [color1, ghost] (x2),
          (v12) -- [black, photon, edge label = {\scriptsize \;$\widehat{\Delta \Omega}_{i,\alpha\beta}^{(\ell+1)}$}, inner sep = 4pt] (b) 
        };
      \end{feynman}
    \end{tikzpicture}\hspace{-0.1cm} &\sim \frac{1}{n_{\ell}} &
    \begin{tikzpicture}[baseline=(b)]
      \begin{feynman}
        \vertex (l) {};
        \vertex[below = 15pt of l, dot, minimum size=3pt, label = {left: {\footnotesize $\alpha^{}$}}] (x1) {};
        \vertex[above = 15pt of l, color1, dot, minimum size=3pt, label = {left: {\footnotesize $\textcolor{color1}{\beta^{}}$}}] (x2) {};
        \vertex[right = 10pt of l, dot, minimum size=0pt] (v12) {};
        \vertex[right = 33pt of v12, dot, minimum size=0pt] (b) {};
        \diagram*{
          (x1) --  (v12) --  [color1, ghost] (x2),
          (v12) -- [color1blackghost, edge label = {\scriptsize \;$\sigma^{(\ell)}_{i,\alpha}\sigma'^{(\ell)}_{i,\textcolor{color1}{\beta}}$}, inner sep = 4pt] (b) 
        };
      \end{feynman}
    \end{tikzpicture}\hspace{-0.1cm} &\sim \frac{C^{(\ell+1)}_{W}}{n_{\ell}}\nonumber\\
    \begin{tikzpicture}[baseline=(b)]
      \begin{feynman}
        \vertex (l) {};
        \vertex[below = 15pt of l, color1, dot, minimum size=3pt, label = {left: {\footnotesize $\textcolor{color1}{\alpha^{}}$}}] (x1) {};
        \vertex[above = 15pt of l, color1, dot, minimum size=3pt, label = {left: {\footnotesize $\textcolor{color1}{\beta^{}}$}}] (x2) {};
        \vertex[right = 10pt of l, dot, minimum size=0pt] (v12) {};
        \vertex[right = 30pt of v12, dot, minimum size=0pt] (b) {};
        \diagram*{
          (x1) --  [color1, ghost] (v12) --  [color1, ghost] (x2),
          (v12) -- [color1doubghost, edge label = {\scriptsize \;$\sigma'^{(\ell)}_{i,\textcolor{color1}{\alpha}}\sigma'^{(\ell)}_{i,\textcolor{color1}{\beta}}$}, inner sep = 4pt] (b) 
        };
      \end{feynman}
    \end{tikzpicture}\hspace{-0.2cm} &\sim \frac{C^{(\ell+1)}_{W}}{n_{\ell}}  &
    \begin{tikzpicture}[baseline=(b)]
      \begin{feynman}
        \vertex (l) {};
        \vertex[below = 15pt of l, color1, dot, minimum size=3pt, label = {left: {\footnotesize $\textcolor{color1}{\alpha^{}}$}}] (x1) {};
        \vertex[above = 15pt of l, color2, dot, minimum size=3pt, label = {left: {\footnotesize $\textcolor{color2}{\beta^{}}$}}] (x2) {};
        \vertex[right = 10pt of l, dot, minimum size=0pt] (v12) {};
        \vertex[right = 30pt of v12, dot, minimum size=0pt] (b) {};
        \diagram*{
          (x1) --  [color1, ghost] (v12) --  [color2, ghost] (x2),
          (v12) -- [color2color1ghost, edge label = {\scriptsize \;$\sigma'^{(\ell)}_{i,\textcolor{color1}{\alpha}}\sigma'^{(\ell)}_{i,\textcolor{color2}{\beta}}$}, inner sep = 4pt] (b) 
        };
      \end{feynman}
    \end{tikzpicture}\hspace{-0.2cm} &\sim \frac{C^{(\ell+1)}_{W}}{n_{\ell}}\label{feynmanrulescubicmain}
  \end{align}
  where $\widehat{\Omega}^{(\ell+1)}_{i,\alpha\beta} = \sigma^{(\ell)}_{i,\alpha}\sigma^{(\ell)}_{i,\beta} + C^{(\ell+1)}_{W}\Theta_{\alpha\beta}^{(\ell)}\sigma'^{(\ell)}_{i,\alpha}\sigma'^{(\ell)}_{i,\beta}$ and $\widehat{\Delta\Omega}_{i,\alpha\beta}^{(\ell+1)}=\widehat{\Omega}_{i,\alpha\beta}^{(\ell+1)}-\langle \widehat{\Omega}_{i,\alpha\beta}^{(\ell+1)} \rangle_{K^{(\ell)}}$. The line without dot is the internal line. We discuss its decoration in the next Feynman rule.  Additional cubic vertices for preactivation, dNTK and ddNTK lines are provided in Appendix~\ref{app:feynman_rules}.
\item The internal lines from the vertices \eqref{feynmanrulescubicmain} are connected to a \emph{propagator} which will be represented by a white blob
  \begin{equation}
    \langle\hspace{3mm}\rangle_{K^{(\ell)}} \equiv
    \begin{tikzpicture}[baseline=-0.1cm]
      \begin{feynman}
        \tikzfeynmanset{every blob = {/tikz/fill=white!50, /tikz/minimum size=15pt}}
        \vertex (l) {};
        \vertex[above = 0pt of l, dot, minimum size=0pt] (v12) {};
        \vertex[above = 0pt of v12, blob] (b) {};
        \vertex[above = 0pt of b, dot, minimum size=0pt] (v34) {};
        \diagram*{
          (v12) -- (b) -- (v34), 
        };
      \end{feynman}
    \end{tikzpicture}   
  \end{equation}
  where $\langle \hspace{3mm}\rangle_{K^{(\ell)}}$ stands for a Gaussian expectation value with mean zero and covariance given by the Gram matrix of the NNGP, as in~\eqref{eq:F}. We take the expectation value over the decorations of the internal lines attached to the propagator, for instance
  \begin{align}
    \begin{tikzpicture}[baseline=(b)]
      \begin{feynman}
        \vertex (l) {};
        \vertex[right = 10pt of l, dot, minimum size=0pt] (v12) {};
        \vertex[right = 40pt of v12, propblob] (b) {};
        \vertex[right = 40pt of b, dot, minimum size=0pt] (v34) {};
        \vertex[right = 8pt of v34] (r) {};
        \diagram*{
          (v12) -- [black, photon, edge label = {\scriptsize $\widehat{\Delta\Omega}_{i,\alpha\beta}$}, inner sep = 4pt] (b) -- [black, photon, edge label = {\scriptsize $\widehat{\Delta\Omega}_{i,\gamma\delta}$}, inner sep = 4pt] (v34)
        };
      \end{feynman}
    \end{tikzpicture}\hspace{-0.2cm}
    \sim \langle \widehat{\Delta\Omega}_{i,\alpha\beta} \widehat{\Delta\Omega}_{i,\gamma\delta}\rangle_{K^{(\ell)}}\,.
  \end{align}
Propagators obey the following set of \emph{selection rules}:
  \begin{enumerate}[label=(\alph*)]
  \item Propagators are only connected to internal lines emanating from the cubic vertices or the internal quartic vetrices introduced below. In particular, propagators cannot be directly connected to other propagators.
  \item Dotted lines attached to a propagator do not appear in the Gaussian expectation value, since they are not decorated.
  \item Each preactivation line decorated with $z_{i}$ acts as a derivative with respect to $z_{i}$ on the argument of the Gaussian expectation value.
  \item The neural indices of all internal lines connected to the propagator have to be equal.
  \item If both dotted and dashed lines of the same color are connected to the propagator, these have to appear in pairs with the same sample index. The lines in a pair will be attached to different vertices. Furthermore, if both vertices connected with one color to the propagator are drawn in the orientation in which they are presented in the Feynman rules, the ordering from top to bottom of the sample indices (and therefore colors) of the lines connected to both vertices have to be the same.
  \item Pairs of dashed lines of the same color connected to the propagator add a factor of $\Theta_{\alpha\beta}$ if they are connected to different vertices. Here, $\alpha$ and $\beta$ are the sample indices of the two lines in the pair.
  \end{enumerate}
\item The 10 rank-four tensors into which we decompose the cumulants will be represented by \emph{quartic interactions} with four external lines. We introduce the following quartic vertices containing NTK and preactivation lines
  \begin{align}
    \begin{tikzpicture}[baseline=(b)]
      \begin{feynman}
        \vertex (l) {};
        \vertex[below = 15pt of l, dot, minimum size=3pt, label = {left: {\footnotesize $\alpha_{1}$}}] (x1) {};
        \vertex[above = 15pt of l, dot, minimum size=3pt, label = {left: {\footnotesize $\alpha_{2}$}}] (x2) {};
        \vertex[right = 20pt of l, quarticblob] (b) {};
        \vertex[below = 25pt of b] {\scriptsize $\frac{1}{n_{\ell}\*}D_{\alpha_{1}\alpha_{2}\textcolor{color1}{\alpha_{3}\alpha_{4}}}^{(\ell+1)}$}; 
        \vertex[right = 20pt of b] (r) {};
        \vertex[above = 15pt of r, dot, color1, minimum size=3pt, label = {right: {\footnotesize $\textcolor{color1}{\alpha_{3}}$}}] (x3) {};
        \vertex[below = 15pt of r, dot, color1, minimum size=3pt, label = {right: {\footnotesize $\textcolor{color1}{\alpha_{4}}$}}] (x4) {};
        \diagram*{
          (x1) -- (b) -- (x2), 
          (x3) -- [color1, ghost] (b) -- [color1, ghost] (x4)
        };
      \end{feynman}
    \end{tikzpicture}
    \hspace{0.5cm}
    \begin{tikzpicture}[baseline=(b)]
      \begin{feynman}
        \vertex (l) {};
        \vertex[below = 15pt of l, dot, minimum size=3pt, label = {left: {\footnotesize $\alpha_{1}$}}] (x1) {};
        \vertex[above = 15pt of l, dot, color1, minimum size=3pt, label = {left: {\footnotesize $\textcolor{color1}{\alpha_{3}}$}}] (x2) {};
        \vertex[right = 20pt of l, quarticblob] (b) {};
        \vertex[below = 25pt of b] {\scriptsize $\frac{1}{n_{\ell}\*}F_{\alpha_{1}\textcolor{color1}{\alpha_{3}}\alpha_{2}\textcolor{color1}{\alpha_{4}}}^{(\ell+1)}$}; 
        \vertex[right = 20pt of b] (r) {};
        \vertex[above = 15pt of r, dot, minimum size=3pt, label = {right: {\footnotesize $\alpha_{2}$}}] (x3) {};
        \vertex[below = 15pt of r, dot, color1, minimum size=3pt, label = {right: {\footnotesize $\textcolor{color1}{\alpha_{4}}$}}] (x4) {};
        \diagram*{
          (x1) -- (b) -- [color1, ghost] (x2), 
          (x3) -- (b) -- [color1, ghost] (x4)
        };
      \end{feynman}
    \end{tikzpicture}\nonumber\\
    \begin{tikzpicture}[baseline=(b)]
      \begin{feynman}
        \vertex (l) {};
        \vertex[below = 15pt of l, dot, color2, minimum size=3pt, label = {left: {\footnotesize $\textcolor{color2}{\alpha_{1}}$}}] (x1) {};
        \vertex[above = 15pt of l, dot, color2, minimum size=3pt, label = {left: {\footnotesize $\textcolor{color2}{\alpha_{2}}$}}] (x2) {};
        \vertex[right = 20pt of l, quarticblob] (b) {};
        \vertex[below = 25pt of b] {\scriptsize $\frac{1}{n_{\ell}\*}A_{\textcolor{color2}{\alpha_{1}}\textcolor{color2}{\alpha_{2}}\textcolor{color1}{\alpha_{3}}\textcolor{color1}{\alpha_{4}}}^{(\ell+1)}$}; 
        \vertex[right = 20pt of b] (r) {};
        \vertex[above = 15pt of r, dot, color1, minimum size=3pt, label = {right: {\footnotesize $\textcolor{color1}{\alpha_{3}}$}}] (x3) {};
        \vertex[below = 15pt of r, dot, color1, minimum size=3pt, label = {right: {\footnotesize $\textcolor{color1}{\alpha_{4}}$}}] (x4) {};
        \diagram*{
          (x1) -- [color2, ghost] (b) -- [color2, ghost] (x2), 
          (x3) -- [color1, ghost] (b) -- [color1, ghost] (x4)
        };
      \end{feynman}
    \end{tikzpicture}
    \hspace{0.5cm}
    \begin{tikzpicture}[baseline=(b)]
      \begin{feynman}
        \vertex (l) {};
        \vertex[below = 15pt of l, dot, color2, minimum size=3pt, label = {left: {\footnotesize $\textcolor{color2}{\alpha_{1}}$}}] (x1) {};
        \vertex[above = 15pt of l, dot, color1, minimum size=3pt, label = {left: {\footnotesize $\textcolor{color1}{\alpha_{3}}$}}] (x2) {};
        \vertex[right = 20pt of l, quarticblob] (b) {};
        \vertex[below = 25pt of b] {\scriptsize $\frac{1}{n_{\ell}\*}B_{\textcolor{color2}{\alpha_{1}}\textcolor{color1}{\alpha_{3}}\textcolor{color2}{\alpha_{2}}\textcolor{color1}{\alpha_{4}}}^{(\ell+1)}$}; 
        \vertex[right = 20pt of b] (r) {};
        \vertex[above = 15pt of r, dot, color2, minimum size=3pt, label = {right: {\footnotesize $\textcolor{color2}{\alpha_{2}}$}}] (x3) {};
        \vertex[below = 15pt of r, dot, color1, minimum size=3pt, label = {right: {\footnotesize $\textcolor{color1}{\alpha_{4}}$}}] (x4) {};
        \diagram*{
          (x1) -- [color2, ghost] (b) -- [color1, ghost] (x2), 
          (x3) -- [color2, ghost] (b) -- [color1, ghost] (x4)
        };
      \end{feynman}
    \end{tikzpicture}
    \label{feynmanrulesquartic}
  \end{align}
  The quartic vertices can also appear with four internal lines instead. In this case, they correspond to tensors in layer $\ell$ instead of $\ell+1$ and are connected to propagators. Further quartic tensors for the remaining tensors involving preactivations, the dNTK and ddNTK are introduced in Appendix~\ref{app:defin-all-tens}. For the mean NNGP and NTK corrections $K^{\{1\}}$ and $\Theta^{\{1\}}$, we similarly introduce \emph{quadratic vertices} of valence two.
\item Internal lines attached to cubic or internal quartic vertices have unassigned neural indices (e.g.\ $i$ in the vertices in~\eqref{feynmanrulescubicmain}). Similarly, internal (solid) preactivation lines have unassigned sample indices. To assemble the term corresponding to a certain diagram, multiply the expressions corresponding to vertices and propagators and sum over all unassigned neural and sample indices. The complete expression for a cumulant is given by summing over all Feynman diagrams with the correct external lines and order in $1/n$.
\end{enumerate}
These Feynman rules satisfy the following theorem:

\begin{restatable}[]{theorem}{firsttheorem}
\label{theoremone}
  The Feynman rules postulated in items $(i)$-$(v)$ uniquely determine the recursion relations governing the layer evolution of the NTK tensors $D$, $F$, $A$, $B$ at order $\frac{1}{n}$.
\end{restatable}
\begin{proof}
  To illustrate how the proof of this statement follows, we will reproduce the recursion relation~\eqref{eq:F} from the above Feynman rules. See Appendix~\ref{app:proofs_one} for a similar analysis applied to the other tensors, completing the proof.

The layer-($\ell+1$) tensor on the LHS of \eqref{eq:F} is given by the second quartic vertex in~\eqref{feynmanrulesquartic}. We will now proceed to draw all Feynman diagrams which have the same external lines as this quartic vertex and are of order $1/n$. First, we notice that the only way to combine a dotted (NTK) line with a solid (preactivation) line in a cubic vertex is to use the second vertex in~\eqref{feynmanrulescubicmain}. We use this vertex twice (for inputs 1, 3 and 2, 4, respectively). One then obtains two internal solid-dashed lines, each decorated with ${\sigma}_{i}^{(\ell)}{\sigma'}_{i}^{(\ell)}$: 
\begin{align}
\begin{tikzpicture}[baseline=(b)]
\begin{feynman}
\vertex (l) {};
\vertex[below = 15pt of l, dot, minimum size=3pt, label = {left: {\footnotesize $1$}}] (x1) {};
\vertex[above = 15pt of l, color1, dot, minimum size=3pt, label = {left: {\footnotesize $\textcolor{color1}{3}$}}] (x2) {};
\vertex[right = 8pt of l, dot, minimum size=0pt] (v12) {};
\tikzfeynmanset{every blob = {/tikz/fill=white!50, /tikz/minimum size=0pt}}
\vertex[right = 30pt of v12, minimum size=0pt] (b1) {};
\vertex[right = 15pt of b1, minimum size=0pt] (b2) {};
\vertex[right = 30pt of b2, dot, minimum size=0pt] (v34) {};
\vertex[right = 8pt of v34] (r) {};
\vertex[above = 15pt of r, dot, minimum size=3pt, label = {right: {\footnotesize $2$}}] (x3) {};
\vertex[below = 15pt of r, color1, dot, minimum size=3pt, label = {right: {\footnotesize $\textcolor{color1}{4}$}}] (x4) {};
\diagram*{
	(x1) -- (v12) -- [color1, ghost] (x2),
	(v12) -- [color1blackghost, edge label = {\scriptsize \;$\sigma_{j}\sigma'_{j}$}, inner sep = 4pt] (b1),
  (b2) -- [blackcolor1ghost, edge label = {\scriptsize \,$\sigma_{k}\sigma'_{k}$}, inner sep = 4pt] (v34), 
	 (x3) -- (v34) -- [color1, ghost] (x4)
};
\end{feynman}
\end{tikzpicture}
\end{align}
This, in turn, gives rise to two possibilities: The input channels for both lines can either be equal or different. If the input channels are equal, they can be connected directly by a propagator:
\begin{align}
\begin{tikzpicture}[baseline=(b)]
\tikzfeynmanset{every blob = {/tikz/fill=white!50, /tikz/minimum size=15pt}}
\begin{feynman}
\vertex (l) {};
\vertex[below = 15pt of l, dot, minimum size=3pt, label = {left: {\footnotesize $1$}}] (x1) {};
\vertex[above = 15pt of l, color1, dot, minimum size=3pt, label = {left: {\footnotesize $\textcolor{color1}{3}$}}] (x2) {};
\vertex[right = 8pt of l, dot, minimum size=0pt] (v12) {};
\vertex[right = 30pt of v12, blob] (b) {};
\vertex[right = 30pt of b, dot, minimum size=0pt] (v34) {};
\vertex[right = 8pt of v34] (r) {};
\vertex[above = 15pt of r, dot, minimum size=3pt, label = {right: {\footnotesize $2$}}] (x3) {};
\vertex[below = 15pt of r, color1, dot, minimum size=3pt, label = {right: {\footnotesize $\textcolor{color1}{4}$}}] (x4) {};
\diagram*{
	(x1) -- (v12) -- [color1, ghost] (x2),
	(v12) -- [color1blackghost, edge label = {\scriptsize \;$\sigma_{j}\sigma'_{j}$}, inner sep = 4pt] (b) -- [blackcolor1ghost, edge label = {\scriptsize \,$\sigma_{j}\sigma'_{j}$}, inner sep = 4pt] (v34), 
	 (x3) -- (v34) -- [color1, ghost] (x4)
};
\end{feynman}
\end{tikzpicture}
\end{align}
If they are different, they can be connected by introducing two propagators and an internal $F$-tensor in layer $\ell$:
\begin{align}
  \begin{tikzpicture}[baseline=(b)]
  \tikzfeynmanset{every blob = {/tikz/fill=white!50, /tikz/minimum size=15pt}}
  \begin{feynman}
    \vertex (l) {};
    \vertex[below = 15pt of l, dot, minimum size=3pt, label = {left: {\footnotesize $1$}}] (x1) {};
    \vertex[above = 15pt of l, color1, dot, minimum size=3pt, label = {left: {\footnotesize $\textcolor{color1}{3}$}}] (x2) {};
    \vertex[right = 8pt of l, dot, minimum size=0pt] (v12) {};
    \vertex[right = 35pt of v12, blob] (b12) {};
    \vertex[right = 16pt of b12, dot, minimum size=0pt] (w12) {};
    \vertex[above = 1pt of w12, label = {above: {\scriptsize \hspace{-10pt} $z_{j_1}$}}] (w12u) {};
    \vertex[below = 8pt of w12] (w12d) {};
    \vertex[left = 10pt of w12d] (w12dl) {};
    \tikzfeynmanset{every blob = {/tikz/fill=gray!50, /tikz/minimum size=15pt}}
    \vertex[right = 3pt of w12, blob , minimum size = 6pt] (b) {};
    \vertex[right = 3pt of b, dot, minimum size=0pt] (w34) {};
    \tikzfeynmanset{every blob = {/tikz/fill=white!50, /tikz/minimum size=15pt}}
    \vertex[right = 16pt of w34, blob] (b34) {};
    \vertex[above = 1pt of w34, label = {above: {\scriptsize \hspace{10pt} $z_{j_2}$}}] (w34u) {};
    \vertex[below = 8pt of w34] (w34d) {};
    \vertex[right = 10pt of w34d] (w34dr) {};
    \vertex[right = 35pt of b34, dot, minimum size=0pt] (v34) {};
    \vertex[right = 8pt of v34] (r) {};
    \vertex[above = 15pt of r, dot, minimum size=3pt, label = {right: {\footnotesize $2$}}] (x3) {};
    \vertex[below = 15pt of r, color1, dot, minimum size=3pt, label = {right: {\footnotesize $\textcolor{color1}{4}$}}] (x4) {};
    \diagram*{
      (x1) -- (v12) -- [color1, ghost] (x2),
      (v12) -- [color1blackghost, edge label = {\scriptsize \;$\sigma_{j_{1}}\sigma'_{j_{1}}$}, inner sep = 4pt] (b12) -- [color1, ghost, quarter left] (w12) -- [quarter left] (b12),
      (b34) -- [color1, ghost, quarter left] (w34) -- [quarter left] (b34) -- [blackcolor1ghost, edge label = {\scriptsize \,$\sigma_{j_{2}}\sigma'_{j_{2}}$}, inner sep = 4pt] (v34),
      (x3) --  (v34) -- [color1, ghost] (x4)
    };
    \draw [decoration={brace}, decorate] (w34dr) -- (w12dl) node [pos=0.5, below = 1pt] {\scriptsize $\frac{1}{n_{\ell-1}\*}F_4^{(\ell)}$};
  \end{feynman}
\end{tikzpicture}
\end{align}
The diagrams corresponding to the expression~\eqref{eq:F} are therefore
%
\begin{align}
  \begin{tikzpicture}[baseline=(b)]
      \begin{feynman}
        \vertex (l) {};
        \vertex[below = 15pt of l, dot, minimum size=3pt, label = {left: {\footnotesize $1$}}] (x1) {};
        \vertex[above = 15pt of l, dot, color1, minimum size=3pt, label = {left: {\footnotesize $\textcolor{color1}{3}$}}] (x2) {};
        \vertex[right = 20pt of l, quarticblob] (b) {};
        \vertex[right = 20pt of b] (r) {};
        \vertex[above = 15pt of r, dot, minimum size=3pt, label = {right: {\footnotesize $2$}}] (x3) {};
        \vertex[below = 15pt of r, dot, color1, minimum size=3pt, label = {right: {\footnotesize $\textcolor{color1}{4}$}}] (x4) {};
        \diagram*{
          (x1) -- (b) -- [color1, ghost] (x2), 
          (x3) -- (b) -- [color1, ghost] (x4)
        };
      \end{feynman}
    \end{tikzpicture}
\!\!&=
\!\!\sum_{j}\!\!
\begin{tikzpicture}[baseline=(b)]
\tikzfeynmanset{every blob = {/tikz/fill=white!50, /tikz/minimum size=15pt}}
\begin{feynman}
\vertex (l) {};
\vertex[below = 15pt of l, dot, minimum size=3pt, label = {left: {\footnotesize $1$}}] (x1) {};
\vertex[above = 15pt of l, color1, dot, minimum size=3pt, label = {left: {\footnotesize $\textcolor{color1}{3}$}}] (x2) {};
\vertex[right = 8pt of l, dot, minimum size=0pt] (v12) {};
\vertex[right = 30pt of v12, blob] (b) {};
\vertex[right = 30pt of b, dot, minimum size=0pt] (v34) {};
\vertex[right = 8pt of v34] (r) {};
\vertex[above = 15pt of r, dot, minimum size=3pt, label = {right: {\footnotesize $2$}}] (x3) {};
\vertex[below = 15pt of r, color1, dot, minimum size=3pt, label = {right: {\footnotesize $\textcolor{color1}{4}$}}] (x4) {};
\diagram*{
	(x1) -- (v12) -- [color1, ghost] (x2),
	(v12) -- [color1blackghost, edge label = {\scriptsize \;$\sigma_{j}\sigma'_{j}$}, inner sep = 4pt] (b) -- [blackcolor1ghost, edge label = {\scriptsize \,$\sigma_{j}\sigma'_{j}$}, inner sep = 4pt] (v34), 
	 (x3) -- (v34) -- [color1, ghost] (x4)
};
\end{feynman}
\end{tikzpicture}\nonumber\\
&\hspace{-4em}+
\sum_{j_1, j_2}\!
\begin{tikzpicture}[baseline=(b)]
  \tikzfeynmanset{every blob = {/tikz/fill=white!50, /tikz/minimum size=15pt}}
  \begin{feynman}
    \vertex (l) {};
    \vertex[below = 15pt of l, dot, minimum size=3pt, label = {left: {\footnotesize $1$}}] (x1) {};
    \vertex[above = 15pt of l, color1, dot, minimum size=3pt, label = {left: {\footnotesize $\textcolor{color1}{3}$}}] (x2) {};
    \vertex[right = 8pt of l, dot, minimum size=0pt] (v12) {};
    \vertex[right = 35pt of v12, blob] (b12) {};
    \vertex[right = 16pt of b12, dot, minimum size=0pt] (w12) {};
    \vertex[above = 1pt of w12, label = {above: {\scriptsize \hspace{-10pt} $z_{j_1}$}}] (w12u) {};
    \vertex[below = 8pt of w12] (w12d) {};
    \vertex[left = 10pt of w12d] (w12dl) {};
    \tikzfeynmanset{every blob = {/tikz/fill=gray!50, /tikz/minimum size=15pt}}
    \vertex[right = 3pt of w12, blob , minimum size = 6pt] (b) {};
    \vertex[right = 3pt of b, dot, minimum size=0pt] (w34) {};
    \tikzfeynmanset{every blob = {/tikz/fill=white!50, /tikz/minimum size=15pt}}
    \vertex[right = 16pt of w34, blob] (b34) {};
    \vertex[above = 1pt of w34, label = {above: {\scriptsize \hspace{10pt} $z_{j_2}$}}] (w34u) {};
    \vertex[below = 8pt of w34] (w34d) {};
    \vertex[right = 10pt of w34d] (w34dr) {};
    \vertex[right = 35pt of b34, dot, minimum size=0pt] (v34) {};
    \vertex[right = 8pt of v34] (r) {};
    \vertex[above = 15pt of r, dot, minimum size=3pt, label = {right: {\footnotesize $2$}}] (x3) {};
    \vertex[below = 15pt of r, color1, dot, minimum size=3pt, label = {right: {\footnotesize $\textcolor{color1}{4}$}}] (x4) {};
    \diagram*{
      (x1) -- (v12) -- [color1, ghost] (x2),
      (v12) -- [color1blackghost, edge label = {\scriptsize \;$\sigma_{j_{1}}\sigma'_{j_{1}}$}, inner sep = 4pt] (b12) -- [color1, ghost, quarter left] (w12) -- [quarter left] (b12),
      (b34) -- [color1, ghost, quarter left] (w34) -- [quarter left] (b34) -- [blackcolor1ghost, edge label = {\scriptsize \,$\sigma_{j_{2}}\sigma'_{j_{2}}$}, inner sep = 4pt] (v34),
      (x3) --  (v34) -- [color1, ghost] (x4)
    };
    \draw [decoration={brace}, decorate] (w34dr) -- (w12dl) node [pos=0.5, below = 1pt] {\scriptsize $\frac{1}{n_{\ell-1}\*}F_4^{(\ell)}$};
  \end{feynman}
\end{tikzpicture}
\label{ftensorfeynmandiagram}
\end{align}
Here, we have added the necessary sums over unassigned neural indices. No more diagrams can be constructed because these would either be at higher order in $1/n$ or be ruled out by the propagator selection rules introduced in item (iii) above. With these rules, one can check that the first and second diagrams on the RHS of~\eqref{ftensorfeynmandiagram} correspond to the first and second terms in~\eqref{eq:F}, respectively.
\end{proof}
Similarly, the full Feynman rules can reproduce the recursion relations of the higher-derivative versions of the NTK as well, as detailed in the following
\begin{restatable}[]{theorem}{secondtheorem}
\label{theoremtwo}
  The set of Feynman rules listed above together with those defined in Appendix ~\ref{app:feynman_rules} recover the recursion relations satisfied by the dNTK and ddNTK tensors: $P$, $Q$ and $R$, $S$, $T$, $U$, respectively, at order $\frac{1}{n}$.
\end{restatable}
\begin{proof} 
See Appendix~\ref{app:proofs_two}.
\end{proof} 

The theorems provided so far prove that the Feynman diagrams yield complete results at order $1/n$. However, they can be extended straightforwardly to all orders in $1/n$ by adding generalizations of the vertices~\eqref{feynmanrulesquartic} with arbitrary numbers of external lines. Importantly, no other modifications are needed as detailed in
\begin{restatable}[]{theorem}{thirdtheorem}
\label{theoremthree}
  The full set of Feynman rules introduced above extended by the higher-order generalizations of the tensors $V$, $D$, $F$, $A$, $B$, $P$, $Q$, $R$, $S$, $T$, $U$ provides a complete characterization of the statistics of the NTK and its descendants at all orders in $\frac{1}{n}$.
\end{restatable}
\begin{proof} 
See Appendix ~\ref{proofoftheoremthree}.
\end{proof}
This remarkable result highlights the power of our diagrammatic approach for systematically computing preactivation and NTK statistics at arbitrary order in \(1/n\) from simple, compact rules. As an illustration, we derive the recursion relation for the $D_{6}$ tensor, relevant at order \(1/n^2\), in Appendix~\ref{app:example_NNLO_tensor}, in sharp contrast to the lengthy and technically demanding algebra required by direct methods.

\section{Applications}\label{sec:applications}
In this section, we will describe three applications of our Feynman rules to problems in deep learning.

\subsection{Recursion Relation for the NTK Mean}
\label{sec:recurs-relat-ntk}
Using our Feynman rules, it is straightforward to compute even complicated recursion relations. To demonstrate this, we will compute the recursion relation for the first-order correction in $1/n$ of the NTK mean $\Theta^{\{1\}}$, which, to the best of our knowledge, has not been derived before. Starting from the quadratic vertex
\begin{tikzpicture}[baseline=(ref)]
\begin{feynman}
        \vertex (l) {};
        \vertex[right = 0pt of l, dot, color1, label = {left: {\textcolor{color1}{$1$}}}] (v12) {};
        \vertex[right = 15pt of v12, quarticblob, minimum size=10pt] (b) {};
        \vertex[below=3pt of b, minimum size=0pt] (ref) {};
        \vertex[right = 15pt of b, dot, color1, label = {right: {\textcolor{color1}{$2$}}}] (v34) {};
        \diagram*{
          (v12) -- [color1, ghost] (b) -- [color1, ghost] (v34)
        };
      \end{feynman}
\end{tikzpicture}      
representing $\Theta^{\{1\}}_{\textcolor{color1}{12}}/n_{\ell}$, we draw all Feynman diagrams with two external NTK lines at order $1/n$. The Feynman rules \eqref{feynmanrulescubicmain} and \eqref{feynmanrulesquartic} as well as the selection rules of the propagator imply the existence of five diagrams for such a tensor $\Theta^{\{1\}(\ell+1)}_{\textcolor{color1}{12}}$: Two diagrams contain the quadratic vertices of $K^{\{1\}}$ and $\Theta^{\{1\}}$ at layer $\ell$, the other three diagrams contain the quartic vertices for $D$, $F$ and $V$,
\allowdisplaybreaks
\begin{align}
  \begin{tikzpicture}[baseline=(b)]
      \begin{feynman}
        \vertex (l) {};
        \vertex[right = 0pt of l, dot, color1, label = {above: {\textcolor{color1}{$1$}}}] (v12) {};
        \vertex[right = 22pt of v12, quarticblob] (b) {};
        \vertex[right = 22pt of b, dot, color1, label = {above: {\textcolor{color1}{$2$}}}] (v34) {};
        \vertex[right = 8pt of v34] (r) {};
        \diagram*{
          (v12) -- [color1, ghost] (b) -- [color1, ghost] (v34)
        };
        \vertex[below = 15pt of b] {{\scriptsize $\frac{1}{n_{\ell}}\Theta^{\{1\}(\ell+1)}_{\textcolor{color1}{12}}$}};
      \end{feynman}
    \end{tikzpicture}
  \hspace{-0.18cm} &= \hspace{-0.18cm}
  \begin{tikzpicture}[baseline=(b)]
    \begin{feynman}
      \vertex[dot, color1, minimum size=3pt, label = {below: {\footnotesize $\textcolor{color1}{1}$}}] (x1) {};
      \vertex[right = 20pt of x1] (v);
      \vertex[right = 20pt of v, dot, color1, minimum size = 3pt, label = {below: {\footnotesize $\textcolor{color1}{2}$}}] (x2) {};
      \vertex[above = 9pt of v] (b) {};
      \vertex[above = 4pt of v, label = {right: \!\*{\scriptsize $\sigma'_{j}\sigma'_{j}$}}] (j){};
      \tikzfeynmanset{every blob = {/tikz/fill=white!50, /tikz/minimum size=15pt}}
      \vertex[above = 16pt of v, blob] (G) {};
      \vertex[above = 22pt of v, dot, minimum size = 0pt] (g) {};
      \vertex[above = 20pt of g, dot, minimum size = 0pt] (w) {};
      \diagram*{
        (x1) -- [color1, ghost] (v) -- [color1, ghost] (x2),
        (v) -- [color1, doubghost] (G) -- (g) -- [color1, ghost, half left] (w) -- [color1, ghost, half left] (g)
      };
      \vertex[above = 0pt of G, blob] (K0){};
      \vertex[above = 20pt of g, smallblob] (K1) {};
      \vertex[above = 10pt of w] (K1) {{\scriptsize $\frac{1}{n_{\ell-1}} \Theta^{\{1\}(\ell)}$}};
      \vertex[right = 15pt of G, label = {}] (Gr) {};
    \end{feynman}
  \end{tikzpicture}
  \hspace{-0.18cm} + \hspace{-0.18cm}
  \begin{tikzpicture}[baseline=(b)]
    \begin{feynman}
      \vertex[dot, color1, minimum size=3pt, label = {below: {\footnotesize $\textcolor{color1}{1}$}}] (x1) {};
      \vertex[right = 20pt of x1] (v);
      \vertex[right = 20pt of v, dot, color1, minimum size = 3pt, label = {below: {\footnotesize $\textcolor{color1}{2}$}}] (x2) {};
      \vertex[above = 9pt of v] (b) {};
      \vertex[above = 4pt of v, label = {right: \!\*{\scriptsize $\widehat{\Delta \Omega}_{j,12}$}}] (j){};
      \vertex[above = 16pt of v, propblob] (G) {};
      \vertex[above = 22pt of v, dot, minimum size = 0pt] (g) {};
      \vertex[above = 20pt of g, dot, minimum size = 0pt] (w) {};
      \diagram*{
        (x1) -- [color1, ghost] (v) -- [color1, ghost] (x2),
        (v) -- [photon] (G) -- (g) -- [half left] (w) -- [half left] (g)
      };
      \vertex[above = 0pt of G, propblob] (K0){};
      \vertex[above = 20pt of g, smallblob] (K1) {};
      \vertex[above = 10pt of w] (K1) {{\scriptsize $\frac{1}{n_{\ell-1}} K^{\{1\}(\ell)}$}};
      \vertex[right = 15pt of G, label = {\scriptsize $z_j$}] (Gr) {};
    \end{feynman}
  \end{tikzpicture}
  \hspace{-0.18cm} + \hspace{-0.18cm}
  \begin{tikzpicture}[baseline=(b)]
    \begin{feynman}
      \vertex[dot, color1, minimum size = 3pt, label = {below: {\footnotesize $\textcolor{color1}{1}$}}] (x1) {};
      \vertex[right = 20pt of x1] (v);
      \vertex[right = 20pt of v, dot, color1, minimum size = 3pt, label = {below: {\footnotesize $\textcolor{color1}{2}$}}] (x2) {};
      \vertex[above = 9pt of v] (b) {};
      \vertex[above = 4pt of v, label = {right: \!{\scriptsize $\widehat{\Delta \Omega}_{j,12}$}}] (j){};
      \tikzfeynmanset{every blob = {/tikz/fill=black!50, /tikz/minimum size=15pt}}
      \vertex[above = 16pt of v, blob] (G) {};
      \vertex[left = 2pt of G, dot, minimum size = 0pt] (Gl) {};
      \vertex[right = 2pt of G, dot, minimum size = 0pt] (Gr) {};
      \vertex[above = 4pt of G, dot, minimum size = 0pt] (Gu) {};
      \vertex[left = 2pt of Gu, dot, minimum size = 0pt] (Gul) {};
      \vertex[right = 2pt of Gu, dot, minimum size = 0pt] (Gur) {};
      \vertex[above = 20pt of G, smallblob] (V4) {};
      \tikzfeynmanset{every blob = {/tikz/fill=white!50, /tikz/minimum size=15pt}}
      \vertex[left = 3pt of V4, dot, minimum size = 0pt] (V4l) {};
      \vertex[right = 3pt of V4, dot, minimum size = 0pt] (V4r) {};
      \diagram*{
        (x1) -- [color1, ghost] (v) -- [color1, ghost] (x2),
        (v) -- [photon] (G) -- (Gl) -- [half left] (V4l) -- [quarter right] (Gul) -- (Gur) -- [quarter right] (V4r) -- [half left] (Gr),
      };
      \vertex[above = 0pt of G, blob] (K0) {};
      \vertex[above = 10pt of V4] (VV4) {{\scriptsize $\frac{1}{n_{\ell-1}} V_4^{(\ell)}$}};
      \vertex[right = 18pt of G, label = {above: {\scriptsize $z_j$}}] (Gr) {};
    \end{feynman}
  \end{tikzpicture}\nonumber\\
  &\quad+ \hspace{-0.15cm}
  \begin{tikzpicture}[baseline=(b)]
    \begin{feynman}
      \vertex[dot, color1, minimum size = 3pt, label = {below: {\footnotesize $\textcolor{color1}{1}$}}] (x1) {};
      \vertex[right = 20pt of x1] (v);
      \vertex[right = 20pt of v, dot, color1, minimum size = 3pt, label = {below: {\footnotesize $\textcolor{color1}{2}$}}] (x2) {};
      \vertex[above = 9pt of v] (b) {};
      \vertex[above = 4pt of v, label = {right: \!{\scriptsize $\sigma'_{j}\sigma'_{j}$}}] (j){};
      \tikzfeynmanset{every blob = {/tikz/fill=black!50, /tikz/minimum size=15pt}}
      \vertex[above = 16pt of v, blob] (G) {};
      \vertex[left = 2pt of G, dot, minimum size = 0pt] (Gl) {};
      \vertex[right = 2pt of G, dot, minimum size = 0pt] (Gr) {};
      \vertex[above = 4pt of G, dot, minimum size = 0pt] (Gu) {};
      \vertex[left = 2pt of Gu, dot, minimum size = 0pt] (Gul) {};
      \vertex[right = 2pt of Gu, dot, minimum size = 0pt] (Gur) {};
      \vertex[above = 20pt of G, smallblob] (V4) {};
      \tikzfeynmanset{every blob = {/tikz/fill=white!50, /tikz/minimum size=15pt}}
      \vertex[left = 3pt of V4, dot, minimum size = 0pt] (V4l) {};
      \vertex[right = 3pt of V4, dot, minimum size = 0pt] (V4r) {};
      \diagram*{
        (x1) -- [color1, ghost] (v) -- [color1, ghost] (x2),
        (v) -- [color1, doubghost] (G) -- (Gl) -- [color1, ghost, half left] (V4l) -- [quarter right] (Gul) -- (Gur) -- [quarter right] (V4r) -- [color1, ghost, half left] (Gr),
      };
      \vertex[above = 0pt of G, blob] (K0) {};
      \vertex[above = 10pt of V4] (VV4) {{\scriptsize $\frac{1}{n_{\ell-1}} D_4^{(\ell)}$}};
      \vertex[right = 18pt of G, label = {above: {\scriptsize $z_j$}}] (Gr) {};
    \end{feynman}
  \end{tikzpicture}
  \hspace{-0.15cm} + \hspace{-0.15cm}
  \begin{tikzpicture}[baseline=(b)]
    \begin{feynman}
      \vertex[dot, color1, minimum size = 3pt, label = {below: {\footnotesize $\textcolor{color1}{1}$}}] (x1) {};
      \vertex[right = 20pt of x1] (v);
      \vertex[right = 20pt of v, dot, color1, minimum size = 3pt, label = {below: {\footnotesize $\textcolor{color1}{2}$}}] (x2) {};
      \vertex[above = 9pt of v] (b) {};
      \vertex[above = 4pt of v, label = {right: \!{\scriptsize $\sigma'_{j}\sigma'_{j}$}}] (j){};
      \tikzfeynmanset{every blob = {/tikz/fill=black!50, /tikz/minimum size=15pt}}
      \vertex[above = 16pt of v, blob] (G) {};
      \vertex[left = 2pt of G, dot, minimum size = 0pt] (Gl) {};
      \vertex[right = 2pt of G, dot, minimum size = 0pt] (Gr) {};
      \vertex[above = 4pt of G, dot, minimum size = 0pt] (Gu) {};
      \vertex[left = 2pt of Gu, dot, minimum size = 0pt] (Gul) {};
      \vertex[right = 2pt of Gu, dot, minimum size = 0pt] (Gur) {};
      \vertex[above = 20pt of G, smallblob] (V4) {};
      \tikzfeynmanset{every blob = {/tikz/fill=white!50, /tikz/minimum size=15pt}}
      \vertex[left = 3pt of V4, dot, minimum size = 0pt] (V4l) {};
      \vertex[right = 3pt of V4, dot, minimum size = 0pt] (V4r) {};
      \diagram*{
        (x1) -- [color1, ghost] (v) -- [color1, ghost] (x2),
        (v) -- [color1, doubghost] (G) -- (Gl) -- [color1, ghost, half left] (V4l) -- [quarter right] (Gul) -- (Gur) -- [color1, ghost, quarter right] (V4r) -- [half left] (Gr),
      };
      \vertex[above = 0pt of G, blob] (K0) {};
      \vertex[above = 10pt of V4] (VV4) {{\scriptsize $\frac{1}{n_{\ell-1}} F_4^{(\ell)}$}};
      \vertex[right = 18pt of G, label = {above: {\scriptsize $z_j$}}] (Gr) {};
    \end{feynman}
  \end{tikzpicture}\label{firstcorrectionntk}
\end{align}
The algebraic expression corresponding to~\eqref{firstcorrectionntk} is given in Appendix~\ref{app:NLO_ntk_mean}.

\subsection{Gradient Stability at Finite Width}\label{sec:stability}
Stabilizing against exploding or vanishing preactivations and gradients is of central importance in deep learning~\cite{glorot2010, he2015a}. In order to analyze the stability of deep networks at infinite width, one decomposes the NNGP into its fixed point $K^\star$ and the deviation $\Delta K^{(\ell)}$ from it: $K^{(\ell)} = K^\star + \Delta K^{(\ell)}$. The NNGP recursion then becomes a recursion for $\Delta K$, and the susceptibility $\chi = \frac{\mathrm{d} \Delta K^{(\ell+1)}}{\mathrm{d} \Delta K^{(\ell)}}$ captures how quickly the NNGP approaches or deviates from the fixed point as depth increases~\cite{poole2016}. The susceptibility also describes the growth behavior of the NTK~\cite{xiao2020} and yields an activation-function dependent phase diagram for the initialization variances of weights and biases. At criticality, when gradients and preactivations neither grow nor vanish exponentially, $\chi=1$.
Recently, the analysis of preactivation stability has been expanded to finite-width. In particular, \cite{banta2024} could show that all higher-order statistics of preactivations at finite width are at criticality if the infinite width is at criticality. Therefore, the forward pass is also stable at finite width. In this section, we will extend this result to statistics involving the NTK, i.e.\ to the backward pass.


The criticality of a tensor can be analyzed by computing how variations of the tensors in layer $\ell$ lead to variations at layer $\ell+1$~\cite{roberts2022, banta2024}.
Our Feynman rules enable a straightforward computation of the variations of tensors involving the NTK. In particular, these variations follow from the observation that NTK external lines propagate through the diagram exclusively via dashed internal lines, which connect to the propagator. These, in turn, generate dotted NTK lines within the diagram, yielding susceptibility factors determined by criticality at infinite width. This leads to the following

\begin{restatable}[]{theorem}{criticalityntk}
\label{theoremfour}
If the NNGP is critical, then any cumulant involving the NTK is also critical.
\end{restatable}
\begin{proof}
See Appendix~\ref{app:criticality_analysis}.
\end{proof}
The proof of this theorem relies fundamentally on the Feynman diagram formalism, providing a clear illustration of how the NTK Feynman rules can be systematically applied to derive powerful results valid to all orders in $\frac{1}{n}$.

\subsection{Absence of NTK-Corrections for Scale-Invariant Nonlinearities}\label{absence-NTK-corrections-ReLU}
Scale-invariant nonlinearities satisfy $\sigma(\lambda z)=\lambda\sigma(z)$ for $\lambda>0$. Important examples include ReLU and LeakyReLU. Remarkably, these activation functions have the property that their NNGP does not receive finite-width corrections for equal inputs, so the infinite-width result is valid for finite-width networks as well, $\widehat K(x,x)=K(x,x)$~\cite{roberts2022}.

We now derive a corresponding result for the finite-width correction to the NTK mean in MLPs with scale-invariant nonlinearities. While the ReLU case was previously established using alternative approaches~\cite{hanin2019,seleznova2022}, the general case follows directly from the NTK Feynman rules of Section~\ref{sec:feynman-rules} together with elementary integration techniques. We thus obtain the following theorem.

\begin{restatable}[]{theorem}{scaleinvariantactivations}
\label{theoremfive}
For scale-invariant activation functions, the diagonal component of the NTK mean does not acquire any finite-width corrections.
\end{restatable}
\begin{proof}
See Appendix~\ref{app:proof-identity-relu}
\end{proof}
This provides yet another instance where the diagrammatic approach proves effective in deriving all-order results in $\frac{1}{n}$.


\begin{figure*}[tb]
  \centering
  \includegraphics[width=0.95\textwidth]{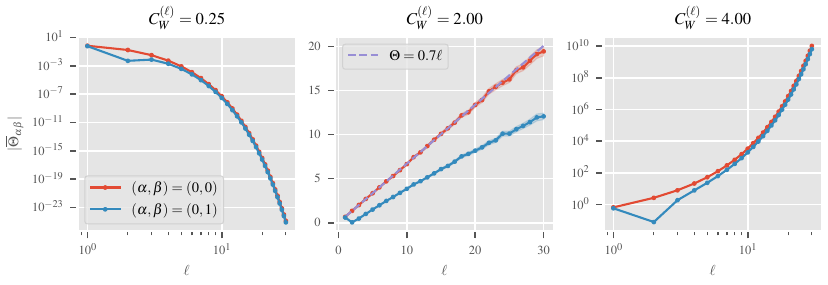}
  \vspace{-0.1cm}
  \caption{\emph{Gradient Stability.} Components of the Monte--Carlo estimated NTK $\overline{\Theta}_{\alpha \beta}$ of a ReLU MLP as a function of layer depth $\ell$ corresponding to single and distinct inputs are shown for three different choices of $C_W^{(\ell)}$. The hidden layers are of size 200. The case for the critical value $C_W^{(\ell)} = C_W^\mathrm{c}=2$ is shown in the middle. For the single input case at criticality (red line in the middle), we also show the expected linear relation~\cite{roberts2022} (purple). Sample means are obtained from \num{1000} initializations. Error bands are standard errors of the mean but mostly too small to be visible.} \label{fig:ntk_comparison}
  \vspace{-0.3cm}
\end{figure*}

\section{Experiments}\label{sec:experiments}
In this section, we summarize our numerical experiments which verify the results from Section~\ref{sec:applications}.


\subsection{Solving the Recursion Relations}


A key question that arises is how well the presented width-dependent contributions capture the statistics of networks at various widths compared to the infinite-width solution. Testing this is non-trivial since it requires solving the coupled recursion relations.
Each one of them contains several Gaussian expectations in up to four dimensions over functions of $z_\alpha$ which do not have an analytic solution for arbitrary activation functions. In favor of generality, we therefore compute the corresponding integrals numerically. A naive implementation at reasonable precision is prohibitively expensive due to the high number of 4$d$ numerical integrals involved. However, most integrals can be reduced to sums of lower-dimensional integrals using the simple marginal distribution of multivariate Gaussian random variables.
The computational cost is further lowered by recognizing those partial derivatives that evaluate to zero, trading symbolic brevity for numerical efficiency.
In addition, applying tensor contraction rules and exploiting tensor symmetries, the number of terms that need to be computed can be reduced significantly. In fact, we implement custom \texttt{SymPy} \cite{sympy2017} routines to handle the involved steps in a general fashion and transform the generated symbolic expressions into numerical functions afterwards. This results in a flexible framework that can easily be extended to other recursion relations and is agnostic to the activation function.

Using this setup, we compute the first-order correction to the NTK, i.e.\  $\Theta^{\{1\}}$ from \eqref{firstcorrectionntk}. This requires simultaneously solving the recursions \eqref{ftensorfeynmandiagram} and \eqref{dtensorfeynman} for $F$ and $D$, respectively, as well as $V$ and $K^{\{1\}}$ given in \eqref{eq:v4_recursion} and \eqref{eq:k1_recursion}.
We compare the frozen NTK $\Theta^{(\ell)}_{\alpha \beta}$ --- computed with the \texttt{neural-tangents} library~\cite{novak2020a} --- and the finite-width corrected NTK $\Theta^{(\ell)}_{\alpha \beta} + \Theta^{\{1\}(\ell)}_{\alpha \beta} / n_\ell$ to the sampled mean kernel. The latter is computed as \mbox{$\overline{\Theta}^{(\ell)}_{\alpha \beta} = \frac{1}{N_\mathrm{net}} \sum_{I=1}^{N_\mathrm{net}} \widehat{\Theta}^{(\ell)}_{I, \alpha \beta}$}, where $\widehat{\Theta}^{(\ell)}_{I, \alpha \beta}$ is the empirical NTK defined in \eqref{eq:10} and $I$ labels the associated independently initialized networks. We choose GeLU as the activation function. The multi-input case for an MLP at layer four is shown in Figure \ref{fig:kernel_finite_width_corrected} as well as the analogous finite-width corrections to the NNGP $K$. Indeed, we observe close agreement between the sampled kernel and the finite-width corrected analytic solution already at widths $n_\ell > 20$.

For further details about the implementation and the experimental setup, we refer the reader to Appendices \ref{app:implementation} and \ref{app:experiments}. In the latter, we also provide additional plots of further kernel components and the solved recursions for $A, B, D, F$ and $V$, confirming strong agreement with the empirical values as well. We also isolate the higher order corrections $\Theta^{\{1\}}$ and $K^{\{1\}}$ from the full NTK and NNGP, respectively, and verify their convergence.

\subsection{Gradient Stability}
We verify numerically for finite-width ReLU-MLPs that the preactivation and NTK statistics are stabilized for the critical initialization $C_W^{(\ell)}=C_W^\mathrm{c}=2$, whereas larger (lower) values lead to exponential growth (decay) with depth. In particular,
we compute the ensemble average \mbox{$\overline{\Theta}^{(\ell)}_{\alpha \beta}$} for up to 30 layers. Figure~\ref{fig:ntk_comparison} shows the linear scaling for both the single and the distinct input case when $C_W^{(\ell)}=C_W^\mathrm{c}$, in contrast to the exponential behavior observed away from this value. In the Appendices~\ref{app:implementation} and \ref{app:experiments} we provide further details and also numerically verify the simultaneous stability of the NNGP $K$, the four-point cumulant $V$ and the tensors $A, B, D$ and $F$ which together describe the complete preactivation and NTK statistics up to rank four. 

\subsection{Absence of Kernel-Corrections for Scale-Invariant Activations}

\begin{figure}[bt]
  \centering
  \includegraphics[width=0.38\textwidth]{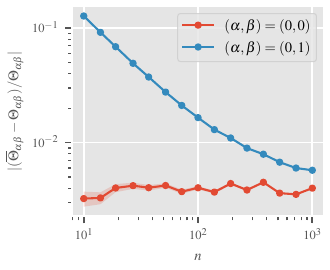}
  \vspace{-0.1cm}
  \caption{\emph{Finite-Width Corrections for ReLU.} Relative deviations of the Monte--Carlo estimated NTK to its infinite-width counterpart as a function of hidden layer width $n=n_\ell$. A four layer ReLU MLP with $C_W=2$ is sampled over \num{5e6} initializations. Error bars of the sample mean are included for both the single and distinct input component.}\label{fig:relu_corrections}
  \vspace{-0.5cm}
\end{figure}

We measure the finite-width corrections to the infinite-width NTK by computing the relative deviation $|(\overline{\Theta}_{\alpha \beta} - \Theta_{\alpha \beta})/\Theta_{\alpha \beta}|$ of the ensemble average to the frozen NTK at different hidden-layer widths $n = n_\ell$. The NNGP corrections are investigated analogously and shown in Figure \ref{fig:nngp_relu_corrections}. The results for the single and distinct input case for a ReLU MLP are shown in Figure~\ref{fig:relu_corrections}. Indeed, the single input component does not receive corrections, while the distinct input case only approaches the infinite-width result as $n \to \infty$.
For further details, we refer to the Appendix~\ref{app:experiments}, where we also check the absence of corrections for LeakyReLU and show that the kernels do receive corrections for GeLU, which is not scale-invariant.


\section{Limitations}\label{sec:conclusion}

Although the framework of computing finite-width corrections to the NTK applies to very general architectures, we have restricted our attention here to MLPs. Extensions to other architectures would require modifications of the Feynman rules, but we see no conceptual obstacle.

Furthermore, our analysis was mostly restricted to order $1/n$. The discussion in Section~\ref{sec:stability} involves all orders in $1/n$ and shows examples of the higher-rank generalizations of the tensors appearing at higher order in $1/n$. We expect such straightforward generalizations to be able to capture all orders in $1/n$, and have performed further tests at order $1/n^{2}$ which showed no contradictions, see Appendix~\ref{app:example_NNLO_tensor}. However, a full analysis of higher-order tensors is beyond the scope of this work.


\FloatBarrier

\section*{Acknowledgments}
We would like to thank Yonatan Kahn and Zhengkang (Kevin) Zhang for inspiring discussions.

This work is supported by the Wallenberg AI, Autonomous Systems and Software Program (WASP) funded by the Knut and Alice Wallenberg Foundation. The computations were enabled by resources provided by the National Academic Infrastructure for Supercomputing in Sweden (NAISS), partially funded by the Swedish Research Council through grant agreement no. 2022-06725.


\section*{Impact Statement}
This paper presents work whose goal is to advance the field of Machine Learning. There are many potential societal consequences of our work, none which we feel must be specifically highlighted here.


\addtolength{\bibsep}{-0.3pt}
\bibliography{icml_2026_content/ntk_feynman}

\bibliographystyle{icml2026}

\ifbool{includeapp}{

\newpage
\appendix
\onecolumn

\section{Cumulants versus Moments}
\label{app:cumulants}

Joint cumulants are generated by the logarithm of the moment generating function. By expanding these generating functions, one arrives at the moment-cumulants formula~\cite{leonov1959}, which specifies how a mixed moment is expanded in terms of joint cumulants,
\begin{align}
  \EE[X_{1}\cdots X_{n}]=\sum_{p\in P_{n}}\prod_{b\in p}\,\EE^{c}\!\!\left[ \prod_{i\in b} X_{i}\right]\,,\label{eq:5}
\end{align}
where $P_{n}$ is the set of partitions of $\{1,\dots,n\}$, hence $b$ is a block in the partition $p$. For instance,
\begin{align}
  \EE[X_{1}X_{2}X_{3}]&=\EE^{c}[X_{1},X_{2},X_{3}]+\EE^{c}[X_{1},X_{2}]\EE^{c}[X_{3}]+\EE^{c}[X_{1},X_{3}]\EE^{c}[X_{2}]+\EE^{c}[X_{2},X_{3}]\EE^{c}[X_{1}]\nonumber\\
  &\qquad+\EE^{c}[X_{1}]\EE^{c}[X_{2}]\EE^{c}[X_{3}]\,.
\end{align}
We will use the decomposition~\eqref{eq:5} to expand mixed moments in terms of cumulants which we then compute using Feynman diagrams.

Intuitively, joint cumulants correspond to subtracting all factorizations of the mixed moment, e.g.
\begin{align}
  \EE^{c}[X_{1},X_{2},X_{3}]&=\EE[X_{1}X_{2}X_{3}]-\EE[X_{1}X_{2}]\EE[X_{3}]-\EE[X_{1}X_{3}]\EE[X_{2}]-\EE[X_{2}X_{3}]\EE[X_{1}]\nonumber\\
  &\qquad+2\EE[X_{1}]\EE[X_{2}]\EE[X_{3}]\,,
\end{align}
where the last term is added twice to compensate for the factorizations in the three terms before. In general, a joint cumulant is given in terms of mixed moments by~\cite{leonov1959}
\begin{align}
  \EE^{c}[X_{1},\cdots, X_{n}]=\sum_{p\in P_{n}}(|p|-1)!(-1)^{|p|-1}\prod_{b\in p}\EE\left[ \prod_{i\in b}X_{i} \right]\,,
\end{align}
where $|p|$ is the number of blocks in the partition $p$.


\section{Network Architecture and Initialization}\label{app:initialization}

We use the NTK parametrization and consider MLPs, i.e.\ the preactivations are given by
\begin{align}
\label{eq:nn_initialization}
z^{(\ell)}_{i}(x) &= \frac{1}{\sqrt{n_{\ell-1}}}\sum_{j=1}^{n_{\ell-1}}W_{ij}^{(\ell)}\sigma(z^{(\ell-1)}_{j}(x))\, \, \, \, \text{for} \, \, \, \, i=1,\ldots,n_{\ell}\,,
\end{align}
where the weights are sampled i.i.d $W_{ij}^{(\ell)} \sim \mathcal{N}(0, C_W^{(\ell)})$. The learning rate does not scale with width.


\section{Definitions of All Tensors}
\label{app:defin-all-tens}

In this section, we provide the definitions for all tensors relevant in this work beyond the NNGP~\eqref{eq:4} and the NTK~\eqref{eq:10}.

The means of the NNGP and NTK receive finite-width corrections which are denoted by
\begin{align}
  \EE[\widehat{K}^{(\ell)}_{\alpha_1 \alpha_2} ]
&= K^{(\ell)}_{\alpha_1 \alpha_2} 
+ \frac{1}{n_{\ell-1}} K^{\{1\}(\ell)}_{\alpha_1 \alpha_2} 
+ O\left( \frac{1}{n^2} \right)\label{k1correction}\\
\EE[\widehat{\Theta}^{(\ell)}_{\alpha_1 \alpha_2} ]
&= \Theta^{(\ell)}_{\alpha_1 \alpha_2} 
+ \frac{1}{n_{\ell-1}} \Theta^{\{1\}(\ell)}_{\alpha_1 \alpha_2} 
+ O\left( \frac{1}{n^2} \right)\,,
\end{align}
where the unhatted quantities correspond to the infinite-width limits. The cumulant of four preactivations decomposes into the $V$ tensor, defining it according to
\begin{align}\label{vfourvertex}
\mathbb{E}_{\theta}^{c} \left[ z^{(\ell)}_{i_1; \alpha_1}, z^{(\ell)}_{i_2; \alpha_2}, z^{(\ell)}_{i_3; \alpha_3}, z^{(\ell)}_{i_4; \alpha_4} \right]
&= \frac{1}{n_{\ell-1}} \Big[
\delta_{i_1 i_2} \delta_{i_3 i_4} V^{(\ell)}_{\alpha_1 \alpha_2\alpha_3 \alpha_4} +
\delta_{i_1 i_3} \delta_{i_2 i_4} V^{(\ell)}_{\alpha_1 \alpha_3\alpha_2 \alpha_4} \notag \\
&\qquad+ \delta_{i_1 i_4} \delta_{i_2 i_3} V^{(\ell)}_{\alpha_1 \alpha_4\alpha_2 \alpha_3} \Big]\,.
\end{align}
The cumulant of two NTK fluctuations decomposes into the $A$ and $B$ tensors,
\begin{align}\label{defabfourpoint}
  \mathbb{E}_{\theta}^{c} \left[ \widehat{\Delta \Theta}^{(\ell)}_{i_1 i_2; \alpha_1 \alpha_2}, \widehat{\Delta \Theta}^{(\ell)}_{i_3 i_4; \alpha_3 \alpha_4} \right] 
  &=
  \frac{1}{n_{\ell-1}} 
  \left[ 
    \delta_{i_1 i_2} \delta_{i_3 i_4} A^{(\ell)}_{\alpha_1\alpha_2\alpha_3\alpha_4}
    + \delta_{i_1 i_3} \delta_{i_2 i_4} B^{(\ell)}_{\alpha_1 \alpha_3 \alpha_2 \alpha_4}\right.\nonumber\\
    &\qquad+ \left.\delta_{i_1 i_4} \delta_{i_2 i_3} B^{(\ell)}_{\alpha_1 \alpha_4 \alpha_2 \alpha_3} \right]\,.
\end{align}
In order to consistently analyze the training dynamics to order $1/n$, the following higher-derivative analogues of the NTK are necessary~\cite{roberts2022}
\begin{align}
  \widehat{\text{d}\Theta}_{i_0 i_1 i_2; \delta_0 \delta_1 \delta_2}^{(\ell)} 
  &= \sum_{\ell_1, \ell_2 = 1}^{\ell} \sum_{\mu_1,\mu_2} 
  \frac{d^2 z_{i_{0};\delta_{0}}^{(\ell)}}{d\theta_{\mu_{1}}^{(\ell_1)} d\theta_{\mu_{2}}^{(\ell_2)}} 
  \frac{dz_{i_{1};\delta_{1}}^{(\ell)}}{d\theta_{\mu_{1}}^{(\ell_1)}}
  \frac{dz_{i_{2};\delta_{2}}^{(\ell)}}{d\theta_{\mu_{2}}^{(\ell_2)}}  \label{eq:12}\\
  \widehat{\text{dd}_{\text{I}}\Theta}_{i_0i_1i_2i_3;\delta_0\delta_1\delta_2\delta_3}^{(\ell)}
  &= \sum_{\ell_1, \ell_2, \ell_3=1}^{\ell} \sum_{\mu_{1},\mu_{2},\mu_{3}}
  \frac{d^3 z_{i_0;\delta_0}^{(\ell)}}{d\theta_{\mu_1}^{(\ell_1)} d\theta_{\mu_2}^{(\ell_2)} d\theta_{\mu_3}^{(\ell_3)}}
  \frac{dz_{i_1;\delta_1}^{(\ell)}}{d\theta_{\mu_1}^{(\ell_1)}}
  \frac{dz_{i_2;\delta_2}^{(\ell)}}{d\theta_{\mu_2}^{(\ell_2)}}
  \frac{dz_{i_3;\delta_3}^{(\ell)}}{d\theta_{\mu_3}^{(\ell_3)}}\label{eq:13}\\
  \widehat{\text{dd}_{\text{II}}\Theta}^{(\ell)}_{i_1 i_2 i_3 i_4; \delta_1 \delta_2 \delta_3 \delta_4} & = \sum_{\ell_1, \ell_2, \ell_3=1}^{\ell} \sum_{\mu_1,\mu_2, \mu_3} 
\frac{d^2 z^{(\ell)}_{i_1; \delta_1}}{d\theta_{\mu_1}^{(\ell_1)} d\theta_{\mu_3}^{(\ell_3)}} \frac{d^2 z^{(\ell)}_{i_2; \delta_2}}{d\theta_{\mu_2}^{(\ell_2)} d\theta_{\mu_3}^{(\ell_3)}} \frac{d z^{(\ell)}_{i_3; \delta_3}}{d\theta_{\mu_1}^{(\ell_1)}} \frac{d z^{(\ell)}_{i_4; \delta_4}}{d\theta_{\mu_2}^{(\ell_2)}}\,.\label{eq:14}
\end{align}
Their the definitions of their cumulants define the tensors $P$, $Q$, $R$, $S$, $T$ and $U$,
\begin{align}
  \mathbb{E}_{\theta}^{c} \left[ \widehat{\mathrm{d}\Theta}^{(\ell)}_{i_0 i_1 i_2; \delta_0 \delta_1 \delta_2 },z^{(\ell)}_{i_3; \delta_3} \right] &{=}
  \frac{1}{n_{\ell-1}} 
  \left[ 
    \delta_{i_0 i_3} \delta_{i_1 i_2} P^{(\ell)}_{\delta_0 \delta_1 \delta_2 \delta_3}
    + 
    \delta_{i_0 i_1} \delta_{i_2 i_3} Q^{(\ell)}_{\delta_0 \delta_5 \delta_6 \delta_7}
    + 
    \delta_{i_0 i_2} \delta_{i_1 i_3} Q^{(\ell)}_{\delta_{0} \delta_{2} \delta_{1} \delta_{3}}
  \right]\\
  \mathbb{E}_{\theta}^{c} \left[ \widehat{\text{dd}_{\text{I}}\Theta}_{i_0i_1i_2i_3;\delta_0\delta_1\delta_2\delta_3}^{(\ell)} \right]
  &{=} \frac{1}{n_{\ell-1}} \left[ \delta_{i_0i_1}\delta_{i_2i_3} R_{\delta_0\delta_1\delta_2\delta_3}^{(\ell)} + \delta_{i_0i_2}\delta_{i_3i_1} R_{\delta_0\delta_2\delta_3\delta_1}^{(\ell)} + \delta_{i_0i_3}\delta_{i_1i_2} R_{\delta_0\delta_3\delta_1\delta_2}^{(\ell)} \right]\\
  \mathbb{E}_{\theta}^{c} \left[ \widehat{\text{dd}_{\text{II}}\Theta}_{i_1i_2i_3i_4;\delta_1\delta_2\delta_3\delta_4}^{(\ell)} \right]
  &{=} \frac{1}{n_{\ell-1}} \left[ \delta_{i_1i_2}\delta_{i_3i_4} S_{\delta_1\delta_2\delta_3\delta_4}^{(\ell)} + \delta_{i_1i_3}\delta_{i_4i_2} T_{\delta_1\delta_3\delta_4\delta_2}^{(\ell)} + \delta_{i_1i_4}\delta_{i_2i_3} U_{\delta_1\delta_4\delta_2\delta_3}^{(\ell)} \right]\,.
\end{align}


\section{Feynman Rules} 
\label{app:feynman_rules}
In this appendix, we provide the complete list of Feynman rules, including those described in the summary in Section~\ref{sec:feynman-rules} above.

\begin{enumerate}[label=(\roman*)]
\item External vertices (which correspond to the quantities which appear in the expectation value) are represented by filled dots.

  \begin{enumerate}[label=(\alph*)]
  \item \emph{Solid external lines} represent preactivations~\cite{banta2024}
    \begin{align}
      &z_{\alpha} \equiv
      \begin{tikzpicture}[baseline=-0.1cm]
        \begin{feynman}
          \vertex (l) {};
          \vertex[right = 0pt of l, dot, minimum size=3pt, label = {left: {\footnotesize $\alpha^{}$}}] (x1) {};
          \vertex[right = 30pt of x1, dot, minimum size=0pt] (b) {};
          \diagram*{
            (x1),
            (x1) -- [inner sep = 4pt] (b) 
          };
        \end{feynman}
      \end{tikzpicture}
    \end{align}
  \item \emph{Dotted} external lines represent NTK fluctuations
    \begin{align}
      \widehat{\Delta \Theta}_{\textcolor{color1}{\alpha\beta}} \equiv
      \begin{tikzpicture}[baseline=0.05cm]
        \begin{feynman}
          \vertex (l) {};
          \vertex[right = 0pt of l, color1, dot, minimum size=3pt, label = {left: {\footnotesize $\textcolor{color1}{\alpha^{}}$}}] (x1) {};
          \vertex[above = 8pt of l, color1, dot, minimum size=3pt, label = {left: {\footnotesize $\textcolor{color1}{\beta^{}}$}}] (x2) {};
          \vertex[right = 30pt of x1, dot, minimum size=0pt] (b) {};
          \vertex[right = 30pt of x2, dot, minimum size=0pt] (bb) {};
          \diagram*{
            (x1), (x2),
            (x1) -- [color1, ghost, inner sep = 4pt] (b),
            (x2) -- [color1, ghost, inner sep = 4pt] (bb)
          };
        \end{feynman}
      \end{tikzpicture}
    \end{align}
    We will use different colors for external dotted lines corresponding to different NTKs.
  \item dNTKs, $\widehat{\mathrm{d} \Theta}_{\delta_0\delta_1\delta_2}
    =\sum_{\mu\nu}\frac{d^2 z_{\delta_0}}{d\theta_{\mu}d\theta_{\nu}}\frac{d z_{\delta_1}}{d\theta_\mu}\frac{d z_{\delta_2}}{d\theta_\nu}$ as defined in~\eqref{eq:12}, are represented by one dotted double line and two dotted single lines. They carry the sample indices of the dNTK according to
    \begin{equation}
      \widehat{\mathrm{d} \Theta}_{\textcolor{color3}{\delta_0}\textcolor{color1}{\delta_1}\textcolor{color2}{\delta_2}} \equiv \begin{tikzpicture}[baseline=(bb)]
        \begin{feynman}
          \vertex (l) {};
          \vertex[above = 12pt of l, color1, dot, minimum size=3pt, label = {left: {\footnotesize $\textcolor{color1}{\delta_1^{}}$}}] (x1) {};
          \vertex[above = 24pt of l, color2, dot, minimum size=3pt, label = {left: {\footnotesize $\textcolor{color2}{\delta_2^{}}$}}] (x2) {};
          \vertex[above = 0pt of l, color3, dot, minimum size=3pt, label = {left: {\footnotesize $\textcolor{color3}{\delta_0}^{}$}}] (x3) {};
          \vertex[right = 30pt of x1, dot, minimum size=0pt] (bb) {};
          \vertex[right = 30pt of x2, dot, minimum size=0pt] (bbb) {};
          \vertex[right = 30pt of x3, dot, minimum size=0pt] (b) {};
          \diagram*{
            (x1), (x2), (x3),
            (x3) -- [color1color2dntknew, inner sep = 4pt] (b),
            (x1) -- [color1, ghost, inner sep = 4pt] (bb),
            (x2) -- [color2, ghost, inner sep = 4pt] (bbb)    
          };
        \end{feynman}
      \end{tikzpicture}
    \end{equation}
    Here, the colors of the dotted lines again correspond to the $\theta$ in the definition of the dNTK. Since now, there are two different $\theta$ (the four $\theta$ appearing in the definition are contracted in pairs), there are two different colors. The double line corresponds to the second derivative, the two single lines correspond to the first derivatives. The colors indicate the contraction pattern between the $\theta$.

  \item The dd$_{\text{I}}$NTK, $
    \widehat{\mathrm{dd_I} \Theta}_{\delta_0\delta_1\delta_2\delta_3}
    =\sum_{\mu\nu\rho}\frac{d^3 z_{\delta_0}}{d\theta_{\mu}d\theta_{\nu}d\theta_\rho}\frac{d z_{\delta_1}}{d\theta_\mu}\frac{d z_{\delta_2}}{d\theta_\nu}\frac{d z_{\delta_3}}{d\theta_\rho}
    $, defined by~\eqref{eq:13} is represented by
    \begin{equation}
      \widehat{\mathrm{dd_I} \Theta}_{\textcolor{color4}{\delta_0}\textcolor{color2}{\delta_1}\textcolor{color1}{\delta_2}\textcolor{color6}{\delta_3}} \equiv \begin{tikzpicture}[baseline=(b)]
        \begin{feynman}
          \vertex (l) {};
          \vertex[above = 12pt of l, color2, dot, minimum size=3pt, label = {left: {\footnotesize $\textcolor{color2}{\delta_1^{}}$}}] (x1) {};
          \vertex[above = 24pt of l, color1, dot, minimum size=3pt, label = {left: {\footnotesize $\textcolor{color1}{\delta_2^{}}$}}] (x2) {};
          \vertex[above = 36pt of l, color6, dot, minimum size=3pt, label = {left: {\footnotesize $\textcolor{color6}{\delta_3^{}}$}}] (x3) {};
          \vertex[right = 0pt of l, color4, dot, minimum size=3pt, label = {left: {\footnotesize $\textcolor{color4}{\delta_0^{}}$}}] (x4) {};
          \vertex[right = 30pt of x1, dot, minimum size=0pt] (b) {};
          \vertex[right = 30pt of x2, dot, minimum size=0pt] (bb) {};
          \vertex[right = 30pt of x3, dot, minimum size=0pt] (bbb) {};
          \vertex[right = 30pt of x4, dot, minimum size=0pt] (bbbb) {};
          \diagram*{
            (x1), (x2), (x3), (x4),
            (x1) -- [color2, ghost, inner sep = 4pt] (b),
            (x2) -- [color1, ghost, inner sep = 4pt] (bb),
            (x3) -- [color6, ghost, inner sep = 4pt] (bbb),
            (x4) -- [color6color1color2ddntknew, inner sep = 4pt] (bbbb)
          };
        \end{feynman}
      \end{tikzpicture}
    \end{equation}
    As before, we use three colored dotted lines due to the presence of three types of $\theta$. The triple line denotes the third derivative, and the three single lines represent the first derivatives. The colors indicate the contraction pattern among the $\theta$.

  \item dd$_{\text{II}}$NTKs, $\widehat{\mathrm{dd_{II}} \Theta}_{\delta_1\delta_2\delta_3\delta_4}
    =\sum_{\mu\nu\rho}\frac{d^2 z_{\delta_1}}{d\theta_{\mu}d\theta_{\nu}}\frac{d^2 z_{\delta_2}}{d\theta_{\rho}d\theta_{\nu}}\frac{d z_{\delta_3}}{d\theta_\mu}\frac{d z_{\delta_4}}{d\theta_\rho}
    $, as defined in~\eqref{eq:14}, are denoted by
    \begin{equation}
      \widehat{\mathrm{dd_{II}} \Theta}_{\textcolor{color5}{\delta_1}\textcolor{color7}{\delta_2}\textcolor{color1}{\delta_3}\textcolor{color2}{\delta_4}} \equiv \begin{tikzpicture}[baseline=(bbbb)]
        \begin{feynman}
          \vertex (l) {};
          \vertex[above = 24pt of l, color1, dot, minimum size=3pt, label = {left: {\footnotesize $\textcolor{color1}{\delta_3^{}}$}}] (x1) {};
          \vertex[above = 36pt of l, color2, dot, minimum size=3pt, label = {left: {\footnotesize $\textcolor{color2}{\delta_4^{}}$}}] (x2) {};
          \vertex[right = 0pt of l, color5, dot, minimum size=3pt, label = {left: {\footnotesize $\textcolor{color5}{\delta_1^{}}$}}] (x3) {};
          \vertex[above = 12pt of l, color7, dot, minimum size=3pt, label = {left: {\footnotesize $\textcolor{color7}{\delta_2^{}}$}}] (x4) {};
          \vertex[right = 30pt of x1, dot, minimum size=0pt] (b) {};
          \vertex[right = 30pt of x2, dot, minimum size=0pt] (bb) {};
          \vertex[right = 30pt of x3, dot, minimum size=0pt] (bbb) {};
          \vertex[right = 30pt of x4, dot, minimum size=0pt] (bbbb) {};
          \diagram*{
            (x1), (x2), (x3), (x4),
            (x1) -- [color1, ghost, inner sep = 4pt] (b),
            (x2) -- [color2, ghost, inner sep = 4pt] (bb),
            (x3) -- [color6color1ddntknew, inner sep = 4pt] (bbb),
            (x4) -- [color6color2ddntknew, inner sep = 4pt] (bbbb)
          };
        \end{feynman}
      \end{tikzpicture}
    \end{equation}
    Here we use again three colored lines for the three types of $\theta$. The two double lines correspond to the second derivatives, and the two single lines represent the first derivatives. The colors indicate the contraction pattern among the $\theta$.
  \end{enumerate}

\item The external lines are attached to \emph{cubic vertices}, which are vertices connected to two external lines and one \emph{internal line}. Two external preactivations are joined in the vertex~\cite{banta2024}
  \begin{align}
    \begin{tikzpicture}[baseline=(b)]
      \begin{feynman}
        \vertex (l) {};
        \vertex[below = 15pt of l, dot, minimum size=3pt, label = {left: {\footnotesize $\alpha^{}$}}] (x1) {};
        \vertex[above = 15pt of l, dot, minimum size=3pt, label = {left: {\footnotesize $\beta^{}$}}] (x2) {};
        \vertex[right = 10pt of l, dot, minimum size=0pt] (v12) {};
        \vertex[right = 30pt of v12, dot, minimum size=0pt] (b) {};
        \diagram*{
          (x1) --  (v12) --  (x2),
          (v12) -- [photon, edge label = {\scriptsize \;$\widehat{\Delta G}^{(\ell)}_{i,\alpha\beta}$}, inner sep = 4pt] (b) 
        };
      \end{feynman}
    \end{tikzpicture} \sim \frac{C^{(\ell+1)}_{W}}{n_{\ell}}
  \end{align}
  where $\widehat{\Delta G}^{(\ell)}_{i,\alpha\beta}=\sigma^{(\ell)}_{i,\alpha}\sigma^{(\ell)}_{i,\beta}-\langle \sigma^{(\ell)}_{i,\alpha}\sigma^{(\ell)}_{i,\beta} \rangle_{K^{(\ell)}}$. The line without dot is the internal line. We discuss its decoration in the next Feynman rule. For cubic interactions involving external NTK lines, we introduce the following Feynman rules
  \begingroup
  \allowdisplaybreaks
  \begin{align}
    \begin{tikzpicture}[baseline=(b)]
      \begin{feynman}
        \vertex (l) {};
        \vertex[below = 15pt of l, color1, dot, minimum size=3pt, label = {left: {\footnotesize $\textcolor{color1}{\alpha^{}}$}}] (x1) {};
        \vertex[above = 15pt of l, color1, dot, minimum size=3pt, label = {left: {\footnotesize $\textcolor{color1}{\beta^{}}$}}] (x2) {};
        \vertex[right = 10pt of l, dot, minimum size=0pt] (v12) {};
        \vertex[right = 30pt of v12, dot, minimum size=0pt] (b) {};
        \diagram*{
          (x1) --  [color1, ghost] (v12) --  [color1, ghost] (x2),
          (v12) -- [black, photon, edge label = {\scriptsize \;$\widehat{\Delta \Omega}_{i,\alpha\beta}^{(\ell+1)}$}, inner sep = 4pt] (b) 
        };
      \end{feynman}
    \end{tikzpicture}\hspace{-0.1cm} &\sim \frac{1}{n_{\ell}} &
    \begin{tikzpicture}[baseline=(b)]
      \begin{feynman}
        \vertex (l) {};
        \vertex[below = 15pt of l, dot, minimum size=3pt, label = {left: {\footnotesize $\alpha^{}$}}] (x1) {};
        \vertex[above = 15pt of l, color1, dot, minimum size=3pt, label = {left: {\footnotesize $\textcolor{color1}{\beta^{}}$}}] (x2) {};
        \vertex[right = 10pt of l, dot, minimum size=0pt] (v12) {};
        \vertex[right = 33pt of v12, dot, minimum size=0pt] (b) {};
        \diagram*{
          (x1) --  (v12) --  [color1, ghost] (x2),
          (v12) -- [color1blackghost, edge label = {\scriptsize \;$\sigma^{(\ell)}_{i,\alpha}\sigma'^{(\ell)}_{i,\textcolor{color1}{\beta}}$}, inner sep = 4pt] (b) 
        };
      \end{feynman}
    \end{tikzpicture}\hspace{-0.1cm} &\sim \frac{C^{(\ell+1)}_{W}}{n_{\ell}}\nonumber\\
    \begin{tikzpicture}[baseline=(b)]
      \begin{feynman}
        \vertex (l) {};
        \vertex[below = 15pt of l, color1, dot, minimum size=3pt, label = {left: {\footnotesize $\textcolor{color1}{\alpha^{}}$}}] (x1) {};
        \vertex[above = 15pt of l, color1, dot, minimum size=3pt, label = {left: {\footnotesize $\textcolor{color1}{\beta^{}}$}}] (x2) {};
        \vertex[right = 10pt of l, dot, minimum size=0pt] (v12) {};
        \vertex[right = 30pt of v12, dot, minimum size=0pt] (b) {};
        \diagram*{
          (x1) --  [color1, ghost] (v12) --  [color1, ghost] (x2),
          (v12) -- [color1doubghost, edge label = {\scriptsize \;$\sigma'^{(\ell)}_{i,\textcolor{color1}{\alpha}}\sigma'^{(\ell)}_{i,\textcolor{color1}{\beta}}$}, inner sep = 4pt] (b) 
        };
      \end{feynman}
    \end{tikzpicture}\hspace{-0.2cm} &\sim \frac{C^{(\ell+1)}_{W}}{n_{\ell}}  &
    \begin{tikzpicture}[baseline=(b)]
      \begin{feynman}
        \vertex (l) {};
        \vertex[below = 15pt of l, color1, dot, minimum size=3pt, label = {left: {\footnotesize $\textcolor{color1}{\alpha^{}}$}}] (x1) {};
        \vertex[above = 15pt of l, color2, dot, minimum size=3pt, label = {left: {\footnotesize $\textcolor{color2}{\beta^{}}$}}] (x2) {};
        \vertex[right = 10pt of l, dot, minimum size=0pt] (v12) {};
        \vertex[right = 30pt of v12, dot, minimum size=0pt] (b) {};
        \diagram*{
          (x1) --  [color1, ghost] (v12) --  [color2, ghost] (x2),
          (v12) -- [color2color1ghost, edge label = {\scriptsize \;$\sigma'^{(\ell)}_{i,\textcolor{color1}{\alpha}}\sigma'^{(\ell)}_{i,\textcolor{color2}{\beta}}$}, inner sep = 4pt] (b) 
        };
      \end{feynman}
    \end{tikzpicture}\hspace{-0.2cm} &\sim \frac{C^{(\ell+1)}_{W}}{n_{\ell}}\label{feynmanrulescubic}
  \end{align}
  \endgroup
  where $\widehat{\Omega}^{(\ell+1)}_{i,\alpha\beta} = \sigma^{(\ell)}_{i,\alpha}\sigma^{(\ell)}_{i,\beta} + C^{(\ell+1)}_{W}\Theta_{\alpha\beta}^{(\ell)}\sigma'^{(\ell)}_{i,\alpha}\sigma'^{(\ell)}_{i,\beta}$ and  $\widehat{\Delta\Omega}_{i,\alpha\beta}^{(\ell+1)}=\widehat{\Omega}_{i,\alpha\beta}^{(\ell+1)}-\langle  \widehat{\Omega}_{i,\alpha\beta}^{(\ell+1)}\rangle_{K^{(\ell)}}$. For the remaining external lines due to the higher-derivative tensors, we introduce the following cubic vertices
  \begingroup
  \allowdisplaybreaks
  \begin{align}
    \begin{tikzpicture}[baseline=(b)]
      \begin{feynman}
        \vertex (l) {};
        \vertex[below = 15pt of l, color3, dot, minimum size=3pt, label = {below: {\footnotesize $\textcolor{color3}{\alpha^{}}$}}] (x1) {};
        \vertex[above = 15pt of l, dot, minimum size=3pt, label = {above: {\footnotesize $\beta$}}] (x2) {};
        \vertex[right = 10pt of l, dot, minimum size=0pt] (v12) {};
        \vertex[right = 30pt of v12, dot, minimum size=0pt] (b) {};
        \diagram*{
          (x1) --  [color2color1dntknew] (v12) -- (x2),
          (v12) -- [color1color2ghost, edge label = {\scriptsize \;$\sigma''^{(\ell)}_{i,\textcolor{color3}{\alpha}}\sigma^{(\ell)}_{i,\beta}$}, inner sep = 4pt] (b) 
        };
      \end{feynman}
    \end{tikzpicture} &\sim \frac{C^{(\ell+1)}_{W}}{n_{\ell}} 
    & \begin{tikzpicture}[baseline=(b)]
      \begin{feynman}
        \vertex (l) {};
        \vertex[below = 15pt of l, color3, dot, minimum size=3pt, label = {below: {\footnotesize $\textcolor{color3}{\alpha^{}}$}}] (x1) {};
        \vertex[above = 15pt of l, dot, minimum size=3pt, label = {above: {\footnotesize $\beta$}}] (x2) {};
        \vertex[right = 10pt of l, dot, minimum size=0pt] (v12) {};
        \vertex[right = 33pt of v12, dot, minimum size=0pt] (b) {};
        \diagram*{
          (x1) --  [color2color1dntknew] (v12) -- (x2),
          (v12) -- [blackcolor1color2ghost, edge label = {\scriptsize \;$\sigma'^{(\ell)}_{i,\textcolor{color3}{\alpha}}\sigma^{(\ell)}_{i,\beta}$}, inner sep = 4pt] (b) 
        };
      \end{feynman}
    \end{tikzpicture} &\sim \frac{C^{(\ell+1)}_{W}}{n_{\ell}} 
    &  \begin{tikzpicture}[baseline=(b)]
      \begin{feynman}
        \vertex (l) {};
        \vertex[below = 15pt of l, color3, dot, minimum size=3pt, label = {below: {\footnotesize $\textcolor{color3}{\alpha}^{}$}}] (x1) {};
        \vertex[above = 15pt of l, color1, dot, minimum size=3pt, label = {above: {\footnotesize $\textcolor{color1}{\beta^{}}$}}] (x2) {};
        \vertex[right = 10pt of l, dot, minimum size=0pt] (v12) {};
        \vertex[right = 33pt of v12, dot, minimum size=0pt] (b) {};
        \diagram*{
          (x1) --  [color2color1dntknew] (v12) -- [color1, ghost] (x2),
          (v12) -- [blackcolor2ghost, edge label = {\scriptsize \;$\sigma'^{(\ell)}_{i,\textcolor{color2}{\alpha}}\sigma^{(\ell)}_{i,\beta}$}, inner sep = 4pt] (b) 
        };
      \end{feynman}
    \end{tikzpicture} &\sim \frac{1}{n_{\ell}}  
    \nonumber\\
    \begin{tikzpicture}[baseline=(b)]
      \begin{feynman}
        \vertex (l) {};
        \vertex[below = 15pt of l, color3, dot, minimum size=3pt, label = {below: {\footnotesize $\textcolor{color3}{\alpha^{}}$}}] (x1) {};
        \vertex[above = 15pt of l, color1, dot, minimum size=3pt, label = {above: {\footnotesize $\textcolor{color1}{\beta^{}}$}}] (x2) {};
        \vertex[right = 10pt of l, dot, minimum size=0pt] (v12) {};
        \vertex[right = 33pt of v12, dot, minimum size=0pt] (b) {};
        \diagram*{
          (x1) --  [color2color1dntknew] (v12) -- [color1, ghost] (x2),
          (v12) -- [blackcolor2ghost, edge label = {\scriptsize \hspace{2mm} \;$\Theta^{(\ell)}_{\textcolor{color1}{\alpha\beta}}\sigma''^{(\ell)}_{i,\textcolor{color2}{\alpha}}\sigma'^{(\ell)}_{i,\beta}$}, inner sep = 4pt] (b) 
        };
      \end{feynman}
    \end{tikzpicture} &\sim \frac{C^{(\ell+1)}_{W}}{n_{\ell}}     
    & \begin{tikzpicture}[baseline=(b)]
      \begin{feynman}
        \vertex (l) {};
        \vertex[below = 15pt of l, color3, dot, minimum size=3pt, label = {below: {\footnotesize $\textcolor{color3}{\alpha^{}}$}}] (x1) {};
        \vertex[above = 15pt of l, color1, dot, minimum size=3pt, label = {above: {\footnotesize $\textcolor{color1}{\beta^{}}$}}] (x2) {};
        \vertex[right = 10pt of l, dot, minimum size=0pt] (v12) {};
        \vertex[right = 33pt of v12, dot, minimum size=0pt] (b) {};
        \diagram*{
          (x1) --  [color2color1dntknew] (v12) -- [color1, ghost] (x2),
          (v12) -- [ntkdntkcolor1color2, edge label = {\scriptsize \;$\sigma'^{(\ell)}_{i,\textcolor{color3}{\alpha}}\sigma'^{(\ell)}_{i,\textcolor{color1}{\beta}}$}, inner sep = 4pt] (b) 
        };
      \end{feynman}
    \end{tikzpicture} &\sim \frac{C^{(\ell+1)}_{W}}{n_{\ell}} &
    \begin{tikzpicture}[baseline=(b)]
      \begin{feynman}
        \vertex (l) {};
        \vertex[below = 15pt of l, color7, dot, minimum size=3pt, label = {below: {\footnotesize $\textcolor{color7}{\alpha^{}}$}}] (x1) {};
        \vertex[above = 15pt of l, color1, dot, minimum size=3pt, label = {above: {\footnotesize $\textcolor{color1}{\beta^{}}$}}] (x2) {};
        \vertex[right = 10pt of l, dot, minimum size=0pt] (v12) {};
        \vertex[right = 33pt of v12, dot, minimum size=0pt] (b) {};
        \diagram*{
          (x1) --  [color6color2ddntknew] (v12) -- [color1, ghost] (x2),
          (v12) -- [ntkdntkblackcolor1color2color6, edge label = {\scriptsize \;$\sigma''^{(\ell)}_{i,\textcolor{color7}{\alpha}}\sigma'^{(\ell)}_{i,\textcolor{color1}{\beta}}$}, inner sep = 4pt] (b) 
        };
      \end{feynman}
    \end{tikzpicture} &\sim \frac{C^{(\ell+1)}_{W}}{n_{\ell}}    
    \nonumber\\ \begin{tikzpicture}[baseline=(b)]
      \begin{feynman}
        \vertex (l) {};
        \vertex[below = 15pt of l, color7, dot, minimum size=3pt, label = {below: {\footnotesize $\textcolor{color7}{\alpha^{}}$}}] (x1) {};
        \vertex[above = 15pt of l, color1, dot, minimum size=3pt, label = {above: {\footnotesize $\textcolor{color1}{\beta^{}}$}}] (x2) {};
        \vertex[right = 10pt of l, dot, minimum size=0pt] (v12) {};
        \vertex[right = 33pt of v12, dot, minimum size=0pt] (b) {};
        \diagram*{
          (x1) --  [color6color2ddntknew] (v12) -- [color1, ghost] (x2),
          (v12) -- [ntkdntkcolor1color2color6, edge label = {\scriptsize \;$\sigma'^{(\ell)}_{i,\textcolor{color7}{\alpha}}\sigma'^{(\ell)}_{i,\textcolor{color1}{\beta}}$}, inner sep = 4pt] (b) 
        };
      \end{feynman}
    \end{tikzpicture} &\sim \frac{C^{(\ell+1)}_{W}}{n_{\ell}}     
    & \begin{tikzpicture}[baseline=(b)]
      \begin{feynman}
        \vertex (l) {};
        \vertex[below = 15pt of l, color7, dot, minimum size=3pt, label = {below: {\footnotesize $\textcolor{color7}{\alpha^{}}$}}] (x1) {};
        \vertex[above = 15pt of l, color5, dot, minimum size=3pt, label = {above: {\footnotesize $\textcolor{color5}{\beta^{}}$}}] (x2) {};
        \vertex[right = 10pt of l, dot, minimum size=0pt] (v12) {};
        \vertex[right = 33pt of v12, dot, minimum size=0pt] (b) {};
        \diagram*{
          (x1) --  [color6color2ddntknew] (v12) -- [color6color1ddntknew] (x2),
          (v12) -- [color1color2ghost, edge label = {\scriptsize \;$\sigma'^{(\ell)}_{i,\textcolor{color2}{\alpha}}\sigma'^{(\ell)}_{i,\textcolor{color1}{\beta}}$}, inner sep = 4pt] (b) 
        };
      \end{feynman}
    \end{tikzpicture} &\sim \frac{1}{n_{\ell}} 
    & \begin{tikzpicture}[baseline=(b)]
      \begin{feynman}
        \vertex (l) {};
        \vertex[below = 15pt of l, color7, dot, minimum size=3pt, label = {below: {\footnotesize $\textcolor{color7}{\alpha^{}}$}}] (x1) {};
        \vertex[above = 15pt of l, color5, dot, minimum size=3pt, label = {above: {\footnotesize $\textcolor{color5}{\beta^{}}$}}] (x2) {};
        \vertex[right = 10pt of l, dot, minimum size=0pt] (v12) {};
        \vertex[right = 33pt of v12, dot, minimum size=0pt] (b) {};
        \diagram*{
          (x1) --  [color6color2ddntknew] (v12) -- [color6color1ddntknew] (x2),
          (v12) -- [color1color2ghost, edge label = {\scriptsize \hspace{2mm} \;$\Theta^{(\ell)}_{\textcolor{color6}{\alpha\beta}}\sigma''^{(\ell)}_{i,\textcolor{color2}{\alpha}}\sigma''^{(\ell)}_{i,\textcolor{color1}{\beta}}$}, inner sep = 4pt] (b) 
        };
      \end{feynman}
    \end{tikzpicture} &\sim \frac{C^{(\ell+1)}_{W}}{n_{\ell}} 
    \nonumber\\
    \begin{tikzpicture}[baseline=(b)]
      \begin{feynman}
        \vertex (l) {};
        \vertex[below = 15pt of l, color7, dot, minimum size=3pt, label = {below: {\footnotesize $\textcolor{color7}{\alpha^{}}$}}] (x1) {};
        \vertex[above = 15pt of l, color5, dot, minimum size=3pt, label = {above: {\footnotesize $\textcolor{color5}{\beta^{}}$}}] (x2) {};
        \vertex[right = 10pt of l, dot, minimum size=0pt] (v12) {};
        \vertex[right = 33pt of v12, dot, minimum size=0pt] (b) {};
        \diagram*{
          (x1) --  [color6color2ddntknew] (v12) -- [color6color1ddntknew] (x2),
          (v12) -- [ddntkddntkcolor1color2ghost, edge label = {\scriptsize \;$\sigma'^{(\ell)}_{i,\textcolor{color7}{\alpha}}\sigma'^{(\ell)}_{i,\textcolor{color5}{\beta}}$}, inner sep = 4pt] (b) 
        };
      \end{feynman}
    \end{tikzpicture} &\sim \frac{C^{(\ell+1)}_{W}}{n_{\ell}} & 
    \begin{tikzpicture}[baseline=(b)]
      \begin{feynman}
        \vertex (l) {};
        \vertex[below = 15pt of l, color4, dot, minimum size=3pt, label = {below: {\footnotesize $\textcolor{color4}{\alpha^{}}$}}] (x1) {};
        \vertex[above = 15pt of l, color6, dot, minimum size=3pt, label = {above: {\footnotesize $\textcolor{color6}{\beta^{}}$}}] (x2) {};
        \vertex[right = 10pt of l, dot, minimum size=0pt] (v12) {};
        \vertex[right = 33pt of v12, dot, minimum size=0pt] (b) {};
        \diagram*{
          (x1) --  [color6color1color2ddntknew] (v12) -- [color6, ghost] (x2),
          (v12) -- [color1color2ghost, edge label = {\scriptsize \;$\sigma''^{(\ell)}_{i,\textcolor{color3}{\alpha}}\sigma^{(\ell)}_{i,\beta}$}, inner sep = 4pt] (b) 
        };
      \end{feynman}
    \end{tikzpicture} &\sim \frac{1}{n_{\ell}} & 
    \begin{tikzpicture}[baseline=(b)]
      \begin{feynman}
        \vertex (l) {};
        \vertex[below = 15pt of l, color4, dot, minimum size=3pt, label = {below: {\footnotesize $\textcolor{color4}{\alpha^{}}$}}] (x1) {};
        \vertex[above = 15pt of l, color6, dot, minimum size=3pt, label = {above: {\footnotesize $\textcolor{color6}{\beta^{}}$}}] (x2) {};
        \vertex[right = 10pt of l, dot, minimum size=0pt] (v12) {};
        \vertex[right = 33pt of v12, dot, minimum size=0pt] (b) {};
        \diagram*{
          (x1) --  [color6color1color2ddntknew] (v12) -- [color6, ghost] (x2),
          (v12) -- [color1color2ghost, edge label = {\scriptsize \hspace{2mm} \;$\Theta^{(\ell)}_{\textcolor{color6}{\alpha\beta}}\sigma'''^{(\ell)}_{i,\textcolor{color3}{\alpha}}\sigma'^{(\ell)}_{i,\beta}$}, inner sep = 4pt] (b) 
        };
      \end{feynman}
    \end{tikzpicture} &\sim \frac{C^{(\ell+1)}_{W}}{n_{\ell}}    
    \nonumber\\
    \begin{tikzpicture}[baseline=(b)]
      \begin{feynman}
        \vertex (l) {};
        \vertex[below = 15pt of l, color4, dot, minimum size=3pt, label = {below: {\footnotesize $\textcolor{color4}{\alpha^{}}$}}] (x1) {};
        \vertex[above = 15pt of l, color6, dot, minimum size=3pt, label = {above: {\footnotesize $\textcolor{color6}{\beta^{}}$}}] (x2) {};
        \vertex[right = 10pt of l, dot, minimum size=0pt] (v12) {};
        \vertex[right = 33pt of v12, dot, minimum size=0pt] (b) {};
        \diagram*{
          (x1) --  [color6color1color2ddntknew] (v12) -- [color6, ghost] (x2),
          (v12) -- [blackcolor1color2ghost, edge label = {\scriptsize \;$\sigma'^{(\ell)}_{i,\textcolor{color3}{\alpha}}\sigma^{(\ell)}_{i,\beta}$}, inner sep = 4pt] (b) 
        };
      \end{feynman}
    \end{tikzpicture} &\sim \frac{1}{n_{\ell}} 
    & \begin{tikzpicture}[baseline=(b)]
      \begin{feynman}
        \vertex (l) {};
        \vertex[below = 15pt of l, color4, dot, minimum size=3pt, label = {below: {\footnotesize $\textcolor{color4}{\alpha^{}}$}}] (x1) {};
        \vertex[above = 15pt of l, color6, dot, minimum size=3pt, label = {above: {\footnotesize $\textcolor{color6}{\beta^{}}$}}] (x2) {};
        \vertex[right = 10pt of l, dot, minimum size=0pt] (v12) {};
        \vertex[right = 33pt of v12, dot, minimum size=0pt] (b) {};
        \diagram*{
          (x1) --  [color6color1color2ddntknew] (v12) -- [color6, ghost] (x2),
          (v12) -- [blackcolor1color2ghost, edge label = {\scriptsize \hspace{1mm}\;$\Theta^{(\ell)}_{\textcolor{color6}{\alpha\beta}}\sigma''^{(\ell)}_{i,\textcolor{color3}{\alpha}}\sigma'^{(\ell)}_{i,\beta}$}, inner sep = 4pt] (b) 
        };
      \end{feynman}
    \end{tikzpicture} &\sim \frac{C^{(\ell+1)}_{W}}{n_{\ell}} &
  \end{align}
  \endgroup
  
\item The internal lines from the vertices \eqref{feynmanrulescubic} are connected to a \emph{propagator} which will be represented by a white blob
  \begin{equation}
    \langle\hspace{3mm}\rangle_{K^{(\ell)}} \equiv
    \begin{tikzpicture}[baseline=-0.1cm]
      \begin{feynman}
        \tikzfeynmanset{every blob = {/tikz/fill=white!50, /tikz/minimum size=15pt}}
        \vertex (l) {};
        \vertex[above = 0pt of l, dot, minimum size=0pt] (v12) {};
        \vertex[above = 0pt of v12, blob] (b) {};
        \vertex[above = 0pt of b, dot, minimum size=0pt] (v34) {};
        \diagram*{
          (v12) -- (b) -- (v34), 
        };
      \end{feynman}
    \end{tikzpicture}   
  \end{equation}
  where $\langle \hspace{3mm}\rangle_{K^{(\ell)}}$ stands for a Gaussian expectation value with mean zero and covariance given by the Gram matrix of the NNGP, as in~\eqref{eq:F}. We take the expectation value over the decorations of the internal lines attached to the propagator, for instance
  \begin{align}
    \begin{tikzpicture}[baseline=(b)]
      \begin{feynman}
        \vertex (l) {};
        \vertex[right = 10pt of l, dot, minimum size=0pt] (v12) {};
        \vertex[right = 40pt of v12, propblob] (b) {};
        \vertex[right = 40pt of b, dot, minimum size=0pt] (v34) {};
        \vertex[right = 8pt of v34] (r) {};
        \diagram*{
          (v12) -- [black, photon, edge label = {\scriptsize $\widehat{\Delta\Omega}_{i,\alpha\beta}$}, inner sep = 4pt] (b) -- [black, photon, edge label = {\scriptsize $\widehat{\Delta\Omega}_{i,\gamma\delta}$}, inner sep = 4pt] (v34)
        };
      \end{feynman}
    \end{tikzpicture}\hspace{-0.2cm}
    \sim \langle \widehat{\Delta\Omega}_{i,\alpha\beta} \widehat{\Delta\Omega}_{i,\gamma\delta}\rangle_{K^{(\ell)}}\,.
  \end{align}
  Propagators obey the following \emph{selection rules}:
  \begin{enumerate}[label=(\alph*)]
  \item Propagators are only connected to internal lines emanating from the cubic vertices or the internal quartic vetrices introduced below. In particular, propagators cannot be directly connected to other propagators.
  \item Dotted lines attached to a propagator do not appear in the Gaussian expectation value, since they are not decorated.
  \item Each preactivation line decorated with $z_{i}$ acts as a derivative with respect to $z_{i}$ on the argument of the Gaussian expectation value.
  \item The neural indices of all internal lines connected to the propagator have to be equal.
  \item If both dotted and dashed lines of the same color are connected to the propagator, these have to appear in pairs with the same sample index. The lines in a pair will be attached to different vertices. Furthermore, if both vertices connected with one color to the propagator are drawn in the orientation in which they are presented in the Feynman rules, the ordering from top to bottom of the sample indices (and therefore colors) of the lines connected to both vertices have to be the same.
  \item Pairs of dashed lines of the same color connected to the propagator add a factor of $\Theta_{\alpha\beta}$ if they are connected to different vertices. Here, $\alpha$ and $\beta$ are the sample indices of the two lines in the pair.
  \end{enumerate}
\item The 10 rank-four tensors into which we decompose the cumulants will be represented by \emph{quartic interactions} with four external lines. For preactivations, we use the vertex $V$~\cite{banta2024}
  \begin{align}
    \begin{tikzpicture}[baseline=(b)]
      \begin{feynman}
        \vertex (l) {};
        \vertex[below = 15pt of l, dot, minimum size=3pt, label = {left: {\footnotesize $1$}}] (x1) {};
        \vertex[above = 15pt of l, dot, minimum size=3pt, label = {left: {\footnotesize $2$}}] (x2) {};
        \vertex[right = 20pt of l, quarticblob] (b) {};
        \vertex[below = 25pt of b] {\scriptsize $\frac{1}{n_{\ell}\*}V_{1234}^{(\ell+1)}$}; 
        \vertex[right = 20pt of b] (r) {};
        \vertex[above = 15pt of r, dot, minimum size=3pt, label = {right: {\footnotesize $3$}}] (x3) {};
        \vertex[below = 15pt of r, dot, minimum size=3pt, label = {right: {\footnotesize $4$}}] (x4) {};
        \diagram*{
          (x1) -- (b) -- (x2), 
          (x3) -- (b) -- (x4)
        };
      \end{feynman}
    \end{tikzpicture}
  \end{align}
  We introduce the following quartic vertices containing NTK and preactivation lines
  \begin{align}
    \begin{tikzpicture}[baseline=(b)]
      \begin{feynman}
        \vertex (l) {};
        \vertex[below = 15pt of l, dot, minimum size=3pt, label = {left: {\footnotesize $\alpha_{1}$}}] (x1) {};
        \vertex[above = 15pt of l, dot, minimum size=3pt, label = {left: {\footnotesize $\alpha_{2}$}}] (x2) {};
        \vertex[right = 20pt of l, quarticblob] (b) {};
        \vertex[below = 25pt of b] {\scriptsize $\frac{1}{n_{\ell}\*}D_{\alpha_{1}\alpha_{2}\textcolor{color1}{\alpha_{3}\alpha_{4}}}^{(\ell+1)}$}; 
        \vertex[right = 20pt of b] (r) {};
        \vertex[above = 15pt of r, dot, color1, minimum size=3pt, label = {right: {\footnotesize $\textcolor{color1}{\alpha_{3}}$}}] (x3) {};
        \vertex[below = 15pt of r, dot, color1, minimum size=3pt, label = {right: {\footnotesize $\textcolor{color1}{\alpha_{4}}$}}] (x4) {};
        \diagram*{
          (x1) -- (b) -- (x2), 
          (x3) -- [color1, ghost] (b) -- [color1, ghost] (x4)
        };
      \end{feynman}
    \end{tikzpicture}
    \hspace{0.5cm}
    \begin{tikzpicture}[baseline=(b)]
      \begin{feynman}
        \vertex (l) {};
        \vertex[below = 15pt of l, dot, minimum size=3pt, label = {left: {\footnotesize $\alpha_{1}$}}] (x1) {};
        \vertex[above = 15pt of l, dot, color1, minimum size=3pt, label = {left: {\footnotesize $\textcolor{color1}{\alpha_{3}}$}}] (x2) {};
        \vertex[right = 20pt of l, quarticblob] (b) {};
        \vertex[below = 25pt of b] {\scriptsize $\frac{1}{n_{\ell}\*}F_{\alpha_{1}\textcolor{color1}{\alpha_{3}}\alpha_{2}\textcolor{color1}{\alpha_{4}}}^{(\ell+1)}$}; 
        \vertex[right = 20pt of b] (r) {};
        \vertex[above = 15pt of r, dot, minimum size=3pt, label = {right: {\footnotesize $\alpha_{2}$}}] (x3) {};
        \vertex[below = 15pt of r, dot, color1, minimum size=3pt, label = {right: {\footnotesize $\textcolor{color1}{\alpha_{4}}$}}] (x4) {};
        \diagram*{
          (x1) -- (b) -- [color1, ghost] (x2), 
          (x3) -- (b) -- [color1, ghost] (x4)
        };
      \end{feynman}
    \end{tikzpicture}
    \hspace{0.5cm}
    \begin{tikzpicture}[baseline=(b)]
      \begin{feynman}
        \vertex (l) {};
        \vertex[below = 15pt of l, dot, color2, minimum size=3pt, label = {left: {\footnotesize $\textcolor{color2}{\alpha_{1}}$}}] (x1) {};
        \vertex[above = 15pt of l, dot, color2, minimum size=3pt, label = {left: {\footnotesize $\textcolor{color2}{\alpha_{2}}$}}] (x2) {};
        \vertex[right = 20pt of l, quarticblob] (b) {};
        \vertex[below = 25pt of b] {\scriptsize $\frac{1}{n_{\ell}\*}A_{\textcolor{color2}{\alpha_{1}}\textcolor{color2}{\alpha_{2}}\textcolor{color1}{\alpha_{3}}\textcolor{color1}{\alpha_{4}}}^{(\ell+1)}$}; 
        \vertex[right = 20pt of b] (r) {};
        \vertex[above = 15pt of r, dot, color1, minimum size=3pt, label = {right: {\footnotesize $\textcolor{color1}{\alpha_{3}}$}}] (x3) {};
        \vertex[below = 15pt of r, dot, color1, minimum size=3pt, label = {right: {\footnotesize $\textcolor{color1}{\alpha_{4}}$}}] (x4) {};
        \diagram*{
          (x1) -- [color2, ghost] (b) -- [color2, ghost] (x2), 
          (x3) -- [color1, ghost] (b) -- [color1, ghost] (x4)
        };
      \end{feynman}
    \end{tikzpicture}
    \hspace{0.5cm}
    \begin{tikzpicture}[baseline=(b)]
      \begin{feynman}
        \vertex (l) {};
        \vertex[below = 15pt of l, dot, color2, minimum size=3pt, label = {left: {\footnotesize $\textcolor{color2}{\alpha_{1}}$}}] (x1) {};
        \vertex[above = 15pt of l, dot, color1, minimum size=3pt, label = {left: {\footnotesize $\textcolor{color1}{\alpha_{3}}$}}] (x2) {};
        \vertex[right = 20pt of l, quarticblob] (b) {};
        \vertex[below = 25pt of b] {\scriptsize $\frac{1}{n_{\ell}\*}B_{\textcolor{color2}{\alpha_{1}}\textcolor{color1}{\alpha_{3}}\textcolor{color2}{\alpha_{2}}\textcolor{color1}{\alpha_{4}}}^{(\ell+1)}$}; 
        \vertex[right = 20pt of b] (r) {};
        \vertex[above = 15pt of r, dot, color2, minimum size=3pt, label = {right: {\footnotesize $\textcolor{color2}{\alpha_{2}}$}}] (x3) {};
        \vertex[below = 15pt of r, dot, color1, minimum size=3pt, label = {right: {\footnotesize $\textcolor{color1}{\alpha_{4}}$}}] (x4) {};
        \diagram*{
          (x1) -- [color2, ghost] (b) -- [color1, ghost] (x2), 
          (x3) -- [color2, ghost] (b) -- [color1, ghost] (x4)
        };
      \end{feynman}
    \end{tikzpicture}
  \end{align}
  For the higher-derivative tensors, we introduce
  \begin{align}
    \begin{tikzpicture}[baseline=(b)]
      \begin{feynman}
        \vertex (l) {};
        \vertex[below = 16pt of l,color1, dot, minimum size=3pt, label = {below: {\footnotesize $\textcolor{color1}{\alpha_1^{}}$}}] (x1) {};
        \vertex[above = 16pt of l, color2, dot, minimum size=3pt, label = {above: {\footnotesize $\textcolor{color2}{\alpha_2^{}}$}}] (x2) {};
        \vertex[right = 8pt of l, dot, minimum size=0pt] (v12) {};
        \vertex[below = 25pt of v12, dot, minimum size=0pt] (v12d) {};
        \vertex[right = 8pt of v12d, dot, minimum size=0pt] (v12dd) {};
        \vertex[right = 20pt of v12, quarticblob] (b) {};
        \vertex[right = 20pt of b, dot, minimum size=0pt] (v34) {};
        \vertex[below = 25pt of v34, dot, minimum size=0pt] (v34d) {};
        \vertex[left = 8pt of v34d, dot, minimum size=0pt] (v34dd) {};
        \vertex[right = 8pt of v34] (r) {};
        \vertex[above = 16pt of r, color3, dot, minimum size=3pt, label = {above: {\footnotesize $\textcolor{color3}{\alpha_3^{}}$}}] (x3) {};
        \vertex[below = 16pt of r, dot, minimum size=3pt, label = {below: {\footnotesize $\alpha_4^{}$}}] (x4) {};
        \diagram*{
          (x1) -- [color1, ghost] (b) -- [color2, ghost] (x2), 
          (x3) -- [color1color2dntknew, ghost] (b) -- (x4)
        };
        \draw [decoration={brace}, decorate] (v12dd) -- (v34dd)
        node [pos=0.5, below = 1pt] {\scriptsize $\frac{1}{n_{\ell}\*}P_{\textcolor{color3}{\alpha_{3}}\textcolor{color1}{\alpha_{1}}\textcolor{color2}{\alpha_{2}}\alpha_{4}}^{(\ell+1)}$};
      \end{feynman}
    \end{tikzpicture}
    & &
    \begin{tikzpicture}[baseline=(b)]
      \begin{feynman}
        \vertex (l) {};
        \vertex[below = 16pt of l, color1!40!color2, dot, minimum size=3pt, label = {below: {\footnotesize $\textcolor{color3}{\alpha_1^{}}$}}] (x1) {};
        \vertex[above = 16pt of l, color1, dot, minimum size=3pt, label = {above: {\footnotesize $\textcolor{color1}{\alpha_3^{}}$}}] (x2) {};
        \vertex[right = 8pt of l, dot, minimum size=0pt] (v12) {};
        \vertex[below = 25pt of v12, dot, minimum size=0pt] (v12d) {};
        \vertex[right = 8pt of v12d, dot, minimum size=0pt] (v12dd) {};
        \vertex[right = 20pt of v12, quarticblob] (b) {};
        \vertex[right = 20pt of b, dot, minimum size=0pt] (v34) {};
        \vertex[below = 25pt of v34, dot, minimum size=0pt] (v34d) {};
        \vertex[left = 8pt of v34d, dot, minimum size=0pt] (v34dd) {};
        \vertex[right = 8pt of v34] (r) {};
        \vertex[above = 16pt of r, color2, dot, minimum size=3pt, label = {above: {\footnotesize $\textcolor{color2}{\alpha_2^{}}$}}] (x3) {};
        \vertex[below = 16pt of r, dot, minimum size=3pt, label = {below: {\footnotesize $\alpha_4^{}$}}] (x4) {};
        \diagram*{
          (x1) -- [color2color1dntknew] (b) -- [color1, ghost] (x2), 
          (x3) -- [color2, ghost] (b) -- (x4)
        };
        \draw [decoration={brace}, decorate] (v12dd) -- (v34dd)
        node [pos=0.5, below = 1pt] {\scriptsize $\frac{1}{n_{\ell}\*}Q_{\textcolor{color3}{\alpha_{1}}\textcolor{color2}{\alpha_{2}}\textcolor{color1}{\alpha_{3}}\alpha_{4}}^{(\ell+1)}$};
      \end{feynman}
    \end{tikzpicture}
    & &
    \begin{tikzpicture}[baseline=(b)]
      \begin{feynman}
        \vertex (l) {};
        \vertex[below = 16pt of l,brown, dot, minimum size=3pt, label = {below: {\footnotesize $\textcolor{color4}{\alpha_1^{}}$}}] (x1) {};
        \vertex[above = 16pt of l, color6, dot, minimum size=3pt, label = {above: {\footnotesize $\textcolor{color6}{\alpha_2^{}}$}}] (x2) {};
        \vertex[right = 8pt of l, dot, minimum size=0pt] (v12) {};
        \vertex[below = 25pt of v12, dot, minimum size=0pt] (v12d) {};
        \vertex[right = 8pt of v12d, dot, minimum size=0pt] (v12dd) {};
        \vertex[right = 20pt of v12, quarticblob] (b) {};
        \vertex[right = 20pt of b, dot, minimum size=0pt] (v34) {};
        \vertex[below = 25pt of v34, dot, minimum size=0pt] (v34d) {};
        \vertex[left = 8pt of v34d, dot, minimum size=0pt] (v34dd) {};
        \vertex[right = 8pt of v34] (r) {};
        \vertex[above = 16pt of r, color2, dot, minimum size=3pt, label = {above: {\footnotesize $\textcolor{color2}{\alpha_3^{}}$}}] (x3) {};
        \vertex[below = 16pt of r, color1, dot, minimum size=3pt, label = {below: {\footnotesize $\textcolor{color1}{\alpha_4^{}}$}}] (x4) {};
        \diagram*{
          (x1) -- [color6color1color2ddntknew] (b) -- [color6, ghost] (x2), 
          (x3) -- [color2, ghost] (b) -- [color1, ghost](x4)
        };
        \draw [decoration={brace}, decorate] (v12dd) -- (v34dd)
        node [pos=0.5, below = 1pt] {\scriptsize $\frac{1}{n_{\ell}\*}R_{\textcolor{color4}{\alpha_{1}}\textcolor{color6}{\alpha_{2}}\textcolor{color2}{\alpha_{3}}\textcolor{color1}{\alpha_{4}}}^{(\ell+1)}$};
      \end{feynman}
    \end{tikzpicture} 
    \nonumber\\
    \begin{tikzpicture}[baseline=(b)]
      \begin{feynman}
        \vertex (l) {};
        \vertex[below = 16pt of l,color7, dot, minimum size=3pt, label = {below: {\footnotesize $\textcolor{color7}{\alpha_1^{}}$}}] (x1) {};
        \vertex[above = 16pt of l, color5, dot, minimum size=3pt, label = {above: {\footnotesize $\textcolor{color5}{\alpha_2^{}}$}}] (x2) {};
        \vertex[right = 8pt of l, dot, minimum size=0pt] (v12) {};
        \vertex[below = 25pt of v12, dot, minimum size=0pt] (v12d) {};
        \vertex[right = 8pt of v12d, dot, minimum size=0pt] (v12dd) {};
        \vertex[right = 20pt of v12, quarticblob] (b) {};
        \vertex[right = 20pt of b, dot, minimum size=0pt] (v34) {};
        \vertex[below = 25pt of v34, dot, minimum size=0pt] (v34d) {};
        \vertex[left = 8pt of v34d, dot, minimum size=0pt] (v34dd) {};
        \vertex[right = 8pt of v34] (r) {};
        \vertex[above = 16pt of r, color2, dot, minimum size=3pt, label = {above: {\footnotesize $\textcolor{color2}{\alpha_3^{}}$}}] (x3) {};
        \vertex[below = 16pt of r, color1, dot, minimum size=3pt, label = {below: {\footnotesize $\textcolor{color1}{\alpha_4^{}}$}}] (x4) {};
        \diagram*{
          (x1) -- [color6color2ddntknew] (b) -- [color6color1ddntknew] (x2), 
          (x3) -- [color2, ghost] (b) -- [color1, ghost] (x4)
        };
        \draw [decoration={brace}, decorate] (v12dd) -- (v34dd)
        node [pos=0.5, below = 1pt] {\scriptsize $\frac{1}{n_{\ell}\*}S_{\textcolor{color7}{\alpha_{1}}\textcolor{color5}{\alpha_{2}}\textcolor{color2}{\alpha_{3}}\textcolor{color1}{\alpha_{4}}}^{(\ell+1)}$};
      \end{feynman}
    \end{tikzpicture}
    & &
    \begin{tikzpicture}[baseline=(b)]
      \begin{feynman}
        \vertex (l) {};
        \vertex[below = 16pt of l, color7, dot, minimum size=3pt, label = {below: {\footnotesize $\textcolor{color7}{\alpha_1^{}}$}}] (x1) {};
        \vertex[above = 16pt of l, color2, dot, minimum size=3pt, label = {above: {\footnotesize $\textcolor{color2}{\alpha_3^{}}$}}] (x2) {};
        \vertex[right = 8pt of l, dot, minimum size=0pt] (v12) {};
        \vertex[below = 25pt of v12, dot, minimum size=0pt] (v12d) {};
        \vertex[right = 8pt of v12d, dot, minimum size=0pt] (v12dd) {};
        \vertex[right = 20pt of v12, quarticblob] (b) {};
        \vertex[right = 20pt of b, dot, minimum size=0pt] (v34) {};
        \vertex[below = 25pt of v34, dot, minimum size=0pt] (v34d) {};
        \vertex[left = 8pt of v34d, dot, minimum size=0pt] (v34dd) {};
        \vertex[right = 8pt of v34] (r) {};
        \vertex[above = 16pt of r, color5, dot, minimum size=3pt, label = {above: {\footnotesize $\textcolor{color5}{\alpha_2^{}}$}}] (x3) {};
        \vertex[below = 16pt of r, color1, dot, minimum size=3pt, label = {below: {\footnotesize $\textcolor{color1}{\alpha_4^{}}$}}] (x4) {};
        \diagram*{
          (x1) -- [color6color2ddntk2new] (b) -- [color2, ghost] (x2), 
          (x3) -- [color6color1ddntk2new, ghost] (b) -- [color1, ghost] (x4)
        };
        \draw [decoration={brace}, decorate] (v12dd) -- (v34dd)
        node [pos=0.5, below = 1pt] {\scriptsize $\frac{1}{n_{\ell}\*}T_{\textcolor{color7}{\alpha_{1}}\textcolor{color2}{\alpha_{3}}\textcolor{color1}{\alpha_{4}}\textcolor{color5}{\alpha_{2}}}^{(\ell+1)}$};
      \end{feynman}
    \end{tikzpicture} 
    & &
    \begin{tikzpicture}[baseline=(b)]
      \begin{feynman}
        \vertex (l) {};
        \vertex[below = 16pt of l, color7, dot, minimum size=3pt, label = {below: {\footnotesize $\textcolor{color7}{\alpha_1^{}}$}}] (x1) {};
        \vertex[above = 16pt of l, color1, dot, minimum size=3pt, label = {above: {\footnotesize $\textcolor{color1}{\alpha_4^{}}$}}] (x2) {};
        \vertex[right = 8pt of l, dot, minimum size=0pt] (v12) {};
        \vertex[below = 25pt of v12, dot, minimum size=0pt] (v12d) {};
        \vertex[right = 8pt of v12d, dot, minimum size=0pt] (v12dd) {};
        \vertex[right = 20pt of v12, quarticblob] (b) {};
        \vertex[right = 20pt of b, dot, minimum size=0pt] (v34) {};
        \vertex[below = 25pt of v34, dot, minimum size=0pt] (v34d) {};
        \vertex[left = 8pt of v34d, dot, minimum size=0pt] (v34dd) {};
        \vertex[right = 8pt of v34] (r) {};
        \vertex[above = 16pt of r, color5, dot, minimum size=3pt, label = {above: {\footnotesize $\textcolor{color5}{\alpha_2^{}}$}}] (x3) {};
        \vertex[below = 16pt of r, color2, dot, minimum size=3pt, label = {below: {\footnotesize $\textcolor{color2}{\alpha_3^{}}$}}] (x4) {};
        \diagram*{
          (x1) -- [color6color2ddntk2new] (b) -- [color1, ghost] (x2), 
          (x3) -- [color6color1ddntk2new, ghost] (b) -- [color2, ghost] (x4)
        };
        \draw [decoration={brace}, decorate] (v12dd) -- (v34dd)
        node [pos=0.5, below = 1pt] {\scriptsize $\frac{1}{n_{\ell}\*}U_{\textcolor{color7}{\alpha_{1}}\textcolor{color1}{\alpha_{4}}\textcolor{color5}{\alpha_{2}}\textcolor{color2}{\alpha_{3}}}^{(\ell+1)}$};
      \end{feynman}
    \end{tikzpicture} 
  \end{align}
  The quartic vertices can also appear with four internal lines instead. In this case, they correspond to tensors in layer $\ell$ instead of $\ell+1$ and are connected to propagators. Further quartic tensors for the remaining tensors involving preactivations, the dNTK and ddNTK are introduced in Appendix~\ref{app:defin-all-tens}. For the mean NNGP and NTK corrections $K^{\{1\}}$ and $\Theta^{\{1\}}$, we similarly introduce \emph{quadratic vertices} of valence two.
\item Internal lines attached to cubic or internal quartic vertices have unassigned neural indices (e.g.\ $i$ in the vertices in~\eqref{feynmanrulescubic}). Similarly, internal (solid) preactivation lines have unassigned sample indices. To assemble the term corresponding to a certain diagram, multiply the expressions corresponding to vertices and propagators and sum over all unassigned neural and sample indices. The complete expression for a cumulant is given by summing over all Feynman diagrams with the correct external lines and order in $1/n$.
\end{enumerate}


\section{Proofs of Main Theorems}\label{app:proofs}

\begingroup
\allowdisplaybreaks


This Appendix contains the rigorous proofs demonstrating how the Feynman rules of Appendix~\ref{app:feynman_rules} yield the algebraic expressions governing the statistics of the NTK, dNTK, and ddNTKs at leading order in the $\frac{1}{n}$ expansion. In addition, it presents a generalization of these results to all orders in $\frac{1}{n}$.

For later convenience, and as a warmup, we begin by reprinting the Feynman diagrams and algebraic expressions for the tensors $V_4$ and $K^{\{1\}}$, as presented in~\cite{banta2024,roberts2022}
\subsection{$V$ and $K^{\{1\}}$} \label{app:proofs_zero}

The cumulant of four preactivations given in \eqref{vfourvertex} can be computed, to first order in $\frac{1}{n}$, from the following Feynman diagrams

\begin{eqnarray}
\begin{tikzpicture}[baseline=(b)]
\begin{feynman}
\vertex (l) {};
\vertex[below = 16pt of l, dot, minimum size=3pt, label = {below: {\footnotesize $1$}}] (x1) {};
\vertex[above = 16pt of l, dot, minimum size=3pt, label = {above: {\footnotesize $2$}}] (x2) {};
\vertex[right = 8pt of l, dot, minimum size=0pt] (v12) {};
\vertex[right = 20pt of v12, quarticblob] (b) {};
\vertex[below = 25pt of b] {\scriptsize $\frac{1}{n_{\ell}}V^{(\ell +1)}_{1234}$}; 
\vertex[right = 20pt of b, dot, minimum size=0pt] (v34) {};
\vertex[right = 8pt of v34] (r) {};
\vertex[above = 16pt of r, dot, minimum size=3pt, label = {above: {\footnotesize $3$}}] (x3) {};
\vertex[below = 16pt of r, dot, minimum size=3pt, label = {below: {\footnotesize $4$}}] (x4) {};
\diagram*{
	(x1) -- (b) -- (x2), 
	(x3) -- (b) -- (x4)
};
\end{feynman}
\end{tikzpicture}
&=& 
\sum_{j}
\begin{tikzpicture}[baseline=(b)]
\begin{feynman}
\tikzfeynmanset{every blob = {/tikz/fill=white!50, /tikz/minimum size=15pt}}
\vertex (l) {};
\vertex[below = 16pt of l, dot, minimum size=3pt, label = {below: {\footnotesize $1$}}] (x1) {};
\vertex[above = 16pt of l, dot, minimum size=3pt, label = {above: {\footnotesize $2$}}] (x2) {};
\vertex[right = 8pt of l, dot, minimum size=0pt] (v12) {};
\vertex[right = 30pt of v12, blob] (b) {};
\vertex[right = 30pt of b, dot, minimum size=0pt] (v34) {};
\vertex[right = 8pt of v34] (r) {};
\vertex[above = 16pt of r, dot, minimum size=3pt, label = {above: {\footnotesize $3$}}] (x3) {};
\vertex[below = 16pt of r, dot, minimum size=3pt, label = {below: {\footnotesize $4$}}] (x4) {};
\diagram*{
	(x1) -- (v12) -- (x2),
	(v12) -- [photon, edge label = {\scriptsize \;$\widehat{\Delta G}_{j}$}, inner sep = 4pt] (b) -- [black, photon, edge label = {\scriptsize \,$\widehat{\Delta G}_{j}$}, inner sep = 4pt] (v34), 
	 (x3) -- (v34) -- (x4)
};
\end{feynman}
\end{tikzpicture}
+ \sum_{j_1, j_2}
\begin{tikzpicture}[baseline=(b)]
\tikzfeynmanset{every blob = {/tikz/fill=white!50, /tikz/minimum size=15pt}}
\begin{feynman}
\vertex (l) {};
\vertex[below = 16pt of l, dot, minimum size=3pt, label = {below: {\footnotesize $1$}}] (x1) {};
\vertex[above = 16pt of l, dot, minimum size=3pt, label = {above: {\footnotesize $2$}}] (x2) {};
\vertex[right = 8pt of l, dot, minimum size=0pt] (v12) {};
\vertex[right = 30pt of v12, blob] (b12) {};
\vertex[right = 16pt of b12, dot, minimum size=0pt] (w12) {};
\vertex[above = 1pt of w12, label = {above: {\scriptsize \hspace{-10pt} $z_{j_1}$}}] (w12u) {};
\vertex[below = 12pt of w12] (w12d) {};
\vertex[left = 10pt of w12d] (w12dl) {};
\tikzfeynmanset{every blob = {/tikz/fill=gray!50, /tikz/minimum size=15pt}}
\vertex[right = 3pt of w12, blob , minimum size = 6pt] (b) {};
\vertex[right = 3pt of b, dot, minimum size=0pt] (w34) {};
\tikzfeynmanset{every blob = {/tikz/fill=white!50, /tikz/minimum size=15pt}}
\vertex[right = 16pt of w34, blob] (b34) {};
\vertex[above = 1pt of w34, label = {above: {\scriptsize \hspace{10pt} $z_{j_2}$}}] (w34u) {};
\vertex[below = 12pt of w34] (w34d) {};
\vertex[right = 10pt of w34d] (w34dr) {};
\vertex[right = 30pt of b34, dot, minimum size=0pt] (v34) {};
\vertex[right = 8pt of v34] (r) {};
\vertex[above = 16pt of r, dot, minimum size=3pt, label = {above: {\footnotesize $3$}}] (x3) {};
\vertex[below = 16pt of r, dot, minimum size=3pt, label = {below: {\footnotesize $4$}}] (x4) {};
\diagram*{
	(x1) -- (v12) -- (x2),
	(v12) -- [photon, edge label = {\scriptsize \;$\widehat{\Delta G}_{j_1}$}, inner sep = 4pt] (b12) -- [quarter left] (w12) -- [quarter left] (b12),
	(b34) -- [quarter left] (w34) -- [quarter left] (b34) -- [black, photon, edge label = {\scriptsize \,$\widehat{\Delta G}_{j_2}$}, inner sep = 4pt] (v34),
	(x3) -- (v34) -- (x4)
};
\draw [decoration={brace}, decorate] (w34dr) -- (w12dl)
node [pos=0.5, below = 1pt] {\scriptsize $\frac{1}{n_{\ell-1}\*}V_4^{(\ell)}$};
\end{feynman}
\end{tikzpicture}
\label{vtensorfeynman}
\end{eqnarray}
with the corresponding algebraic expression given by
\begin{align}
    \frac{1}{n_{\ell}}V^{(\ell+1)}_{1234}
= & \frac{1}{n_{\ell}}\left(C_{W}^{(\ell+1)}\right)^{2} \left[ \langle \sigma^{(\ell)}_1\sigma^{(\ell)}_2\sigma^{(\ell)}_3\sigma^{(\ell)}_4 \rangle_{K^{(\ell)}} - \langle \sigma^{(\ell)}_{1}\sigma^{(\ell)}_{2} \rangle_{K^{(\ell)}} \langle \sigma^{(\ell)}_{3}\sigma^{(\ell)}_{4} \rangle_{K^{(\ell)}} \right] \nonumber \\
  & + \frac{1}{n_{\ell-1}} \frac{\left(C_{W}^{(\ell+1)}\right)^{2}}{4} \sum_{\beta_{1},\ldots,\beta_{4} \in \{1,2,3,4\}} V^{(\ell)}_{\beta_{1}\beta_{2}\beta_{3}\beta_{4}} \langle \frac{d^{2}(\widehat{\Delta G}^{(\ell)}_{12})}{dz^{(\ell)}_{\beta_{1}}dz^{(\ell)}_{\beta_{2}}}\rangle_{K^{(\ell)}} \langle \frac{d^{2}(\widehat{\Delta G}^{(\ell)}_{34})}{dz^{(\ell)}_{\beta_{3}}dz^{(\ell)}_{\beta_{4}}} \rangle_{K^{(\ell)}},
  \label{eq:v4_recursion}
\end{align}

Likewise, the recursion relation obeyed by the first-order correction to the NNGP $K^{\{1\}}$, given in \eqref{k1correction}, follows from the diagrammatic structure
\begin{align}
	 \begin{tikzpicture}[baseline=(b)]
      \begin{feynman}
        \vertex (l) {};
        \vertex[right = 0pt of l, dot, label = {above: {$1$}}] (v12) {};
        \vertex[right = 22pt of v12, quarticblob] (b) {};
        \vertex[right = 22pt of b, dot, label = {above: {$2$}}] (v34) {};
        \vertex[right = 8pt of v34] (r) {};
        \diagram*{
          (v12) -- (b) --  (v34)
        };
        \vertex[below = 15pt of b] {{\scriptsize $\frac{1}{n_{\ell}}K^{\{1\}(\ell+1)}_{12}$}};
      \end{feynman}
    \end{tikzpicture} &=
\begin{tikzpicture}[baseline=(b)]
    \begin{feynman}
      \vertex[dot, minimum size=3pt, label = {below: {\footnotesize $1$}}] (x1) {};
      \vertex[right = 20pt of x1] (v);
      \vertex[right = 20pt of v, dot, minimum size = 3pt, label = {below: {\footnotesize $2$}}] (x2) {};
      \vertex[above = 9pt of v] (b) {};
      \vertex[above = 4pt of v, label = {right: \!\*{\scriptsize $\widehat{\Delta G}_{j,12}$}}] (j){};
      \vertex[above = 16pt of v, propblob] (G) {};
      \vertex[above = 22pt of v, dot, minimum size = 0pt] (g) {};
      \vertex[above = 20pt of g, dot, minimum size = 0pt] (w) {};
      \diagram*{
        (x1) -- (v) -- (x2),
        (v) -- [photon] (G) -- (g) -- [half left] (w) -- [half left] (g)
      };
      \vertex[above = 0pt of G, propblob] (K0){};
      \vertex[above = 20pt of g, smallblob] (K1) {};
      \vertex[above = 10pt of w] (K1) {{\scriptsize $\frac{1}{n_{\ell-1}} K^{\{1\}(\ell)}$}};
      \vertex[right = 15pt of G, label = {\scriptsize $z_j$}] (Gr) {};
    \end{feynman}
  \end{tikzpicture}
  + \begin{tikzpicture}[baseline=(b)]
    \begin{feynman}
	\vertex[dot, minimum size = 3pt, label = {below: {\footnotesize $1$}}] (x1) {};
      \vertex[right = 20pt of x1] (v);
      \vertex[right = 20pt of v, dot, minimum size = 3pt, label = {below: {\footnotesize $2$}}] (x2) {};
      \vertex[above = 9pt of v] (b) {};
      \vertex[above = 4pt of v, label = {right: \!{\scriptsize $\widehat{\Delta G}_{j,12}$}}] (j){};
      \tikzfeynmanset{every blob = {/tikz/fill=black!50, /tikz/minimum size=15pt}}
      \vertex[above = 16pt of v, blob] (G) {};
      \vertex[left = 2pt of G, dot, minimum size = 0pt] (Gl) {};
      \vertex[right = 2pt of G, dot, minimum size = 0pt] (Gr) {};
      \vertex[above = 4pt of G, dot, minimum size = 0pt] (Gu) {};
      \vertex[left = 2pt of Gu, dot, minimum size = 0pt] (Gul) {};
      \vertex[right = 2pt of Gu, dot, minimum size = 0pt] (Gur) {};
      \vertex[above = 20pt of G, smallblob] (V4) {};
      \tikzfeynmanset{every blob = {/tikz/fill=white!50, /tikz/minimum size=15pt}}
      \vertex[left = 3pt of V4, dot, minimum size = 0pt] (V4l) {};
      \vertex[right = 3pt of V4, dot, minimum size = 0pt] (V4r) {};
      \diagram*{
        (x1) -- (v) -- (x2),
        (v) -- [photon] (G) -- (Gl) -- [half left] (V4l) -- [quarter right] (Gul) -- (Gur) -- [quarter right] (V4r) -- [half left] (Gr),
      };
      \vertex[above = 0pt of G, blob] (K0) {};
      \vertex[above = 10pt of V4] (VV4) {{\scriptsize $\frac{1}{n_{\ell-1}} V_4^{(\ell)}$}};
      \vertex[right = 18pt of G, label = {above: {\scriptsize $z_j$}}] (Gr) {};
    \end{feynman}
  \end{tikzpicture}
\end{align}
which algebraically takes the form
\begin{eqnarray}
    \frac{1}{n_{\ell}}K^{\{1\}(\ell+1)}_{12}  &=&  \frac{C_{W}^{(\ell+1)}}{2 n_{\ell-1}}\sum_{\beta_{1},\beta_{2} \in \{1,2\}} K^{\{1\}(\ell)}_{\beta_{1}\beta_{2}}\langle \frac{d^{2} (\widehat{\Delta G}^{(\ell)}_{12})}{d z^{(\ell)}_{\beta_{1}} d z^{(\ell)}_{\beta_{2}}}\rangle_{K^{(\ell)}} \nonumber \\
    && + \frac{C_{W}^{(\ell+1)}}{8n_{\ell-1}}\sum_{\beta_{1},\beta_{2},\beta_{3},\beta_{4}\in \{1,2\}}V^{(\ell)}_{(\beta_{1}\beta_{2})(\beta_{3}\beta_{4})}\langle \frac{d^{4}(\widehat{\Delta G}^{(\ell)}_{12})}{d z^{(\ell)}_{\beta_{1}}d z^{(\ell)}_{\beta_{2}}d z^{(\ell)}_{\beta_{3}}d z^{(\ell)}_{\beta_{4}}}\rangle_{K^{(\ell)}}\,.
    \label{eq:k1_recursion}
\end{eqnarray}

\subsection{Proof of Theorem \ref{theoremone}}\label{app:proofs_one}

\firsttheorem*

The proof will proceed case-by-case for the joint NTK-preactivation cumulants and the NTK variance as follows.

\subsubsection{Joint NTK-Preactivation Cumulants}

\begin{proof}

As discussed in Section~\ref{sec-finitewidth-corrections-NTK}, the first non-trivial cumulant involving preactivations and the NTK can be written as
\begin{align}
  &\EE^{c}_{\theta}[z^{(\ell+1)}_{i_{1}}(x_{1}),z^{(\ell+1)}_{i_{2}}(x_{2}),\widehat{\Delta\Theta}^{(\ell+1)}_{i_{3}i_{4}}(\textcolor{color1}{x_{3}},\textcolor{color1}{x_{4}})]\nonumber\\
  &\qquad=\frac{1}{n_{\ell}}\!\!\left(\! D^{(\ell+1)}_{12\textcolor{color1}{34}}\delta_{i_{1}i_{2}}\delta_{i_{3}i_{4}}{+}F^{(\ell+1)}_{1\textcolor{color1}{3}2\textcolor{color1}{4}}\delta_{i_{1}i_{3}}\delta_{i_{2}i_{4}}{+}F^{(\ell+1)}_{1\textcolor{color1}{4}2\textcolor{color1}{3}}\delta_{i_{1}i_{4}}\delta_{i_{2}i_{3}} \!\right)\,.
\end{align}
The Feynman rules we studied in the main text then allow us to write the following diagrams for the tensor $D^{(\ell+1)}_{12\textcolor{color1}{34}}$:
\begin{eqnarray}
\begin{tikzpicture}[baseline=(b)]
\begin{feynman}
\vertex (l) {};
\vertex[below = 16pt of l, dot, minimum size=3pt, label = {below: {\footnotesize $1$}}] (x1) {};
\vertex[above = 16pt of l, dot, minimum size=3pt, label = {above: {\footnotesize $2$}}] (x2) {};
\vertex[right = 8pt of l, dot, minimum size=0pt] (v12) {};
\vertex[right = 20pt of v12, quarticblob] (b) {};
\vertex[below = 25pt of b] {\scriptsize $\frac{1}{n_{\ell}}D^{(\ell +1)}_{12\textcolor{color1}{3}\textcolor{color1}{4}}$}; 
\vertex[right = 20pt of b, dot, minimum size=0pt] (v34) {};
\vertex[right = 8pt of v34] (r) {};
\vertex[above = 16pt of r, color1, dot, minimum size=3pt, label = {above: {\footnotesize $\textcolor{color1}{3}$}}] (x3) {};
\vertex[below = 16pt of r, color1, dot, minimum size=3pt, label = {below: {\footnotesize $\textcolor{color1}{4}$}}] (x4) {};
\diagram*{
	(x1) -- (b) -- (x2), 
	 (x3) -- [color1, ghost] (b) -- [color1, ghost] (x4)
};
\end{feynman}
\end{tikzpicture}
&=& 
\sum_{j}
\begin{tikzpicture}[baseline=(b)]
\begin{feynman}
\tikzfeynmanset{every blob = {/tikz/fill=white!50, /tikz/minimum size=15pt}}
\vertex (l) {};
\vertex[below = 16pt of l, dot, minimum size=3pt, label = {below: {\footnotesize $1$}}] (x1) {};
\vertex[above = 16pt of l, dot, minimum size=3pt, label = {above: {\footnotesize $2$}}] (x2) {};
\vertex[right = 8pt of l, dot, minimum size=0pt] (v12) {};
\vertex[right = 30pt of v12, blob] (b) {};
\vertex[right = 30pt of b, dot, minimum size=0pt] (v34) {};
\vertex[right = 8pt of v34] (r) {};
\vertex[above = 16pt of r, color1, dot, minimum size=3pt, label = {above: {\footnotesize $\textcolor{color1}{3}$}}] (x3) {};
\vertex[below = 16pt of r, color1, dot, minimum size=3pt, label = {below: {\footnotesize $\textcolor{color1}{4}$}}] (x4) {};
\diagram*{
	(x1) -- (v12) -- (x2),
	(v12) -- [photon, edge label = {\scriptsize \;$\widehat{\Delta G}_{j}$}, inner sep = 4pt] (b) -- [black, photon, edge label = {\scriptsize \,$\widehat{\Delta \Omega}_{j}$}, inner sep = 4pt] (v34), 
	 (x3) -- [color1, ghost] (v34) -- [color1, ghost] (x4)
};
\end{feynman}
\end{tikzpicture}
+ \sum_{j_1, j_2}
\begin{tikzpicture}[baseline=(b)]
\tikzfeynmanset{every blob = {/tikz/fill=white!50, /tikz/minimum size=15pt}}
\begin{feynman}
\vertex (l) {};
\vertex[below = 16pt of l, dot, minimum size=3pt, label = {below: {\footnotesize $1$}}] (x1) {};
\vertex[above = 16pt of l, dot, minimum size=3pt, label = {above: {\footnotesize $2$}}] (x2) {};
\vertex[right = 8pt of l, dot, minimum size=0pt] (v12) {};
\vertex[right = 30pt of v12, blob] (b12) {};
\vertex[right = 16pt of b12, dot, minimum size=0pt] (w12) {};
\vertex[above = 1pt of w12, label = {above: {\scriptsize \hspace{-10pt} $z_{j_1}$}}] (w12u) {};
\vertex[below = 12pt of w12] (w12d) {};
\vertex[left = 10pt of w12d] (w12dl) {};
\tikzfeynmanset{every blob = {/tikz/fill=gray!50, /tikz/minimum size=15pt}}
\vertex[right = 3pt of w12, blob , minimum size = 6pt] (b) {};
\vertex[right = 3pt of b, dot, minimum size=0pt] (w34) {};
\tikzfeynmanset{every blob = {/tikz/fill=white!50, /tikz/minimum size=15pt}}
\vertex[right = 16pt of w34, blob] (b34) {};
\vertex[above = 1pt of w34, label = {above: {\scriptsize \hspace{10pt} $z_{j_2}$}}] (w34u) {};
\vertex[below = 12pt of w34] (w34d) {};
\vertex[right = 10pt of w34d] (w34dr) {};
\vertex[right = 30pt of b34, dot, minimum size=0pt] (v34) {};
\vertex[right = 8pt of v34] (r) {};
\vertex[above = 16pt of r, color1, dot, minimum size=3pt, label = {above: {\footnotesize $\textcolor{color1}{3}$}}] (x3) {};
\vertex[below = 16pt of r, color1, dot, minimum size=3pt, label = {below: {\footnotesize $\textcolor{color1}{4}$}}] (x4) {};
\diagram*{
	(x1) -- (v12) -- (x2),
	(v12) -- [photon, edge label = {\scriptsize \;$\widehat{\Delta G}_{j_1}$}, inner sep = 4pt] (b12) -- [quarter left] (w12) -- [quarter left] (b12),
	(b34) -- [quarter left] (w34) -- [quarter left] (b34) -- [black, photon, edge label = {\scriptsize \,$\widehat{\Delta \Omega}_{j_2}$}, inner sep = 4pt] (v34),
	(x3) -- [color1, ghost] (v34) -- [color1, ghost] (x4)
};
\draw [decoration={brace}, decorate] (w34dr) -- (w12dl)
node [pos=0.5, below = 1pt] {\scriptsize $\frac{1}{n_{\ell-1}\*}V_4^{(\ell)}$};
\end{feynman}
\end{tikzpicture}\nonumber\\
&& 
+ \sum_{j_1, j_2}
\begin{tikzpicture}[baseline=(b)]
\tikzfeynmanset{every blob = {/tikz/fill=white!50, /tikz/minimum size=15pt}}
\begin{feynman}
\vertex (l) {};
\vertex[below = 16pt of l, dot, minimum size=3pt, label = {below: {\footnotesize $1$}}] (x1) {};
\vertex[above = 16pt of l, dot, minimum size=3pt, label = {above: {\footnotesize $2$}}] (x2) {};
\vertex[right = 8pt of l, dot, minimum size=0pt] (v12) {};
\vertex[right = 30pt of v12, blob] (b12) {};
\vertex[right = 16pt of b12, dot, minimum size=0pt] (w12) {};
\vertex[above = 1pt of w12, label = {above: {\scriptsize \hspace{-10pt} $z_{j_1}$}}] (w12u) {};
\vertex[below = 12pt of w12] (w12d) {};
\vertex[left = 10pt of w12d] (w12dl) {};
\tikzfeynmanset{every blob = {/tikz/fill=gray!50, /tikz/minimum size=15pt}}
\vertex[right = 3pt of w12, blob , minimum size = 6pt] (b) {};
\vertex[right = 3pt of b, dot, minimum size=0pt] (w34) {};
\tikzfeynmanset{every blob = {/tikz/fill=white!50, /tikz/minimum size=15pt}}
\vertex[right = 35pt of w34, blob] (b34) {};
\vertex[above = 5pt of w34, label = {above: {\scriptsize \hspace{30pt} $\textcolor{color1}{3}$}}] (w34u) {};
\vertex[below = 12pt of w34, label = {above: {\scriptsize \hspace{30pt} $\textcolor{color1}{4}$}}] (w34d) {};
\vertex[right = 10pt of w34d] (w34dr) {};
\vertex[right = 30pt of b34, dot, minimum size=0pt] (v34) {};
\vertex[right = 30pt of b34, dot, minimum size=0pt] (vv34) {};
\vertex[right = 8pt of v34] (r) {};
\vertex[above = 16pt of r, color1, dot, minimum size=3pt, label = {above: {\footnotesize $\textcolor{color1}{3}$}}] (x3) {};
\vertex[below = 16pt of r, color1, dot, minimum size=3pt, label = {below: {\footnotesize $\textcolor{color1}{4}$}}] (x4) {};
\diagram*{
	(x1) -- (v12) -- (x2),
	(v12) -- [photon, edge label = {\scriptsize \;$\widehat{\Delta G}_{j_1}$}, inner sep = 4pt] (b12) -- [quarter left] (w12) -- [quarter left] (b12),
	(b34) -- [color1, ghost, quarter left] (w34) -- [color1, ghost, quarter left] (b34) -- [color1doubghost, edge label = {\scriptsize \,$\sigma'_{j_{2}}\sigma'_{j_{2}}$}, inner sep = 4pt] (v34), 
	(x3) -- [color1, ghost] (v34) -- [color1, ghost] (x4)
};
\draw [decoration={brace}, decorate] (w34dr) -- (w12dl)
node [pos=0.5, below = 1pt] {\scriptsize $\frac{1}{n_{\ell-1}\*}D_4^{(\ell)}$};
\end{feynman}
\end{tikzpicture}
\label{dtensorfeynman}
\end{eqnarray}
whose algebraic representation reads 
\begin{eqnarray}
  \frac{1}{n_{\ell}}D^{(\ell +1)}_{12\textcolor{color1}{3}\textcolor{color1}{4}} &=& \frac{C_{W}^{(\ell +1)}}{n_{\ell}}\langle \widehat{\Delta G}^{(\ell)}_{12}\widehat{\Delta \Omega}_{34}^{(\ell +1)}\rangle_{K^{(\ell)}} \nonumber\\*
  && + \frac{C_{W}^{(\ell +1)}}{4n_{\ell-1}}\sum_{\beta_{1},\beta_{2},\beta_{3},\beta_{4} \in \{1,2,3,4\}}V^{(\ell)}_{(\beta_{1}\beta_{2})(\beta_{3}\beta_{4})}\langle \frac{d^{2}(\widehat{\Delta G}^{(\ell)}_{12})}{d z^{(\ell)}_{\beta_{1}}d z^{(\ell)}_{\beta_{2}}}\rangle_{K^{(\ell)}}\langle \frac{d^{2}(\widehat{\Delta \Omega}^{(\ell +1)}_{34})}{d z^{(\ell)}_{\beta_{1}}d z^{(\ell)}_{\beta_{2}}}\rangle_{K^{(\ell)}}\nonumber\\*
  && + \frac{(C_{W}^{(\ell +1)})^{2}}{2n_{\ell-1}}\sum_{\beta_{1},\beta_{2} \in \{1,2,3,4\}}\langle \frac{d^{2}(\widehat{\Delta G}^{(\ell)}_{12})}{d z^{(\ell)}_{\beta_{1}}d z^{(\ell)}_{\beta_{2}}}\rangle_{K^{(\ell)}} \langle \sigma'^{(\ell)}_{\textcolor{color1}{3}}\sigma'^{(\ell)}_{\textcolor{color1}{4}}\rangle_{K^{(\ell)}}D^{(\ell)}_{\beta_{1}\beta_{2}\textcolor{color1}{34}}\,.
\end{eqnarray}
This expression agrees with the expression found in~\cite{roberts2022} using laborious algebraic manipulations.

The tensor $F^{(\ell+1)}$ was studied in detail in the main text, thus we omit its discussion.

\end{proof}

\subsubsection{NTK Variance}

\begin{proof}
As shown in \cite{roberts2022}, the NTK variance can be written in the compact form
\begin{align}
  &\EE^{c}_{\theta}[\widehat{\Delta\Theta}^{(\ell+1)}_{i_{1}i_{2}}(\textcolor{color1}{x_{1}},\textcolor{color1}{x_{2}}),\widehat{\Delta\Theta}^{(\ell+1)}_{i_{3}i_{4}}(\textcolor{color2}{x_{3}},\textcolor{color2}{x_{4}})]\nonumber\\
  &\qquad=\frac{1}{n_{\ell}}\!\!\left(\! A^{(\ell+1)}_{\textcolor{color1}{12}\textcolor{color2}{34}}\delta_{i_{1}i_{2}}\delta_{i_{3}i_{4}}{+}B^{(\ell+1)}_{\textcolor{color1}{1}\textcolor{color2}{3}\textcolor{color1}{2}\textcolor{color2}{4}}\delta_{i_{1}i_{3}}\delta_{i_{2}i_{4}}{+}B^{(\ell+1)}_{\textcolor{color1}{1}\textcolor{color2}{4}\textcolor{color1}{2}\textcolor{color2}{3}}\delta_{i_{1}i_{4}}\delta_{i_{2}i_{3}} \!\right)\,.
\end{align}
The use of the NTK Feynman rules of Section~\ref{feynman-diagrams-NTKs}, then simply dictate the recursion relation of the tensor $A^{(\ell+1)}_{\textcolor{color1}{12}\textcolor{color2}{34}}$: 
\begin{eqnarray}
\begin{tikzpicture}[baseline=(b)]
\begin{feynman}
\vertex (l) {};
\vertex[below = 16pt of l, color1, dot, minimum size=3pt, label = {below: {\footnotesize $\textcolor{color1}{1}$}}] (x1) {};
\vertex[above = 16pt of l, color1, dot, minimum size=3pt, label = {above: {\footnotesize $\textcolor{color1}{2}$}}] (x2) {};
\vertex[right = 8pt of l, dot, minimum size=0pt] (v12) {};
\vertex[right = 20pt of v12, quarticblob] (b) {};
\vertex[below = 25pt of b] {\scriptsize $\frac{1}{n_{\ell}}A^{(\ell +1)}_{\textcolor{color1}{12}\textcolor{color2}{34}}$}; 
\vertex[right = 20pt of b, dot, minimum size=0pt] (v34) {};
\vertex[right = 8pt of v34] (r) {};
\vertex[above = 16pt of r, color2, dot, minimum size=3pt, label = {above: {\footnotesize $\textcolor{color2}{3}$}}] (x3) {};
\vertex[below = 16pt of r, color2, dot, minimum size=3pt, label = {below: {\footnotesize $\textcolor{color2}{4}$}}] (x4) {};
\diagram*{
	(x1) -- [color1, ghost] (b) -- [color1, ghost](x2),
    (x3) -- [color2, ghost] (b) -- [color2, ghost] (x4)
};
\end{feynman}
\end{tikzpicture}
&=& 
\sum_{j}
\begin{tikzpicture}[baseline=(b)]
\begin{feynman}
\tikzfeynmanset{every blob = {/tikz/fill=white!50, /tikz/minimum size=15pt}}
\vertex (l) {};
\vertex[below = 16pt of l, color1, dot, minimum size=3pt, label = {below: {\footnotesize $\textcolor{color1}{1}$}}] (x1) {};
\vertex[above = 16pt of l, color1, dot, minimum size=3pt, label = {above: {\footnotesize $\textcolor{color1}{2}$}}] (x2) {};
\vertex[right = 8pt of l, dot, minimum size=0pt] (v12) {};
\vertex[right = 30pt of v12, blob] (b) {};
\vertex[right = 30pt of b, dot, minimum size=0pt] (v34) {};
\vertex[right = 8pt of v34] (r) {};
\vertex[above = 16pt of r, color2, dot, minimum size=3pt, label = {above: {\footnotesize $\textcolor{color2}{3}$}}] (x3) {};
\vertex[below = 16pt of r, color2, dot, minimum size=3pt, label = {below: {\footnotesize $\textcolor{color2}{4}$}}] (x4) {};
\diagram*{
	(x1) -- [color1, ghost] (v12) -- [color1, ghost] (x2),
	(v12) -- [black, photon, edge label = {\scriptsize \;$\widehat{\Delta \Omega}_{j}$}, inner sep = 4pt] (b) -- [black, photon, edge label = {\scriptsize \,$\widehat{\Delta \Omega}_{j}$}, inner sep = 4pt] (v34), 
	 (x3) -- [color2, ghost] (v34) -- [color2, ghost] (x4)
};
\end{feynman}
\end{tikzpicture}
 + \sum_{j_1, j_2}
\begin{tikzpicture}[baseline=(b)]
\tikzfeynmanset{every blob = {/tikz/fill=white!50, /tikz/minimum size=15pt}}
\begin{feynman}
\vertex (l) {};
\vertex[below = 16pt of l, color1, dot, minimum size=3pt, label = {below: {\footnotesize $\textcolor{color1}{1}$}}] (x1) {};
\vertex[above = 16pt of l, color1, dot, minimum size=3pt, label = {above: {\footnotesize $\textcolor{color1}{2}$}}] (x2) {};
\vertex[right = 8pt of l, dot, minimum size=0pt] (v12) {};
\vertex[right = 35pt of v12, blob] (b12) {};
\vertex[right = 16pt of b12, dot, minimum size=0pt] (w12) {};
\vertex[above = 1pt of w12, label = {above: {\scriptsize \hspace{-10pt} $z_{j_1}$}}] (w12u) {};
\vertex[below = 12pt of w12] (w12d) {};
\vertex[left = 10pt of w12d] (w12dl) {};
\tikzfeynmanset{every blob = {/tikz/fill=gray!50, /tikz/minimum size=15pt}}
\vertex[right = 3pt of w12, blob , minimum size = 6pt] (b) {};
\vertex[right = 3pt of b, dot, minimum size=0pt] (w34) {};
\tikzfeynmanset{every blob = {/tikz/fill=white!50, /tikz/minimum size=15pt}}
\vertex[right = 16pt of w34, blob] (b34) {};
\vertex[above = 1pt of w34, label = {above: {\scriptsize \hspace{10pt} $z_{j_2}$}}] (w34u) {};
\vertex[below = 12pt of w34] (w34d) {};
\vertex[right = 10pt of w34d] (w34dr) {};
\vertex[right = 35pt of b34, dot, minimum size=0pt] (v34) {};
\vertex[right = 8pt of v34] (r) {};
\vertex[above = 16pt of r, color2, dot, minimum size=3pt, label = {above: {\footnotesize $\textcolor{color2}{3}$}}] (x3) {};
\vertex[below = 16pt of r, color2, dot, minimum size=3pt, label = {below: {\footnotesize $\textcolor{color2}{4}$}}] (x4) {};
\diagram*{
	(x1) -- [color1, ghost] (v12) -- [color1, ghost](x2),
	(v12) -- [black, photon, edge label = {\scriptsize \;$\widehat{\Delta \Omega}_{j_1}$}, inner sep = 4pt] (b12) -- [quarter left] (w12) -- [quarter left] (b12),
	(b34) -- [quarter left] (w34) -- [quarter left] (b34) -- [black, photon, edge label = {\scriptsize \,$\widehat{\Delta \Omega}_{j_2}$}, inner sep = 4pt] (v34),
	(x3) -- [color2, ghost] (v34) -- [color2, ghost] (x4)
};
\draw [decoration={brace}, decorate] (w34dr) -- (w12dl)
node [pos=0.5, below = 1pt] {\scriptsize $\frac{1}{n_{\ell-1}\*}V_4^{(\ell)}$};
\end{feynman}
\end{tikzpicture}\nonumber\\
&& + 
\sum_{j_1,j_2}
\begin{tikzpicture}[baseline=(b)]
\tikzfeynmanset{every blob = {/tikz/fill=white!50, /tikz/minimum size=15pt}}
\begin{feynman}
\vertex (l) {};
\vertex[below = 16pt of l, color1, dot, minimum size=3pt, label = {below: {\footnotesize $\textcolor{color1}{1}$}}] (x1) {};
\vertex[above = 16pt of l, color1, dot, minimum size=3pt, label = {above: {\footnotesize $\textcolor{color1}{2}$}}] (x2) {};
\vertex[right = 8pt of l, dot, minimum size=0pt] (v12) {};
\vertex[right = 35pt of v12, blob] (b12) {};
\vertex[right = 16pt of b12, dot, minimum size=0pt] (w12) {};
\vertex[above = 1pt of w12, label = {above: {\scriptsize \hspace{-10pt} $z_{j_1}$}}] (w12u) {};
\vertex[below = 12pt of w12] (w12d) {};
\vertex[left = 10pt of w12d] (w12dl) {};
\tikzfeynmanset{every blob = {/tikz/fill=gray!50, /tikz/minimum size=15pt}}
\vertex[right = 3pt of w12, blob , minimum size = 6pt] (b) {};
\vertex[right = 3pt of b, dot, minimum size=0pt] (w34) {};
\tikzfeynmanset{every blob = {/tikz/fill=white!50, /tikz/minimum size=15pt}}
\vertex[right = 35pt of w34, blob] (b34) {};
\vertex[above = 5pt of w34, label = {above: {\scriptsize \hspace{30pt} $\textcolor{color2}{3}$}}] (w34u) {};
\vertex[below = 12pt of w34, label = {above: {\scriptsize \hspace{30pt} $\textcolor{color2}{4}$}}] (w34d) {};
\vertex[right = 10pt of w34d] (w34dr) {};
\vertex[right = 35pt of b34, dot, minimum size=0pt] (v34) {};
\vertex[right = 8pt of v34] (r) {};
\vertex[above = 16pt of r, color2, dot, minimum size=3pt, label = {above: {\footnotesize $\textcolor{color2}{3}$}}] (x3) {};
\vertex[below = 16pt of r, color2, dot, minimum size=3pt, label = {below: {\footnotesize $\textcolor{color2}{4}$}}] (x4) {};
\diagram*{
	(x1) -- [color1, ghost] (v12) -- [color1, ghost](x2),
	(v12) -- [black, photon, edge label = {\scriptsize \;$\widehat{\Delta \Omega}_{j_1}$}, inner sep = 4pt] (b12) -- [quarter left] (w12) -- [quarter left] (b12),
	(b34) -- [color2, ghost, quarter left] (w34) -- [color2, ghost, quarter left] (b34) -- [color2doubghost, edge label = {\scriptsize \,$\sigma'_{j_2}\sigma'_{j_2}$}, inner sep = 4pt] (v34),
	(x3) -- [color2, ghost] (v34) -- [color2, ghost] (x4)
};
\draw [decoration={brace}, decorate] (w34dr) -- (w12dl)
node [pos=0.5, below = 1pt] {\scriptsize $\frac{1}{n_{\ell-1}\*}D_4^{(\ell)}$};
\end{feynman}
\end{tikzpicture} \nonumber\\
&& + 
\sum_{j_1,j_2}
\begin{tikzpicture}[baseline=(b)]
\tikzfeynmanset{every blob = {/tikz/fill=white!50, /tikz/minimum size=15pt}}
\begin{feynman}
\vertex (l) {};
\vertex[below = 16pt of l, color1, dot, minimum size=3pt, label = {below: {\footnotesize $\textcolor{color1}{1}$}}] (x1) {};
\vertex[above = 16pt of l, color1, dot, minimum size=3pt, label = {above: {\footnotesize $\textcolor{color1}{2}$}}] (x2) {};
\vertex[right = 8pt of l, dot, minimum size=0pt] (v12) {};
\vertex[right = 35pt of v12, blob] (b12) {};
\vertex[right = 35pt of b12, dot, minimum size=0pt] (w12) {};
\vertex[above = 5pt of w12, label = {above: {\scriptsize \hspace{-35pt} $\textcolor{color1}{2}$}}] (w12u) {};
\vertex[below = 12pt of w12, label = {above: {\scriptsize \hspace{-35pt} $\textcolor{color1}{1}$}}] (w12d) {};
\vertex[left = 10pt of w12d] (w12dl) {};
\tikzfeynmanset{every blob = {/tikz/fill=gray!50, /tikz/minimum size=15pt}}
\vertex[right = 3pt of w12, blob , minimum size = 6pt] (b) {};
\vertex[right = 3pt of b, dot, minimum size=0pt] (w34) {};
\tikzfeynmanset{every blob = {/tikz/fill=white!50, /tikz/minimum size=15pt}}
\vertex[right = 16pt of w34, blob] (b34) {};
\vertex[above = 1pt of w34, label = {above: {\scriptsize \hspace{10pt} $z_{j_2}$}}] (w34u) {};
\vertex[below = 12pt of w34] (w34d) {};
\vertex[right = 10pt of w34d] (w34dr) {};
\vertex[right = 35pt of b34, dot, minimum size=0pt] (v34) {};
\vertex[right = 8pt of v34] (r) {};
\vertex[above = 16pt of r, color2, dot, minimum size=3pt, label = {above: {\footnotesize $\textcolor{color2}{3}$}}] (x3) {};
\vertex[below = 16pt of r, color2, dot, minimum size=3pt, label = {below: {\footnotesize $\textcolor{color2}{4}$}}] (x4) {};
\diagram*{
	(x1) -- [color1, ghost] (v12) -- [color1, ghost](x2),
	(v12) -- [color1doubghost, edge label = {\scriptsize \;$\sigma'_{j_{1}}\sigma'_{j_{1}}$}, inner sep = 4pt] (b12) -- [color1, ghost, quarter left] (w12) -- [color1, ghost, quarter left] (b12),
	(b34) -- [quarter left] (w34) -- [quarter left] (b34) -- [black, photon, edge label = {\scriptsize \,$\widehat{\Delta \Omega}_{j_{2}}$}, inner sep = 4pt] (v34),
	(x3) -- [color2, ghost] (v34) -- [color2, ghost] (x4)
};
\draw [decoration={brace}, decorate] (w34dr) -- (w12dl)
node [pos=0.5, below = 1pt] {\scriptsize $\frac{1}{n_{\ell-1}\*}D_4^{(\ell)}$};
\end{feynman}
\end{tikzpicture}\nonumber\\
&& + \sum_{j_1, j_2}
\begin{tikzpicture}[baseline=(b)]
\tikzfeynmanset{every blob = {/tikz/fill=white!50, /tikz/minimum size=15pt}}
\begin{feynman}
\vertex (l) {};
\vertex[below = 16pt of l, color1, dot, minimum size=3pt, label = {below: {\footnotesize $\textcolor{color1}{1}$}}] (x1) {};
\vertex[above = 16pt of l, color1, dot, minimum size=3pt, label = {above: {\footnotesize $\textcolor{color1}{2}$}}] (x2) {};
\vertex[right = 8pt of l, dot, minimum size=0pt] (v12) {};
\vertex[right = 35pt of v12, blob] (b12) {};
\vertex[right = 35pt of b12, dot, minimum size=0pt] (w12) {};
\vertex[above = 5pt of w12, label = {above: {\scriptsize \hspace{-35pt} $\textcolor{color1}{2}$}}] (w12u) {};
\vertex[below = 12pt of w12, label = {above: {\scriptsize \hspace{-35pt} $\textcolor{color1}{1}$}}] (w12d) {};
\vertex[left = 10pt of w12d] (w12dl) {};
\tikzfeynmanset{every blob = {/tikz/fill=gray!50, /tikz/minimum size=15pt}}
\vertex[right = 3pt of w12, blob , minimum size = 6pt] (b) {};
\vertex[right = 3pt of b, dot, minimum size=0pt] (w34) {};
\tikzfeynmanset{every blob = {/tikz/fill=white!50, /tikz/minimum size=15pt}}
\vertex[right = 35pt of w34, blob] (b34) {};
\vertex[above = 5pt of w34, label = {above: {\scriptsize \hspace{30pt} $\textcolor{color2}{3}$}}] (w34u) {};
\vertex[below = 12pt of w34, label = {above: {\scriptsize \hspace{30pt} $\textcolor{color2}{4}$}}] (w34d) {};
\vertex[right = 10pt of w34d] (w34dr) {};
\vertex[right = 35pt of b34, dot, minimum size=0pt] (v34) {};
\vertex[right = 8pt of v34] (r) {};
\vertex[above = 16pt of r, color2, dot, minimum size=3pt, label = {above: {\footnotesize $\textcolor{color2}{3}$}}] (x3) {};
\vertex[below = 16pt of r, color2, dot, minimum size=3pt, label = {below: {\footnotesize $\textcolor{color2}{4}$}}] (x4) {};
\diagram*{
	(x1) -- [color1, ghost] (v12) -- [color1, ghost] (x2),
	(v12) -- [color1doubghost, edge label = {\scriptsize \;$\sigma'_{j_1}\sigma'_{j_1}$}, inner sep = 4pt] (b12) -- [color1, ghost, quarter left] (w12) -- [color1, ghost, quarter left] (b12),
	(b34) -- [color2, ghost, quarter left] (w34) -- [color2, ghost, quarter left] (b34) -- [color2doubghost, edge label = {\scriptsize \,$\sigma'_{j_{2}}\sigma'_{j_{2}}$}, inner sep = 4pt] (v34),
	(x3) -- [color2, ghost] (v34) -- [color2, ghost] (x4)
};
\draw [decoration={brace}, decorate] (w34dr) -- (w12dl)
node [pos=0.5, below = 1pt] {\scriptsize $\frac{1}{n_{\ell-1}\*}A_4^{(\ell)}$};
\end{feynman}
\end{tikzpicture}
\end{eqnarray}
which translates into the analytical expression
\begin{eqnarray}
	\frac{1}{n_{\ell}}A^{(\ell +1)}_{\textcolor{color1}{12}\textcolor{color2}{34}} &=& \frac{1}{n_{\ell}}\langle \widehat{\Delta \Omega}^{(\ell + 1)}_{12}\widehat{\Delta \Omega}_{34}^{(\ell +1)}\rangle_{K^{(\ell)}} \nonumber\\
	&& + \frac{1}{4n_{\ell-1}}\sum_{\beta_{1},\beta_{2},\beta_{3},\beta_{4} \in \{1,2,3,4\}}V^{(\ell)}_{(\beta_{1}\beta_{2})(\beta_{3}\beta_{4})}\langle \frac{d^{2}(\widehat{\Delta \Omega}^{(\ell + 1)}_{12})}{d z^{(\ell)}_{\beta_{1}}d z^{(\ell)}_{\beta_{2}}}\rangle_{K^{(\ell)}}\langle \frac{d^{2}(\widehat{\Delta \Omega}^{(\ell +1)}_{34})}{d z^{(\ell)}_{\beta_{1}}d z^{(\ell)}_{\beta_{2}}}\rangle_{K^{(\ell)}}\nonumber\\
	&& + \frac{C_{W}^{(\ell +1)}}{2n_{\ell-1}}\sum_{\beta_{1},\beta_{2} \in \{1,2,3,4\}}\langle \frac{d^{2}(\widehat{\Delta \Omega}^{(\ell + 1)}_{12})}{d z^{(\ell)}_{\beta_{1}}d z^{(\ell)}_{\beta_{2}}}\rangle_{K^{(\ell)}} \langle \sigma'^{(\ell)}_{\textcolor{color2}{3}}\sigma'^{(\ell)}_{\textcolor{color2}{4}}\rangle_{K^{(\ell)}}D^{(\ell)}_{\beta_{1}\beta_{2}\textcolor{color2}{34}} \nonumber\\
	&& + \frac{C_{W}^{(\ell +1)}}{2n_{\ell-1}}\sum_{\beta_{3},\beta_{4} \in \{1,2,3,4\}}\langle \frac{d^{2}(\widehat{\Delta \Omega}^{(\ell + 1)}_{34})}{d z^{(\ell)}_{\beta_{3}}d z^{(\ell)}_{\beta_{4}}}\rangle_{K^{(\ell)}} \langle \sigma'^{(\ell)}_{\textcolor{color1}{1}}\sigma'^{(\ell)}_{\textcolor{color1}{2}}\rangle_{K^{(\ell)}}D^{(\ell)}_{\textcolor{color1}{12}\beta_{3}\beta_{4}}\nonumber\\
	&& + \frac{(C_{W}^{(\ell +1)})^{2}}{n_{\ell-1}}\langle \sigma'^{(\ell)}_{\textcolor{color1}{1}}\sigma'^{(\ell)}_{\textcolor{color1}{2}}\rangle_{K^{(\ell)}} \langle \sigma'^{(\ell)}_{\textcolor{color2}{3}}\sigma'^{(\ell)}_{\textcolor{color2}{4}}\rangle_{K^{(\ell)}}A^{(\ell)}_{\textcolor{color1}{12}\textcolor{color2}{34}}\,. 
    \label{eq:A_recursion_algebraic}
\end{eqnarray}
      This expression agrees with the expression found in~\cite{roberts2022} using algebraic manipulations.

Similarly, one can find the tensor $B^{(\ell + 1)}$ from the following diagrammatic relation:
\begin{eqnarray}
\begin{tikzpicture}[baseline=(b)]
\begin{feynman}
\vertex (l) {};
\vertex[below = 16pt of l, color1, dot, minimum size=3pt, label = {below: {\footnotesize $\textcolor{color1}{1}$}}] (x1) {};
\vertex[above = 16pt of l, color2, dot, minimum size=3pt, label = {above: {\footnotesize $\textcolor{color2}{3}$}}] (x2) {};
\vertex[right = 8pt of l, dot, minimum size=0pt] (v12) {};
\vertex[right = 20pt of v12, quarticblob] (b) {};
\vertex[below = 25pt of b] {\scriptsize $\frac{1}{n_{\ell}}B^{(\ell +1)}_{\textcolor{color1}{1}\textcolor{color2}{3}\textcolor{color1}{2}\textcolor{color2}{4}}$}; 
\vertex[right = 20pt of b, dot, minimum size=0pt] (v34) {};
\vertex[right = 8pt of v34] (r) {};
\vertex[above = 16pt of r, color1, dot, minimum size=3pt, label = {above: {\footnotesize $\textcolor{color1}{2}$}}] (x3) {};
\vertex[below = 16pt of r, color2, dot, minimum size=3pt, label = {below: {\footnotesize $\textcolor{color2}{4}$}}] (x4) {};
\diagram*{
	(x1) -- [color1, ghost] (b) -- [color2, ghost](x2),
	(x3) -- [color1, ghost] (b) -- [color2, ghost] (x4)
};
\end{feynman}
\end{tikzpicture}
&=&
\sum_{j}
\begin{tikzpicture}[baseline=(b)]
\tikzfeynmanset{every blob = {/tikz/fill=white!50, /tikz/minimum size=15pt}}
\begin{feynman}
\vertex (l) {};
\vertex[below = 16pt of l, color1, dot, minimum size=3pt, label = {below: {\footnotesize $\textcolor{color1}{1}$}}] (x1) {};
\vertex[above = 16pt of l, color2, dot, minimum size=3pt, label = {above: {\footnotesize $\textcolor{color2}{3}$}}] (x2) {};
\vertex[right = 8pt of l, dot, minimum size=0pt] (v12) {};
\vertex[right = 30pt of v12, blob] (b) {};
\vertex[right = 30pt of b, dot, minimum size=0pt] (v34) {};
\vertex[right = 8pt of v34] (r) {};
\vertex[above = 16pt of r, color1, dot, minimum size=3pt, label = {above: {\footnotesize $\textcolor{color1}{2}$}}] (x3) {};
\vertex[below = 16pt of r, color2, dot, minimum size=3pt, label = {below: {\footnotesize $\textcolor{color2}{4}$}}] (x4) {};
\diagram*{
	(x1) -- [color1, ghost] (v12) -- [color2, ghost] (x2),
	(v12) -- [color2color1ghost, edge label = {\scriptsize \;$\sigma'_{j}\sigma'_{j}$}, inner sep = 4pt] (b) -- [color1color2ghost, edge label = {\scriptsize \,$\sigma'_{j}\sigma'_{j}$}, inner sep = 4pt] (v34), 
	 (x3) -- [color1, ghost] (v34) -- [color2, ghost] (x4)
};
\end{feynman}
\end{tikzpicture}
+ \sum_{j_1, j_2}
\begin{tikzpicture}[baseline=(b)]
\tikzfeynmanset{every blob = {/tikz/fill=white!50, /tikz/minimum size=15pt}}
\begin{feynman}
\vertex (l) {};
\vertex[below = 16pt of l, color1, dot, minimum size=3pt, label = {below: {\footnotesize $\textcolor{color1}{1}$}}] (x1) {};
\vertex[above = 16pt of l, color2, dot, minimum size=3pt, label = {above: {\footnotesize $\textcolor{color2}{3}$}}] (x2) {};
\vertex[right = 8pt of l, dot, minimum size=0pt] (v12) {};
\vertex[right = 35pt of v12, blob] (b12) {};
\vertex[right = 35pt of b12, dot, minimum size=0pt] (w12) {};
\vertex[above = 5pt of w12, label = {above: {\scriptsize \hspace{-35pt} $\textcolor{color2}{3}$}}] (w12u) {};
\vertex[below = 12pt of w12, label = {above: {\scriptsize \hspace{-35pt} $\textcolor{color1}{1}$}}] (w12d) {};
\vertex[left = 10pt of w12d] (w12dl) {};
\tikzfeynmanset{every blob = {/tikz/fill=gray!50, /tikz/minimum size=15pt}}
\vertex[right = 3pt of w12, blob , minimum size = 6pt] (b) {};
\vertex[right = 3pt of b, dot, minimum size=0pt] (w34) {};
\tikzfeynmanset{every blob = {/tikz/fill=white!50, /tikz/minimum size=15pt}}
\vertex[right = 35pt of w34, blob] (b34) {};
\vertex[above = 5pt of w34, label = {above: {\scriptsize \hspace{30pt} $\textcolor{color1}{2}$}}] (w34u) {};
\vertex[below = 12pt of w34, label = {above: {\scriptsize \hspace{30pt} $\textcolor{color2}{4}$}}] (w34d) {};
\vertex[right = 10pt of w34d] (w34dr) {};
\vertex[right = 35pt of b34, dot, minimum size=0pt] (v34) {};
\vertex[right = 8pt of v34] (r) {};
\vertex[above = 16pt of r, color1, dot, minimum size=3pt, label = {above: {\footnotesize $\textcolor{color1}{2}$}}] (x3) {};
\vertex[below = 16pt of r, color2, dot, minimum size=3pt, label = {below: {\footnotesize $\textcolor{color2}{4}$}}] (x4) {};
\diagram*{
	(x1) -- [color1, ghost] (v12) -- [color2, ghost] (x2),
	(v12) -- [color2color1ghost, edge label = {\scriptsize \;$\sigma'_{j_1}\sigma'_{j_1}$}, inner sep = 4pt] (b12) -- [color2, ghost, quarter left] (w12) -- [color1, ghost, quarter left] (b12),
	(b34) -- [color2, ghost, quarter left] (w34) -- [color1, ghost, quarter left] (b34) -- [color1color2ghost, edge label = {\scriptsize \,$\sigma'_{j_{2}}\sigma'_{j_{2}}$}, inner sep = 4pt] (v34),
	(x3) -- [color1, ghost] (v34) -- [color2, ghost] (x4)
};
\draw [decoration={brace}, decorate] (w34dr) -- (w12dl)
node [pos=0.5, below = 1pt] {\scriptsize $\frac{1}{n_{\ell-1}\*}B_4^{(\ell)}$};
\end{feynman}
\end{tikzpicture}\nonumber\\
\end{eqnarray}
whose algebraic interpretation reads
\begin{eqnarray}
\frac{1}{n_{\ell}}B^{(\ell +1)}_{\textcolor{color1}{1}\textcolor{color2}{3}\textcolor{color1}{2}\textcolor{color2}{4}} &=& \frac{(C_{W}^{(\ell +1)})^{2}}{n_{\ell}}\Theta^{(\ell)}_{\textcolor{color1}{12}}\Theta^{(\ell)}_{\textcolor{color2}{34}}\langle \sigma'^{(\ell)}_{\textcolor{color1}{1}}\sigma'^{(\ell)}_{\textcolor{color1}{2}}\sigma'^{(\ell)}_{\textcolor{color2}{3}}\sigma'^{(\ell)}_{\textcolor{color2}{4}}\rangle_{K^{(\ell)}} \nonumber\\
&& + \frac{(C_{W}^{(\ell +1)})^{2}}{n_{\ell-1}}\langle \sigma'^{(\ell)}_{\textcolor{color1}{1}}\sigma'^{(\ell)}_{\textcolor{color2}{3}}\rangle_{K^{(\ell)}}\langle \sigma'^{(\ell)}_{\textcolor{color1}{2}}\sigma'^{(\ell)}_{\textcolor{color2}{4}}\rangle_{K^{(\ell)}}B^{(\ell)}_{\textcolor{color1}{1}\textcolor{color2}{3}\textcolor{color1}{2}\textcolor{color2}{4}}\,.
\label{eq:B_recursion_algebraic}
\end{eqnarray}
This expression agrees with the expression found in~\cite{roberts2022} using algebraic manipulations.
\end{proof}

\subsection{Proof of Theorem \ref{theoremtwo}}\label{app:proofs_two}
\secondtheorem*

The proof will proceed case-by-case for the joint dNTK-preactivation cumulant and the ddNTK means as follows.

\subsubsection{Joint \scabbr{dNTK}-Preactivation Cumulant}
\begin{proof}
The first relevant cumulant at order $\frac{1}{n}$ including the dNTK is given by \cite{roberts2022}
\begin{align}
  \mathbb{E}^{c}_{\theta} \left[ \widehat{\mathrm{d}\Theta}^{(\ell + 1)}_{i_0 i_1 i_2}( \textcolor{color3}{x_0}, \textcolor{color1}{x_1}, \textcolor{color2}{x_2}), z^{(\ell + 1)}_{i_3}(x_3)\right] &=
  \frac{1}{n_{\ell}} 
  \left[ 
    \delta_{i_0 i_3} \delta_{i_1 i_2} P_{\textcolor{color3}{0}\textcolor{color1}{1}\textcolor{color2}{2}3}^{(\ell +1)}
    + 
    \delta_{i_0 i_1} \delta_{i_2 i_3} Q_{\textcolor{color3}{0}\textcolor{color1}{1}\textcolor{color2}{2}3}^{(\ell +1)}
    + 
    \delta_{i_0 i_2} \delta_{i_1 i_3} Q_{\textcolor{color3}{0}\textcolor{color2}{2}\textcolor{color1}{1}3}^{(\ell +1)}
  \right]
\end{align}
The Feynman rules introduced in Appendix \ref{app:feynman_rules}, then lead to the following expression for the tensor $P_{\textcolor{color3}{0}\textcolor{color1}{1}\textcolor{color2}{2}3}^{(\ell +1)}$:
\begin{eqnarray}
\begin{tikzpicture}[baseline=(b)]
\begin{feynman}
\vertex (l) {};
\vertex[below = 16pt of l, color1, dot, minimum size=3pt, label = {below: {\footnotesize $\textcolor{color1}{1}$}}] (x1) {};
\vertex[above = 16pt of l, color2, dot, minimum size=3pt, label = {above: {\footnotesize $\textcolor{color2}{2}$}}] (x2) {};
\vertex[right = 8pt of l, dot, minimum size=0pt] (v12) {};
\vertex[right = 20pt of v12, quarticblob] (b) {};
\vertex[below = 25pt of b] {\scriptsize $\frac{1}{n_{\ell}\*}P_{\textcolor{color3}{0}\textcolor{color1}{1}\textcolor{color2}{2}3}^{(\ell +1)}$}; 
\vertex[right = 20pt of b, dot, minimum size=0pt] (v34) {};
\vertex[right = 8pt of v34] (r) {};
\vertex[above = 16pt of r, color3, dot, minimum size=3pt, label = {above: {\footnotesize $\textcolor{color3}{0}$}}] (x3) {};
\vertex[below = 16pt of r, dot, minimum size=3pt, label = {below: {\footnotesize $3$}}] (x4) {};
\diagram*{
	(x1) -- [color1, ghost] (b) -- [color2, ghost](x2),
	(x3) -- [color1color2dntknew] (b) -- (x4)
};
\end{feynman}
\end{tikzpicture}
&=&
\sum_{j}
\begin{tikzpicture}[baseline=(b)]
\tikzfeynmanset{every blob = {/tikz/fill=white!50, /tikz/minimum size=15pt}}
\begin{feynman}
\vertex (l) {};
\vertex[below = 16pt of l, color1, dot, minimum size=3pt, label = {below: {\footnotesize $\textcolor{color1}{1}$}}] (x1) {};
\vertex[above = 16pt of l, color2, dot, minimum size=3pt, label = {above: {\footnotesize $\textcolor{color2}{2}$}}] (x2) {};
\vertex[right = 8pt of l, dot, minimum size=0pt] (v12) {};
\vertex[right = 30pt of v12, blob] (b) {};
\vertex[right = 30pt of b, dot, minimum size=0pt] (v34) {};
\vertex[right = 8pt of v34] (r) {};
\vertex[above = 16pt of r, color3, dot, minimum size=3pt, label = {above: {\footnotesize $\textcolor{color3}{0}$}}] (x3) {};
\vertex[below = 16pt of r, dot, minimum size=3pt, label = {below: {\footnotesize $3$}}] (x4) {};
\diagram*{
	(x1) -- [color1, ghost] (v12) -- [color2, ghost] (x2),
	(v12) -- [color2color1ghost, edge label = {\scriptsize \;$\sigma'_{j}\sigma'_{j}$}, inner sep = 4pt] (b) -- [color1color2ghost, edge label = {\scriptsize \,$\sigma''_{j}\sigma_{j}$}, inner sep = 4pt] (v34), 
	 (x3) -- [color1color2dntknew] (v34) -- (x4)
};
\end{feynman}
\end{tikzpicture}
+ \sum_{j_1, j_2}
\begin{tikzpicture}[baseline=(b)]
\tikzfeynmanset{every blob = {/tikz/fill=white!50, /tikz/minimum size=15pt}}
\begin{feynman}
\vertex (l) {};
\vertex[below = 16pt of l, color1, dot, minimum size=3pt, label = {below: {\footnotesize $\textcolor{color1}{1}$}}] (x1) {};
\vertex[above = 16pt of l, color2, dot, minimum size=3pt, label = {above: {\footnotesize $\textcolor{color2}{2}$}}] (x2) {};
\vertex[right = 8pt of l, dot, minimum size=0pt] (v12) {};
\vertex[right = 35pt of v12, blob] (b12) {};
\vertex[right = 35pt of b12, dot, minimum size=0pt] (w12) {};
\vertex[above = 5pt of w12, label = {above: {\scriptsize \hspace{-35pt} $\textcolor{color2}{2}$}}] (w12u) {};
\vertex[below = 12pt of w12, label = {above: {\scriptsize \hspace{-35pt} $\textcolor{color1}{1}$}}] (w12d) {};
\vertex[left = 10pt of w12d] (w12dl) {};
\tikzfeynmanset{every blob = {/tikz/fill=gray!50, /tikz/minimum size=15pt}}
\vertex[right = 3pt of w12, blob , minimum size = 6pt] (b) {};
\vertex[right = 3pt of b, dot, minimum size=0pt] (w34) {};
\tikzfeynmanset{every blob = {/tikz/fill=white!50, /tikz/minimum size=15pt}}
\vertex[right = 35pt of w34, blob] (b34) {};
\vertex[above = 5pt of w34, label = {above: {\scriptsize \hspace{30pt} $\textcolor{color1}{0}$}}] (w34u) {};
\vertex[below = 12pt of w34, label = {above: {\scriptsize \hspace{30pt} $\textcolor{color2}{0}$}}] (w34d) {};
\vertex[right = 10pt of w34d] (w34dr) {};
\vertex[right = 35pt of b34, dot, minimum size=0pt] (v34) {};
\vertex[right = 8pt of v34] (r) {};
\vertex[above = 16pt of r, color3, dot, minimum size=3pt, label = {above: {\footnotesize $\textcolor{color3}{0}$}}] (x3) {};
\vertex[below = 16pt of r, dot, minimum size=3pt, label = {below: {\footnotesize $3$}}] (x4) {};
\diagram*{
	(x1) -- [color1, ghost] (v12) -- [color2, ghost] (x2),
	(v12) -- [color2color1ghost, edge label = {\scriptsize \;$\sigma'_{j_1}\sigma'_{j_1}$}, inner sep = 4pt] (b12) -- [color2, ghost, quarter left] (w12) -- [color1, ghost, quarter left] (b12),
	(b34) -- [color2, ghost, quarter left] (w34) -- [color1, ghost, quarter left] (b34) -- [color1color2ghost, edge label = {\scriptsize \,$\sigma''_{j_{2}}\sigma_{j_{2}}$}, inner sep = 4pt] (v34),
	(x3) -- [color1color2dntknew] (v34) -- (x4)
};
\draw [decoration={brace}, decorate] (w34dr) -- (w12dl)
node [pos=0.5, below = 1pt] {\scriptsize $\frac{1}{n_{\ell-1}\*}B_4^{(\ell)}$};
\end{feynman}
\end{tikzpicture}\nonumber\\
&& 
+ \sum_{j_1, j_2}
\begin{tikzpicture}[baseline=(b)]
\tikzfeynmanset{every blob = {/tikz/fill=white!50, /tikz/minimum size=15pt}}
\begin{feynman}
\vertex (l) {};
\vertex[below = 16pt of l, color1, dot, minimum size=3pt, label = {below: {\footnotesize $\textcolor{color1}{1}$}}] (x1) {};
\vertex[above = 16pt of l, color2, dot, minimum size=3pt, label = {above: {\footnotesize $\textcolor{color2}{2}$}}] (x2) {};
\vertex[right = 8pt of l, dot, minimum size=0pt] (v12) {};
\vertex[right = 35pt of v12, blob] (b12) {};
\vertex[right = 35pt of b12, dot, minimum size=0pt] (w12) {};
\vertex[above = 5pt of w12, label = {above: {\scriptsize \hspace{-35pt} $\textcolor{color2}{2}$}}] (w12u) {};
\vertex[below = 12pt of w12, label = {above: {\scriptsize \hspace{-35pt} $\textcolor{color1}{1}$}}] (w12d) {};
\vertex[left = 10pt of w12d] (w12dl) {};
\tikzfeynmanset{every blob = {/tikz/fill=gray!50, /tikz/minimum size=15pt}}
\vertex[right = 3pt of w12, blob , minimum size = 6pt] (b) {};
\vertex[right = 3pt of b, dot, minimum size=0pt] (w34) {};
\tikzfeynmanset{every blob = {/tikz/fill=white!50, /tikz/minimum size=15pt}}
\vertex[right = 35pt of w34, blob] (b34) {};
\vertex[above = 5pt of w34, label = {above: {\scriptsize \hspace{30pt} $\textcolor{color3}{0}$}}] (w34u) {};
\vertex[below = 12pt of w34, label = {above: {\scriptsize \hspace{30pt} $z_{j_{2},\beta_{3}}$}}] (w34d) {};
\vertex[right = 10pt of w34d] (w34dr) {};
\vertex[right = 35pt of b34, dot, minimum size=0pt] (v34) {};
\vertex[right = 8pt of v34] (r) {};
\vertex[above = 16pt of r, color3, dot, minimum size=3pt, label = {above: {\footnotesize $\textcolor{color3}{0}$}}] (x3) {};
\vertex[below = 16pt of r, dot, minimum size=3pt, label = {below: {\footnotesize $3$}}] (x4) {};
\diagram*{
	(x1) -- [color1, ghost] (v12) -- [color2, ghost] (x2),
	(v12) -- [color2color1ghost, edge label = {\scriptsize \;$\sigma'_{j_1}\sigma'_{j_1}$}, inner sep = 4pt] (b12) -- [color2, ghost, quarter left] (w12) -- [color1, ghost, quarter left] (b12),
	(b34) -- [quarter left] (w34) -- [color2color1dntknew, quarter left] (b34) -- [color1color2blackghost, edge label = {\scriptsize \,$\sigma'_{j_{2}}\sigma_{j_{2}}$}, inner sep = 4pt] (v34),
	(x3) -- [color1color2dntknew] (v34) -- (x4)
};
\draw [decoration={brace}, decorate] (w34dr) -- (w12dl)
node [pos=0.5, below = 1pt] {\scriptsize $\frac{1}{n_{\ell-1}\*}P_4^{(\ell)}$};
\end{feynman}
\end{tikzpicture} 
\end{eqnarray}
or, equivalently
\begin{eqnarray}
	\frac{1}{n_{\ell}}P_{\textcolor{color3}{0}\textcolor{color1}{1}\textcolor{color2}{2}3}^{(\ell +1)} &=& \frac{(C_{W}^{(\ell +1)})^{2}}{n_{\ell}}\Theta^{(\ell)}_{\textcolor{color1}{01}}\Theta^{(\ell)}_{\textcolor{color2}{02}}\langle\sigma'^{(\ell)}_{\textcolor{color1}{1}}\sigma'^{(\ell)}_{\textcolor{color2}{2}}\sigma''^{(\ell)}_{\textcolor{color3}{0}}\sigma^{(\ell)}_{3}\rangle_{K^{(\ell)}}\nonumber\\
	&& + \frac{(C_{W}^{(\ell +1)})^{2}}{n_{\ell-1}}\langle \sigma'^{(\ell)}_{\textcolor{color1}{1}}\sigma'^{(\ell)}_{\textcolor{color2}{2}}\rangle_{K^{(\ell)}}\langle \sigma''^{(\ell)}_{\textcolor{color1}{0}}\sigma^{(\ell)}_{\textcolor{color2}{0}}\rangle_{K^{(\ell)}}B^{(\ell)}_{\textcolor{color1}{0}\textcolor{color2}{0}\textcolor{color1}{1}\textcolor{color2}{2}}\nonumber\\
	&& + \frac{(C_{W}^{(\ell +1)})^{2}}{n_{\ell-1}}\langle \sigma'^{(\ell)}_{\textcolor{color1}{1}}\sigma'^{(\ell)}_{\textcolor{color2}{2}}\rangle_{K^{(\ell)}}\sum_{\beta_{3}\in \{1,2,3,4\}}\langle \frac{d (\sigma'^{(\ell)}_{\textcolor{color3}{0}}\sigma^{(\ell)}_{3})}{d z^{(\ell)}_{\beta_{3}}}\rangle_{K^{(\ell)}}  P^{(\ell)}_{\textcolor{color3}{0}\textcolor{color1}{1}\textcolor{color2}{2}\beta_{3}}\,.
      \end{eqnarray}
      This expression agrees with the expression found in~\cite{roberts2022} using algebraic manipulations.

The application of these same rules to the construction of the tensor $Q_{\textcolor{color3}{0}\textcolor{color1}{1}\textcolor{color2}{2}3}^{(\ell +1)}$ yields:
\begin{eqnarray}
\begin{tikzpicture}[baseline=(b)]
\begin{feynman}
\vertex (l) {};
\vertex[below = 16pt of l, color3, dot, minimum size=3pt, label = {below: {\footnotesize $\textcolor{color3}{0}$}}] (x1) {};
\vertex[above = 16pt of l, color1, dot, minimum size=3pt, label = {above: {\footnotesize $\textcolor{color1}{1}$}}] (x2) {};
\vertex[right = 8pt of l, dot, minimum size=0pt] (v12) {};
\vertex[right = 20pt of v12, quarticblob] (b) {};
\vertex[below = 25pt of b] {\scriptsize $\frac{1}{n_{\ell}\*}Q_{\textcolor{color3}{0}\textcolor{color1}{1}\textcolor{color2}{2}3}^{(\ell +1)}$}; 
\vertex[right = 20pt of b, dot, minimum size=0pt] (v34) {};
\vertex[right = 8pt of v34] (r) {};
\vertex[above = 16pt of r, color2, dot, minimum size=3pt, label = {above: {\footnotesize $\textcolor{color2}{2}$}}] (x3) {};
\vertex[below = 16pt of r, dot, minimum size=3pt, label = {below: {\footnotesize $3$}}] (x4) {};
\diagram*{
	(x1) -- [color2color1dntknew] (b) -- [color1, ghost](x2),
	(x3) -- [color2, ghost] (b) -- (x4)
};
\end{feynman}
\end{tikzpicture}
&=&
\frac{1}{C_{W}^{(\ell +1)}} \begin{tikzpicture}[baseline=(b)]
\begin{feynman}
\vertex (l) {};
\vertex[below = 16pt of l, color2, dot, minimum size=3pt, label = {below: {\footnotesize $\textcolor{color2}{0}$}}] (x1) {};
\vertex[above = 16pt of l, dot, minimum size=3pt, label = {above: {\footnotesize $1$}}] (x2) {};
\vertex[right = 8pt of l, dot, minimum size=0pt] (v12) {};
\vertex[right = 20pt of v12, quarticblob] (b) {};
\vertex[below = 25pt of b] {\scriptsize $\frac{1}{n_{\ell}\*}F_{\textcolor{color2}{0}1\textcolor{color2}{2}3}^{(\ell +1)}$}; 
\vertex[right = 20pt of b, dot, minimum size=0pt] (v34) {};
\vertex[right = 8pt of v34] (r) {};
\vertex[above = 16pt of r, color2, dot, minimum size=3pt, label = {above: {\footnotesize $\textcolor{color2}{2}$}}] (x3) {};
\vertex[below = 16pt of r, dot, minimum size=3pt, label = {below: {\footnotesize $3$}}] (x4) {};
\diagram*{
	(x1)  -- [color2, ghost] (b) -- (x2),
	(x3) -- [color2, ghost] (b) -- (x4)
};
\end{feynman}
\end{tikzpicture}
+ \sum_{j}
\begin{tikzpicture}[baseline=(b)]
\tikzfeynmanset{every blob = {/tikz/fill=white!50, /tikz/minimum size=15pt}}
\begin{feynman}
\vertex (l) {};
\vertex[below = 16pt of l, color3, dot, minimum size=3pt, label = {below: {\footnotesize $\textcolor{color3}{0}$}}] (x1) {};
\vertex[above = 16pt of l, color1, dot, minimum size=3pt, label = {above: {\footnotesize $\textcolor{color1}{1}$}}] (x2) {};
\vertex[right = 8pt of l, dot, minimum size=0pt] (v12) {};
\vertex[right = 30pt of v12, blob] (b) {};
\vertex[right = 30pt of b, dot, minimum size=0pt] (v34) {};
\vertex[right = 8pt of v34] (r) {};
\vertex[above = 16pt of r, color2, dot, minimum size=3pt, label = {above: {\footnotesize $\textcolor{color2}{2}$}}] (x3) {};
\vertex[below = 16pt of r, dot, minimum size=3pt, label = {below: {\footnotesize $3$}}] (x4) {};
\diagram*{
	(x1) -- [color2color1dntknew] (v12) -- [color1, ghost] (x2),
	(v12) -- [blackcolor2ghost, edge label = {\scriptsize \;$\sigma''_{j}\sigma'_{j}$}, inner sep = 4pt] (b) -- [color2blackghost, edge label = {\scriptsize \,$\sigma'_{j}\sigma_{j}$}, inner sep = 4pt] (v34), 
	 (x3) -- [color2, ghost] (v34) -- (x4)
};
\end{feynman}
\end{tikzpicture}\nonumber\\
&& + \sum_{j_1, j_2}
\begin{tikzpicture}[baseline=(b)]
\tikzfeynmanset{every blob = {/tikz/fill=white!50, /tikz/minimum size=15pt}}
\begin{feynman}
\vertex (l) {};
\vertex[below = 16pt of l, color3, dot, minimum size=3pt, label = {below: {\footnotesize $\textcolor{color3}{0}$}}] (x1) {};
\vertex[above = 16pt of l, color1, dot, minimum size=3pt, label = {above: {\footnotesize $\textcolor{color1}{1}$}}] (x2) {};
\vertex[right = 8pt of l, dot, minimum size=0pt] (v12) {};
\vertex[right = 35pt of v12, blob] (b12) {};
\vertex[right = 35pt of b12, dot, minimum size=0pt] (w12) {};
\vertex[above = 5pt of w12, label = {above: {\scriptsize \hspace{-35pt} $z_{j_1,\lambda_{1}}$}}] (w12u) {};
\vertex[below = 12pt of w12, label = {above: {\scriptsize \hspace{-35pt} $\textcolor{color2}{0}$}}] (w12d) {};
\vertex[left = 10pt of w12d] (w12dl) {};
\tikzfeynmanset{every blob = {/tikz/fill=gray!50, /tikz/minimum size=15pt}}
\vertex[right = 3pt of w12, blob , minimum size = 6pt] (b) {};
\vertex[right = 3pt of b, dot, minimum size=0pt] (w34) {};
\tikzfeynmanset{every blob = {/tikz/fill=white!50, /tikz/minimum size=15pt}}
\vertex[right = 35pt of w34, blob] (b34) {};
\vertex[above = 5pt of w34, label = {above: {\scriptsize \hspace{30pt} $\textcolor{color2}{2}$}}] (w34u) {};
\vertex[below = 12pt of w34, label = {above: {\scriptsize \hspace{30pt} $z_{j_2,\lambda_{3}}$}}] (w34d) {};
\vertex[right = 10pt of w34d] (w34dr) {};
\vertex[right = 35pt of b34, dot, minimum size=0pt] (v34) {};
\vertex[right = 8pt of v34] (r) {};
\vertex[above = 16pt of r, color2, dot, minimum size=3pt, label = {above: {\footnotesize $\textcolor{color2}{2}$}}] (x3) {};
\vertex[below = 16pt of r, dot, minimum size=3pt, label = {below: {\footnotesize $3$}}] (x4) {};
\diagram*{
	(x1) -- [color2color1dntknew] (v12) -- [color1, ghost] (x2),
	(v12) -- [blackcolor2ghost, edge label = {\scriptsize \;$\sigma''_{j_1}\sigma'_{j_1}$}, inner sep = 4pt] (b12) -- [quarter left] (w12) -- [color2, ghost, quarter left] (b12),
	(b34) -- [quarter left] (w34) -- [color2, ghost, quarter left] (b34) -- [color2blackghost, edge label = {\scriptsize \,$\sigma'_{j_{2}}\sigma_{j_{2}}$}, inner sep = 4pt] (v34),
	(x3) -- [color2, ghost] (v34) -- (x4)
};
\draw [decoration={brace}, decorate] (w34dr) -- (w12dl)
node [pos=0.5, below = 1pt] {\scriptsize $\frac{1}{n_{\ell-1}\*}F_4^{(\ell)}$};
\end{feynman}
\end{tikzpicture}\nonumber\\
&& 
+ \sum_{j_1, j_2}
\begin{tikzpicture}[baseline=(b)]
\tikzfeynmanset{every blob = {/tikz/fill=white!50, /tikz/minimum size=15pt}}
\begin{feynman}
\vertex (l) {};
\vertex[below = 16pt of l, color3, dot, minimum size=3pt, label = {below: {\footnotesize $\textcolor{color3}{0}$}}] (x1) {};
\vertex[above = 16pt of l, color1, dot, minimum size=3pt, label = {above: {\footnotesize $\textcolor{color1}{1}$}}] (x2) {};
\vertex[right = 8pt of l, dot, minimum size=0pt] (v12) {};
\vertex[right = 35pt of v12, blob] (b12) {};
\vertex[right = 35pt of b12, dot, minimum size=0pt] (w12) {};
\vertex[above = 5pt of w12, label = {above: {\scriptsize \hspace{-35pt} $\textcolor{color1}{1}$}}] (w12u) {};
\vertex[below = 12pt of w12, label = {above: {\scriptsize \hspace{-35pt} $\textcolor{color3}{0}$}}] (w12d) {};
\vertex[left = 10pt of w12d] (w12dl) {};
\tikzfeynmanset{every blob = {/tikz/fill=gray!50, /tikz/minimum size=15pt}}
\vertex[right = 3pt of w12, blob , minimum size = 6pt] (b) {};
\vertex[right = 3pt of b, dot, minimum size=0pt] (w34) {};
\tikzfeynmanset{every blob = {/tikz/fill=white!50, /tikz/minimum size=15pt}}
\vertex[right = 35pt of w34, blob] (b34) {};
\vertex[above = 5pt of w34, label = {above: {\scriptsize \hspace{30pt} $\textcolor{color2}{2}$}}] (w34u) {};
\vertex[below = 12pt of w34, label = {above: {\scriptsize \hspace{30pt} $z_{j_2,\beta_{3}}$}}] (w34d) {};
\vertex[right = 10pt of w34d] (w34dr) {};
\vertex[right = 35pt of b34, dot, minimum size=0pt] (v34) {};
\vertex[right = 8pt of v34] (r) {};
\vertex[above = 16pt of r, color2, dot, minimum size=3pt, label = {above: {\footnotesize $\textcolor{color2}{2}$}}] (x3) {};
\vertex[below = 16pt of r, dot, minimum size=3pt, label = {below: {\footnotesize $3$}}] (x4) {};
\diagram*{
	(x1) -- [color2color1dntknew] (v12) -- [color1, ghost] (x2),
	(v12) -- [ntkdntkcolor1color2, edge label = {\scriptsize \;$\sigma'_{j_1}\sigma'_{j_1}$}, inner sep = 4pt] (b12) -- [color1, ghost, quarter left] (w12) -- [color1color2dntknew, quarter left] (b12),
	(b34) -- [quarter left] (w34) -- [color2, ghost, quarter left] (b34) -- [color2blackghost, edge label = {\scriptsize \,$\sigma'_{j_{2}}\sigma_{j_{2}}$}, inner sep = 4pt] (v34),
	(x3) -- [color2, ghost] (v34) -- (x4)
};
\draw [decoration={brace}, decorate] (w34dr) -- (w12dl)
node [pos=0.5, below = 1pt] {\scriptsize $\frac{1}{n_{\ell-1}\*}Q_4^{(\ell)}$};
\end{feynman}
\end{tikzpicture} 
\end{eqnarray}
which, in algebraic notation, reads
\begin{eqnarray}
	\frac{1}{n_{\ell}}Q^{(\ell +1)}_{\textcolor{color3}{0}\textcolor{color1}{1}\textcolor{color2}{2}3} &=& \frac{1}{C_{W}^{(\ell +1)}n_{\ell}}F^{(\ell +1)}_{\textcolor{color2}{0}1\textcolor{color2}{2}3} + \frac{(C_{W}^{(\ell +1)})^{2}}{n_{\ell}}\Theta^{(\ell)}_{\textcolor{color1}{0}\textcolor{color1}{1}}\Theta^{(\ell)}_{\textcolor{color2}{0}\textcolor{color2}{2}}\langle \sigma''^{(\ell)}_{\textcolor{color2}{0}}\sigma'^{(\ell)}_{1}\sigma'^{(\ell)}_{\textcolor{color2}{2}}\sigma^{(\ell)}_{3}\rangle_{K^{(\ell)}}\nonumber\\
	&& + \frac{(C_{W}^{(\ell +1)})^{2}}{n_{\ell-1}}\Theta^{(\ell)}_{\textcolor{color1}{01}}\sum_{\lambda_{1},\lambda_{3}\in\{0,1,2,3\}}\langle \frac{d (\sigma''^{(\ell)}_{\textcolor{color2}{0}}\sigma'^{(\ell)}_{1})}{d z^{(\ell)}_{\lambda_{1}}}\rangle_{K^{(\ell)}}\langle \frac{d (\sigma'^{(\ell)}_{\textcolor{color2}{2}}\sigma^{(\ell)}_{3})}{d z^{(\ell)}_{\lambda_{3}}}\rangle_{K^{(\ell)}}F^{(\ell)}_{\lambda_{1}\textcolor{color2}{0}\lambda_{3}\textcolor{color2}{2}}\nonumber\\
	&& + \frac{(C_{W}^{(\ell +1)})^{2}}{n_{\ell-1}}\langle \sigma'^{(\ell)}_{\textcolor{color3}{0}}\sigma'^{(\ell)}_{\textcolor{color1}{1}}\rangle_{K^{(\ell)}}\sum_{\beta_{3} \in \{0,1,2,3\}} \langle \frac{d (\sigma'^{(\ell)}_{\textcolor{color2}{2}}\sigma^{(\ell)}_{3})}{d z^{(\ell)}_{\beta_{3}}}\rangle_{K^{(\ell)}}Q^{(\ell)}_{\textcolor{color3}{0}\textcolor{color1}{1}\textcolor{color2}{2}\beta_{3}}\,.
      \end{eqnarray}
      This expression agrees with the expression found in~\cite{roberts2022} using algebraic manipulations.
\end{proof}

\subsubsection{\scabbr{dd$_{\text{I}}$NTK} Mean}
\begin{proof}
Unlike the previous cases, the first non-trivial cumulant containing the ddNTK is its mean itself. For the first type, one finds that \cite{roberts2022}:
\begin{align}
  \mathbb{E}^{c}_{\theta} \left[ \widehat{\text{dd}_{\text{I}}\Theta}_{i_0i_1i_2i_3}^{(\ell + 1)}(\textcolor{color4}{x_0},\textcolor{color6}{x_1},\textcolor{color2}{x_2}, \textcolor{color1}{x_3}) \right]
  &= \frac{1}{n_{\ell}} \left[ \delta_{i_0i_1}\delta_{i_2i_3} R_{\textcolor{color4}{0}\textcolor{color6}{1}\textcolor{color2}{2}\textcolor{color1}{3}}^{(\ell+1)} + \delta_{i_0i_2}\delta_{i_3i_1} R_{\textcolor{color4}{0}\textcolor{color2}{2}\textcolor{color6}{1}\textcolor{color1}{3}}^{(\ell+1)} + \delta_{i_0i_3}\delta_{i_1i_2} R_{\textcolor{color4}{0}\textcolor{color1}{3}\textcolor{color6}{1}\textcolor{color2}{2}}^{(\ell+1)} \right]
\end{align}

Again, the use of the Feynman rules \ref{app:feynman_rules}, imply the following structure for the tensor $R_{\textcolor{color4}{0}\textcolor{color6}{1}\textcolor{color2}{2}\textcolor{color1}{3}}^{(\ell+1)}$:

\begin{eqnarray}
\begin{tikzpicture}[baseline=(b)]
\begin{feynman}
\vertex (l) {};
\vertex[below = 16pt of l, color4, dot, minimum size=3pt, label = {below: {\footnotesize $\textcolor{color4}{0}$}}] (x1) {};
\vertex[above = 16pt of l, color6, dot, minimum size=3pt, label = {above: {\footnotesize $\textcolor{color6}{1}$}}] (x2) {};
\vertex[right = 8pt of l, color1, dot, minimum size=0pt] (v12) {};
\vertex[right = 20pt of v12, quarticblob] (b) {};
\vertex[below = 25pt of b] {\scriptsize $\frac{1}{n_{\ell}}R^{(\ell +1)}_{\textcolor{color4}{0}\textcolor{color6}{1}\textcolor{color2}{2}\textcolor{color1}{3}}$}; 
\vertex[right = 20pt of b, dot, minimum size=0pt] (v34) {};
\vertex[right = 8pt of v34] (r) {};
\vertex[above = 16pt of r, color2, dot, minimum size=3pt, label = {above: {\footnotesize $\textcolor{color2}{2}$}}] (x3) {};
\vertex[below = 16pt of r, color1, dot, minimum size=3pt, label = {below: {\footnotesize $\textcolor{color1}{3}$}}] (x4) {};
\diagram*{
	(x1) -- [color6color1color2ddntknew] (b) -- [color6, ghost](x2),
	(x3) -- [color2, ghost] (b) -- [color1, ghost] (x4)
};
\end{feynman}
\end{tikzpicture}
&=& \sum_{j}
\begin{tikzpicture}[baseline=(b)]
\tikzfeynmanset{every blob = {/tikz/fill=white!50, /tikz/minimum size=15pt}}
\begin{feynman}
\vertex (l) {};
\vertex[below = 16pt of l, color4, dot, minimum size=3pt, label = {below: {\footnotesize $\textcolor{color4}{0}$}}] (x1) {};
\vertex[above = 16pt of l, color6, dot, minimum size=3pt, label = {above: {\footnotesize $\textcolor{color6}{1}$}}] (x2) {};
\vertex[right = 8pt of l, dot, minimum size=0pt] (v12) {};
\vertex[right = 30pt of v12, blob] (b) {};
\vertex[right = 30pt of b, dot, minimum size=0pt] (v34) {};
\vertex[right = 8pt of v34] (r) {};
\vertex[above = 16pt of r, color2, dot, minimum size=3pt, label = {above: {\footnotesize $\textcolor{color2}{2}$}}] (x3) {};
\vertex[below = 16pt of r, color1, dot, minimum size=3pt, label = {below: {\footnotesize $\textcolor{color1}{3}$}}] (x4) {};
\diagram*{
	(x1) -- [color6color1color2ddntknew] (v12) -- [color6, ghost] (x2),
	(v12) -- [color1color2ghost, edge label = {\scriptsize \;$\sigma''_{j}\sigma_{j}$}, inner sep = 4pt] (b) -- [color2color1ghost, edge label = {\scriptsize \,$\sigma'_{j}\sigma'_{j}$}, inner sep = 4pt] (v34), 
	 (x3) -- [color2, ghost] (v34) -- [color1, ghost](x4)
};
\end{feynman}
\end{tikzpicture} + \sum_{j_1, j_2}
\begin{tikzpicture}[baseline=(b)]
\tikzfeynmanset{every blob = {/tikz/fill=white!50, /tikz/minimum size=15pt}}
\begin{feynman}
\vertex (l) {};
\vertex[below = 16pt of l, color4, dot, minimum size=3pt, label = {below: {\footnotesize $\textcolor{color4}{0}$}}] (x1) {};
\vertex[above = 16pt of l, color6, dot, minimum size=3pt, label = {above: {\footnotesize $\textcolor{color6}{1}$}}] (x2) {};
\vertex[right = 8pt of l, dot, minimum size=0pt] (v12) {};
\vertex[right = 35pt of v12, blob] (b12) {};
\vertex[right = 35pt of b12, dot, minimum size=0pt] (w12) {};
\vertex[above = 5pt of w12, label = {above: {\scriptsize \hspace{-35pt} $\textcolor{color1}{0}$}}] (w12u) {};
\vertex[below = 12pt of w12, label = {above: {\scriptsize \hspace{-35pt} $\textcolor{color2}{0}$}}] (w12d) {};
\vertex[left = 10pt of w12d] (w12dl) {};
\tikzfeynmanset{every blob = {/tikz/fill=gray!50, /tikz/minimum size=15pt}}
\vertex[right = 3pt of w12, blob , minimum size = 6pt] (b) {};
\vertex[right = 3pt of b, dot, minimum size=0pt] (w34) {};
\tikzfeynmanset{every blob = {/tikz/fill=white!50, /tikz/minimum size=15pt}}
\vertex[right = 35pt of w34, blob] (b34) {};
\vertex[above = 5pt of w34, label = {above: {\scriptsize \hspace{30pt} $\textcolor{color2}{2}$}}] (w34u) {};
\vertex[below = 12pt of w34, label = {above: {\scriptsize \hspace{30pt} $\textcolor{color1}{3}$}}] (w34d) {};
\vertex[right = 10pt of w34d] (w34dr) {};
\vertex[right = 35pt of b34, dot, minimum size=0pt] (v34) {};
\vertex[right = 8pt of v34] (r) {};
\vertex[above = 16pt of r, color2, dot, minimum size=3pt, label = {above: {\footnotesize $\textcolor{color2}{2}$}}] (x3) {};
\vertex[below = 16pt of r, color1, dot, minimum size=3pt, label = {below: {\footnotesize $\textcolor{color1}{3}$}}] (x4) {};
\diagram*{
	(x1) -- [color6color1color2ddntknew] (v12) -- [color6, ghost] (x2),
	(v12) -- [color1color2ghost, edge label = {\scriptsize \,$\sigma''_{j_{1}}\sigma_{j_{1}}$}, inner sep = 4pt] (b12) -- [color1, ghost, quarter left] (w12) -- [color2, ghost, quarter left] (b12),
	(b34) -- [color1, ghost, quarter left] (w34) -- [color2, ghost, quarter left] (b34) -- [color2color1ghost, edge label = {\scriptsize \,$\sigma'_{j_{2}}\sigma'_{j_{2}}$}, inner sep = 4pt] (v34),
	(x3) -- [color2, ghost] (v34) -- [color1, ghost](x4)
};
\draw [decoration={brace}, decorate] (w34dr) -- (w12dl)
node [pos=0.5, below = 1pt] {\scriptsize $\frac{1}{n_{\ell-1}\*}B_4^{(\ell)}$};
\end{feynman}
\end{tikzpicture}\nonumber\\
&& + \sum_{j_1, j_2}
\begin{tikzpicture}[baseline=(b)]
\tikzfeynmanset{every blob = {/tikz/fill=white!50, /tikz/minimum size=15pt}}
\begin{feynman}
\vertex (l) {};
\vertex[below = 16pt of l, color4, dot, minimum size=3pt, label = {below: {\footnotesize $\textcolor{color4}{0}$}}] (x1) {};
\vertex[above = 16pt of l, color6, dot, minimum size=3pt, label = {above: {\footnotesize $\textcolor{color6}{1}$}}] (x2) {};
\vertex[right = 8pt of l, dot, minimum size=0pt] (v12) {};
\vertex[right = 35pt of v12, blob] (b12) {};
\vertex[right = 35pt of b12, dot, minimum size=0pt] (w12) {};
\vertex[above = 5pt of w12, label = {above: {\scriptsize \hspace{-35pt} $z_{j_1,\lambda_{1}}$}}] (w12u) {};
\vertex[below = 12pt of w12, label = {above: {\scriptsize \hspace{-35pt} $\textcolor{color3}{0}$}}] (w12d) {};
\vertex[left = 10pt of w12d] (w12dl) {};
\tikzfeynmanset{every blob = {/tikz/fill=gray!50, /tikz/minimum size=15pt}}
\vertex[right = 3pt of w12, blob , minimum size = 6pt] (b) {};
\vertex[right = 3pt of b, dot, minimum size=0pt] (w34) {};
\tikzfeynmanset{every blob = {/tikz/fill=white!50, /tikz/minimum size=15pt}}
\vertex[right = 35pt of w34, blob] (b34) {};
\vertex[above = 5pt of w34, label = {above: {\scriptsize \hspace{30pt} $\textcolor{color2}{2}$}}] (w34u) {};
\vertex[below = 12pt of w34, label = {above: {\scriptsize \hspace{30pt} $\textcolor{color1}{3}$}}] (w34d) {};
\vertex[right = 10pt of w34d] (w34dr) {};
\vertex[right = 35pt of b34, dot, minimum size=0pt] (v34) {};
\vertex[right = 8pt of v34] (r) {};
\vertex[above = 16pt of r, color2, dot, minimum size=3pt, label = {above: {\footnotesize $\textcolor{color2}{2}$}}] (x3) {};
\vertex[below = 16pt of r, color1, dot, minimum size=3pt, label = {below: {\footnotesize $\textcolor{color1}{3}$}}] (x4) {};
\diagram*{
	(x1) -- [color6color1color2ddntknew] (v12) -- [color6, ghost] (x2),
	(v12) -- [blackcolor1color2ghost, edge label = {\scriptsize \,$\sigma'_{j_{1}}\sigma_{j_{1}}$}, inner sep = 4pt] (b12) -- [quarter left] (w12) -- [color1color2dntknew, quarter left] (b12),
	(b34) -- [color1, ghost, quarter left] (w34) -- [color2, ghost, quarter left] (b34) -- [color2color1ghost, edge label = {\scriptsize \,$\sigma'_{j_{2}}\sigma'_{j_{2}}$}, inner sep = 4pt] (v34),
	(x3) -- [color2, ghost] (v34) -- [color1, ghost](x4)
};
\draw [decoration={brace}, decorate] (w34dr) -- (w12dl)
node [pos=0.5, below = 1pt] {\scriptsize $\frac{1}{n_{\ell-1}\*}P_4^{(\ell)}$};
\end{feynman}
\end{tikzpicture}\nonumber\\
&& + \sum_{j}
\begin{tikzpicture}[baseline=(b)]
\tikzfeynmanset{every blob = {/tikz/fill=white!50, /tikz/minimum size=15pt}}
\begin{feynman}
\vertex (l) {};
\vertex[below = 16pt of l, color4, dot, minimum size=3pt, label = {below: {\footnotesize $\textcolor{color4}{0}$}}] (x1) {};
\vertex[above = 16pt of l, color6, dot, minimum size=3pt, label = {above: {\footnotesize $\textcolor{color6}{1}$}}] (x2) {};
\vertex[right = 8pt of l, dot, minimum size=0pt] (v12) {};
\vertex[right = 30pt of v12, blob] (b) {};
\vertex[right = 30pt of b, dot, minimum size=0pt] (v34) {};
\vertex[right = 8pt of v34] (r) {};
\vertex[above = 16pt of r, color2, dot, minimum size=3pt, label = {above: {\footnotesize $\textcolor{color2}{2}$}}] (x3) {};
\vertex[below = 16pt of r, color1, dot, minimum size=3pt, label = {below: {\footnotesize $\textcolor{color1}{3}$}}] (x4) {};
\diagram*{
	(x1) -- [color6color1color2ddntknew] (v12) -- [color6, ghost] (x2),
	(v12) -- [color1color2ghost, edge label = {\scriptsize \;$\sigma'''_{j}\sigma'_{j}$}, inner sep = 4pt] (b) -- [color2color1ghost, edge label = {\scriptsize \,$\sigma'_{j}\sigma'_{j}$}, inner sep = 4pt] (v34), 
	 (x3) -- [color2, ghost] (v34) -- [color1, ghost](x4)
};
\end{feynman}
\end{tikzpicture}
+ \sum_{j_1, j_2}
\begin{tikzpicture}[baseline=(b)]
\tikzfeynmanset{every blob = {/tikz/fill=white!50, /tikz/minimum size=15pt}}
\begin{feynman}
\vertex (l) {};
\vertex[below = 16pt of l, color4, dot, minimum size=3pt, label = {below: {\footnotesize $\textcolor{color4}{0}$}}] (x1) {};
\vertex[above = 16pt of l, color6, dot, minimum size=3pt, label = {above: {\footnotesize $\textcolor{color6}{1}$}}] (x2) {};
\vertex[right = 8pt of l, dot, minimum size=0pt] (v12) {};
\vertex[right = 35pt of v12, blob] (b12) {};
\vertex[right = 35pt of b12, dot, minimum size=0pt] (w12) {};
\vertex[above = 5pt of w12, label = {above: {\scriptsize \hspace{-35pt} $\textcolor{color1}{0}$}}] (w12u) {};
\vertex[below = 12pt of w12, label = {above: {\scriptsize \hspace{-35pt} $\textcolor{color2}{0}$}}] (w12d) {};
\vertex[left = 10pt of w12d] (w12dl) {};
\tikzfeynmanset{every blob = {/tikz/fill=gray!50, /tikz/minimum size=15pt}}
\vertex[right = 3pt of w12, blob , minimum size = 6pt] (b) {};
\vertex[right = 3pt of b, dot, minimum size=0pt] (w34) {};
\tikzfeynmanset{every blob = {/tikz/fill=white!50, /tikz/minimum size=15pt}}
\vertex[right = 35pt of w34, blob] (b34) {};
\vertex[above = 5pt of w34, label = {above: {\scriptsize \hspace{30pt} $\textcolor{color2}{2}$}}] (w34u) {};
\vertex[below = 12pt of w34, label = {above: {\scriptsize \hspace{30pt} $\textcolor{color1}{3}$}}] (w34d) {};
\vertex[right = 10pt of w34d] (w34dr) {};
\vertex[right = 35pt of b34, dot, minimum size=0pt] (v34) {};
\vertex[right = 8pt of v34] (r) {};
\vertex[above = 16pt of r, color2, dot, minimum size=3pt, label = {above: {\footnotesize $\textcolor{color2}{2}$}}] (x3) {};
\vertex[below = 16pt of r, color1, dot, minimum size=3pt, label = {below: {\footnotesize $\textcolor{color1}{3}$}}] (x4) {};
\diagram*{
	(x1) -- [color6color1color2ddntknew] (v12) -- [color6, ghost] (x2),
	(v12) -- [color1color2ghost, edge label = {\scriptsize \,$\sigma'''_{j_{1}}\sigma'_{j_{1}}$}, inner sep = 4pt] (b12) -- [color1, ghost, quarter left] (w12) -- [color2, ghost, quarter left] (b12),
	(b34) -- [color1, ghost, quarter left] (w34) -- [color2, ghost, quarter left] (b34) -- [color2color1ghost, edge label = {\scriptsize \,$\sigma'_{j_{2}}\sigma'_{j_{2}}$}, inner sep = 4pt] (v34),
	(x3) -- [color2, ghost] (v34) -- [color1, ghost](x4)
};
\draw [decoration={brace}, decorate] (w34dr) -- (w12dl)
node [pos=0.5, below = 1pt] {\scriptsize $\frac{1}{n_{\ell-1}\*}B_4^{(\ell)}$};
\end{feynman}
\end{tikzpicture}\nonumber\\
&& + \sum_{j_1, j_2}
\begin{tikzpicture}[baseline=(b)]
\tikzfeynmanset{every blob = {/tikz/fill=white!50, /tikz/minimum size=15pt}}
\begin{feynman}
\vertex (l) {};
\vertex[below = 16pt of l, color4, dot, minimum size=3pt, label = {below: {\footnotesize $\textcolor{color4}{0}$}}] (x1) {};
\vertex[above = 16pt of l, color6, dot, minimum size=3pt, label = {above: {\footnotesize $\textcolor{color6}{1}$}}] (x2) {};
\vertex[right = 8pt of l, dot, minimum size=0pt] (v12) {};
\vertex[right = 35pt of v12, blob] (b12) {};
\vertex[right = 35pt of b12, dot, minimum size=0pt] (w12) {};
\vertex[above = 5pt of w12, label = {above: {\scriptsize \hspace{-35pt} $z_{j_1,\lambda_{1}}$}}] (w12u) {};
\vertex[below = 12pt of w12, label = {above: {\scriptsize \hspace{-35pt} $\textcolor{color3}{0}$}}] (w12d) {};
\vertex[left = 10pt of w12d] (w12dl) {};
\tikzfeynmanset{every blob = {/tikz/fill=gray!50, /tikz/minimum size=15pt}}
\vertex[right = 3pt of w12, blob , minimum size = 6pt] (b) {};
\vertex[right = 3pt of b, dot, minimum size=0pt] (w34) {};
\tikzfeynmanset{every blob = {/tikz/fill=white!50, /tikz/minimum size=15pt}}
\vertex[right = 35pt of w34, blob] (b34) {};
\vertex[above = 5pt of w34, label = {above: {\scriptsize \hspace{30pt} $\textcolor{color2}{2}$}}] (w34u) {};
\vertex[below = 12pt of w34, label = {above: {\scriptsize \hspace{30pt} $\textcolor{color1}{3}$}}] (w34d) {};
\vertex[right = 10pt of w34d] (w34dr) {};
\vertex[right = 35pt of b34, dot, minimum size=0pt] (v34) {};
\vertex[right = 8pt of v34] (r) {};
\vertex[above = 16pt of r, color2, dot, minimum size=3pt, label = {above: {\footnotesize $\textcolor{color2}{2}$}}] (x3) {};
\vertex[below = 16pt of r, color1, dot, minimum size=3pt, label = {below: {\footnotesize $\textcolor{color1}{3}$}}] (x4) {};
\diagram*{
	(x1) -- [color6color1color2ddntknew] (v12) -- [color6, ghost] (x2),
	(v12) -- [blackcolor1color2ghost, edge label = {\scriptsize \,$\sigma''_{j_{1}}\sigma'_{j_{1}}$}, inner sep = 4pt] (b12) -- [quarter left] (w12) -- [color1color2dntknew, quarter left] (b12),
	(b34) -- [color1, ghost, quarter left] (w34) -- [color2, ghost, quarter left] (b34) -- [color2color1ghost, edge label = {\scriptsize \,$\sigma'_{j_{2}}\sigma'_{j_{2}}$}, inner sep = 4pt] (v34),
	(x3) -- [color2, ghost] (v34) -- [color1, ghost](x4)
};
\draw [decoration={brace}, decorate] (w34dr) -- (w12dl)
node [pos=0.5, below = 1pt] {\scriptsize $\frac{1}{n_{\ell-1}\*}P_4^{(\ell)}$};
\end{feynman}
\end{tikzpicture}\nonumber\\
&& + \sum_{j_1, j_2}
\begin{tikzpicture}[baseline=(b)]
\tikzfeynmanset{every blob = {/tikz/fill=white!50, /tikz/minimum size=15pt}}
\begin{feynman}
\vertex (l) {};
\vertex[below = 16pt of l, color4, dot, minimum size=3pt, label = {below: {\footnotesize $\textcolor{color4}{0}$}}] (x1) {};
\vertex[above = 16pt of l, color6, dot, minimum size=3pt, label = {above: {\footnotesize $\textcolor{color6}{1}$}}] (x2) {};
\vertex[right = 8pt of l, dot, minimum size=0pt] (v12) {};
\vertex[right = 35pt of v12, blob] (b12) {};
\vertex[right = 35pt of b12, dot, minimum size=0pt] (w12) {};
\vertex[above = 5pt of w12, label = {above: {\scriptsize \hspace{-35pt} $\textcolor{color6}{1}$}}] (w12u) {};
\vertex[below = 12pt of w12, label = {above: {\scriptsize \hspace{-35pt} $\textcolor{color4}{0}$}}] (w12d) {};
\vertex[left = 10pt of w12d] (w12dl) {};
\tikzfeynmanset{every blob = {/tikz/fill=gray!50, /tikz/minimum size=15pt}}
\vertex[right = 3pt of w12, blob , minimum size = 6pt] (b) {};
\vertex[right = 3pt of b, dot, minimum size=0pt] (w34) {};
\tikzfeynmanset{every blob = {/tikz/fill=white!50, /tikz/minimum size=15pt}}
\vertex[right = 35pt of w34, blob] (b34) {};
\vertex[above = 5pt of w34, label = {above: {\scriptsize \hspace{30pt} $\textcolor{color2}{2}$}}] (w34u) {};
\vertex[below = 12pt of w34, label = {above: {\scriptsize \hspace{30pt} $\textcolor{color1}{3}$}}] (w34d) {};
\vertex[right = 10pt of w34d] (w34dr) {};
\vertex[right = 35pt of b34, dot, minimum size=0pt] (v34) {};
\vertex[right = 8pt of v34] (r) {};
\vertex[above = 16pt of r, color2, dot, minimum size=3pt, label = {above: {\footnotesize $\textcolor{color2}{2}$}}] (x3) {};
\vertex[below = 16pt of r, color1, dot, minimum size=3pt, label = {below: {\footnotesize $\textcolor{color1}{3}$}}] (x4) {};
\diagram*{
	(x1) -- [color6color1color2ddntknew] (v12) -- [color6, ghost] (x2),
	(v12) -- [color6color6color1color2ghost, edge label = {\scriptsize \,$\sigma'_{j_{1}}\sigma'_{j_{1}}$}, inner sep = 4pt] (b12) -- [color6, ghost, quarter left] (w12) -- [color6color1color2ddntk2new, quarter left] (b12),
	(b34) -- [color1, ghost, quarter left] (w34) -- [color2, ghost, quarter left] (b34) -- [color2color1ghost, edge label = {\scriptsize \,$\sigma'_{j_{2}}\sigma'_{j_{2}}$}, inner sep = 4pt] (v34),
	(x3) -- [color2, ghost] (v34) -- [color1, ghost](x4)
};
\draw [decoration={brace}, decorate] (w34dr) -- (w12dl)
node [pos=0.5, below = 1pt] {\scriptsize $\frac{1}{n_{\ell-1}\*}R_4^{(\ell)}$};
\end{feynman}
\end{tikzpicture}
\end{eqnarray}

or, in analytical form
\begin{eqnarray}
	\frac{1}{n_{\ell}}R^{(\ell +1)}_{\textcolor{color4}{0}\textcolor{color6}{1}\textcolor{color2}{2}\textcolor{color1}{3}} &=& \frac{C_{W}^{(\ell +1)}}{n_{\ell}}\Theta^{(\ell)}_{\textcolor{color2}{0}\textcolor{color2}{2}}\Theta^{(\ell)}_{\textcolor{color1}{0}\textcolor{color1}{3}}\langle \sigma''^{(\ell)}_{\textcolor{color3}{0}}\sigma^{(\ell)}_{1}\sigma'^{(\ell)}_{\textcolor{color2}{2}}\sigma'^{(\ell)}_{\textcolor{color1}{3}}\rangle_{K^{(\ell)}} + \frac{C_{W}^{(\ell +1)}}{n_{\ell-1}}\langle \sigma''^{(\ell)}_{\textcolor{color3}{0}}\sigma^{(\ell)}_{1}\rangle_{K^{(\ell)}}\langle \sigma'^{(\ell)}_{\textcolor{color2}{2}}\sigma'^{(\ell)}_{\textcolor{color1}{3}}\rangle_{K^{(\ell)}}B^{(\ell)}_{\textcolor{color2}{0}\textcolor{color1}{0}\textcolor{color2}{2}\textcolor{color1}{3}}\nonumber\\
	&& + \frac{C_{W}^{(\ell +1)}}{n_{\ell-1}}\langle \sigma'^{(\ell)}_{\textcolor{color2}{2}}\sigma'^{(\ell)}_{\textcolor{color1}{3}}\rangle_{K^{(\ell)}}\sum_{\lambda_{1} \in \{0,1,2,3\}}\langle \frac{d (\sigma'^{(\ell)}_{\textcolor{color3}{0}}\sigma^{(\ell)}_{1})}{d z^{(\ell)}_{\lambda_{1}}}\rangle_{K^{(\ell)}}P^{(\ell)}_{\textcolor{color3}{0}\textcolor{color2}{2}\textcolor{color1}{3}\lambda_{1}} \nonumber\\
	&& + \frac{(C_{W}^{(\ell +1)})^{2}}{n_{\ell}}\Theta^{(\ell)}_{\textcolor{color6}{0}\textcolor{color6}{1}}\Theta^{(\ell)}_{\textcolor{color2}{0}\textcolor{color2}{2}}\Theta^{(\ell)}_{\textcolor{color1}{0}\textcolor{color1}{3}}\langle \sigma'''^{(\ell)}_{\textcolor{color3}{0}}\sigma^{(\ell)}_{1}\sigma'^{(\ell)}_{\textcolor{color2}{2}}\sigma'^{(\ell)}_{\textcolor{color1}{3}}\rangle_{K^{(\ell)}} \nonumber\\
        && + \frac{(C_{W}^{(\ell +1)})^{2}}{n_{\ell-1}}\Theta^{(\ell)}_{\textcolor{color6}{0}\textcolor{color6}{1}}\langle \sigma'''^{(\ell)}_{\textcolor{color3}{0}}\sigma^{(\ell)}_{1}\rangle_{K^{(\ell)}}\langle \sigma'^{(\ell)}_{\textcolor{color2}{2}}\sigma'^{(\ell)}_{\textcolor{color1}{3}}\rangle_{K^{(\ell)}}B^{(\ell)}_{\textcolor{color2}{0}\textcolor{color1}{0}\textcolor{color2}{2}\textcolor{color1}{3}}\nonumber\\
	&& + \frac{(C_{W}^{(\ell +1)})^{2}}{n_{\ell-1}}\Theta^{(\ell)}_{\textcolor{color6}{0}\textcolor{color6}{1}}\langle \sigma'^{(\ell)}_{\textcolor{color2}{2}}\sigma'^{(\ell)}_{\textcolor{color1}{3}}\rangle_{K^{(\ell)}}\sum_{\lambda_{1} \in \{0,1,2,3\}}\langle \frac{d (\sigma''^{(\ell)}_{\textcolor{color3}{0}}\sigma^{(\ell)}_{1})}{d z^{(\ell)}_{\lambda_{1}}}\rangle_{K^{(\ell)}}P^{(\ell)}_{\textcolor{color3}{0}\textcolor{color2}{2}\textcolor{color1}{3}\lambda_{1}} \nonumber\\
	&& + \frac{(C_{W}^{(\ell +1)})^{2}}{n_{\ell-1}}\langle \sigma'^{(\ell)}_{\textcolor{color4}{0}}\sigma'^{(\ell)}_{\textcolor{color6}{1}}\rangle_{K^{(\ell)}}\langle\sigma'^{(\ell)}_{\textcolor{color2}{2}}\sigma'^{(\ell)}_{\textcolor{color1}{3}}\rangle_{K^{(\ell)}}R^{(\ell)}_{\textcolor{color4}{0}\textcolor{color6}{1}\textcolor{color2}{2}\textcolor{color1}{3}}\,.
      \end{eqnarray}
      This expression agrees with the expression found in~\cite{roberts2022} using algebraic manipulations.
\end{proof}
\subsubsection{\scabbr{dd$_{\text{II}}$NTK} Mean}

\begin{proof}
Analogously, the mean of the second type of the ddNTK can be shown to have the structure \cite{roberts2022}:

\begin{align}\label{ddiintkmean}
  \mathbb{E}^{c}_{\theta} \left[ \widehat{\text{dd}_{\text{II}}\Theta}_{i_0i_1i_2i_3}^{(\ell + 1)}(\textcolor{color7}{x_1}\textcolor{color5}{x_2}\textcolor{color2}{x_3}\textcolor{color1}{x_4}) \right]
  &= \frac{1}{n_{\ell}} \left[ \delta_{i_1i_2}\delta_{i_3i_4} S_{\textcolor{color7}{1}\textcolor{color5}{2}\textcolor{color2}{3}\textcolor{color1}{4}}^{(\ell+1)} + \delta_{i_1i_3}\delta_{i_2i_4} T_{\textcolor{color7}{1}\textcolor{color2}{3}\textcolor{color5}{2}\textcolor{color1}{4}}^{(\ell+1)} + \delta_{i_1i_4}\delta_{i_2i_3} U_{\textcolor{color7}{1}\textcolor{color1}{4}\textcolor{color5}{2}\textcolor{color2}{3}}^{(\ell+1)} \right]
\end{align}

Once more, the set of Feynman rules \ref{app:feynman_rules} dictates the definition of the tensors on the r.h.s of \eqref{ddiintkmean}. For the tensor $S_{\textcolor{color7}{1}\textcolor{color5}{2}\textcolor{color2}{3}\textcolor{color1}{4}}^{(\ell+1)}$, one finds
\begin{eqnarray}
\begin{tikzpicture}[baseline=(b)]
\begin{feynman}
\vertex (l) {};
\vertex[below = 16pt of l, color7, dot, minimum size=3pt, label = {below: {\footnotesize $\textcolor{color7}{1}$}}] (x1) {};
\vertex[above = 16pt of l, color5, dot, minimum size=3pt, label = {above: {\footnotesize $\textcolor{color5}{2}$}}] (x2) {};
\vertex[right = 8pt of l, color1, dot, minimum size=0pt] (v12) {};
\vertex[right = 20pt of v12, quarticblob] (b) {};
\vertex[below = 25pt of b] {\scriptsize $\frac{1}{n_{\ell}}S^{(\ell+1)}_{\textcolor{color7}{1}\textcolor{color5}{2}\textcolor{color2}{3}\textcolor{color1}{4}}$}; 
\vertex[right = 20pt of b, dot, minimum size=0pt] (v34) {};
\vertex[right = 8pt of v34] (r) {};
\vertex[above = 16pt of r, color2, dot, minimum size=3pt, label = {above: {\footnotesize $\textcolor{color2}{3}$}}] (x3) {};
\vertex[below = 16pt of r, color1, dot, minimum size=3pt, label = {below: {\footnotesize $\textcolor{color1}{4}$}}] (x4) {};
\diagram*{
	(x1) -- [color6color2ddntknew] (b) -- [color6color1ddntknew](x2),
	(x3) -- [color2, ghost] (b) -- [color1, ghost] (x4)
};
\end{feynman}
\end{tikzpicture}
&=&  \sum_{j}
\begin{tikzpicture}[baseline=(b)]
\tikzfeynmanset{every blob = {/tikz/fill=white!50, /tikz/minimum size=15pt}}
\begin{feynman}
\vertex (l) {};
\vertex[below = 16pt of l, color7, dot, minimum size=3pt, label = {below: {\footnotesize $\textcolor{color7}{1}$}}] (x1) {};
\vertex[above = 16pt of l, color5, dot, minimum size=3pt, label = {above: {\footnotesize $\textcolor{color5}{2}$}}] (x2) {};
\vertex[right = 8pt of l, dot, minimum size=0pt] (v12) {};
\vertex[right = 30pt of v12, blob] (b) {};
\vertex[right = 30pt of b, dot, minimum size=0pt] (v34) {};
\vertex[right = 8pt of v34] (r) {};
\vertex[above = 16pt of r, color2, dot, minimum size=3pt, label = {above: {\footnotesize $\textcolor{color2}{3}$}}] (x3) {};
\vertex[below = 16pt of r, color1, dot, minimum size=3pt, label = {below: {\footnotesize $\textcolor{color1}{4}$}}] (x4) {};
\diagram*{
	(x1) -- [color6color2ddntknew] (v12) -- [color6color1ddntknew] (x2),
	(v12) -- [color1color2ghost, edge label = {\scriptsize \;$\sigma'_{j}\sigma'_{j}$}, inner sep = 4pt] (b) -- [color2color1ghost, edge label = {\scriptsize \,$\sigma'_{j}\sigma'_{j}$}, inner sep = 4pt] (v34), 
	 (x3) -- [color2, ghost] (v34) -- [color1, ghost](x4)
};
\end{feynman}
\end{tikzpicture}
 + \sum_{j_1, j_2}
\begin{tikzpicture}[baseline=(b)]
\tikzfeynmanset{every blob = {/tikz/fill=white!50, /tikz/minimum size=15pt}}
\begin{feynman}
\vertex (l) {};
\vertex[below = 16pt of l, color7, dot, minimum size=3pt, label = {below: {\footnotesize $\textcolor{color7}{1}$}}] (x1) {};
\vertex[above = 16pt of l, color5, dot, minimum size=3pt, label = {above: {\footnotesize $\textcolor{color5}{2}$}}] (x2) {};
\vertex[right = 8pt of l, dot, minimum size=0pt] (v12) {};
\vertex[right = 35pt of v12, blob] (b12) {};
\vertex[right = 35pt of b12, dot, minimum size=0pt] (w12) {};
\vertex[above = 5pt of w12, label = {above: {\scriptsize \hspace{-35pt} $\textcolor{color1}{2}$}}] (w12u) {};
\vertex[below = 12pt of w12, label = {above: {\scriptsize \hspace{-35pt} $\textcolor{color2}{1}$}}] (w12d) {};
\vertex[left = 10pt of w12d] (w12dl) {};
\tikzfeynmanset{every blob = {/tikz/fill=gray!50, /tikz/minimum size=15pt}}
\vertex[right = 3pt of w12, blob , minimum size = 6pt] (b) {};
\vertex[right = 3pt of b, dot, minimum size=0pt] (w34) {};
\tikzfeynmanset{every blob = {/tikz/fill=white!50, /tikz/minimum size=15pt}}
\vertex[right = 35pt of w34, blob] (b34) {};
\vertex[above = 5pt of w34, label = {above: {\scriptsize \hspace{30pt} $\textcolor{color2}{3}$}}] (w34u) {};
\vertex[below = 12pt of w34, label = {above: {\scriptsize \hspace{30pt} $\textcolor{color1}{4}$}}] (w34d) {};
\vertex[right = 10pt of w34d] (w34dr) {};
\vertex[right = 35pt of b34, dot, minimum size=0pt] (v34) {};
\vertex[right = 8pt of v34] (r) {};
\vertex[above = 16pt of r, color2, dot, minimum size=3pt, label = {above: {\footnotesize $\textcolor{color2}{3}$}}] (x3) {};
\vertex[below = 16pt of r, color1, dot, minimum size=3pt, label = {below: {\footnotesize $\textcolor{color1}{4}$}}] (x4) {};
\diagram*{
	(x1) -- [color6color2ddntknew] (v12) -- [color6color1ddntknew] (x2),
	(v12) -- [color1color2ghost, edge label = {\scriptsize \,$\sigma'_{j_{1}}\sigma'_{j_{1}}$}, inner sep = 4pt] (b12) -- [color1, ghost, quarter left] (w12) -- [color2, ghost, quarter left] (b12),
	(b34) -- [color1, ghost, quarter left] (w34) -- [color2, ghost, quarter left] (b34) -- [color2color1ghost, edge label = {\scriptsize \,$\sigma'_{j_{2}}\sigma'_{j_{2}}$}, inner sep = 4pt] (v34),
	(x3) -- [color2, ghost] (v34) -- [color1, ghost](x4)
};
\draw [decoration={brace}, decorate] (w34dr) -- (w12dl)
node [pos=0.5, below = 1pt] {\scriptsize $\frac{1}{n_{\ell-1}\*}B_4^{(\ell)}$};
\end{feynman}
\end{tikzpicture}\nonumber\\
&& + \sum_{j}
\begin{tikzpicture}[baseline=(b)]
\tikzfeynmanset{every blob = {/tikz/fill=white!50, /tikz/minimum size=15pt}}
\begin{feynman}
\vertex (l) {};
\vertex[below = 16pt of l, color7, dot, minimum size=3pt, label = {below: {\footnotesize $\textcolor{color7}{1}$}}] (x1) {};
\vertex[above = 16pt of l, color5, dot, minimum size=3pt, label = {above: {\footnotesize $\textcolor{color5}{2}$}}] (x2) {};
\vertex[right = 8pt of l, dot, minimum size=0pt] (v12) {};
\vertex[right = 30pt of v12, blob] (b) {};
\vertex[right = 30pt of b, dot, minimum size=0pt] (v34) {};
\vertex[right = 8pt of v34] (r) {};
\vertex[above = 16pt of r, color2, dot, minimum size=3pt, label = {above: {\footnotesize $\textcolor{color2}{3}$}}] (x3) {};
\vertex[below = 16pt of r, color1, dot, minimum size=3pt, label = {below: {\footnotesize $\textcolor{color1}{4}$}}] (x4) {};
\diagram*{
	(x1) -- [color6color2ddntknew] (v12) -- [color6color1ddntknew] (x2),
	(v12) -- [color1color2ghost, edge label = {\scriptsize \;$\sigma''_{j}\sigma''_{j}$}, inner sep = 4pt] (b) -- [color2color1ghost, edge label = {\scriptsize \,$\sigma'_{j}\sigma'_{j}$}, inner sep = 4pt] (v34), 
	 (x3) -- [color2, ghost] (v34) -- [color1, ghost](x4)
};
\end{feynman}
\end{tikzpicture}
+ \sum_{j_1, j_2}
\begin{tikzpicture}[baseline=(b)]
\tikzfeynmanset{every blob = {/tikz/fill=white!50, /tikz/minimum size=15pt}}
\begin{feynman}
\vertex (l) {};
\vertex[below = 16pt of l, color7, dot, minimum size=3pt, label = {below: {\footnotesize $\textcolor{color7}{1}$}}] (x1) {};
\vertex[above = 16pt of l, color5, dot, minimum size=3pt, label = {above: {\footnotesize $\textcolor{color5}{2}$}}] (x2) {};
\vertex[right = 8pt of l, dot, minimum size=0pt] (v12) {};
\vertex[right = 35pt of v12, blob] (b12) {};
\vertex[right = 35pt of b12, dot, minimum size=0pt] (w12) {};
\vertex[above = 5pt of w12, label = {above: {\scriptsize \hspace{-35pt} $\textcolor{color1}{2}$}}] (w12u) {};
\vertex[below = 12pt of w12, label = {above: {\scriptsize \hspace{-35pt} $\textcolor{color2}{1}$}}] (w12d) {};
\vertex[left = 10pt of w12d] (w12dl) {};
\tikzfeynmanset{every blob = {/tikz/fill=gray!50, /tikz/minimum size=15pt}}
\vertex[right = 3pt of w12, blob , minimum size = 6pt] (b) {};
\vertex[right = 3pt of b, dot, minimum size=0pt] (w34) {};
\tikzfeynmanset{every blob = {/tikz/fill=white!50, /tikz/minimum size=15pt}}
\vertex[right = 35pt of w34, blob] (b34) {};
\vertex[above = 5pt of w34, label = {above: {\scriptsize \hspace{30pt} $\textcolor{color2}{3}$}}] (w34u) {};
\vertex[below = 12pt of w34, label = {above: {\scriptsize \hspace{30pt} $\textcolor{color1}{4}$}}] (w34d) {};
\vertex[right = 10pt of w34d] (w34dr) {};
\vertex[right = 35pt of b34, dot, minimum size=0pt] (v34) {};
\vertex[right = 8pt of v34] (r) {};
\vertex[above = 16pt of r, color2, dot, minimum size=3pt, label = {above: {\footnotesize $\textcolor{color2}{3}$}}] (x3) {};
\vertex[below = 16pt of r, color1, dot, minimum size=3pt, label = {below: {\footnotesize $\textcolor{color1}{4}$}}] (x4) {};
\diagram*{
	(x1) -- [color6color2ddntknew] (v12) -- [color6color1ddntknew] (x2),
	(v12) -- [color1color2ghost, edge label = {\scriptsize \,$\sigma''_{j_{1}}\sigma''_{j_{1}}$}, inner sep = 4pt] (b12) -- [color1, ghost, quarter left] (w12) -- [color2, ghost, quarter left] (b12),
	(b34) -- [color1, ghost, quarter left] (w34) -- [color2, ghost, quarter left] (b34) -- [color2color1ghost, edge label = {\scriptsize \,$\sigma'_{j_{2}}\sigma'_{j_{2}}$}, inner sep = 4pt] (v34),
	(x3) -- [color2, ghost] (v34) -- [color1, ghost](x4)
};
\draw [decoration={brace}, decorate] (w34dr) -- (w12dl)
node [pos=0.5, below = 1pt] {\scriptsize $\frac{1}{n_{\ell-1}\*}B_4^{(\ell)}$};
\end{feynman}
\end{tikzpicture}\nonumber\\
&& + \sum_{j_1, j_2}
\begin{tikzpicture}[baseline=(b)]
\tikzfeynmanset{every blob = {/tikz/fill=white!50, /tikz/minimum size=15pt}}
\begin{feynman}
\vertex (l) {};
\vertex[below = 16pt of l, color7, dot, minimum size=3pt, label = {below: {\footnotesize $\textcolor{color7}{1}$}}] (x1) {};
\vertex[above = 16pt of l, color5, dot, minimum size=3pt, label = {above: {\footnotesize $\textcolor{color5}{2}$}}] (x2) {};
\vertex[right = 8pt of l, dot, minimum size=0pt] (v12) {};
\vertex[right = 35pt of v12, blob] (b12) {};
\vertex[right = 35pt of b12, dot, minimum size=0pt] (w12) {};
\vertex[above = 5pt of w12, label = {above: {\scriptsize \hspace{-35pt} $\textcolor{color5}{2}$}}] (w12u) {};
\vertex[below = 12pt of w12, label = {above: {\scriptsize \hspace{-35pt} $\textcolor{color7}{1}$}}] (w12d) {};
\vertex[left = 10pt of w12d] (w12dl) {};
\tikzfeynmanset{every blob = {/tikz/fill=gray!50, /tikz/minimum size=15pt}}
\vertex[right = 3pt of w12, blob , minimum size = 6pt] (b) {};
\vertex[right = 3pt of b, dot, minimum size=0pt] (w34) {};
\tikzfeynmanset{every blob = {/tikz/fill=white!50, /tikz/minimum size=15pt}}
\vertex[right = 35pt of w34, blob] (b34) {};
\vertex[above = 5pt of w34, label = {above: {\scriptsize \hspace{30pt} $\textcolor{color2}{3}$}}] (w34u) {};
\vertex[below = 12pt of w34, label = {above: {\scriptsize \hspace{30pt} $\textcolor{color1}{4}$}}] (w34d) {};
\vertex[right = 10pt of w34d] (w34dr) {};
\vertex[right = 35pt of b34, dot, minimum size=0pt] (v34) {};
\vertex[right = 8pt of v34] (r) {};
\vertex[above = 16pt of r, color2, dot, minimum size=3pt, label = {above: {\footnotesize $\textcolor{color2}{3}$}}] (x3) {};
\vertex[below = 16pt of r, color1, dot, minimum size=3pt, label = {below: {\footnotesize $\textcolor{color1}{4}$}}] (x4) {};
\diagram*{
	(x1) -- [color6color2ddntknew] (v12) -- [color6color1ddntknew] (x2),
	(v12) -- [ddntkddntkcolor1color2ghost, edge label = {\scriptsize \,$\sigma'_{j_{1}}\sigma'_{j_{1}}$}, inner sep = 4pt] (b12) -- [color6color1ddntk2new, quarter left] (w12) -- [color6color2ddntk2new, quarter left] (b12),
	(b34) -- [color1,ghost, quarter left] (w34) -- [color2, ghost, quarter left] (b34) -- [color2color1ghost, edge label = {\scriptsize \,$\sigma'_{j_{2}}\sigma'_{j_{2}}$}, inner sep = 4pt] (v34),
	(x3) -- [color2, ghost] (v34) -- [color1, ghost](x4)
};
\draw [decoration={brace}, decorate] (w34dr) -- (w12dl)
node [pos=0.5, below = 1pt] {\scriptsize $\frac{1}{n_{\ell-1}\*}S_4^{(\ell)}$};
\end{feynman}
\end{tikzpicture}
\end{eqnarray}
or, in algebraic form
\begin{eqnarray}
	\frac{1}{n_{\ell}}S^{(\ell+1)}_{\textcolor{color7}{1}\textcolor{color5}{2}\textcolor{color2}{3}\textcolor{color1}{4}} &=& \frac{C_{W}^{(\ell+1)}}{n_{\ell}} \langle \sigma'^{(\ell)}_{\textcolor{color2}{1}}\sigma'^{(\ell)}_{\textcolor{color1}{2}}\sigma'^{(\ell)}_{\textcolor{color2}{3}}\sigma'^{(\ell)}_{\textcolor{color1}{4}}\rangle_{K^{(\ell)}}\Theta^{(\ell)}_{\textcolor{color2}{1}\textcolor{color2}{3}}\Theta^{(\ell)}_{\textcolor{color1}{2}\textcolor{color1}{4}} + \frac{C_{W}^{(\ell+1)}}{n_{\ell-1}}\langle \sigma'^{(\ell)}_{\textcolor{color2}{1}}\sigma'^{(\ell)}_{\textcolor{color1}{2}}\rangle_{K^{(\ell)}}\langle \sigma'^{(\ell)}_{\textcolor{color2}{3}}\sigma'^{(\ell)}_{\textcolor{color1}{4}}\rangle_{K^{(\ell)}}B^{(\ell)}_{\textcolor{color2}{1}\textcolor{color1}{2}\textcolor{color2}{3}\textcolor{color1}{4}}\nonumber\\
	&&  + \frac{(C_{W}^{(\ell+1)})^{2}}{n_{\ell-1}}\Theta^{(\ell)}_{\textcolor{color6}{1}\textcolor{color6}{2}}\Theta^{(\ell)}_{\textcolor{color2}{1}\textcolor{color2}{3}}\Theta^{(\ell)}_{\textcolor{color1}{2}\textcolor{color1}{4}}\langle \sigma''^{(\ell)}_{\textcolor{color2}{1}}\sigma''^{(\ell)}_{\textcolor{color1}{2}}\sigma'^{(\ell)}_{\textcolor{color2}{3}}\sigma'^{(\ell)}_{\textcolor{color1}{4}}\rangle_{K^{(\ell)}}\nonumber\\
	&& + \frac{(C_{W}^{(\ell+1)})^{2}}{n_{\ell-1}}\Theta^{(\ell)}_{\textcolor{color6}{1}\textcolor{color6}{2}}\langle \sigma''^{(\ell)}_{\textcolor{color2}{1}}\sigma''^{(\ell)}_{\textcolor{color1}{2}}\rangle_{K^{(\ell)}}\langle \sigma'^{(\ell)}_{\textcolor{color2}{3}}\sigma'^{(\ell)}_{\textcolor{color1}{4}}\rangle_{K^{(\ell)}}B^{(\ell)}_{\textcolor{color2}{1}\textcolor{color1}{2}\textcolor{color2}{3}\textcolor{color1}{4}}\nonumber\\
	&& + \frac{(C_{W}^{(\ell+1)})^{2}}{n_{\ell-1}}\langle \sigma'^{(\ell)}_{\textcolor{color7}{1}}\sigma'^{(\ell)}_{\textcolor{color5}{2}}\rangle_{K^{\ell}}\langle \sigma'^{(\ell)}_{\textcolor{color2}{3}}\sigma'^{(\ell)}_{\textcolor{color1}{4}}\rangle_{K^{(\ell)}}S^{(\ell)}_{\textcolor{color7}{1}\textcolor{color5}{2}\textcolor{color2}{3}\textcolor{color1}{4}}\,.
\end{eqnarray}
This expression agrees with the expression found in~\cite{roberts2022} using algebraic manipulations.

For the tensor $T_{\textcolor{color7}{1}\textcolor{color2}{3}\textcolor{color5}{2}\textcolor{color1}{4}}^{(\ell+1)}$, one obtains

\begin{eqnarray}

\end{eqnarray}

with algebraic representation given by
\begin{eqnarray}
  \frac{1}{n_{\ell}}T^{(\ell+1)}_{\textcolor{color7}{1}\textcolor{color2}{3}\textcolor{color5}{2}\textcolor{color1}{4}} &=& \frac{1}{n_{\ell}}\langle \sigma'^{(\ell)}_{\textcolor{color6}{1}}\sigma^{(\ell)}_{3}\sigma'^{(\ell)}_{\textcolor{color6}{2}}\sigma^{(\ell)}_{4}\rangle_{K^{(\ell)}}\Theta^{(\ell)}_{\textcolor{color6}{12}} \nonumber\\
  && + \frac{1}{n_{\ell-1}}\sum_{\lambda_{3},\lambda_{4}\in\{1,2,3,4\}}\langle \frac{d (\sigma'^{(\ell)}_{\textcolor{color6}{1}}\sigma^{(\ell)}_{3})}{d z^{(\ell)}_{\lambda_{3}}}\rangle_{K^{(\ell)}}\langle \frac{d (\sigma'^{(\ell)}_{\textcolor{color6}{2}}\sigma^{(\ell)}_{4})}{d z^{(\ell)}_{\lambda_{4}}}\rangle_{K^{(\ell)}}F^{(\ell)}_{\lambda_{3}\textcolor{color6}{1}\lambda_{4}\textcolor{color6}{2}}\nonumber\\
&& + \frac{C_{W}^{(\ell+1)}}{n_{\ell}}\Theta^{(\ell)}_{\textcolor{color6}{12}}\Theta^{(\ell)}_{\textcolor{color1}{24}}\langle \sigma'^{(\ell)}_{\textcolor{color6}{1}} \sigma^{(\ell)}_{3}\sigma''^{(\ell)}_{\textcolor{color6}{2}} \sigma'^{(\ell)}_{4} \rangle_{K^{(\ell)}} \nonumber\\
&& + \frac{C_{W}^{(\ell+1)}}{n_{\ell-1}}\Theta^{(\ell)}_{\textcolor{color1}{24}}\sum_{\lambda_{3},\lambda_{4}\in\{1,2,3,4\}}\langle \frac{d (\sigma'^{(\ell)}_{\textcolor{color6}{1}}\sigma^{(\ell)}_{3})}{d z^{(\ell)}_{\lambda_{3}}}\rangle_{K^{(\ell)}}\langle \frac{d (\sigma''^{(\ell)}_{\textcolor{color6}{2}}\sigma'^{(\ell)}_{4})}{d z^{(\ell)}_{\lambda_{4}}}\rangle_{K^{(\ell)}}F^{(\ell)}_{\lambda_{3}\textcolor{color6}{1}\lambda_{4}\textcolor{color6}{2}}\nonumber\\
&& + \frac{C_{W}^{(\ell+1)}}{n_{\ell-1}}\langle \sigma'^{(\ell)}_{\textcolor{color5}{2}}\sigma'^{(\ell)}_{\textcolor{color1}{4}}\rangle_{K^{(\ell)}}\sum_{\lambda_{3} \in \{1,2,3,4\}}\langle \frac{d (\sigma'^{(\ell)}_{\textcolor{color6}{1}}\sigma'^{(\ell)}_{3})}{d z^{(\ell)}_{\lambda_{3}}}\rangle_{K^{(\ell)}}Q^{(\ell)}_{\textcolor{color5}{2}\textcolor{color1}{4}\textcolor{color6}{1}\lambda_{3}}\nonumber\\
&& + \frac{C_{W}^{(\ell+1)}}{n_{\ell}}\Theta^{(\ell)}_{\textcolor{color6}{12}}\Theta^{(\ell)}_{\textcolor{color2}{13}}\langle \sigma''^{(\ell)}_{\textcolor{color6}{1}} \sigma'^{(\ell)}_{3}\sigma'^{(\ell)}_{\textcolor{color6}{2}} \sigma^{(\ell)}_{4} \rangle_{K^{(\ell)}} \nonumber\\
&& + \frac{C_{W}^{(\ell+1)}}{n_{\ell-1}}\Theta^{(\ell)}_{\textcolor{color2}{13}}\sum_{\lambda_{3},\lambda_{4}\in\{1,2,3,4\}}\langle \frac{d (\sigma''^{(\ell)}_{\textcolor{color6}{1}}\sigma'^{(\ell)}_{3})}{d z^{(\ell)}_{\lambda_{3}}}\rangle_{K^{(\ell)}}\langle \frac{d (\sigma'^{(\ell)}_{\textcolor{color6}{2}}\sigma^{(\ell)}_{4})}{d z^{(\ell)}_{\lambda_{4}}}\rangle_{K^{(\ell)}}F^{(\ell)}_{\lambda_{4}\textcolor{color6}{2}\lambda_{3}\textcolor{color6}{1}}\nonumber\\
&& + \frac{C_{W}^{(\ell+1)}}{n_{\ell-1}}\langle \sigma'^{(\ell)}_{\textcolor{color7}{1}}\sigma'^{(\ell)}_{\textcolor{color2}{3}}\rangle_{K^{(\ell)}}\sum_{\lambda_{4} \in \{1,2,3,4\}}\langle \frac{d (\sigma'^{(\ell)}_{\textcolor{color6}{2}}\sigma'^{(\ell)}_{4})}{d z^{(\ell)}_{\lambda_{4}}}\rangle_{K^{(\ell)}}Q^{(\ell)}_{\textcolor{color7}{1}\textcolor{color2}{3}\textcolor{color6}{2}\lambda_{4}}\nonumber\\
&& + \frac{(C_{W}^{\ell+1})^{2}}{n_{\ell}}\Theta^{(\ell)}_{\textcolor{color2}{13}}\Theta^{(\ell)}_{\textcolor{color1}{24}}\Theta^{(\ell)}_{\textcolor{color6}{12}}\langle \sigma''^{(\ell)}_{\textcolor{color6}{1}}\sigma'^{(\ell)}_{3}\sigma''^{(\ell)}_{\textcolor{color6}{2}}\sigma'^{(\ell)}_{4}\rangle_{K^{(\ell)}}\nonumber\\
&& + \frac{(C_{W}^{\ell+1})^{2}}{n_{\ell-1}}\Theta^{(\ell)}_{\textcolor{color2}{13}}\Theta^{(\ell)}_{\textcolor{color1}{24}}\sum_{\lambda_{3},\lambda_{4} \in \{1,2,3,4\}}\langle\frac{d(\sigma''^{(\ell)}_{\textcolor{color6}{1}}\sigma'^{(\ell)}_{3})}{d z^{(\ell)}_{\lambda_{3}}}\rangle_{K^{(\ell)}}\langle\frac{d(\sigma''^{(\ell)}_{\textcolor{color6}{2}}\sigma'^{(\ell)}_{4})}{d z^{(\ell)}_{\lambda_{4}}}\rangle_{K^{(\ell)}}F^{(\ell)}_{\lambda_{3}\textcolor{color6}{1}\lambda_{4}\textcolor{color6}{2}}\nonumber\\
&&  + \frac{(C_{W}^{\ell+1})^{2}}{n_{\ell-1}}\Theta^{(\ell)}_{\textcolor{color2}{13}}\langle \sigma'^{(\ell)}_{\textcolor{color5}{2}}\sigma'^{(\ell)}_{\textcolor{color1}{4}}\rangle_{K^{(\ell)}}\sum_{\lambda_{3}\in\{1,2,3,4\}}\langle \frac{d(\sigma''^{(\ell)}_{\textcolor{color6}{1}}\sigma'^{(\ell)}_{3})}{d z^{(\ell)}_{\lambda_{3}}}\rangle_{K^{(\ell)}}Q^{(\ell)}_{\textcolor{color5}{2}\textcolor{color1}{4}\textcolor{color6}{1}\lambda_{3}} \nonumber\\
&& + \frac{(C_{W}^{\ell+1})^{2}}{n_{\ell-1}}\Theta^{(\ell)}_{\textcolor{color1}{24}}\langle \sigma'^{(\ell)}_{\textcolor{color7}{1}}\sigma'^{(\ell)}_{\textcolor{color2}{3}}\rangle_{K^{(\ell)}}\sum_{\lambda_{4}\in\{1,2,3,4\}}\langle \frac{d(\sigma''^{(\ell)}_{\textcolor{color6}{2}}\sigma'^{(\ell)}_{4})}{d z^{(\ell)}_{\lambda_{4}}}\rangle_{K^{(\ell)}}Q^{(\ell)}_{\textcolor{color7}{1}\textcolor{color2}{3}\textcolor{color6}{2}\lambda_{4}} \nonumber\\
&& + \frac{(C_{w}^{(\ell+1)})^{2}}{n_{\ell-1}}\langle \sigma'^{(\ell)}_{\textcolor{color7}{1}}\sigma'^{(\ell)}_{\textcolor{color2}{3}}\rangle_{K^{(\ell)}}\langle \sigma'^{(\ell)}_{\textcolor{color5}{2}}\sigma'^{(\ell)}_{\textcolor{color1}{4}}\rangle_{K^{(\ell)}}T^{(\ell)}_{\textcolor{color7}{1}\textcolor{color2}{3}\textcolor{color5}{2}\textcolor{color1}{4}}\,.
\end{eqnarray}
This expression agrees with the expression found in~\cite{roberts2022} using algebraic manipulations.

Finally, for the tensor $U_{\textcolor{color7}{1}\textcolor{color1}{4}\textcolor{color5}{2}\textcolor{color2}{3}}^{(\ell+1)}$, one learns

\begin{eqnarray}
\begin{tikzpicture}[baseline=(b)]
\begin{feynman}
\vertex (l) {};
\vertex[below = 16pt of l, color7, dot, minimum size=3pt, label = {below: {\footnotesize $\textcolor{color7}{1}$}}] (x1) {};
\vertex[above = 16pt of l, color1, dot, minimum size=3pt, label = {above: {\footnotesize $\textcolor{color1}{4}$}}] (x2) {};
\vertex[right = 8pt of l, dot, minimum size=0pt] (v12) {};
\vertex[right = 20pt of v12, quarticblob] (b) {};
\vertex[below = 25pt of b] {\scriptsize $\frac{1}{n_{\ell}}U^{(\ell+1)}_{\textcolor{color7}{1}\textcolor{color1}{4}\textcolor{color5}{2}\textcolor{color2}{3}}$}; 
\vertex[right = 20pt of b, dot, minimum size=0pt] (v34) {};
\vertex[right = 8pt of v34] (r) {};
\vertex[above = 16pt of r, color5, dot, minimum size=3pt, label = {above: {\footnotesize $\textcolor{color5}{2}$}}] (x3) {};
\vertex[below = 16pt of r, color2, dot, minimum size=3pt, label = {below: {\footnotesize $\textcolor{color2}{3}$}}] (x4) {};
\diagram*{
	(x1) -- [color6color2ddntk2new] (b) -- [color1, ghost](x2),
	(x3) -- [color6color1ddntk2new] (b) -- [color2, ghost] (x4)
};
\end{feynman}
\end{tikzpicture} &=& \sum_{j}
\begin{tikzpicture}[baseline=(b)]
\tikzfeynmanset{every blob = {/tikz/fill=white!50, /tikz/minimum size=15pt}}
\begin{feynman}
\vertex (l) {};
\vertex[below = 16pt of l, color7, dot, minimum size=3pt, label = {below: {\footnotesize $\textcolor{color7}{1}$}}] (x1) {};
\vertex[above = 16pt of l, color1, dot, minimum size=3pt, label = {above: {\footnotesize $\textcolor{color1}{4}$}}] (x2) {};
\vertex[right = 8pt of l, dot, minimum size=0pt] (v12) {};
\vertex[right = 30pt of v12, blob] (b) {};
\vertex[right = 30pt of b, dot, minimum size=0pt] (v34) {};
\vertex[right = 8pt of v34] (r) {};
\vertex[above = 16pt of r, color5, dot, minimum size=3pt, label = {above: {\footnotesize $\textcolor{color5}{2}$}}] (x3) {};
\vertex[below = 16pt of r, color2, dot, minimum size=3pt, label = {below: {\footnotesize $\textcolor{color2}{3}$}}] (x4) {};
\diagram*{
	(x1) -- [color6color2ddntk2new] (v12) -- [color1, ghost] (x2),
	(v12) -- [ntkdntkblackcolor1color2color6, edge label = {\scriptsize \;$\sigma''_{j}\sigma'_{j}$}, inner sep = 4pt] (b) -- [ntkdntkcolor6color1color2black, edge label = {\scriptsize \,$\sigma''_{j}\sigma'_{j}$}, inner sep = 4pt] (v34), 
	 (x3) -- [color6color1ddntk2new] (v34) -- [color2, ghost] (x4)
};
\end{feynman}
\end{tikzpicture}\nonumber\\
&& + \sum_{j_1, j_2}
\begin{tikzpicture}[baseline=(b)]
\tikzfeynmanset{every blob = {/tikz/fill=white!50, /tikz/minimum size=15pt}}
\begin{feynman}
\vertex (l) {};
\vertex[below = 16pt of l, color7, dot, minimum size=3pt, label = {below: {\footnotesize $\textcolor{color7}{1}$}}] (x1) {};
\vertex[above = 16pt of l, color1, dot, minimum size=3pt, label = {above: {\footnotesize $\textcolor{color1}{4}$}}] (x2) {};
\vertex[right = 8pt of l, dot, minimum size=0pt] (v12) {};
\vertex[right = 35pt of v12, blob] (b12) {};
\vertex[right = 35pt of b12, dot, minimum size=0pt] (w12) {};
\vertex[above = 5pt of w12, label = {above: {\scriptsize \hspace{-35pt} $\textcolor{color1}{4}$}}] (w12u) {};
\vertex[below = 12pt of w12, label = {above: {\scriptsize \hspace{-35pt} $\textcolor{color7}{1}$}}] (w12d) {};
\vertex[left = 10pt of w12d] (w12dl) {};
\tikzfeynmanset{every blob = {/tikz/fill=gray!50, /tikz/minimum size=15pt}}
\vertex[right = 3pt of w12, blob , minimum size = 6pt] (b) {};
\vertex[right = 3pt of b, dot, minimum size=0pt] (w34) {};
\tikzfeynmanset{every blob = {/tikz/fill=white!50, /tikz/minimum size=15pt}}
\vertex[right = 35pt of w34, blob] (b34) {};
\vertex[above = 5pt of w34, label = {above: {\scriptsize \hspace{30pt} $\textcolor{color5}{2}$}}] (w34u) {};
\vertex[below = 12pt of w34, label = {above: {\scriptsize \hspace{30pt} $\textcolor{color2}{3}$}}] (w34d) {};
\vertex[right = 10pt of w34d] (w34dr) {};
\vertex[right = 35pt of b34, dot, minimum size=0pt] (v34) {};
\vertex[right = 8pt of v34] (r) {};
\vertex[above = 16pt of r, color5, dot, minimum size=3pt, label = {above: {\footnotesize $\textcolor{color5}{2}$}}] (x3) {};
\vertex[below = 16pt of r, color2, dot, minimum size=3pt, label = {below: {\footnotesize $\textcolor{color2}{3}$}}] (x4) {};
\diagram*{
	(x1) -- [color6color2ddntk2new] (v12) -- [color1, ghost] (x2),
	(v12) -- [ntkdntkcolor1color2color6, edge label = {\scriptsize \,$\sigma'_{j_{2}}\sigma'_{j_{2}}$}, inner sep = 4pt] (b12) -- [color1, ghost, quarter left] (w12) -- [color6color2ddntknew, quarter left] (b12),
	(b34) -- [color2, ghost, quarter left] (w34) -- [color6color1ddntknew, quarter left] (b34) -- [ntkdntkcolor6color1color2, edge label = {\scriptsize \,$\sigma'_{j_{2}}\sigma'_{j_{2}}$}, inner sep = 4pt] (v34),
	(x3) -- [color6color1ddntk2new, ghost] (v34) -- [color2, ghost] (x4)
};
\draw [decoration={brace}, decorate] (w34dr) -- (w12dl)
node [pos=0.5, below = 1pt] {\scriptsize $\frac{1}{n_{\ell-1}\*}U_4^{(\ell)}$};
\end{feynman}
\end{tikzpicture}
\end{eqnarray}
whose analytical expression reads
\begin{eqnarray}
	\frac{1}{n_{\ell}}U^{(\ell+1)}_{\textcolor{color7}{1}\textcolor{color1}{4}\textcolor{color5}{2}\textcolor{color2}{3}} &=& \frac{(C_{W}^{(\ell+1)})^{2}}{n_{\ell}}\Theta^{(\ell)}_{\textcolor{color6}{12}}\Theta^{(\ell)}_{\textcolor{color1}{24}}\Theta^{(\ell)}_{\textcolor{color2}{13}}\langle \sigma''^{(\ell)}_{\textcolor{color7}{1}}\sigma'^{(\ell)}_{\textcolor{color1}{4}}\sigma''^{(\ell)}_{\textcolor{color5}{2}}\sigma'^{(\ell)}_{\textcolor{color2}{3}}\rangle_{K^{(\ell)}} \nonumber\\
    && + \frac{(C_{W}^{(\ell+1)})^{2}}{n_{\ell-1}}\langle \sigma'^{(\ell)}_{\textcolor{color7}{1}}\sigma'^{(\ell)}_{\textcolor{color1}{4}}\rangle_{K^{(\ell)}}\langle \sigma'^{(\ell)}_{\textcolor{color5}{2}}\sigma'^{(\ell)}_{\textcolor{color2}{3}}\rangle_{K^{(\ell)}}U^{(\ell)}_{\textcolor{color7}{1}\textcolor{color1}{4}\textcolor{color5}{2}\textcolor{color2}{3}}\,.
  \end{eqnarray}
  This expression agrees with the expression found in~\cite{roberts2022} using algebraic manipulations.
\end{proof}

\subsection{Proof of Theorem \ref{theoremthree}}\label{proofoftheoremthree}
\thirdtheorem*
\begin{proof}

Three key facts are essential for the validity of the statement above:
\begin{enumerate}
	\item The number of external lines is necessarily even. This follows directly from the definition of the NTK and its descendants, together with the Gaussian nature of the parameter distributions.
	
	\item Any expectation value at layer $(\ell+1)$ is generated solely by cubic vertices involving one pair of external lines and a single internal line. This becomes evident upon analyzing the neural index structure resulting from Wick contractions of the parameters inside the expectation value under consideration. For instance, an expectation value involving $2N$ preactivations produces a rank-$2N$ tensor accompanied by $N$ Kronecker deltas carrying $2N$ external neural indices. These Kronecker deltas correspond precisely to the pairs of external lines attached to the cubic vertices described above.
	
    \item The presence of internal lines induce the emergence of higher-order tensors at the preceding layer $\ell$, which constitute natural generalizations of the quartic vertices introduced in Appendix \ref{app:feynman_rules}. This fact arises from the $\frac{1}{n}$-expansion of the preactivation probability distribution in conjunction with the definitions of the tensors in question. For instance, at order $\frac{1}{n^{2}}$, the quartic vertex $V_{4}^{(\ell+1)} = V^{(\ell+1)}$ receives contributions from the sextic vertex at the preceding layer $V_{6}^{(\ell)}$:
	\begin{eqnarray}
\begin{tikzpicture}[baseline=(b)]
\begin{feynman}
\vertex (l) {};
\vertex[below = 16pt of l, dot, minimum size=3pt, label = {below: {\footnotesize $1$}}] (x1) {};
\vertex[above = 16pt of l, dot, minimum size=3pt, label = {above: {\footnotesize $2$}}] (x2) {};
\vertex[right = 8pt of l, dot, minimum size=0pt] (v12) {};
\vertex[right = 20pt of v12, quarticblob] (b) {};
\vertex[below = 25pt of b] {\scriptsize $\frac{1}{n_{\ell}}V^{(\ell +1)}_{4}$}; 
\vertex[right = 20pt of b, dot, minimum size=0pt] (v34) {};
\vertex[right = 8pt of v34] (r) {};
\vertex[above = 16pt of r, dot, minimum size=3pt, label = {above: {\footnotesize $3$}}] (x3) {};
\vertex[below = 16pt of r, dot, minimum size=3pt, label = {below: {\footnotesize $4$}}] (x4) {};
\diagram*{
	(x1) -- (b) -- (x2), 
	 (x3) -- (b) -- (x4)
};
\end{feynman}
\end{tikzpicture}
&=& 
\sum_{j_1, j_2}
\begin{tikzpicture}[baseline=(b)]
\tikzfeynmanset{every blob = {/tikz/fill=white!50, /tikz/minimum size=15pt}}
\begin{feynman}
\vertex (l) {};
\vertex[below = 16pt of l, dot, minimum size=3pt, label = {below: {\footnotesize $1$}}] (x1) {};
\vertex[above = 16pt of l, dot, minimum size=3pt, label = {above: {\footnotesize $2$}}] (x2) {};
\vertex[right = 8pt of l, dot, minimum size=0pt] (v12) {};
\vertex[right = 30pt of v12, blob] (b12) {};
\vertex[right = 16pt of b12, dot, minimum size=0pt] (w12) {};
\vertex[above = 1pt of w12, label = {above: {\scriptsize \hspace{-10pt} $z_{j_1}$}}] (w12u) {};
\vertex[below = 12pt of w12] (w12d) {};
\vertex[left = 10pt of w12d] (w12dl) {};
\tikzfeynmanset{every blob = {/tikz/fill=gray!50, /tikz/minimum size=15pt}}
\vertex[right = 3pt of w12, blob , minimum size = 6pt] (b) {};
\vertex[right = 3pt of b, dot, minimum size=0pt] (w34) {};
\tikzfeynmanset{every blob = {/tikz/fill=white!50, /tikz/minimum size=15pt}}
\vertex[right = 16pt of w34, blob] (b34) {};
\vertex[above = 1pt of w34, label = {above: {\scriptsize \hspace{10pt} $z_{j_2}$}}] (w34u) {};
\vertex[below = 12pt of w34] (w34d) {};
\vertex[right = 10pt of w34d] (w34dr) {};
\vertex[right = 30pt of b34, dot, minimum size=0pt] (v34) {};
\vertex[right = 8pt of v34] (r) {};
\vertex[above = 16pt of r, dot, minimum size=3pt, label = {above: {\footnotesize $3$}}] (x3) {};
\vertex[below = 16pt of r, dot, minimum size=3pt, label = {below: {\footnotesize $4$}}] (x4) {};
\diagram*{
	(x1) -- (v12) -- (x2),
	(v12) -- [photon, edge label = {\scriptsize \;$\widehat{\Delta G}_{j_1}$}, inner sep = 4pt] (b12) -- (w12) -- (b12) -- [quarter left] (w12) -- [quarter left] (b12),
	(b34) -- (w34) -- (b34) -- [quarter left] (w34) -- [quarter left] (b34) -- [photon, edge label = {\scriptsize \,$\widehat{\Delta G}_{j_2}$}, inner sep = 4pt] (v34),
	(x3) -- (v34) -- (x4)
};
\draw [decoration={brace}, decorate] (w34dr) -- (w12dl)
node [pos=0.5, below = 1pt] {\scriptsize $\frac{1}{n^{2}_{\ell-1}\*}V_6^{(\ell)}$};
\end{feynman} 
\end{tikzpicture}
+ \ldots
\end{eqnarray}
where, as anticipated, $V_{6}^{(\ell)}$ corresponds to the joint cumulant of six preactivations evaluated at layer $\ell$, while the ellipsis denotes terms involving $V_{4}^{(\ell)}$ and $K^{\{1\}(\ell)}$.

Similarly, at the same order, the NTK tensor $F_{4}^{(l+1)} = F^{(l+1)}$ receives corrections from its higher-order analogue $F_{6}^{(l+1)}$
\begin{align}
  \begin{tikzpicture}[baseline=(b)]
      \begin{feynman}
        \vertex (l) {};
        \vertex[below = 15pt of l, dot, minimum size=3pt, label = {left: {\footnotesize $1$}}] (x1) {};
        \vertex[above = 15pt of l, dot, color1, minimum size=3pt, label = {left: {\footnotesize $\textcolor{color1}{3}$}}] (x2) {};
        \vertex[right = 20pt of l, quarticblob] (b) {};
        \vertex[below = 25pt of b] {\scriptsize $\frac{1}{n_{\ell}}F^{(\ell+1)}_{4}$}; 
        \vertex[right = 20pt of b] (r) {};
        \vertex[above = 15pt of r, dot, minimum size=3pt, label = {right: {\footnotesize $2$}}] (x3) {};
        \vertex[below = 15pt of r, dot, color1, minimum size=3pt, label = {right: {\footnotesize $\textcolor{color1}{4}$}}] (x4) {};
        \diagram*{
          (x1) -- (b) -- [color1, ghost] (x2), 
          (x3) -- (b) -- [color1, ghost] (x4)
        };
      \end{feynman}
    \end{tikzpicture}
&=& \sum_{j_1, j_2}\!
\begin{tikzpicture}[baseline=(b)]
  \tikzfeynmanset{every blob = {/tikz/fill=white!50, /tikz/minimum size=15pt}}
  \begin{feynman}
    \vertex (l) {};
    \vertex[below = 15pt of l, dot, minimum size=3pt, label = {left: {\footnotesize $1$}}] (x1) {};
    \vertex[above = 15pt of l, color1, dot, minimum size=3pt, label = {left: {\footnotesize $\textcolor{color1}{3}$}}] (x2) {};
    \vertex[right = 8pt of l, dot, minimum size=0pt] (v12) {};
    \vertex[right = 35pt of v12, blob] (b12) {};
    \vertex[right = 16pt of b12, dot, minimum size=0pt] (w12) {};
    \vertex[above = 1pt of w12, label = {above: {\scriptsize \hspace{-10pt} $z_{j_1}$}}] (w12u) {};
    \vertex[below = 8pt of w12] (w12d) {};
    \vertex[left = 10pt of w12d] (w12dl) {};
    \tikzfeynmanset{every blob = {/tikz/fill=gray!50, /tikz/minimum size=15pt}}
    \vertex[right = 3pt of w12, blob , minimum size = 6pt] (b) {};
    \vertex[right = 3pt of b, dot, minimum size=0pt] (w34) {};
    \tikzfeynmanset{every blob = {/tikz/fill=white!50, /tikz/minimum size=15pt}}
    \vertex[right = 16pt of w34, blob] (b34) {};
    \vertex[above = 1pt of w34, label = {above: {\scriptsize \hspace{10pt} $z_{j_2}$}}] (w34u) {};
    \vertex[below = 8pt of w34] (w34d) {};
    \vertex[right = 10pt of w34d] (w34dr) {};
    \vertex[right = 35pt of b34, dot, minimum size=0pt] (v34) {};
    \vertex[right = 8pt of v34] (r) {};
    \vertex[above = 15pt of r, dot, minimum size=3pt, label = {right: {\footnotesize $2$}}] (x3) {};
    \vertex[below = 15pt of r, color1, dot, minimum size=3pt, label = {right: {\footnotesize $\textcolor{color1}{4}$}}] (x4) {};
    \diagram*{
      (x1) -- (v12) -- [color1, ghost] (x2),
      (v12) -- [color1blackghost, edge label = {\scriptsize \;$\sigma_{j_{1}}\sigma'_{j_{1}}$}, inner sep = 4pt] (b12)  -- (w12) -- (b12)-- [color1, ghost, quarter left] (w12) -- [quarter left] (b12),
      (b34) -- (w34) -- (b34) -- [color1, ghost, quarter left] (w34) -- [quarter left] (b34) -- [blackcolor1ghost, edge label = {\scriptsize \,$\sigma_{j_{2}}\sigma'_{j_{2}}$}, inner sep = 4pt] (v34),
      (x3) --  (v34) -- [color1, ghost] (x4)
    };
    \draw [decoration={brace}, decorate] (w34dr) -- (w12dl) node [pos=0.5, below = 1pt] {\scriptsize $\frac{1}{n^{2}_{\ell-1}\*}F_6^{(\ell)}$};
  \end{feynman}
\end{tikzpicture} + \ldots
\end{align}
where $F_{6}^{(\ell)}$ denotes the join cumulant of four preactivations and one NTK at layer $\ell$, see our example below for details, and the ellipsis represents contributions containing $F^{(\ell)}_{4}$ and $D^{(\ell)}_{4} = D^{(\ell)}$.

In general, at higher order in $\frac{1}{n}$, tensors involving novel combinations and orderings of preactivation, NTK, dNTK and ddNTK lines attached to a solid blob will appear. However, these just need to be enumerated and do not come with new Feynman rules. In order to solve the recursions, they have to be added to the list of tensors to keep track of from layer to layer.
\end{enumerate}
Having established these facts, the final step in proving this theorem is to demonstrate the uniqueness of the Feynman rules for the cubic vertices defined in Appendix~\ref{app:feynman_rules} at all orders in $\tfrac{1}{n}$. This uniqueness is a direct consequence of the definitions of the NTK, dNTK, and ddNTK tensors in terms of joint cumulants. To make this more explicit, consider the tensor $V_{4}^{(\ell)}$
\begin{align}
\frac{1}{n_{\ell}}V_{1234}^{(\ell+1)} &= \left(C_{W}^{(\ell+1)}\right)^{2}\mathbb{E}[\widehat{\Delta K}_{12}^{(\ell+1)}\widehat{\Delta K}_{34}^{(\ell+1)}]\label{eq:7}
\end{align}
where $\widehat{\Delta K}^{(\ell + 1)}_{12} = \frac{1}{n_{\ell}}\sum_{i=1}^{n_{\ell}}\left(\sigma^{(\ell)}_{i,1}\sigma^{(\ell)}_{i,2} - \mathbb{E}[\sigma^{(\ell)}_{i,1}\sigma^{(\ell)}_{i,2}]\right)$. Expanding out the expectation value $\mathbb{E}[\sigma^{(\ell)}_{i,1}\sigma^{(\ell)}_{i,2}]$ will result in a contribution of the form $\widehat{\Delta G}^{(\ell + 1)}_{i,12} = \sigma^{(\ell)}_{i,1}\sigma^{(\ell)}_{i,2} - \langle\sigma^{(\ell)}_{i,1}\sigma^{(\ell)}_{i,2}\rangle_{K^{(\ell)}}$ plus higher-order corrections which are all proportional to factors of $V_{4}$ or its higher-order generalizations $V_{6}$, $V_{8}$ etc. Therefore, the only cubic vertex from which this joint cumulant is built takes the form
  \begin{align}
    \begin{tikzpicture}[baseline=(b)]
      \begin{feynman}
        \vertex (l) {};
        \vertex[below = 15pt of l, dot, minimum size=3pt, label = {left: {\footnotesize $1$}}] (x1) {};
        \vertex[above = 15pt of l, dot, minimum size=3pt, label = {left: {\footnotesize $2$}}] (x2) {};
        \vertex[right = 10pt of l, dot, minimum size=0pt] (v12) {};
        \vertex[right = 30pt of v12, dot, minimum size=0pt] (b) {};
        \diagram*{
          (x1) --  (v12) --  (x2),
          (v12) -- [photon, edge label = {\scriptsize \;$\widehat{\Delta G}^{(\ell)}_{i,12}$}, inner sep = 4pt] (b) 
        };
      \end{feynman}
    \end{tikzpicture} \sim \frac{C^{(\ell+1)}_{W}}{n_{\ell}}
  \end{align}
  which is exactly the vertex from item $(ii)$ of Appendix~\ref{app:feynman_rules}. This means that the cubic vertex characterizing two NTK lines and one preactivation does not change at higher order in $\frac{1}{n}$. Therefore, one can replace $\widehat{\Delta K}_{i,12}^{(\ell+1)} \rightarrow \widehat{\Delta G}_{i,12}^{(\ell+1)}$ when constructing the Feynman diagrams for~\eqref{eq:7}.
  

A similar line of reasoning applies to the NTK tensors. Consider, for instance, the tensor $D_{4}^{(\ell)}$ \cite{roberts2022}
\begin{align}
\frac{1}{n_{\ell}} D_{12\textcolor{color1}{3}\textcolor{color1}{4}}^{(\ell+1)} &= 
C_{W}^{(\ell+1)}\mathbb{E} \Big[ \widehat{\Delta K}^{(\ell+1)}_{12}  \widehat{\Delta \omega}^{(\ell+1)}_{34} \Big] 
+ \left(C_{W}^{(\ell+1)}\right)^{2} 
\left( \frac{1}{n_{\ell}} \sum_{j=1}^{n} \mathbb{E} \Big[ \widehat{\Delta K}^{(\ell+1)}_{12} \sigma'^{(\ell)}_{j,\textcolor{color1}{3}} \sigma'^{(\ell)}_{j,\textcolor{color1}{4}}\widehat{\Delta \Theta}^{(\ell)}_{jj,\textcolor{color1}{3}\textcolor{color1}{4}}\Big] \right) \label{defdtensor}
\end{align}
with $\widehat{\Delta \omega}^{(\ell + 1)}_{34} = \widehat{\Delta K}^{(\ell+1)}_{34} + C_{W}^{(\ell+1)}\Theta^{(\ell)}_{34}\left(\sigma'^{(\ell)}_{3}\sigma'^{(\ell)}_{4} - \mathbb{E}[\sigma'^{(\ell)}_{3}\sigma'^{(\ell)}_{4}]\right)$. In this case, there are two cubic vertices involving a pair of NTK lines $\textcolor{color1}{34}$, which, as above, correspond to the Gaussian analogues of their full expectation values.
%
Specifically, two internal lines may arise in these cubic vertices: the first corresponds to the Gaussian analogue of $\widehat{\Delta \omega}_{i,34}^{(\ell+1)}$, namely $\widehat{\Delta \Omega}_{i,34}^{(\ell+1)} = \widehat{\Omega}_{i,34}^{(\ell+1)} - \langle \widehat{\Omega}_{i,34}^{(\ell+1)} \rangle_{K^{(\ell)}}$; the second is given by the product $\sigma'^{(\ell)}_{i,\textcolor{color1}{3}}\sigma'^{(\ell)}_{i,\textcolor{color1}{4}}$. These are exactly the two vertices with two external dotted lines of the same color in~\eqref{feynmanrulescubic}. Once again, no new tensor structures arise beyond those generated by the $\frac{1}{n}$-expansion of the full probability distribution, which, as discussed in the third point above, are fully accounted for by the higher-order generalizations of the tensors examined in the present work. Therefore, the cubic vertices involving two NTK lines remain invariant to all orders in $\frac{1}{n}$.

The same reasoning extends to the remaining tensors, thereby establishing the uniqueness of the Feynman rules presented in Appendix~\ref{app:feynman_rules}.
\end{proof}
\endgroup

\section{$D$ Tensor at Order $1/n^{2}$ from Feynman Diagrams}
\label{app:example_NNLO_tensor}
\begingroup
\allowdisplaybreaks

In this Appendix we display all the possible diagrams that can be built out of the Feynman rules defined in \ref{app:feynman_rules}, for the specific case of the tensor D in the expectation value
 \begin{align}\label{higher-order-D-F}
     \mathbb{E}^{c}_{\theta}[z^{(\ell+1)}_{i_{1},1},&\,z^{(\ell+1)}_{i_{2},2},z^{(\ell+1)}_{i_{3},3},z^{(\ell+1)}_{i_{4},4}, \widehat{\Delta \Theta}^{(\ell+1)}_{i_{5}i_{6},\textcolor{color1}{5}\textcolor{color1}{6}}] \nonumber\\ 
      &{=} \frac{1}{n_{\ell}^{2}}\!\bigg[\delta_{i_{1}i_{2}}\delta_{i_{3}i_{4}}\delta_{i_{5}i_{6}}D^{(\ell+1)}_{1234\textcolor{color1}{56}} {+} \delta_{i_{1}i_{2}}\delta_{i_{3}i_{5}}\delta_{i_{4}i_{6}}F^{(\ell+1)}_{123\textcolor{color1}{5}4\textcolor{color1}{6}} {+} \ldots\!\bigg]
\end{align}

The diagrams describing the tensor $D$ at layer $\ell + 1$ are then given by 
\begin{align}
\frac{1}{n_{\ell}}& D^{(\ell + 1)}_{1234\textcolor{color1}{56}} =    \sum_j

\end{align}
\endgroup
where $P(a_{1}a_{2},a_{3}a_{4},\ldots,a_{2k-1}a_{2k})$ denotes the complement of the set of permutations generated by the pairs $(a_{1}a_{2}),(a_{3}a_{4}),\ldots, (a_{2k-1}a_{2k})$, e.g. $P(12,34) = \{(34,12)\}$.

We have checked that these diagrams sucessfully reproduce the formulae obtained from perturbation theory and functional integration when applied to the study of the statistics of the NTK at order $\frac{1}{n^{2}}$. 

\section{NTK Mean at Order $1/n$}
\label{app:NLO_ntk_mean}
As discussed in Section~\ref{sec-finitewidth-corrections-NTK}, the Feynman rules introduced in the main text constrain the set of diagrams that can be constructed for two external NTK lines at order $\frac{1}{n}$ to those shown in~\eqref{firstcorrectionntk}. 
Next, we provide the algebraic formulae associated to each diagram in such expansion:
\begingroup
\allowdisplaybreaks
\begin{align}
\begin{tikzpicture}[baseline=(b)]
    \begin{feynman}
      \vertex[dot, color1, minimum size=3pt, label = {below: {\footnotesize $\textcolor{color1}{1}$}}] (x1) {};
      \vertex[right = 20pt of x1] (v);
      \vertex[right = 20pt of v, dot, color1, minimum size = 3pt, label = {below: {\footnotesize $\textcolor{color1}{2}$}}] (x2) {};
      \vertex[above = 9pt of v] (b) {};
      \vertex[above = 4pt of v, label = {right: \!\*{\scriptsize $\sigma'_{j}\sigma'_{j}$}}] (j){};
      \tikzfeynmanset{every blob = {/tikz/fill=white!50, /tikz/minimum size=15pt}}
      \vertex[above = 16pt of v, blob] (G) {};
      \vertex[above = 22pt of v, dot, minimum size = 0pt] (g) {};
      \vertex[above = 20pt of g, dot, minimum size = 0pt] (w) {};
      \diagram*{
        (x1) -- [color1, ghost] (v) -- [color1, ghost] (x2),
        (v) -- [color1, doubghost] (G) -- (g) -- [color1, ghost, half left] (w) -- [color1, ghost, half left] (g)
      };
      \vertex[above = 0pt of G, blob] (K0){};
      \vertex[above = 20pt of g, smallblob] (K1) {};
      \vertex[above = 10pt of w] (K1) {{\scriptsize $\frac{1}{n_{\ell-1}} \Theta^{\{1\}(\ell)}$}};
      \vertex[right = 15pt of G, label = {}] (Gr) {};
    \end{feynman}
  \end{tikzpicture}
  &= \frac{C_{W}^{(\ell+1)}}{n_{\ell-1}} \Theta^{\{1\}(\ell)}_{\textcolor{color1}{12}}\langle \sigma'^{(\ell)}_{\textcolor{color1}{1}}\sigma'^{(\ell)}_{\textcolor{color1}{2}}\rangle_{K^{(\ell)}}\nonumber\\
  \begin{tikzpicture}[baseline=(b)]
    \begin{feynman}
      \vertex[dot, color1, minimum size=3pt, label = {below: {\footnotesize $\textcolor{color1}{1}$}}] (x1) {};
      \vertex[right = 20pt of x1] (v);
      \vertex[right = 20pt of v, dot, color1, minimum size = 3pt, label = {below: {\footnotesize $\textcolor{color1}{2}$}}] (x2) {};
      \vertex[above = 9pt of v] (b) {};
      \vertex[above = 4pt of v, label = {right: \!\*{\scriptsize $\widehat{\Delta \Omega}_{j,12}$}}] (j){};
      \vertex[above = 16pt of v, propblob] (G) {};
      \vertex[above = 22pt of v, dot, minimum size = 0pt] (g) {};
      \vertex[above = 20pt of g, dot, minimum size = 0pt] (w) {};
      \diagram*{
        (x1) -- [color1, ghost] (v) -- [color1, ghost] (x2),
        (v) -- [photon] (G) -- (g) -- [half left] (w) -- [half left] (g)
      };
      \vertex[above = 0pt of G, propblob] (K0){};
      \vertex[above = 20pt of g, smallblob] (K1) {};
      \vertex[above = 10pt of w] (K1) {{\scriptsize $\frac{1}{n_{\ell-1}} K^{\{1\}(\ell)}$}};
      \vertex[right = 15pt of G, label = {\scriptsize $z_j$}] (Gr) {};
    \end{feynman}
  \end{tikzpicture}
  &=  \frac{1}{2 n_{\ell-1}}\sum_{\beta_{1},\beta_{2} \in \{1,2\}} K^{\{1\}(\ell)}_{\beta_{1}\beta_{2}}\langle \frac{d^{2} (\widehat{\Delta \Omega}^{(\ell+1)}_{12})}{d z^{(\ell)}_{\beta_{1}} d z^{(\ell)}_{\beta_{2}}}\rangle_{K^{(\ell)}} \nonumber\\
  \begin{tikzpicture}[baseline=(b)]
    \begin{feynman}
      \vertex[dot, color1, minimum size = 3pt, label = {below: {\footnotesize $\textcolor{color1}{1}$}}] (x1) {};
      \vertex[right = 20pt of x1] (v);
      \vertex[right = 20pt of v, dot, color1, minimum size = 3pt, label = {below: {\footnotesize $\textcolor{color1}{2}$}}] (x2) {};
      \vertex[above = 9pt of v] (b) {};
      \vertex[above = 4pt of v, label = {right: \!{\scriptsize $\widehat{\Delta \Omega}_{j,12}$}}] (j){};
      \tikzfeynmanset{every blob = {/tikz/fill=black!50, /tikz/minimum size=15pt}}
      \vertex[above = 16pt of v, blob] (G) {};
      \vertex[left = 2pt of G, dot, minimum size = 0pt] (Gl) {};
      \vertex[right = 2pt of G, dot, minimum size = 0pt] (Gr) {};
      \vertex[above = 4pt of G, dot, minimum size = 0pt] (Gu) {};
      \vertex[left = 2pt of Gu, dot, minimum size = 0pt] (Gul) {};
      \vertex[right = 2pt of Gu, dot, minimum size = 0pt] (Gur) {};
      \vertex[above = 20pt of G, smallblob] (V4) {};
      \tikzfeynmanset{every blob = {/tikz/fill=white!50, /tikz/minimum size=15pt}}
      \vertex[left = 3pt of V4, dot, minimum size = 0pt] (V4l) {};
      \vertex[right = 3pt of V4, dot, minimum size = 0pt] (V4r) {};
      \diagram*{
        (x1) -- [color1, ghost] (v) -- [color1, ghost] (x2),
        (v) -- [photon] (G) -- (Gl) -- [half left] (V4l) -- [quarter right] (Gul) -- (Gur) -- [quarter right] (V4r) -- [half left] (Gr),
      };
      \vertex[above = 0pt of G, blob] (K0) {};
      \vertex[above = 10pt of V4] (VV4) {{\scriptsize $\frac{1}{n_{\ell-1}} V_4^{(\ell)}$}};
      \vertex[right = 18pt of G, label = {above: {\scriptsize $z_j$}}] (Gr) {};
    \end{feynman}
  \end{tikzpicture}
  &= \frac{1}{8n_{\ell-1}}\sum_{\beta_{1},\beta_{2},\beta_{3},\beta_{4}\in \{1,2\}}V^{(\ell)}_{(\beta_{1}\beta_{2})(\beta_{3}\beta_{4})}\langle \frac{d^{4}(\widehat{\Delta \Omega}^{(\ell+1)}_{12})}{d z^{(\ell)}_{\beta_{1}}d z^{(\ell)}_{\beta_{2}}d z^{(\ell)}_{\beta_{3}}d z^{(\ell)}_{\beta_{4}}}\rangle_{K^{(\ell)}}\nonumber\\
  \begin{tikzpicture}[baseline=(b)]
    \begin{feynman}
      \vertex[dot, color1, minimum size = 3pt, label = {below: {\footnotesize $\textcolor{color1}{1}$}}] (x1) {};
      \vertex[right = 20pt of x1] (v);
      \vertex[right = 20pt of v, dot, color1, minimum size = 3pt, label = {below: {\footnotesize $\textcolor{color1}{2}$}}] (x2) {};
      \vertex[above = 9pt of v] (b) {};
      \vertex[above = 4pt of v, label = {right: \!{\scriptsize $\sigma'_{j}\sigma'_{j}$}}] (j){};
      \tikzfeynmanset{every blob = {/tikz/fill=black!50, /tikz/minimum size=15pt}}
      \vertex[above = 16pt of v, blob] (G) {};
      \vertex[left = 2pt of G, dot, minimum size = 0pt] (Gl) {};
      \vertex[right = 2pt of G, dot, minimum size = 0pt] (Gr) {};
      \vertex[above = 4pt of G, dot, minimum size = 0pt] (Gu) {};
      \vertex[left = 2pt of Gu, dot, minimum size = 0pt] (Gul) {};
      \vertex[right = 2pt of Gu, dot, minimum size = 0pt] (Gur) {};
      \vertex[above = 20pt of G, smallblob] (V4) {};
      \tikzfeynmanset{every blob = {/tikz/fill=white!50, /tikz/minimum size=15pt}}
      \vertex[left = 3pt of V4, dot, minimum size = 0pt] (V4l) {};
      \vertex[right = 3pt of V4, dot, minimum size = 0pt] (V4r) {};
      \diagram*{
        (x1) -- [color1, ghost] (v) -- [color1, ghost] (x2),
        (v) -- [color1, doubghost] (G) -- (Gl) -- [color1, ghost, half left] (V4l) -- [quarter right] (Gul) -- (Gur) -- [quarter right] (V4r) -- [color1, ghost, half left] (Gr),
      };
      \vertex[above = 0pt of G, blob] (K0) {};
      \vertex[above = 10pt of V4] (VV4) {{\scriptsize $\frac{1}{n_{\ell-1}} D_4^{(\ell)}$}};
      \vertex[right = 18pt of G, label = {above: {\scriptsize $z_j$}}] (Gr) {};
    \end{feynman}
  \end{tikzpicture}
  &= \frac{C_{W}^{(\ell+1)}}{2 n_{\ell-1}}\sum_{\beta_{1},\beta_{2} \in \{1,2\}}\langle \frac{d^{2} (\sigma'^{(\ell)}_{\textcolor{color1}{1}}\sigma'^{(\ell)}_{\textcolor{color1}{2}})}{d z^{(\ell)}_{\beta_{1}}d z^{(\ell)}_{\beta_{2}}}\rangle_{K^{(\ell)}} D^{(\ell)}_{\beta_{1}\beta_{2}\textcolor{color1}{12}} \nonumber\\
  \begin{tikzpicture}[baseline=(b)]
    \begin{feynman}
      \vertex[dot, color1, minimum size = 3pt, label = {below: {\footnotesize $\textcolor{color1}{1}$}}] (x1) {};
      \vertex[right = 20pt of x1] (v);
      \vertex[right = 20pt of v, dot, color1, minimum size = 3pt, label = {below: {\footnotesize $\textcolor{color1}{2}$}}] (x2) {};
      \vertex[above = 9pt of v] (b) {};
      \vertex[above = 4pt of v, label = {right: \!{\scriptsize $\sigma'_{j}\sigma'_{j}$}}] (j){};
      \tikzfeynmanset{every blob = {/tikz/fill=black!50, /tikz/minimum size=15pt}}
      \vertex[above = 16pt of v, blob] (G) {};
      \vertex[left = 2pt of G, dot, minimum size = 0pt] (Gl) {};
      \vertex[right = 2pt of G, dot, minimum size = 0pt] (Gr) {};
      \vertex[above = 4pt of G, dot, minimum size = 0pt] (Gu) {};
      \vertex[left = 2pt of Gu, dot, minimum size = 0pt] (Gul) {};
      \vertex[right = 2pt of Gu, dot, minimum size = 0pt] (Gur) {};
      \vertex[above = 20pt of G, smallblob] (V4) {};
      \tikzfeynmanset{every blob = {/tikz/fill=white!50, /tikz/minimum size=15pt}}
      \vertex[left = 3pt of V4, dot, minimum size = 0pt] (V4l) {};
      \vertex[right = 3pt of V4, dot, minimum size = 0pt] (V4r) {};
      \diagram*{
        (x1) -- [color1, ghost] (v) -- [color1, ghost] (x2),
        (v) -- [color1, doubghost] (G) -- (Gl) -- [color1, ghost, half left] (V4l) -- [quarter right] (Gul) -- (Gur) -- [color1, ghost, quarter right] (V4r) -- [half left] (Gr),
      };
      \vertex[above = 0pt of G, blob] (K0) {};
      \vertex[above = 10pt of V4] (VV4) {{\scriptsize $\frac{1}{n_{\ell-1}} F_4^{(\ell)}$}};
      \vertex[right = 18pt of G, label = {above: {\scriptsize $z_j$}}] (Gr) {};
    \end{feynman}
  \end{tikzpicture}
  &= \frac{C_{W}^{(\ell+1)}}{n_{\ell-1}}\sum_{\beta_{1},\beta_{2} \in \{1,2\}}\langle \frac{d^{2} (\sigma'^{(\ell)}_{\textcolor{color1}{1}}\sigma'^{(\ell)}_{\textcolor{color1}{2}})}{d z^{(\ell)}_{\beta_{1}}d z^{(\ell)}_{\beta_{2}}}\rangle_{K^{(\ell)}} F^{(\ell)}_{\beta_{1}\textcolor{color1}{1}\beta_{2}\textcolor{color1}{2}}
\end{align}
\endgroup


\section{Proof of Theorem \ref{theoremfour}}
\label{app:criticality_analysis}
In this Appendix, we will show that the tensors $F$, $D$, $A$ and $B$, which describe the NTK statistics at order $\frac{1}{n}$, as well as their higher-order gerenalizations, are also set to criticality once the susceptibilities of section \ref{sec:stability} are constrained to satisfy ~\cite{banta2024}
\begin{align}\label{criticalityconstraintsequiv}
    (\chi^{(\ell+1)}_{\perp})_{\alpha\beta} \equiv C_{W}^{(\ell+1)}\langle \sigma'^{(\ell)}_{\alpha}\sigma'^{(\ell)}_{\beta}\rangle_{K^{(\ell)}} &= 1 &,
    && (\chi^{(\ell+1)}_{\parallel})_{\alpha\beta} \equiv C_{W}^{(\ell+1)}\langle \sigma''^{(\ell)}_{\alpha}\sigma^{(\ell)}_{\beta}\rangle_{K^{(\ell)}} &= 0 \,.
\end{align}

\criticalityntk*

\begin{proof}
The proof will be presented following the sequence outlined above.

\begin{enumerate}[label=(\roman*)]

    \item We begin with the tensor $F_{1\textcolor{color1}{3}2\textcolor{color1}{4}}^{(\ell+1)}$ studied in section \ref{sec:feynman-rules}. Consider the variation of the second term in~\eqref{eq:F} at criticality (the infinite-width NTK $\Theta$ in the first term will not lead to exponential behavior and can hence be ignored),
\begin{align}\label{variationftensor}
    \frac{1}{n_{l}}\delta F_{1\textcolor{color1}{3}2\textcolor{color1}{4}}^{(\ell+1)} = \frac{(C_{W}^{(\ell+1)})^{2}}{n_{\ell-1}}\sum_{\alpha,\beta,\gamma,\delta=1}^{4}\langle \sigma^{(\ell)}_{1}\sigma'^{(\ell)}_{\textcolor{color1}{3}}z_{\alpha}^{(\ell)}\rangle_{K^{(\ell)}}\langle \sigma^{(\ell)}_{2}\sigma'^{(\ell)}_{\textcolor{color1}{4}}z_{\beta}^{(\ell)}\rangle_{K^{(\ell)}}K_{(\ell)}^{\alpha\gamma}K_{(\ell)}^{\beta\delta}\delta F^{(\ell)}_{\gamma\textcolor{color1}{3}\delta\textcolor{color1}{4}}\,.
\end{align}
To prevent exponential behavior, the multiplicative factor on the RHS of~\eqref{variationftensor} should be set to unity. Using integration by parts in the Gaussian expectation value, we find
\begin{align}
    \sum_{\alpha=1}^{4}\langle \sigma^{(\ell)}_{\epsilon}\sigma'^{(\ell)}_{\lambda}z^{(\ell)}_{\alpha} \rangle_{K^{(\ell)}} K^{\alpha\gamma}_{(\ell)} = \left\langle \frac{\mathrm{d}(\sigma^{(\ell)}_{\epsilon}\sigma'^{(\ell)}_{\lambda})}{\mathrm{d} z^{(\ell)}_{\gamma}}\right\rangle_{K^{(\ell)}}\hspace{-0.3cm}= \bigg[\delta_{\epsilon\gamma}\langle\sigma'^{(\ell)}_{\epsilon}\sigma'^{(\ell)}_{\lambda}\rangle_{K^{(\ell)}} + \delta_{\gamma\lambda}\langle \sigma_{\epsilon}\sigma''_{\lambda}\rangle_{K^{(\ell)}}\bigg] \,.
\end{align}
According to~\eqref{criticalityconstraintsequiv}, at criticality we hence find
\begin{align}\label{criticalityconditionchicirc}
   (\chi^{(\ell+1)}_{\circ})_{\epsilon\lambda,\gamma} \equiv C_{W}^{(\ell+1)} \sum_{\alpha=1}^{4}\langle \sigma^{(\ell)}_{\epsilon}\sigma'^{(\ell)}_{\lambda}z^{(\ell)}_{\alpha} \rangle_{K^{(\ell)}} K^{\alpha\gamma}_{(\ell)} = \delta_{\epsilon\gamma}\
 \end{align}
 and therefore $F$ is stable as well. Therefore, the criticality condition found by setting the NNGP to criticality at infinite width is sufficient to set $F$ to criticality at order $1/n$ as well.

    \item Similarly, once the NNGP, quartic vertex, and NTK have been established as stable, the stability of the tensor $D$ follows directly from its variation in the preceding layer. In diagrams:  
\begin{eqnarray}\label{d-diagram}
\begin{tikzpicture}[baseline=(d.base)]
\begin{feynman}
\vertex (d) {};
\vertex[left = 4pt of d, dot, minimum size=0pt] (v12) {};
\vertex[left = 16pt of v12, dot, minimum size=0pt] (l) {};
\vertex[below = 16pt of l, dot, minimum size=3pt, label = {below: {\footnotesize $1$}}] (x1) {};
\vertex[above = 16pt of l, dot, minimum size=3pt, label = {above: {\footnotesize $2$}}] (x2) {};
\vertex[right = 4pt of d, dot, minimum size=0pt] (v34) {};
\vertex[right = 16pt of v34, dot, minimum size=0pt] (r) {};
\vertex[above = 16pt of r, color1, dot, minimum size=3pt, label = {above: {\footnotesize $\textcolor{color1}{3}$}}] (x3) {};
\vertex[below = 16pt of r, color1, dot, minimum size=3pt, label = {below: {\footnotesize $\textcolor{color1}{4}$}}] (x4) {};
\diagram*{
	(x1) -- (v12) -- (x2), 
	 (x3) -- [color1, ghost] (v34) -- [color1, ghost] (x4)
};
\tikzfeynmanset{every blob = {/tikz/fill=black!50, /tikz/minimum size=15pt}}
\vertex[right = 0pt of d, blob, minimum size = 8pt] (dd) {{\scriptsize $\delta$}};
\end{feynman}
\end{tikzpicture}
&=& \sum_{j_1, j_2}
\begin{tikzpicture}[baseline=(b)]
\tikzfeynmanset{every blob = {/tikz/fill=white!50, /tikz/minimum size=15pt}}
\begin{feynman}
\vertex (l) {};
\vertex[below = 16pt of l, dot, minimum size=3pt, label = {below: {\footnotesize $1$}}] (x1) {};
\vertex[above = 16pt of l, dot, minimum size=3pt, label = {above: {\footnotesize $2$}}] (x2) {};
\vertex[right = 8pt of l, dot, minimum size=0pt] (v12) {};
\vertex[right = 30pt of v12, blob] (b12) {};
\vertex[right = 16pt of b12, dot, minimum size=0pt] (w12) {};
\vertex[above = 1pt of w12, label = {above: {\scriptsize \hspace{-10pt} $z_{j_1}$}}] (w12u) {};
\vertex[below = 12pt of w12] (w12d) {};
\vertex[left = 10pt of w12d] (w12dl) {};
\tikzfeynmanset{every blob = {/tikz/fill=black!50, /tikz/minimum size=15pt}}
\vertex[right = 3pt of w12, blob , minimum size = 7pt] (b) {{\scriptsize $\delta$}};
\vertex[right = 3pt of b, dot, minimum size=0pt] (w34) {};
\tikzfeynmanset{every blob = {/tikz/fill=white!50, /tikz/minimum size=15pt}}
\vertex[right = 35pt of w34, blob] (b34) {};
\vertex[above = 5pt of w34, label = {above: {\scriptsize \hspace{30pt} $\textcolor{color1}{3}$}}] (w34u) {};
\vertex[below = 12pt of w34, label = {above: {\scriptsize \hspace{30pt} $\textcolor{color1}{4}$}}] (w34d) {};
\vertex[right = 10pt of w34d] (w34dr) {};
\vertex[right = 30pt of b34, dot, minimum size=0pt] (v34) {};
\vertex[right = 30pt of b34, dot, minimum size=0pt] (vv34) {};
\vertex[right = 8pt of v34] (r) {};
\vertex[above = 16pt of r, color1, dot, minimum size=3pt, label = {above: {\footnotesize $\textcolor{color1}{3}$}}] (x3) {};
\vertex[below = 16pt of r, color1, dot, minimum size=3pt, label = {below: {\footnotesize $\textcolor{color1}{4}$}}] (x4) {};
\diagram*{
	(x1) -- (v12) -- (x2),
	(v12) -- [photon, edge label = {\scriptsize \;$\widehat{\Delta G}_{j_1}$}, inner sep = 4pt] (b12) -- [quarter left] (w12) -- [quarter left] (b12),
	(b34) -- [color1, ghost, quarter left] (w34) -- [color1, ghost, quarter left] (b34) -- [color1doubghost, edge label = {\scriptsize \,$\sigma'_{j_{2}}\sigma'_{j_{2}}$}, inner sep = 4pt] (v34), 
	(x3) -- [color1, ghost] (v34) -- [color1, ghost] (x4)
};
\end{feynman}
\end{tikzpicture}
\end{eqnarray}
where the $\delta$-blob denotes $\delta D$, and we used the notation $\widehat{\Delta G}_{j,\alpha\beta}=\sigma_{j,\alpha}\sigma_{j,\beta}-\langle \sigma_{j,\alpha}\sigma_{j,\beta} \rangle$ from~\cite{banta2024}. In algebraic form, the diagram \eqref{d-diagram} reads
\begin{eqnarray}
\frac{1}{n_{\ell}}\delta D^{(\ell)}_{12\textcolor{color1}{34}} &=& \frac{1}{n_{\ell-1}}\sum_{\beta_{1},\beta_{2} \in \{1,2,3,4\}}(\chi_{||}^{(\ell+1)})_{12,\beta_{1}\beta_{2}}(\chi_{\perp}^{(\ell+1)})_{\textcolor{color1}{34}}\delta D^{(\ell)}_{\beta_{1}\beta_{2}\textcolor{color1}{34}}
\end{eqnarray}
In this manner, one can see that the NNGP criticality conditions in the infinite-width limit, namely $(\chi_{||}^{(\ell+1)})_{12,\beta_{1}\beta_{2}} = \delta_{1\beta_{1}}\delta_{2\beta_{2}}$, $(\chi_{\perp}^{(\ell+1)})_{\textcolor{color1}{34}} = 1$, immediately imply the stability of $D$. 

\item Analogously, the NTK variance can be examined under the criticality conditions, in which case the variation of the tensor $A$ at layer $\ell + 1$ readily follows from its corresponding recursion relation. In diagrams: 
\begin{eqnarray}
\begin{tikzpicture}[baseline=(d.base)]
\begin{feynman}
\vertex (d) {};
\vertex[left = 4pt of d, dot, minimum size=0pt] (v12) {};
\vertex[left = 16pt of v12, dot, minimum size=0pt] (l) {};
\vertex[below = 16pt of l, color1, dot, minimum size=3pt, label = {below: {\footnotesize $\textcolor{color1}{1}$}}] (x1) {};
\vertex[above = 16pt of l, color1, dot, minimum size=3pt, label = {above: {\footnotesize $\textcolor{color1}{2}$}}] (x2) {};
\vertex[right = 4pt of d, dot, minimum size=0pt] (v34) {};
\vertex[right = 16pt of v34, dot, minimum size=0pt] (r) {};
\vertex[above = 16pt of r, color2, dot, minimum size=3pt, label = {above: {\footnotesize $\textcolor{color2}{3}$}}] (x3) {};
\vertex[below = 16pt of r, color2, dot, minimum size=3pt, label = {below: {\footnotesize $\textcolor{color2}{4}$}}] (x4) {};
\diagram*{
	(x1) -- [color1, ghost] (v12) -- [color1, ghost] (x2), 
	 (x3) -- [color2, ghost] (v34) -- [color2, ghost] (x4)
};
\tikzfeynmanset{every blob = {/tikz/fill=black!50, /tikz/minimum size=15pt}}
\vertex[right = 0pt of d, blob, minimum size = 8pt] (dd) {{\scriptsize $\delta$}};
\end{feynman}
\end{tikzpicture}
&=& \sum_{j_1, j_2}
\begin{tikzpicture}[baseline=(b)]
\tikzfeynmanset{every blob = {/tikz/fill=white!50, /tikz/minimum size=15pt}}
\begin{feynman}
\vertex (l) {};
\vertex[below = 16pt of l, dot, color1, minimum size=3pt, label = {below: {\footnotesize $\textcolor{color1}{1}$}}] (x1) {};
\vertex[above = 16pt of l, dot, color1, minimum size=3pt, label = {above: {\footnotesize $\textcolor{color1}{2}$}}] (x2) {};
\vertex[right = 8pt of l, dot, minimum size=0pt] (v12) {};
\vertex[right = 35pt of v12, blob] (b12) {};
\vertex[right = 35pt of b12, dot, minimum size=0pt] (w12) {};
\vertex[above = 5pt of w12, label = {above: {\scriptsize \hspace{-35pt} $ \textcolor{color1}{2}$}}] (w12u) {};
\vertex[below = 12pt of w12, label = {above: {\scriptsize \hspace{-35pt} $\textcolor{color1}{1}$}}] (w12d) {};
\vertex[left = 10pt of w12d] (w12dl) {};
\tikzfeynmanset{every blob = {/tikz/fill=black!50, /tikz/minimum size=15pt}}
\vertex[right = 3pt of w12, blob , minimum size = 7pt] (b) {{\scriptsize $\delta$}};
\vertex[right = 3pt of b, dot, minimum size=0pt] (w34) {};
\tikzfeynmanset{every blob = {/tikz/fill=white!50, /tikz/minimum size=15pt}}
\vertex[right = 35pt of w34, blob] (b34) {};
\vertex[above = 5pt of w34, label = {above: {\scriptsize \hspace{30pt} $\textcolor{color2}{3}$}}] (w34u) {};
\vertex[below = 12pt of w34, label = {above: {\scriptsize \hspace{30pt} $\textcolor{color2}{4}$}}] (w34d) {};
\vertex[right = 10pt of w34d] (w34dr) {};
\vertex[right = 30pt of b34, dot, minimum size=0pt] (v34) {};
\vertex[right = 30pt of b34, dot, minimum size=0pt] (vv34) {};
\vertex[right = 8pt of v34] (r) {};
\vertex[above = 16pt of r, color2, dot, minimum size=3pt, label = {above: {\footnotesize $\textcolor{color2}{3}$}}] (x3) {};
\vertex[below = 16pt of r, color2, dot, minimum size=3pt, label = {below: {\footnotesize $\textcolor{color2}{4}$}}] (x4) {};
\diagram*{
	(x1) -- [color1, ghost] (v12) -- [color1, ghost] (x2),
	(v12) -- [color1doubghost, edge label = {\scriptsize \;$\sigma'_{j_1}\sigma'_{j_1}$}, inner sep = 4pt] (b12) -- [color1, ghost, quarter left] (w12) -- [color1, ghost, quarter left] (b12),
	(b34) -- [color2, ghost, quarter left] (w34) -- [color2, ghost, quarter left] (b34) -- [color2doubghost, edge label = {\scriptsize \,$\sigma'_{j_{2}}\sigma'_{j_{2}}$}, inner sep = 4pt] (v34), 
	(x3) -- [color2, ghost] (v34) -- [color2, ghost] (x4)
};
\end{feynman}
\end{tikzpicture}
\end{eqnarray}
or, equivalently
\begin{eqnarray}
\frac{1}{n_{\ell}}\delta A^{(\ell)}_{\textcolor{color1}{12}\textcolor{color2}{34}} &=& \frac{1}{n_{\ell-1}}(\chi_{\perp}^{(\ell+1)})_{\textcolor{color1}{12}}(\chi_{\perp}^{(\ell+1)})_{\textcolor{color2}{34}}\delta D^{(\ell)}_{\textcolor{color1}{12}\textcolor{color2}{34}}
\end{eqnarray}
Hence, we see that the criticality condition $(\chi_{\perp}^{\ell+1})_{\alpha\beta} = 1$, directly demands the stability of the tensor $A$.

\item In a similar fashion, the variation of the tensor $B$ at criticality is given by
\begin{eqnarray}
\begin{tikzpicture}[baseline=(d.base)]
\begin{feynman}
\vertex (d) {};
\vertex[left = 4pt of d, dot, minimum size=0pt] (v12) {};
\vertex[left = 16pt of v12, dot, minimum size=0pt] (l) {};
\vertex[below = 16pt of l, color1, dot, minimum size=3pt, label = {below: {\footnotesize $\textcolor{color1}{1}$}}] (x1) {};
\vertex[above = 16pt of l, color2, dot, minimum size=3pt, label = {above: {\footnotesize $\textcolor{color2}{3}$}}] (x2) {};
\vertex[right = 4pt of d, dot, minimum size=0pt] (v34) {};
\vertex[right = 16pt of v34, dot, minimum size=0pt] (r) {};
\vertex[above = 16pt of r, color1, dot, minimum size=3pt, label = {above: {\footnotesize $\textcolor{color1}{2}$}}] (x3) {};
\vertex[below = 16pt of r, color2, dot, minimum size=3pt, label = {below: {\footnotesize $\textcolor{color2}{4}$}}] (x4) {};
\diagram*{
	(x1) -- [color1, ghost] (v12) -- [color2, ghost] (x2), 
	 (x3) -- [color1, ghost] (v34) -- [color2, ghost] (x4)
};
\tikzfeynmanset{every blob = {/tikz/fill=black!50, /tikz/minimum size=15pt}}
\vertex[right = 0pt of d, blob, minimum size = 8pt] (dd) {{\scriptsize $\delta$}};
\end{feynman}
\end{tikzpicture}
&=& \sum_{j_1, j_2}
\begin{tikzpicture}[baseline=(b)]
\tikzfeynmanset{every blob = {/tikz/fill=white!50, /tikz/minimum size=15pt}}
\begin{feynman}
\vertex (l) {};
\vertex[below = 16pt of l, dot, color1, minimum size=3pt, label = {below: {\footnotesize $\textcolor{color1}{1}$}}] (x1) {};
\vertex[above = 16pt of l, dot, color2, minimum size=3pt, label = {above: {\footnotesize $\textcolor{color2}{3}$}}] (x2) {};
\vertex[right = 8pt of l, dot, minimum size=0pt] (v12) {};
\vertex[right = 35pt of v12, blob] (b12) {};
\vertex[right = 35pt of b12, dot, minimum size=0pt] (w12) {};
\vertex[above = 5pt of w12, label = {above: {\scriptsize \hspace{-35pt} $ \textcolor{color2}{3}$}}] (w12u) {};
\vertex[below = 12pt of w12, label = {above: {\scriptsize \hspace{-35pt} $\textcolor{color1}{1}$}}] (w12d) {};
\vertex[left = 10pt of w12d] (w12dl) {};
\tikzfeynmanset{every blob = {/tikz/fill=black!50, /tikz/minimum size=15pt}}
\vertex[right = 3pt of w12, blob , minimum size = 7pt] (b) {{\scriptsize $\delta$}};
\vertex[right = 3pt of b, dot, minimum size=0pt] (w34) {};
\tikzfeynmanset{every blob = {/tikz/fill=white!50, /tikz/minimum size=15pt}}
\vertex[right = 35pt of w34, blob] (b34) {};
\vertex[above = 5pt of w34, label = {above: {\scriptsize \hspace{30pt} $\textcolor{color1}{2}$}}] (w34u) {};
\vertex[below = 12pt of w34, label = {above: {\scriptsize \hspace{30pt} $\textcolor{color2}{4}$}}] (w34d) {};
\vertex[right = 10pt of w34d] (w34dr) {};
\vertex[right = 30pt of b34, dot, minimum size=0pt] (v34) {};
\vertex[right = 30pt of b34, dot, minimum size=0pt] (vv34) {};
\vertex[right = 8pt of v34] (r) {};
\vertex[above = 16pt of r, color1, dot, minimum size=3pt, label = {above: {\footnotesize $\textcolor{color1}{2}$}}] (x3) {};
\vertex[below = 16pt of r, color2, dot, minimum size=3pt, label = {below: {\footnotesize $\textcolor{color2}{4}$}}] (x4) {};
\diagram*{
	(x1) -- [color1, ghost] (v12) -- [color2, ghost] (x2),
	(v12) -- [color2color1ghost, edge label = {\scriptsize \;$\sigma'_{j_1}\sigma'_{j_1}$}, inner sep = 4pt] (b12) -- [color2, ghost, quarter left] (w12) -- [color1, ghost, quarter left] (b12),
	(b34) -- [color2, ghost, quarter left] (w34) -- [color1, ghost, quarter left] (b34) -- [color1color2ghost, edge label = {\scriptsize \,$\sigma'_{j_{2}}\sigma'_{j_{2}}$}, inner sep = 4pt] (v34), 
	(x3) -- [color1, ghost] (v34) -- [color2, ghost] (x4)
};
\end{feynman}
\end{tikzpicture}
\end{eqnarray}
or, explicitly
\begin{eqnarray}
\frac{1}{n_{\ell}}\delta B^{(\ell)}_{\textcolor{color1}{1}\textcolor{color2}{3}\textcolor{color1}{2}\textcolor{color2}{4}} &=& \frac{1}{n_{\ell-1}}(\chi_{\perp}^{(\ell+1)})_{\textcolor{color1}{1}\textcolor{color2}{3}}(\chi_{\perp}^{(\ell+1)})_{\textcolor{color1}{2}\textcolor{color2}{4}}\delta B^{(\ell)}_{\textcolor{color1}{1}\textcolor{color2}{3}\textcolor{color1}{2}\textcolor{color2}{4}}
\end{eqnarray}
Once more, one concludes that criticality, $(\chi_{\perp}^{\ell+1})_{\alpha\beta} = 1$, sets the stability of $B$.

\end{enumerate}

This reasoning extends to the analysis of higher-rank tensors associated with cumulants involving an arbitrary number of preactivations and NTKs, using the Feynman rules introduced in Section~\ref{sec:feynman-rules}. For example, we define higher-rank analogues of the tensors $D$, $F$, $A$, $B$ from~\eqref{eq:3}, \eqref{defabfourpoint} as
\begin{align}
     &\mathbb{E}^{c}_{\theta}[ z^{(\ell+1)}_{i_{1},1},z^{(\ell+1)}_{i_{2},2}, z^{(\ell+1)}_{i_{3},3},z^{(\ell+1)}_{i_{4},4},\ldots, \widehat{\Delta \Theta}^{(\ell+1)}_{i_{m-1}i_{m},\textcolor{color1}{(m-1)}\textcolor{color1}{m}}] \nonumber\\ 
      &{=} \frac{1}{n_{\ell}^{\frac{m}{2}-1}}\!\bigg[\delta_{i_{1}i_{2}}\delta_{i_{3}i_{4}}\ldots \delta_{i_{m-1}i_{m}}D^{(\ell+1)}_{1234\ldots \textcolor{color1}{(m-1)m}} {+} \delta_{i_{1}i_{2}}\delta_{i_{3}i_{(m-1)}}\ldots \delta_{i_{4}i_{m}}F^{(\ell+1)}_{123\textcolor{color1}{(m-1)}\ldots 4\textcolor{color1}{m}} {+} \ldots\!\bigg] \label{higher-order-tensors} \\
     &\mathbb{E}^{c}_{\theta}[z^{(\ell+1)}_{i_{1},1},z^{(\ell+1)}_{i_{2},2},\widehat{\Delta \Theta}^{(\ell+1)}_{i_{3}i_{4},\textcolor{color1}{3}\textcolor{color1}{4}} ,\ldots, \widehat{\Delta \Theta}^{(\ell+1)}_{i_{m-1}i_{m},\textcolor{color2}{(m-1)}\textcolor{color2}{m}}] \nonumber\\ 
      &{=} \frac{1}{n_{\ell}^{\frac{m}{2}-1}}\!\bigg[\delta_{i_{1}i_{2}}\delta_{i_{3}i_{4}}\ldots \delta_{i_{m-1}i_{m}}A^{(\ell+1)}_{12\textcolor{color1}{34}\ldots \textcolor{color2}{(m-1)m}} {+} \delta_{i_{1}i_{2}}\delta_{i_{3}i_{(m-1)}}\ldots \delta_{i_{4}i_{m}}B^{(\ell+1)}_{12\textcolor{color1}{3}\textcolor{color2}{(m-1)}\ldots \textcolor{color1}{4}\textcolor{color2}{m}} {+} \ldots\!\bigg] \label{higher-order-tensors_AB}
\end{align}

\begin{enumerate}[label=(\roman*)]

\item Following our bootstrap strategy, we assume that lower-order tensors have been tuned to criticality. The variation of the tensor $F^{(\ell+1)}_{123\textcolor{color1}{(m-1)}\ldots 4\textcolor{color1}{m}}$ then follows directly from the diagram

\begin{eqnarray}
    \begin{tikzpicture}[baseline=(d.base)]
        \begin{feynman}
        \vertex (d) {};
        \vertex[left = 4pt of d, dot, minimum size = 0pt] (v12) {};
        \vertex[left = 16pt of v12] (l) {};
        \vertex[below = 8pt of l, dot, minimum size = 3pt, label = {left: {\footnotesize $3$}}] (x1) {};
        \vertex[above = 8pt of l, dot, color1, minimum size = 3pt, label = {left: {\footnotesize $\textcolor{color1}{m-1}$}}] (x2) {};
        \vertex[below = 4pt of d, dot, minimum size = 0pt] (v34) {};
        \vertex[below = 16pt of v34] (u) {};
        \vertex[left = 8pt of u, dot, minimum size = 3pt, label = {below: {\footnotesize $2$}}] (x3) {};
        \vertex[right = 8pt of u, dot, minimum size = 3pt, label = {below: {\footnotesize $1$}}] (x4) {};
        \vertex[right = 4pt of d, dot, minimum size = 0pt] (v56) {};
        \vertex[right = 16pt of v56, dot, minimum size = 0pt] (r) {};
        \vertex[above = 8pt of r, dot, minimum size = 3pt, label = {right: {\footnotesize $4$}}] (x5) {};
        \vertex[below = 8pt of r, dot, color1, minimum size = 3pt, label = {right: {\footnotesize $\textcolor{color1}{m}$}}] (x6) {};
        \diagram*{
            (x1) -- (v12) --[color1, ghost] (x2),
            (x3) -- (v34) -- (x4),
            (x5) -- (v56) -- [color1, ghost] (x6)
        };
        \tikzfeynmanset{every blob = {/tikz/fill=white!70!black, /tikz/minimum size=15pt}}
        \vertex[right = 0pt of d, blob, minimum size = 8pt] (dd) {{\scriptsize $\delta$}};
        \vertex[above = 15pt of d, dot, minimum size = 1pt] (d2) {};
        \vertex[left = 8pt of d2] (d2l) {};
        \vertex[below = 2pt of d2l, dot, minimum size = 1pt] (d1) {};
        \vertex[right = 8pt of d2] (d2r) {};
        \vertex[below = 2pt of d2r, dot, minimum size = 1pt] (d3) {};
        \end{feynman}
        \end{tikzpicture}
        \;\;\,=\; \sum_{j_1, \dots, j_k}
        \begin{tikzpicture}[baseline=(d.base)]
        \begin{feynman}
        \vertex (d) {};
        \vertex[left = 4pt of d, dot, minimum size = 0pt] (w12) {};
        \vertex[below = 11pt of w12, label = {above: {\scriptsize \hspace{-22pt} $z_{j_2}$}}] (w12d) {};
        \tikzfeynmanset{every blob = {/tikz/fill=white!50, /tikz/minimum size=15pt}}
        \vertex[left = 25pt of w12, blob] (b12) {};
        \vertex[left = 25pt of b12, dot, minimum size = 0pt] (v12) {};
        \vertex[left = 8pt of v12] (l) {};
        \vertex[below = 16pt of l, dot, minimum size = 3pt, label = {left: {\footnotesize $3$}}] (x1) {};
        \vertex[above = 16pt of l, dot, color1, minimum size = 3pt, label = {left: {\footnotesize $\textcolor{color1}{m-1}$}}] (x2) {};
        \vertex[below = 4pt of d, dot, minimum size = 0pt] (w34) {};
        \vertex[right = 13pt of w34, label = {above: {\scriptsize $ $}}, inner sep = -1pt] (w34r) {};
        \vertex[below = 25pt of w34, propblob] (b34) {};
        \vertex[below = 25pt of b34, dot, minimum size = 0pt] (v34) {};
        \vertex[below = 8pt of v34] (u) {};
        \vertex[left = 16pt of u, dot, minimum size = 3pt, label = {above: {\footnotesize $2$}}] (x3) {};
        \vertex[right = 16pt of u, dot, minimum size = 3pt, label = {above: {\footnotesize $1$}}] (x4) {};
        \vertex[right = 4pt of d, dot, minimum size = 0pt] (w56) {};
        \vertex[below = 1pt of w56, label = {below: {\scriptsize \hspace{18pt} $z_{j_1}$}}] (w56d) {};
        \vertex[right = 25pt of w56, propblob] (b56) {};
        \vertex[above = 2pt of w56, label = {above: {\scriptsize \hspace{22pt} $z_{j_k}$}}] (w56dd) {};
        \vertex[right = 25pt of b56, dot, minimum size = 0pt] (v56) {};
        \vertex[right = 8 pt of v56, dot, minimum size = 0pt] (r) {};
        \vertex[above = 16pt of r, dot, minimum size = 3pt, label = {right: {\footnotesize $4$}}] (x5) {};
        \vertex[below = 16pt of r, dot, color1, minimum size = 3pt, label = {right: {\footnotesize $\textcolor{color1}{m}$}}] (x6) {};
        \diagram*{
            (x1) -- (v12) -- [color1, ghost] (x2),
            (x3) -- (v34) -- (x4),
            (x5) -- (v56) -- [color1, ghost] (x6),
            (v12) -- [color1blackghost, edge label' = {\scriptsize \;$\sigma_{j_2}\sigma'_{j_2}$}, inner sep = 4pt] (b12) -- [color1, ghost, quarter left] (w12) -- [quarter left] (b12),
            (v34) -- [photon, edge label = {\scriptsize $\widehat{\Delta G}_{j_1}$}, inner sep = 2pt] (b34) -- [quarter left] (w34) -- [quarter left] (b34),
            (v56) -- [blackcolor1ghost, edge label = {\scriptsize \,$\sigma_{j_k}\sigma'_{j_k}$}, inner sep = 4pt] (b56) -- [color1, ghost, quarter left] (w56) -- [quarter left] (b56)
        };
        \tikzfeynmanset{every blob = {/tikz/fill=white!70!black, /tikz/minimum size=15pt}}
        \vertex[right = 0pt of d, blob, minimum size = 8pt] (dd) {{\scriptsize $\delta$}};
        \vertex[above = 22pt of d, dot, minimum size = 1pt] (d2) {};
        \vertex[left = 10pt of d2] (d2l) {};
        \vertex[below = 3pt of d2l, dot, minimum size = 1pt] (d1) {};
        \vertex[right = 10pt of d2] (d2r) {};
        \vertex[below = 3pt of d2r, dot, minimum size = 1pt] (d3) {};
        \end{feynman}
      \end{tikzpicture}
      \label{eq:1}
\end{eqnarray}
Algebraically, this diagram evaluates to
\begin{align}
  \delta F^{(\ell+1)}_{123\textcolor{color1}{(m-1)}\ldots 4\textcolor{color1}{m}} {=} \sum_{\beta_{k}}(\chi^{(\ell+1)}_{\parallel})_{12,\beta_{1}\beta_{2}}(\chi^{(\ell+1)}_{\circ})_{3 \textcolor{color1}{(m-1)},\beta_{3}}\ldots (\chi^{(\ell+1)}_{\circ})_{4\textcolor{color1}{m},\beta_{4}}\delta F^{(\ell)}_{\beta_{1}\beta_{2}\beta_{3}\textcolor{color1}{(m-1)}\ldots \beta_{4}\textcolor{color1}{m}}\,,\label{eq:8}
\end{align}
where $(\chi^{(\ell+1)}_{\parallel})_{\alpha\beta,\gamma\delta}=\frac{C_{W}^{(\ell+1)}}{2}\langle \frac{\mathrm{d}^{2}(\sigma_{\alpha}\sigma_{\beta})}{\mathrm{d} z_{\gamma}\mathrm{d} z_{\delta}} \rangle_{K^{(\ell)}}=\delta_{\alpha\gamma}\delta_{\beta\delta}$ at criticality~\cite{banta2024}, setting the prefactor in~\eqref{eq:8} to 1. Once more, one finds that the criticality conditions entirely determine the stability of the higher-rank $F$ tensor.

\item Analogously, $\delta D^{(\ell+1)}_{1234\ldots \textcolor{color1}{(m-1)}\textcolor{color1}{m}}$ in \eqref{higher-order-tensors}, can easily be deduced from the diagrammatic relation
\begin{eqnarray}
    \begin{tikzpicture}[baseline=(d.base)]
        \begin{feynman}
        \vertex (d) {};
        \vertex[left = 4pt of d, dot, minimum size = 0pt] (v12) {};
        \vertex[left = 16pt of v12] (l) {};
        \vertex[below = 8pt of l, dot, minimum size = 3pt, label = {above: {\footnotesize $3$}}] (x1) {};
        \vertex[above = 8pt of l, dot, minimum size = 3pt, label = {above: {\footnotesize $4$}}] (x2) {};
        \vertex[below = 4pt of d, dot, minimum size = 0pt] (v34) {};
        \vertex[below = 16pt of v34] (u) {};
        \vertex[left = 8pt of u, dot, minimum size = 3pt, label = {above: {\footnotesize $2$}}] (x3) {};
        \vertex[right = 8pt of u, dot, minimum size = 3pt, label = {above: {\footnotesize $1$}}] (x4) {};
        \vertex[right = 4pt of d, dot, minimum size = 0pt] (v56) {};
        \vertex[right = 16pt of v56, dot, minimum size = 0pt] (r) {};
        \vertex[above = 8pt of r, dot, color1, minimum size = 3pt, label = {above: {\footnotesize \hspace{7pt} $\textcolor{color1}{(m-1)}$}}] (x5) {};
        \vertex[below = 8pt of r, dot, color1, minimum size = 3pt, label = {below: {\footnotesize \hspace{7pt} $\textcolor{color1}{m}$}}] (x6) {};
        \diagram*{
            (x1) -- (v12) -- (x2),
            (x3) -- (v34) -- (x4),
            (x5) -- [color1, ghost] (v56) -- [color1, ghost] (x6)
        };
        \tikzfeynmanset{every blob = {/tikz/fill=black!50, /tikz/minimum size=15pt}}
        \vertex[right = 0pt of d, blob, minimum size = 8pt] (dd) {{\scriptsize $\delta$}};
        \vertex[above = 15pt of d, dot, minimum size = 1pt] (d2) {};
        \vertex[left = 8pt of d2] (d2l) {};
        \vertex[below = 2pt of d2l, dot, minimum size = 1pt] (d1) {};
        \vertex[right = 8pt of d2] (d2r) {};
        \vertex[below = 2pt of d2r, dot, minimum size = 1pt] (d3) {};
        \end{feynman}
        \end{tikzpicture}
        \;\;\,=\; \sum_{j_1, \dots, j_k}
        \begin{tikzpicture}[baseline=(d.base)]
        \begin{feynman}
        \vertex (d) {};
        \vertex[left = 4pt of d, dot, minimum size = 0pt] (w12) {};
        \vertex[above = 1pt of w12, label = {above: {\scriptsize \hspace{-10pt} $z_{j_2}$}}] (w12d) {};
        \tikzfeynmanset{every blob = {/tikz/fill=white!50, /tikz/minimum size=15pt}}
        \vertex[left = 25pt of w12, blob] (b12) {};
        \vertex[left = 25pt of b12, dot, minimum size = 0pt] (v12) {};
        \vertex[left = 8pt of v12] (l) {};
        \vertex[below = 16pt of l, dot, minimum size = 3pt, label = {below: {\footnotesize $3$}}] (x1) {};
        \vertex[above = 16pt of l, dot, minimum size = 3pt, label = {above: {\footnotesize $4$}}] (x2) {};
        \vertex[below = 4pt of d, dot, minimum size = 0pt] (w34) {};
        \vertex[right = 13pt of w34, label = {above: {\scriptsize $ $}}, inner sep = -1pt] (w34r) {};
        \tikzfeynmanset{every blob = {/tikz/fill=white!50, /tikz/minimum size=15pt}}
        \vertex[below = 25pt of w34, blob] (b34) {};
        \vertex[below = 25pt of b34, dot, minimum size = 0pt] (v34) {};
        \vertex[below = 8pt of v34] (u) {};
        \vertex[left = 16pt of u, dot, minimum size = 3pt, label = {above: {\footnotesize $2$}}] (x3) {};
        \vertex[right = 16pt of u, dot, minimum size = 3pt, label = {above: {\footnotesize $1$}}] (x4) {};
        \vertex[right = 4pt of d, dot, minimum size = 0pt] (w56) {};
        \vertex[below = 1pt of w56, label = {below: {\scriptsize \hspace{18pt} $z_{j_1}$}}] (w56d) {};
        \tikzfeynmanset{every blob = {/tikz/fill=white!50, /tikz/minimum size=15pt}}
        \vertex[right = 25pt of w56, blob] (b56) {};
        \vertex[right = 25pt of b56, dot, minimum size = 0pt] (v56) {};
        \vertex[right = 8 pt of v56, dot, minimum size = 0pt] (r) {};
        \vertex[above = 16pt of r, dot, color1, minimum size = 3pt, label = {above: {\footnotesize $\textcolor{color1}{(m-1)}$}}] (x5) {};
        \vertex[below = 16pt of r, dot, color1, minimum size = 3pt, label = {below: {\footnotesize $\textcolor{color1}{m}$}}] (x6) {};
        \diagram*{
            (x1) -- (v12) -- (x2),
            (x3) -- (v34) -- (x4),
            (x5) -- [color1, ghost] (v56) -- [color1, ghost] (x6),
            (v12) -- [photon, edge label' = {\scriptsize \;$\widehat{\Delta G}_{j_2}$}, inner sep = 4pt] (b12) -- [quarter left] (w12) -- [quarter left] (b12),
            (v34) -- [photon, edge label = {\scriptsize $\widehat{\Delta G}_{j_1}$}, inner sep = 4pt] (b34) -- [quarter left] (w34) -- [quarter left] (b34),
            (v56) -- [color1doubghost, edge label = {\scriptsize \,$\sigma'_{j_k}\sigma'_{j_k}$}, inner sep = 4pt] (b56) -- [color1, ghost, quarter left] (w56) -- [color1, ghost, quarter left] (b56)
        };
        \tikzfeynmanset{every blob = {/tikz/fill=black!50, /tikz/minimum size=15pt}}
        \vertex[right = 0pt of d, blob, minimum size = 8pt] (dd) {{\scriptsize $\delta$}};
        \vertex[above = 25pt of d, dot, minimum size = 1pt] (d2) {};
        \vertex[left = 10pt of d2] (d2l) {};
        \vertex[below = 3pt of d2l, dot, minimum size = 1pt] (d1) {};
        \vertex[right = 10pt of d2] (d2r) {};
        \vertex[below = 3pt of d2r, dot, minimum size = 1pt] (d3) {};
        \end{feynman}
        \end{tikzpicture}
\end{eqnarray}
which algebraically reads
\begin{eqnarray}\label{dequationcriticality}
    \delta D^{(\ell+1)}_{1234\ldots \textcolor{color1}{(m-1)}\textcolor{color1}{m}} \! \! \! \! \! \! &=& \! \! \! \! \! \! \sum_{\beta_{k}}(\chi^{(\ell+1)}_{\parallel})_{12,\beta_{1}\beta_{2}}(\chi^{(\ell+1)}_{\parallel})_{34,\beta_{3}\beta_{4}}\ldots (\chi^{(\ell+1)}_{\perp})_{\textcolor{color1}{(m-1)}\textcolor{color1}{m}}\delta D^{(\ell)}_{\beta_{1}\beta_{2}\beta_{3}\beta_{4}\ldots \textcolor{color1}{(m-1)}\textcolor{color1}{m}}
\end{eqnarray}
Remarkably, one sees that the criticality conditions for $\chi_{\parallel}$ in~\eqref{criticalityconstraintsequiv} and $\chi_{\perp}$ below~\eqref{eq:8} are again sufficient to fix the stability of the tensor $D$.

\item Likewise, the variation of the NTK-variance tensor $A$ can be written as
\begin{eqnarray}
    \begin{tikzpicture}[baseline=(d.base)]
        \begin{feynman}
        \vertex (d) {};
        \vertex[left = 4pt of d, dot, minimum size = 0pt] (v12) {};
        \vertex[left = 16pt of v12] (l) {};
        \vertex[below = 8pt of l, dot, color1, minimum size = 3pt, label = {above: {\footnotesize $\textcolor{color1}{3}$}}] (x1) {};
        \vertex[above = 8pt of l, dot, color1, minimum size = 3pt, label = {above: {\footnotesize $\textcolor{color1}{4}$}}] (x2) {};
        \vertex[below = 4pt of d, dot, minimum size = 0pt] (v34) {};
        \vertex[below = 16pt of v34] (u) {};
        \vertex[left = 8pt of u, dot, minimum size = 3pt, label = {above: {\footnotesize $2$}}] (x3) {};
        \vertex[right = 8pt of u, dot, minimum size = 3pt, label = {above: {\footnotesize $1$}}] (x4) {};
        \vertex[right = 4pt of d, dot, minimum size = 0pt] (v56) {};
        \vertex[right = 16pt of v56, dot, minimum size = 0pt] (r) {};
        \vertex[above = 8pt of r, dot, color2, minimum size = 3pt, label = {above: {\footnotesize \hspace{7pt} $\textcolor{color2}{(m-1)}$}}] (x5) {};
        \vertex[below = 8pt of r, dot, color2, minimum size = 3pt, label = {below: {\footnotesize \hspace{7pt} $\textcolor{color2}{m}$}}] (x6) {};
        \diagram*{
            (x1) -- [color1, ghost] (v12) -- [color1, ghost] (x2),
            (x3) -- (v34) -- (x4),
            (x5) -- [color2, ghost] (v56) -- [color2, ghost] (x6)
        };
        \tikzfeynmanset{every blob = {/tikz/fill=black!50, /tikz/minimum size=15pt}}
        \vertex[right = 0pt of d, blob, minimum size = 8pt] (dd) {{\scriptsize $\delta$}};
        \vertex[above = 15pt of d, dot, minimum size = 1pt] (d2) {};
        \vertex[left = 8pt of d2] (d2l) {};
        \vertex[below = 2pt of d2l, dot, minimum size = 1pt] (d1) {};
        \vertex[right = 8pt of d2] (d2r) {};
        \vertex[below = 2pt of d2r, dot, minimum size = 1pt] (d3) {};
        \end{feynman}
        \end{tikzpicture}
        \;\;\,=\; \sum_{j_1, \dots, j_k}
        \begin{tikzpicture}[baseline=(d.base)]
        \begin{feynman}
        \vertex (d) {};
        \vertex[left = 4pt of d, dot, minimum size = 0pt] (w12) {};
        \vertex[above = 1pt of w12, label = {above: {\scriptsize \hspace{-10pt} $ $}}] (w12d) {};
        \tikzfeynmanset{every blob = {/tikz/fill=white!50, /tikz/minimum size=15pt}}
        \vertex[left = 25pt of w12, blob] (b12) {};
        \vertex[left = 25pt of b12, dot, minimum size = 0pt] (v12) {};
        \vertex[left = 8pt of v12] (l) {};
        \vertex[below = 16pt of l, dot, color1, minimum size = 3pt, label = {below: {\footnotesize $\textcolor{color1}{3}$}}] (x1) {};
        \vertex[above = 16pt of l, dot, color1, minimum size = 3pt, label = {above: {\footnotesize $\textcolor{color1}{4}$}}] (x2) {};
        \vertex[below = 4pt of d, dot, minimum size = 0pt] (w34) {};
        \vertex[right = 13pt of w34, label = {above: {\scriptsize $ $}}, inner sep = -1pt] (w34r) {};
        \tikzfeynmanset{every blob = {/tikz/fill=white!50, /tikz/minimum size=15pt}}
        \vertex[below = 25pt of w34, blob] (b34) {};
        \vertex[below = 25pt of b34, dot, minimum size = 0pt] (v34) {};
        \vertex[below = 8pt of v34] (u) {};
        \vertex[left = 16pt of u, dot, minimum size = 3pt, label = {above: {\footnotesize $2$}}] (x3) {};
        \vertex[right = 16pt of u, dot, minimum size = 3pt, label = {above: {\footnotesize $1$}}] (x4) {};
        \vertex[right = 4pt of d, dot, minimum size = 0pt] (w56) {};
        \vertex[below = 1pt of w56, label = {below: {\scriptsize \hspace{18pt} $z_{j_1} $}}] (w56d) {};
        \tikzfeynmanset{every blob = {/tikz/fill=white!50, /tikz/minimum size=15pt}}
        \vertex[right = 25pt of w56, blob] (b56) {};
        \vertex[right = 25pt of b56, dot, minimum size = 0pt] (v56) {};
        \vertex[right = 8 pt of v56, dot, minimum size = 0pt] (r) {};
        \vertex[above = 16pt of r, dot, color2, minimum size = 3pt, label = {above: {\footnotesize $\textcolor{color2}{(m-1)}$}}] (x5) {};
        \vertex[below = 16pt of r, dot, color2, minimum size = 3pt, label = {below: {\footnotesize $\textcolor{color2}{m}$}}] (x6) {};
        \diagram*{
            (x1) -- [color1, ghost] (v12) -- [color1, ghost] (x2),
            (x3) -- (v34) -- (x4),
            (x5) -- [color2, ghost] (v56) -- [color2, ghost] (x6),
            (v12) -- [color1doubghost, edge label' = {\scriptsize \;$\sigma'_{j_2}\sigma'_{j_2}$}, inner sep = 4pt] (b12) -- [color1, ghost, quarter left] (w12) -- [color1, ghost, quarter left] (b12),
            (v34) -- [photon, edge label = {\scriptsize $\widehat{\Delta G}_{j_1}$}, inner sep = 4pt] (b34) -- [quarter left] (w34) -- [quarter left] (b34),
            (v56) -- [color2doubghost, edge label = {\scriptsize \,$\sigma'_{j_k}\sigma'_{j_k}$}, inner sep = 4pt] (b56) -- [color2, ghost, quarter left] (w56) -- [color2, ghost, quarter left] (b56)
        };
        \tikzfeynmanset{every blob = {/tikz/fill=black!50, /tikz/minimum size=15pt}}
        \vertex[right = 0pt of d, blob, minimum size = 8pt] (dd) {{\scriptsize $\delta$}};
        \vertex[above = 25pt of d, dot, minimum size = 1pt] (d2) {};
        \vertex[left = 10pt of d2] (d2l) {};
        \vertex[below = 3pt of d2l, dot, minimum size = 1pt] (d1) {};
        \vertex[right = 10pt of d2] (d2r) {};
        \vertex[below = 3pt of d2r, dot, minimum size = 1pt] (d3) {};
        \end{feynman}
        \end{tikzpicture}
\end{eqnarray}
or, in analytical form
\begin{eqnarray}
    \delta A^{(\ell+1)}_{12\textcolor{color1}{34}\ldots \textcolor{color2}{(m-1)m}} &=& \sum_{\beta_{k}}(\chi_{||}^{(\ell+1)})_{12,\beta_{1}\beta_{2}}(\chi_{\perp}^{(\ell+1)})_{\textcolor{color1}{34}}\ldots (\chi_{\perp}^{(\ell+1)})_{\textcolor{color2}{(m-1)m}}\delta A^{(\ell)}_{\beta_{1}\beta_{2}\textcolor{color1}{34}\ldots \textcolor{color2}{(m-1)m}}
\end{eqnarray}
which shows $A$ stays stable once the criticality conditions studied previously are satisfied. 

\item Similarly, the variation of the tensor $B$ takes the form
\begin{eqnarray}
    \begin{tikzpicture}[baseline=(d.base)]
        \begin{feynman}
        \vertex (d) {};
        \vertex[left = 4pt of d, dot, minimum size = 0pt] (v12) {};
        \vertex[left = 16pt of v12] (l) {};
        \vertex[below = 8pt of l, dot, color1, minimum size = 3pt, label = {above: {\footnotesize $\textcolor{color1}{3}$}}] (x1) {};
        \vertex[above = 8pt of l, dot, color2, minimum size = 3pt, label = {above: {\footnotesize $\textcolor{color2}{(m-1)}$}}] (x2) {};
        \vertex[below = 4pt of d, dot, minimum size = 0pt] (v34) {};
        \vertex[below = 16pt of v34] (u) {};
        \vertex[left = 8pt of u, dot, minimum size = 3pt, label = {above: {\footnotesize $2$}}] (x3) {};
        \vertex[right = 8pt of u, dot, minimum size = 3pt, label = {above: {\footnotesize $1$}}] (x4) {};
        \vertex[right = 4pt of d, dot, minimum size = 0pt] (v56) {};
        \vertex[right = 16pt of v56, dot, minimum size = 0pt] (r) {};
        \vertex[above = 8pt of r, dot, color1, minimum size = 3pt, label = {above: {\footnotesize \hspace{7pt} $\textcolor{color1}{4}$}}] (x5) {};
        \vertex[below = 8pt of r, dot, color2, minimum size = 3pt, label = {below: {\footnotesize \hspace{7pt} $\textcolor{color2}{m}$}}] (x6) {};
        \diagram*{
            (x1) -- [color1, ghost] (v12) -- [color2, ghost] (x2),
            (x3) -- (v34) -- (x4),
            (x5) --[color1, ghost] (v56) -- [color2, ghost] (x6)
        };
        \tikzfeynmanset{every blob = {/tikz/fill=black!50, /tikz/minimum size=15pt}}
        \vertex[right = 0pt of d, blob, minimum size = 8pt] (dd) {{\scriptsize $\delta$}};
        \vertex[above = 15pt of d, dot, minimum size = 1pt] (d2) {};
        \vertex[left = 8pt of d2] (d2l) {};
        \vertex[below = 2pt of d2l, dot, minimum size = 1pt] (d1) {};
        \vertex[right = 8pt of d2] (d2r) {};
        \vertex[below = 2pt of d2r, dot, minimum size = 1pt] (d3) {};
        \end{feynman}
        \end{tikzpicture}
        \;\;\,=\; \sum_{j_1, \dots, j_k}
        \begin{tikzpicture}[baseline=(d.base)]
        \begin{feynman}
        \vertex (d) {};
        \vertex[left = 4pt of d, dot, minimum size = 0pt] (w12) {};
        \vertex[above = 1pt of w12, label = {above: {\scriptsize \hspace{-10pt} $ $}}] (w12d) {};
        \tikzfeynmanset{every blob = {/tikz/fill=white!50, /tikz/minimum size=15pt}}
        \vertex[left = 25pt of w12, blob] (b12) {};
        \vertex[left = 25pt of b12, dot, minimum size = 0pt] (v12) {};
        \vertex[left = 8pt of v12] (l) {};
        \vertex[below = 16pt of l, dot, color1, minimum size = 3pt, label = {below: {\footnotesize $\textcolor{color1}{3}$}}] (x1) {};
        \vertex[above = 16pt of l, dot, color2, minimum size = 3pt, label = {above: {\footnotesize $\textcolor{color2}{(m-1)}$}}] (x2) {};
        \vertex[below = 4pt of d, dot, minimum size = 0pt] (w34) {};
        \vertex[right = 13pt of w34, label = {above: {\scriptsize $ $}}, inner sep = -1pt] (w34r) {};
        \tikzfeynmanset{every blob = {/tikz/fill=white!50, /tikz/minimum size=15pt}}
        \vertex[below = 25pt of w34, blob] (b34) {};
        \vertex[below = 25pt of b34, dot, minimum size = 0pt] (v34) {};
        \vertex[below = 8pt of v34] (u) {};
        \vertex[left = 16pt of u, dot, minimum size = 3pt, label = {above: {\footnotesize $2$}}] (x3) {};
        \vertex[right = 16pt of u, dot, minimum size = 3pt, label = {above: {\footnotesize $1$}}] (x4) {};
        \vertex[right = 4pt of d, dot, minimum size = 0pt] (w56) {};
        \vertex[below = 1pt of w56, label = {below: {\scriptsize \hspace{18pt} $z_{j_1} $}}] (w56d) {};
        \tikzfeynmanset{every blob = {/tikz/fill=white!50, /tikz/minimum size=15pt}}
        \vertex[right = 25pt of w56, blob] (b56) {};
        \vertex[right = 25pt of b56, dot, minimum size = 0pt] (v56) {};
        \vertex[right = 8 pt of v56, dot, minimum size = 0pt] (r) {};
        \vertex[above = 16pt of r, dot, color1, minimum size = 3pt, label = {above: {\footnotesize $\textcolor{color1}{4}$}}] (x5) {};
        \vertex[below = 16pt of r, dot, color2, minimum size = 3pt, label = {below: {\footnotesize $\textcolor{color2}{m}$}}] (x6) {};
        \diagram*{
            (x1) -- [color1, ghost] (v12) -- [color2, ghost] (x2),
            (x3) -- (v34) -- (x4),
            (x5) -- [color1, ghost] (v56) -- [color2, ghost] (x6),
            (v12) -- [color2color1ghost, edge label' = {\scriptsize \;$\sigma'_{j_2}\sigma'_{j_2}$}, inner sep = 4pt] (b12) -- [color2, ghost, quarter left] (w12) -- [color1, ghost, quarter left] (b12),
            (v34) -- [photon, edge label = {\scriptsize $\widehat{\Delta G}_{j_1}$}, inner sep = 4pt] (b34) -- [quarter left] (w34) -- [quarter left] (b34),
            (v56) -- [color1color2ghost, edge label = {\scriptsize \,$\sigma'_{j_k}\sigma'_{j_k}$}, inner sep = 4pt] (b56) -- [color2, ghost, quarter left] (w56) -- [color1, ghost, quarter left] (b56)
        };
        \tikzfeynmanset{every blob = {/tikz/fill=black!50, /tikz/minimum size=15pt}}
        \vertex[right = 0pt of d, blob, minimum size = 8pt] (dd) {{\scriptsize $\delta$}};
        \vertex[above = 25pt of d, dot, minimum size = 1pt] (d2) {};
        \vertex[left = 10pt of d2] (d2l) {};
        \vertex[below = 3pt of d2l, dot, minimum size = 1pt] (d1) {};
        \vertex[right = 10pt of d2] (d2r) {};
        \vertex[below = 3pt of d2r, dot, minimum size = 1pt] (d3) {};
        \end{feynman}
        \end{tikzpicture}
\end{eqnarray}
which translates into the formal expression
\begin{eqnarray}
        \delta B^{(\ell+1)}_{12\textcolor{color1}{3}\textcolor{color2}{(m-1)}\ldots \textcolor{color1}{4}\textcolor{color2}{m}} &=& \sum_{\beta_{k}}(\chi_{||}^{(\ell+1)})_{12,\beta_{1}\beta_{2}}(\chi_{\perp}^{(\ell+1)})_{\textcolor{color1}{3}\textcolor{color2}{(m-1)}}\ldots (\chi_{\perp}^{(\ell+1)})_{\textcolor{color1}{4}\textcolor{color2}{m}}\delta B^{(\ell)}_{\beta_{1}\beta_{2}\textcolor{color1}{3}\textcolor{color2}{(m-1)}\ldots \textcolor{color1}{4}\textcolor{color2}{m}}
\end{eqnarray}
One sees again that the criticality conditions guarantee the stability of the higher-order tensor $B$.

\end{enumerate}

The leading-order contributions to the rank-$k$ tensors are $\mathcal{O}(1/n^{(k-2)/2})$. However, since the internal lines and propagators at leading order already exhaust all possible contributions to the variations and these were set to 1 by the infinite-width criticality conditions, we conclude that all higher order contributions to the tensors discussed here are stabilized as well.

Therefore, all preactivation- and NTK statistics are stabilized to all orders in $1/n$. This analysis demonstrates how the NTK Feynman rules can be used to obtain general all-order results.
\end{proof}

\section{Proof of Theorem~\ref{theoremfive}}
\label{app:proof-identity-relu}

\scaleinvariantactivations*

\begin{proof}
Scale-invariant activation functions are defined by
\begin{eqnarray}
    \sigma(z) =  \left\{
\begin{array}{ll}
      a_{+}z \, ,& z \geq 0 \\
      a_{-}z \, ,& z < 0 \\
\end{array} 
\right. 
\end{eqnarray}
Due to its simplicity, one can evaluate Gaussian expectation values containing polynomials of these functions. For instance, the Gaussians $\langle \sigma(z)\sigma(z)\rangle_{K}$, $\langle \sigma'(z)\sigma'(z)\rangle_{K}$, which define the tensor $\widehat{\Omega}$, can be exactly computed through the use of ordinary algebraic manipulations. The former takes the form
\begin{eqnarray}\label{sigmasigmascaleinv}
    \langle\sigma(z)\sigma(z)\rangle_{K} &=& \frac{1}{\sqrt{2\pi K}}\int_{-\infty}^{+\infty}e^{-\frac{z^{2}}{2K}}\sigma^{2}(z)dz  = \left(\frac{a_{+}^{2} + a_{-}^{2}}{2}\right)\frac{1}{\sqrt{2\pi K}}\int_{-\infty}^{+\infty}e^{-\frac{z^{2}}{2K}}z^{2}dz\nonumber\\
    &=& A\, K
\end{eqnarray}
where $A = \left(\frac{a_{+}^{2} + a_{-}^{2}}{2}\right)$, while the latter results in
\begin{eqnarray}\label{sigmaprimesigmaprimescaleinv}
    \langle\sigma'(z)\sigma'(z)\rangle_{K} &=& \frac{1}{\sqrt{2\pi K}}\int_{-\infty}^{+\infty}e^{-\frac{z^{2}}{2K}}\sigma'^{2}(z)dz  = \left(\frac{a_{+}^{2} - a_{-}^{2}}{2}\right)\frac{1}{\sqrt{2\pi K}}\int_{-\infty}^{+\infty}e^{-\frac{z^{2}}{2K}}dz\nonumber\\
    &=& \tilde{A}
\end{eqnarray} 
with $\tilde{A} = \left(\frac{a_{+}^{2} - a_{-}^{2}}{2}\right)$. 

These results immediately imply that the leading order correction in $\frac{1}{n}$ to the diagonal component of the NTK mean, given in~\eqref{firstcorrectionntk}, vanishes for the case under discussion: The second diagram in~\eqref{firstcorrectionntk} does not contribute since $K^{\{1\}(\ell)}=0$ for scale-invariant activation functions. The third diagram vanishes due to the property
\begin{align}
    2K^{2}\frac{d}{dK}\bigg[2K^{2}\frac{d}{dK}\langle \widehat{\Omega}(z)\rangle_{K}\bigg] - 8K^{3}\frac{d}{dK}\langle \widehat{\Omega}(z)\rangle_{K} &= 0\label{eq:11}
\end{align}
satisfied by the kernel $K=K(x,x)$ of scale-invariant activation functions, and the use of~\eqref{sigmasigmascaleinv} and ~\eqref{sigmaprimesigmaprimescaleinv}. Indeed, the Gaussian expectation value of the tensor $\widehat{\Omega}(z) = \sigma(z)\sigma(z) + C_{W} \Theta^{(l)} \sigma'(z)\sigma'(z)$ can easily be computed to be
\begin{eqnarray}
\langle \widehat{\Omega}(z)\rangle_{K} &=& A K + C_{W}\tilde{A} \Theta
\end{eqnarray}
which implies that $\frac{d}{dK}\langle \widehat{\Omega}(z)\rangle = A$. Consequently, \eqref{eq:11} holds identically. The fourth and fifth diagrams in~\eqref{firstcorrectionntk} vanish as a consequence of
\begin{align}
\langle \sigma'(z) \sigma'(z) (z^{2} - K)\rangle_{K} &=2K^{2} \frac{d}{dK}\langle \sigma'(z) \sigma'(z)\rangle_{K}\label{eq:9}
\end{align}
which, in turn, follows from ordinary integration techniques
\begin{eqnarray}\label{identitysecondcase}
    2K^{2}\frac{d}{dK}\langle \sigma'(z)\sigma'(z)\rangle_{K} &=& 2K^{2}\frac{d}{dK}\left( \frac{1}{\sqrt{2\pi K}}\int_{-\infty}^{+\infty}e^{-\frac{z^{2}}{2K}}\sigma'^{2}(z)dz\right)\nonumber\\
    &=& \int_{-\infty}^{+\infty}\frac{2K^{2}}{\sqrt{2\pi K}}\left(-\frac{1}{2K} + \frac{z^{2}}{2K^{2}}\right)\sigma'(z)^{2}dz\nonumber\\
    &=& \langle (z^{2} - K)\sigma'(z)\sigma'(z)\rangle_{K}
\end{eqnarray}
Applying~\eqref{sigmaprimesigmaprimescaleinv} then yields $\langle (z^{2} - K)\sigma'(z)\sigma'(z)\rangle_{K} = 0$ for scale-invariant nonlinearities. Therefore, the only diagram which contributes on the RHS of~\eqref{firstcorrectionntk} is that proportional to $\Theta^{\{1\}}$. Since the first-layer NTK mean does not develop any finite-width correction, one concludes by induction that the NTK mean does not receive finite-width corrections for scale-invariant nonlinearities.


The Feynman diagram approach now allows this argument to be extended to arbitrary order in $\frac{1}{n}$. To see this, as pointed out in~\ref{proofoftheoremthree}, observe that any internal lines connected to a propagator must appear in pairs, that is, the number of such lines is always even. According to Feynman rule (iii)-(c) in Appendix~\ref{app:feynman_rules}, each of these lines corresponds to a derivative with respect to the pre-activation variable $z$. In other words, the effect of attaching internal lines is to apply derivatives of the form $\frac{d^{2m}}{dz^{2m}}$ to the function $\Phi(z)$ inside a Gaussian integral. This operation can be translated into a more tractable form using the identity
\begin{align}
K^{2m}\left\langle \frac{d^{2m}}{dz^{2m}}\Phi(z)\right\rangle_{K} &= 2^{m}K^{2m}\frac{d^{m}}{dK^{m}}\left\langle \Phi(z)\right\rangle_{K} 
\end{align}
where $\langle \cdot \rangle_{K}$ denotes the Gaussian expectation with variance $K$. This formula allows one to express high-order derivatives with respect to $z$ as derivatives with respect to $K$, simplifying the calculation. It generalizes equations~\eqref{eq:11} and~\eqref{identitysecondcase}, which correspond to the cases $m=2$ and $m=1$, respectively.

Next, we consider which interactions contribute to the NTK mean. Among the cubic interactions, only two types involve two external NTK-dotted lines: one where $\Phi = \widehat{\Delta \Omega}$, and another where $\Phi = \sigma'\sigma'$. For scale-invariant activation functions, the Gaussian expectation values of both of these expressions are constants, or equivalently, they do not depend on $K$. As a consequence, their derivatives with respect to $K$ vanish, and so do all corresponding higher-order diagrammatic contributions.

The only surviving contribution is the one proportional to the diagonal component of the NTK mean in the previous layer. This is consistent with the first-order correction shown in equation~\eqref{firstcorrectionntk}, and it implies that any higher-order correction to the NTK mean at order $\frac{1}{n^{k}}$ is simply proportional to the same order correction in the preceding layer.

Finally, since the NTK in the first layer is deterministic and does not fluctuate, there is no correction to propagate. We therefore conclude that the diagonal component of the NTK mean remains exact at all orders in $\frac{1}{n}$ for scale-invariant activation functions.
\end{proof}



\section{Implementation Details}
\label{app:implementation}

\subsection{Recursion Relations}
As laid out in Section \ref{sec:experiments}, we first define the recursions symbolically with custom \texttt{SymPy} classes and perform several manipulations in favor of numerical performance. Afterwards, the expressions are dynamically translated into numerical functions that call \texttt{NumPy} \cite{numpy2020} and \texttt{SciPy} \cite{scipy2020} routines.

\textbf{Symbolic Definition and Manipulations.} In addition to the classes provided by \texttt{SymPy}, we implement a specialized class for the Gaussian expectations appearing in all recursions as well as a distinct matrix class for the covariance matrix $K^{(\ell)}$. First, it is checked if the Gaussian expectation can be simplified using integration by parts. This was already performed in all recursions presented here but e.g.\ not in the recursions presented in \cite{roberts2022}. Afterwards, where applicable, derivatives as well as matrix multiplications of $K^{(\ell)}$ with its inverse are rewritten in terms of Kronecker deltas. Lastly, all present sums are checked if they collapse as a result of Kronecker deltas. These contractions increase the length of the symbolic expression but lead to fewer terms that need to be evaluated.

\textbf{Dynamic Conversion to Numerical Integration.} \texttt{SymPy} provides the \texttt{lambdify} function to translate its symbolic expressions into numerical functions using e.g. \texttt{SciPy}. We build on this functionality but introduce a custom translator, a so-called \texttt{Printer} in the \texttt{SymPy} package. Its main purpose is to replace the Gaussian expectation with a custom numerical function that detects integrand symmetry, marginalizes the normal distribution over dimensions that the integrand does not depend on and caches all results to speed up possible repeated evaluation of the same expectation. The numerical integration is then handled by the \texttt{scipy.integrate.cubature} function.

\subsection{Monte--Carlo Estimation}

Individual networks are initialized in \texttt{JAX} using \texttt{neural-tangent}'s \texttt{stax} layers. The computation of NTK-related statistics requires the Jacobian $\frac{\partial z_i^{(\ell)}(x_\alpha)}{\partial \theta_\mu}$ and can be computed using \texttt{JAX}'s automatic differentiation. However, since the Jacobian is expensive both in computation and in memory for wider networks, we compute each layer's statistics one at a time and then continue recursively to deeper layers using the chain rule. The sampling is parallelized over multiple initializations using \texttt{vmap}.

\textbf{The Kernels.} The mean NTK and NNGP are estimated by averaging over $N_\text{net}$ initializations of the network, i.e.\
\begin{align}
  \overline{K}^{(\ell)}_{\alpha \beta} &= \frac{1}{N_\text{net}} \sum_{I=1}^{N_\text{net}} z^{(\ell)}_{I;i,\alpha} z^{(\ell)}_{I;i,\beta} \,, \label{eq:nngp_mc}\\
  \overline{\Theta}^{(\ell)}_{\alpha \beta} &= \frac{1}{N_\text{net}} \sum_{I=1}^{N_\text{net}} \left( \sum_{\mu}\frac{\partial z^{(\ell)}_{I;i,\alpha}}{\partial\theta_{\mu}}\frac{\partial z^{(\ell)}_{I; i, \beta}}{\partial\theta_{\mu}}\right) \,, \label{eq:ntk_mc}
\end{align}
where $i$ is an arbitrary but fixed channel in the corresponding layer and $I$ labels the different initializations. The error bars correspond to the standard deviation of the mean and are estimated by dividing the sample standard deviation of the individual samples by $\sqrt{N_\text{net}}$. 
To improve sample efficiency, we exploit the $i$ independence for more expensive experiments and average over the diagonal
\begin{align}
  \overline{K}^{(\ell)}_{\alpha \beta} &= \frac{1}{N_\text{net}} \sum_{I=1}^{N_\text{net}} \frac{1}{n_\ell} \sum_{i=1}^{n_{\ell}}z^{(\ell)}_{I;i,\alpha} z^{(\ell)}_{I;i,\beta} \, ,\label{eq:nngp_trace_avg} \\
  \overline{\Theta}^{(\ell)}_{\alpha \beta} &= \frac{1}{N_\text{net}} \sum_{I=1}^{N_\text{net}} \left( \frac{1}{n_\ell} \sum_{i=1}^{n_\ell} \left( \sum_{\mu}\frac{\partial z^{(\ell)}_{I;i,\alpha}}{\partial\theta_{\mu}}\frac{\partial z^{(\ell)}_{I; i, \beta}}{\partial\theta_{\mu}}\right) \right) \,.
  \label{eq:ntk_trace_avg}
\end{align}

\textbf{The Four-Point Cumulant $V$.} Using \eqref{eq:nngp_mc}, we estimate the four-point cumulant $V$ to first order by
\begin{equation}
    \overline{V}^{(\ell)}_{\alpha \beta \gamma \delta} = \left(\frac{1}{N_\text{net}} \sum_{I=1}^{N_\text{net}} \frac{n_{\ell -1}}{n_{\ell} (n_{\ell}-1)} \sum_{\substack{i, j = 1 \\i \ne j}}^{n_\ell} z^{(\ell)}_{I;i,\alpha} z^{(\ell)}_{I;i,\beta} z^{(\ell)}_{I;j,\gamma} z^{(\ell)}_{I;j,\delta}\right) - n_{\ell-1}\overline{K}^{(\ell)}_{\alpha \beta} \overline{K}^{(\ell)}_{\gamma \delta}+ \mathcal{O}\left(\frac{1}{n}\right) \,. \label{eq:v4_mc}
\end{equation}
In order to obtain statistics of this estimate, we repeat the Monte--Carlo sampling $N_\text{stats}$ times (for both $V$ and $K$) and take the corresponding mean and standard deviation as the estimate for $V$ and its error, respectively. Here, we use the diagonal-averaged version to estimate $\overline{K}^{(\ell)}$ given in \eqref{eq:nngp_trace_avg}.

\textbf{The Tensors $A, B, D$ and $F$.} Similarly to $V$, the tensors are estimated to first order by
\begin{align}
  \overline{A}^{(\ell)}_{\alpha\beta\gamma\delta} &= \frac{1}{N_\text{net}}\sum_{I=1}^{N_\text{net}}\frac{n_{\ell-1}}{n_{\ell}^2}\sum_{i,j=1}^{n_{\ell}}\widehat{\Delta\Theta}^{(\ell)}_{I;ii,\alpha\beta}\widehat{\Delta \Theta}_{I;jj,\gamma\delta}^{(\ell)} + \mathcal{O}\left(\frac{1}{n}\right) \, , \label{eq:A_mc}\\
  \overline{B}^{(\ell)}_{\alpha\gamma\beta\delta} &= \frac{1}{N_\text{net}}\sum_{I=1}^{N_\text{net}}\frac{n_{\ell-1}}{n_{\ell}^2}\sum_{i,j=1}^{n_{\ell}}\widehat{\Delta\Theta}^{(\ell)}_{I;ij,\alpha\beta}\widehat{\Delta \Theta}_{I;ij,\gamma\delta}^{(\ell)} + \mathcal{O}\left(\frac{1}{n}\right) \, , \label{eq:B_mc}\\
  \overline{D}^{(\ell)}_{\alpha\beta\gamma\delta} &= \frac{1}{N_\text{net}}\sum_{I=1}^{N_\text{net}}\frac{n_{\ell-1}}{n_{\ell}^2}\sum_{i,j=1}^{n_{\ell}}z_{I;i,\alpha}^{(\ell)}z_{I;i,\beta}^{(\ell)}\widehat{\Delta \Theta}_{I;jj,\gamma\delta}^{(\ell)} + \mathcal{O}\left(\frac{1}{n}\right) \, , \label{eq:D_mc}\\
  \overline{F}^{(\ell)}_{\alpha\gamma\beta\delta} &= \frac{1}{N_\text{net}}\sum_{I=1}^{N_\text{net}}\frac{n_{\ell-1}}{n_{\ell}^2}\sum_{i,j=1}^{n_{\ell}}z_{I;i,\alpha}^{(\ell)}z_{I;j,\beta}^{(\ell)}\widehat{\Delta \Theta}_{I;ij,\gamma\delta}^{(\ell)} + \mathcal{O}\left(\frac{1}{n}\right) \, ,\label{eq:F_mc}
\end{align}
where the NTK fluctuation $\widehat{\Delta \Theta}^{(\ell)}_{ij,\alpha \beta}$ was already introduced in Section \ref{sec-finitewidth-corrections-NTK} and is computed with respect to the diagonal-averaged mean from \eqref{eq:ntk_trace_avg}. Again, the computation is repeated $N_\text{stats}$ times to obtain the mean and standard deviation of the estimate.

\section{Additional Experiments and Details}
\label{app:experiments}

In this section, we provide all the details to the experiments presented in Section \ref{sec:experiments} and additional results supporting our theory.
For all results shown below, the empirical quantities were sampled from an ensemble of feedforward neural networks initialized according to \eqref{eq:nn_initialization}.

\subsection{Computation of the Finite-Width Recursions}
In the following, we provide network and input details to the experiment shown in Figure \ref{fig:kernel_finite_width_corrected}. In addition, we also show the results of the solved recursions for the remaining tensors at leading order $\frac{1}{n}$.

\textbf{Setup.} All recursions are solved for a four layer GeLU MLP at various widths up to $n_\ell = \num{220}$. The weights' variance is set to $C_W^{(\ell)} = \num{1.98305826}$ to improve numerical stability with depth. The numerical integration of the Gaussian expectations is performed with \texttt{SciPy}'s \texttt{cubature} function using the parameters \texttt{rtol=0.001}, \texttt{max\_subdivisions=10000} and \texttt{rule="genz-malik"} for multidimensional integrals and \texttt{rule="gauss-kronrod"} for 1$d$ integrals. The input components were sampled from a standard normal distribution and rescaled by a factor of \num{0.53}, resulting in
\begin{alignat}{4}
  &x_0 = ( && +\num{0.6540248765053858}, && -\num{1.3592788739383235}, \nonumber \\
  & && -\num{0.3791179316445409}, && +\num{0.34191025226280697}), \\
  &x_1 = (&& -\num{0.6786403121150224}, && -\num{0.12161213502515096}, \nonumber \\
  & && +\num{0.21716372787141658}, && -\num{1.4709063616118982}), \\
  &x_2 = (&& +\num{0.2684047246058049}, && +\num{0.19352982216329817}, \nonumber \\
  & && -\num{0.18546381464689993}, && -\num{0.8116446646575691}), \\
  &x_3 = (&& +\num{0.06801939825322936}, && -\num{0.25387687345791304}, \nonumber \\
  & && -\num{0.8409923450288397}, && -\num{0.6282339955619969}).
\end{alignat}
The rescaling factor renders the diagonal kernel corrections more pronounced, see the GeLU related discussion in subsection \ref{ssub_app:kernel_corrections}. Since the empirical estimates tend to suffer from numerical noise at larger widths, we initialized the networks to double precision.

\textbf{Single-Input NNGP and NTK.} In addition to the multi-input case shown in Figure \ref{fig:kernel_finite_width_corrected} we also show the convergence in the diagonal components in Figure \ref{fig:both_kernels_finite_width_corrected_diagonal}. Also here we observe close agreement between the sampled and the corrected kernels at widths $n_\ell > 20$. For sampling, equations \eqref{eq:ntk_mc} and \eqref{eq:nngp_mc} were used.
\begin{figure}
  \begin{center}
    \includegraphics[width=0.85\textwidth]{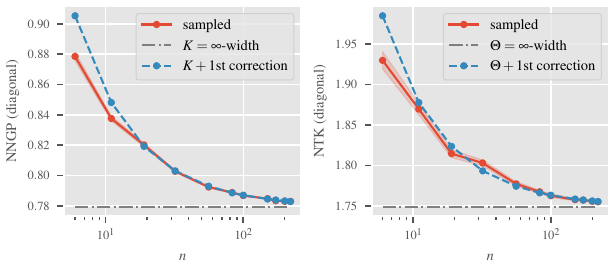}
  \end{center}
  \caption{\emph{Finite-Width Corrected Kernels (Single-Input).} The Monte--Carlo (MC) estimated diagonal entry NNGP $\overline{K}_{00}$ and NTK $\overline{\Theta}_{00}$ (red) at the fourth layer of a GeLU-MLP are shown at different hidden layer widths $n=n_\ell$ and compared to the first-order corrected finite-width solution $K^{(\ell)}_{00} +K^{\{1\}(\ell)}_{00} / n_\ell$ and $\Theta^{(\ell)}_{00} + \Theta^{\{1\}(\ell)}_{00} / n_\ell$ (blue), respectively, as well as to infinite-width results (gray). Sample sizes for the MC estimates of the NNGP and NTK are \num{e6} and \num{e5}, respectively. Error bars are included, but mostly covered by the mean line.}
  \label{fig:both_kernels_finite_width_corrected_diagonal}
\end{figure}

\textbf{First-Order Corrections to the NNGP and NTK in Isolation.} Although Figure \ref{fig:kernel_finite_width_corrected} and \ref{fig:both_kernels_finite_width_corrected_diagonal} already show highly improved accuracy of the corrected analytic kernels, the exactness of the first-order corrections $\Theta^{\{1\}}$ and $K^{\{1\}}$ is difficult to evaluate in those plots due to the suppression of the higher order term by $1/n$ compared to the infinite-width term. Thus, we isolate the correction by estimating
\begin{equation}
    n_\ell \cdot \mathbb{E}\left[\widehat{\Theta}^{(\ell)}_{\alpha \beta} - \Theta^{(\ell)}_{\alpha \beta}\right] = \Theta^{\{1\}(\ell)}_{\alpha \beta} + \mathcal{O}\left(\frac{1}{n}\right) \quad \text{and} \quad n_\ell \cdot \mathbb{E}\left[\widehat{K}^{(\ell)}_{\alpha \beta} - K^{(\ell)}_{\alpha \beta}\right] = K^{\{1\}(\ell)}_{\alpha \beta} + \mathcal{O}\left(\frac{1}{n}\right)\, .
\end{equation}
The result is shown in Figure \ref{fig:kernel_corrections} and further confirms the correctness of the first-order correction. Note that the width-convergence behaviour is expected because the deviation from the infinite-width solution contains \emph{all} higher-order terms. We also emphasize that the sampling is becoming numerically sensitive at larger widths which is further amplified by the multiplication of the deviations by $n_\ell$ in particular towards the right side of the plots.
\begin{figure}
  \begin{center}
    \includegraphics[width=0.85\textwidth]{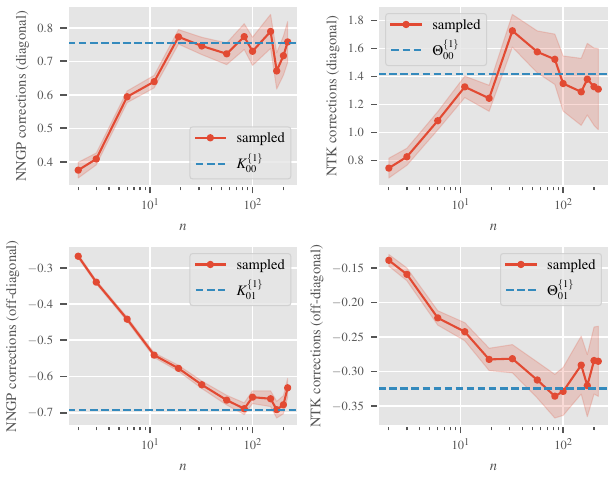}
  \end{center}
  \caption{\emph{Convergence of Finite-Width Deviations to the First-Order Corrections.}  The Monte--Carlo estimated finite width deviations of the NNGP and NTK  $n \cdot (\overline{K}_{\alpha \beta} - K_{\alpha \beta})$ and $n \cdot (\overline{\Theta}_{\alpha \beta} - \Theta_{\alpha \beta})$ (red) at the fourth layer of a GeLU-MLP are shown at different hidden layer widths $n=n_\ell$ and compared to the first order corrections $K^{\{1\}(\ell)}_{\alpha \beta}$ and $\Theta^{\{1\}(\ell)}_{\alpha \beta}$ (blue), respectively. The sample sizes are \num{e6} for the NNGP and \num{e5} for the NTK.}
  \label{fig:kernel_corrections}
\end{figure}

\textbf{Additional Recursions.} For completeness, we provide the convergence results of the tensors $V, D$ and $F$, necessary to solve the recursions for the NNGP and NTK corrections in Figures \ref{fig:V4_ana_vs_stat}, \ref{fig:D_ana_vs_stat} and \ref{fig:F_ana_vs_stat}. In addition, the convergence to the solved recursions for tensors $A$ and $B$, by using \eqref{eq:A_recursion_algebraic} and \eqref{eq:B_recursion_algebraic}, respectively, are provided in Figures \ref{fig:A_ana_vs_stat} and \ref{fig:B_ana_vs_stat}. The empirical estimates are computed using equations \eqref{eq:v4_mc} -- \eqref{eq:F_mc}.

\begin{figure}
  \begin{center}
    \includegraphics[width=0.99\textwidth]{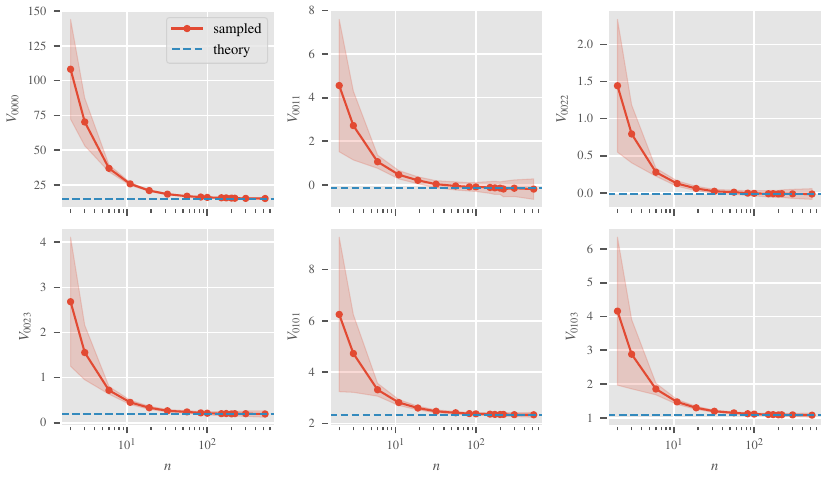}
  \end{center}
  \caption{\emph{$V$ Tensor Convergence.} Selected components of the Monte--Carlo estimated $\overline{V}$ tensor (red) at the fourth layer of a GeLU-MLP are shown at different hidden layer widths $n=n_\ell$ and compared to its analytic leading order result of $V$ (blue). $\overline{V}$ is computed from $N_\text{net} = 10^5$ initializations and its mean and error bars are estimated from $N_\text{stats} = 100$ repetitions.}
  \label{fig:V4_ana_vs_stat}
\end{figure}

\begin{figure}
  \begin{center}
    \includegraphics[width=0.99\textwidth]{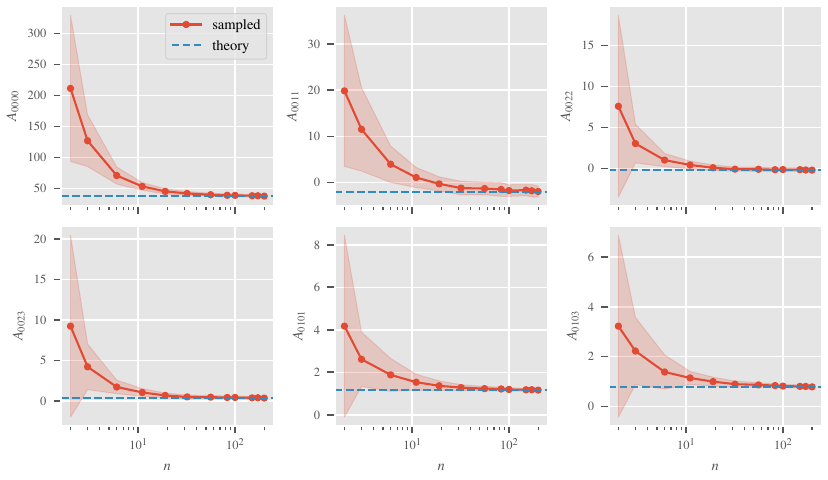}
  \end{center}
  \caption{\emph{$A$ Tensor Convergence.} Selected components of the Monte--Carlo estimated $\overline{A}$ tensor (red) at the fourth layer of a GeLU-MLP are shown at different hidden layer widths $n=n_\ell$ and compared to its analytic leading order result $A$ (blue). $\overline{A}$ is computed from $N_\text{net} = 1000$ initializations and its mean and error bars are estimated from $N_\text{stats} = 100$ repetitions.}
  \label{fig:A_ana_vs_stat}
\end{figure}

\begin{figure}
  \begin{center}
    \includegraphics[width=0.99\textwidth]{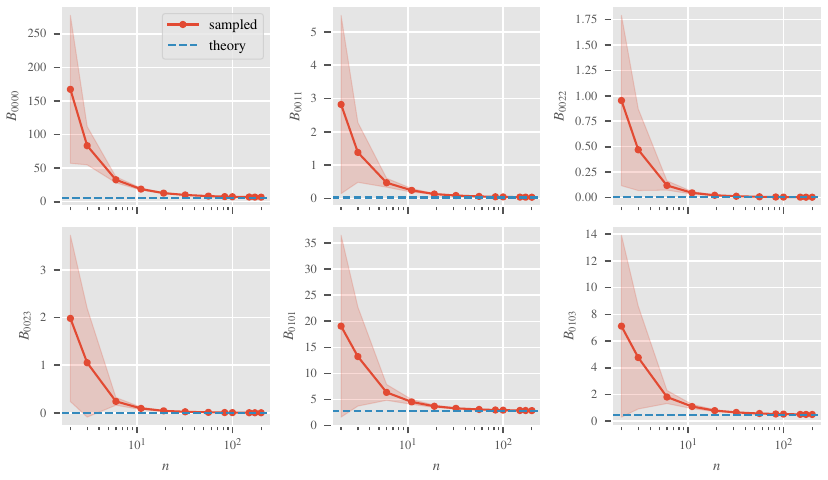}
  \end{center}
  \caption{\emph{$B$ Tensor Convergence.} Selected components of the Monte--Carlo estimated $\overline{B}$ tensor (red) at the fourth layer of a GeLU-MLP are shown at different hidden layer widths $n=n_\ell$ and compared to its analytic leading order result $B$ (blue). $\overline{B}$ is computed from $N_\text{net} = 1000$ initializations and its mean and error bars are estimated from $N_\text{stats} = 100$ repetitions.}
  \label{fig:B_ana_vs_stat}
\end{figure}

 \begin{figure}
  \begin{center}
    \includegraphics[width=0.99\textwidth]{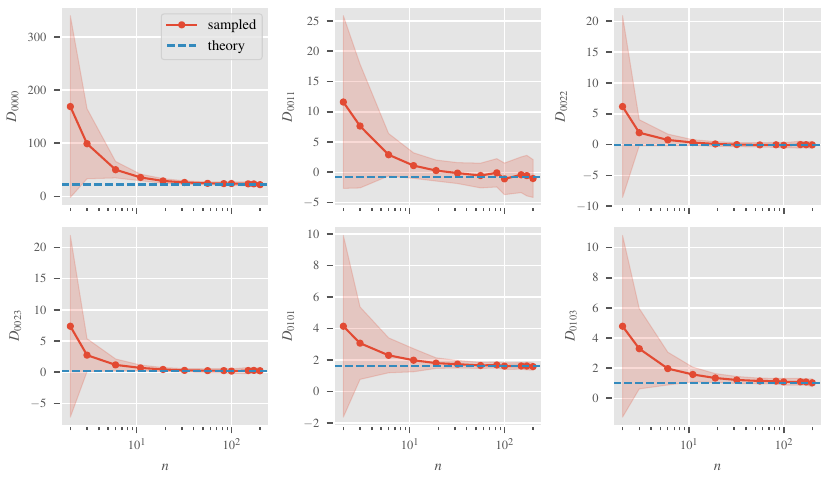}
  \end{center}
  \caption{\emph{$D$ Tensor Convergence.} Selected components of the Monte--Carlo estimated $\overline{D}$ tensor (red) at the fourth layer of a GeLU-MLP are shown at different hidden layer widths $n=n_\ell$ and compared to its analytic leading order result $D$ (blue). $\overline{D}$ is computed from $N_\text{net} = 1000$ initializations and its mean and error bars are estimated from $N_\text{stats} = 100$ repetitions.}
  \label{fig:D_ana_vs_stat}
\end{figure}

\begin{figure}
  \begin{center}
    \includegraphics[width=0.99\textwidth]{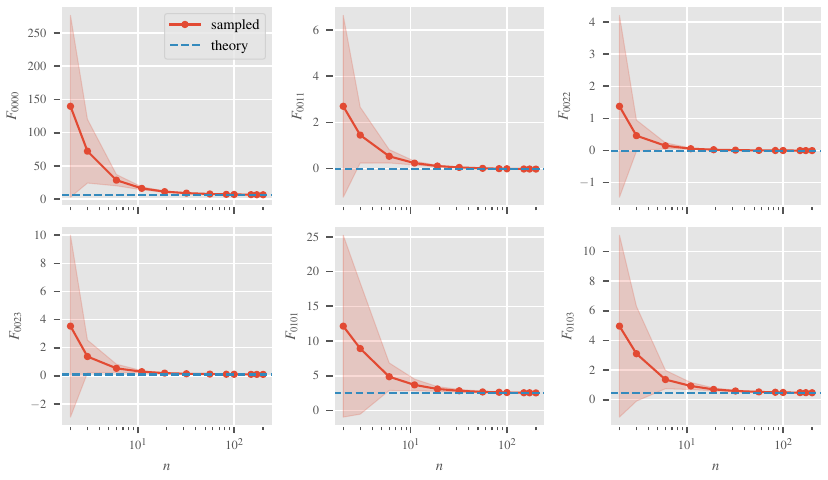}
  \end{center}
  \caption{\emph{$F$ Tensor Convergence.} Selected components of the Monte--Carlo estimated $\overline{F}$ tensor (red) at the fourth layer of a GeLU-MLP are shown at different hidden layer widths $n=n_\ell$ and compared to its analytic leading order result $F$ (blue). $\overline{F}$ is computed from $N_\text{net} = 1000$ initializations and its mean and error bars are estimated from $N_\text{stats} = 100$ repetitions.}
  \label{fig:F_ana_vs_stat}
\end{figure}

\subsection{Stability Analysis}

In this subsection we provide additional experiments verifying the simultaneous stability of the NNGP $K$, the four-point cumulant $V$ as well as the four tensors $A, B, D$ and $F$ with increasing network depth $\ell$. Furthermore, we provide details of the numerical setup.

\textbf{Setup.} All experiments in this section are performed on a ReLU MLP without biases.
For the inputs, each component of all vectors was drawn from a standard normal distribution. For the tensors $A, B, D$ and $F$ we thus used
\begin{alignat}{4}
  & x_0 = ( &&-\num{0.9895229339599609}, && -\num{0.5992491841316223}), \label{eq:x0_input} \\
  & x_1 = ( &&-\num{0.17877478897571564}, && +\num{2.253682851791382}), \label{eq:x1_input} \\
  & x_2 = ( &&+\num{1.0237634181976318}, && -\num{0.4618060886859894}), \label{eq:x2_input} \\
  & x_3 = ( &&-\num{0.5364212393760681}, && +\num{1.9298086166381836}) \label{eq:x3_input}.
\end{alignat}

For the two-dimensional tensors $K$ and $\Theta$, only $x_0$ and $x_1$ were used. The $V$ computation involved 4-dimensional inputs resulting in
\begin{alignat}{4}
  &x_0 = ( && -\num{0.9895229339599609}, && -\num{0.5992491841316223}, \nonumber \\
  & && -\num{0.17877478897571564}, && +\num{2.253682851791382}), \\
  &x_1 = (&& +\num{1.0237634181976318}, && -\num{0.4618060886859894}, \nonumber \\
  & && -\num{0.5364212393760681}, && +\num{1.9298086166381836}), \\
  &x_2 = (&& -\num{1.95197594165802}, && -\num{1.7220025062561035}, \nonumber \\
  & && +\num{0.8821542859077454}, && -\num{1.1963286399841309}), \\
  &x_3 = (&& -\num{0.1944112777709961}, && -\num{0.540934145450592}, \nonumber \\ 
  & && +\num{0.3308846354484558}, && +\num{0.0907512903213501}).
\end{alignat}


\textbf{Linear Scaling of the NTK $\Theta$ at Criticality.} The purple line in Figure \ref{fig:ntk_comparison} (middle) shows the predicted linear scaling of the single-input NTK component for scale-invariant activation functions when setting $C_W^{(\ell)} = C_W^\mathrm{c}$ \cite{roberts2022}. For a bias-free ReLU MLP, it is given by
\begin{equation}
  \Theta^{(\ell)}_{\alpha \alpha} = \frac{1}{n_0} \|x_\alpha\|^2  \ell \,.
\end{equation}
The NTK was estimated using equation \eqref{eq:ntk_mc}.

\begin{figure}
    \begin{center}
        \includegraphics[width=0.99\textwidth]{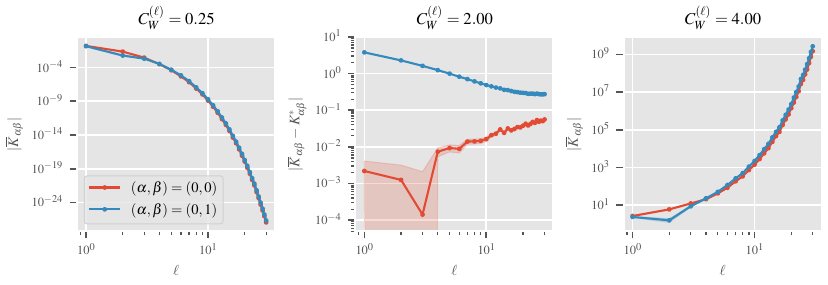}
    \end{center}
    \caption{\emph{Variance Stability.} Components of the Monte--Carlo estimated NNGP $\overline{K}_{\alpha \beta}$ of a ReLU MLP as a function of layer depth $\ell$ corresponding to single and dinstinct inputs are shown for three different choices of $C_W^{(\ell)}$. The hidden layers are of size \num{200}. The middle plot shows the deviation of the NNGP component to the predicted fixed point kernel $K^*_{\alpha \beta}$ at the critical value $C_W^{(\ell)} = C_{W}^\mathrm{c} = 2$ (see text). Sample means are obtained from \num{1000} initializations for the non-critical cases (left and right), whereas the result in the middle is obtained from a sample size of $N_\text{net} = \num{e6}$. Error bars are included in all three plots (see text).}\label{fig:nngp_comparison}
\end{figure}
\textbf{Covariance (NNGP) Stability.} For completeness, in Figure \ref{fig:nngp_comparison} we show the stability analysis of the covariance corresponding to the same setup that was used for the gradient analysis shown in Figure \ref{fig:ntk_comparison}, verifying the simultaneous stability of the NNGP and the NTK at the critical value $C_W^\mathrm{c}$. The NNGP was estimated according to equation \eqref{eq:nngp_mc}. Note that the covariance has already been studied numerically in \cite{banta2024}. As argued in \cite{roberts2022,halverson2021,banta2024}, there exists a non-trivial fixed point kernel $K^*_{\alpha \beta} \notin \{0, \infty\}$ for a ReLU MLP at criticality. It is given by
\begin{equation}
  K^*_{\alpha \beta} = \frac{C_W^\mathrm{c}}{n_0} \|x_\alpha\| \|x_\beta\| \, ,
  \label{eq:fixed_point_kernel_relu}
\end{equation}
where $n_0$ denotes the dimension of the input space, and is subtracted from the estimated NNGP in Figure \ref{fig:nngp_comparison} (middle).

\textbf{Four-Point Cumulant $V$.} Due to the symmetries
\begin{equation}
  V^{(\ell)}_{\alpha \beta \gamma \delta} = V^{(\ell)}_{\beta \alpha \gamma \delta} = V^{(\ell)}_{\alpha \beta \delta \gamma} = V^{(\ell)}_{\gamma \delta \alpha \beta} \, ,
  \label{eq:v4_symmetries}
\end{equation}
there are 55 independent components of this tensor. We compute the same subset as presented in \cite{banta2024} following equation \eqref{eq:v4_mc}.
The results for up to $\ell = 30$ are shown in Figure \ref{fig:v4_comparison_critical} for the critical case and in Figure \ref{fig:v4_comparison_non_critical} for the non-critical case. In addition to the result in \cite{banta2024}, we also provide regressions of the asymptotic power laws in the critical case, confirming the stability of $V$.
\begin{figure}
    \begin{center}
        \includegraphics[width=0.99\textwidth]{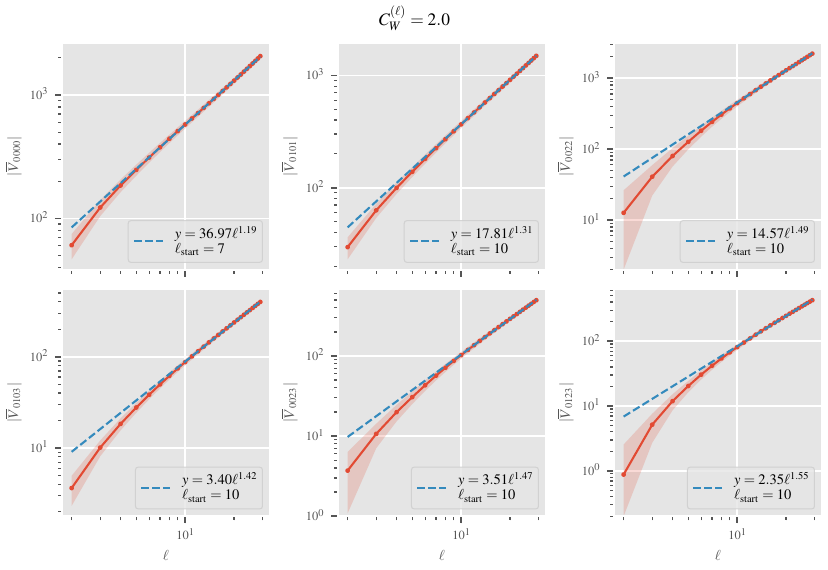}
    \end{center}
    \caption{\emph{Stability of the Four-Point Cumulant $V$ at Criticality.} Selected components of the Monte--Carlo estimated tensor $\overline{V}$ of a ReLU MLP as a function of layer depth $\ell$ are shown for the critical value $C_W^{(\ell)} = C_W^\mathrm{c} = 2$. The hidden layers are of size \num{300}. A regression of the asymptotic power law is shown in blue. The lowest layer included in the regression is given by $\ell_\text{start}$. $\overline{V}$ is computed from $N_\text{net} = \num{e4}$ initializations and its mean and error bars are estimated from $N_\text{stats} = \num{1000}$ repetitions (see text).}\label{fig:v4_comparison_critical}
\end{figure}

\begin{figure}
  \begin{center}
    \includegraphics[width=0.95\textwidth]{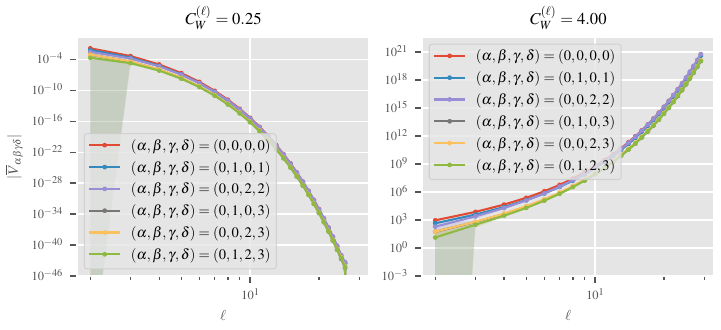}
  \end{center}
\caption{\emph{Instability of the Four-Point Cumulant $V$ Away from Criticality.} Selected components of the Monte--Carlo estimated tensor $\overline{V}$ are shown for $C_W^{(\ell} < C_W^\mathrm{c}$ (left) and $C_W^{(\ell} > C_W^\mathrm{c}$ (right). $\overline{V}$ is computed from $N_\text{net} = \num{e4}$ initializations and its mean and error bars are estimated from  $N_\text{stats} = \num{1000}$ repetitions (see text).}\label{fig:v4_comparison_non_critical}
\end{figure}

 \textbf{Tensors $A, B, D$ and $F$.}
 For each of these tensors, the critical and non-critical cases are computed for the same input indices as for $V$ up to $\ell = 30$ following equations \eqref{eq:A_mc} -- \eqref{eq:F_mc}. The results for tensor $A$ are shown in Figures \ref{fig:A_comparison_critical} \& \ref{fig:A_comparison_non_critical}, for tensor $B$ in Figures \ref{fig:B_comparison_critical} \& \ref{fig:B_comparison_non_critical}, for tensor $D$ in Figures \ref{fig:D_comparison_critical} \& \ref{fig:D_comparison_non_critical} and for tensor $F$ in Figures \ref{fig:F_comparison_critical} \& \ref{fig:F_comparison_non_critical}. The regression of the asymptotic power laws is shown in blue for the critical cases. The power-law behavior at criticality for all tested components confirms that the NTK-tensors $A$, $B$, $D$ and $F$ are also stabilized by the critical initialization variance of the infinite width limit.
\begin{figure}
    \begin{center}
        \includegraphics[width=0.99\textwidth]{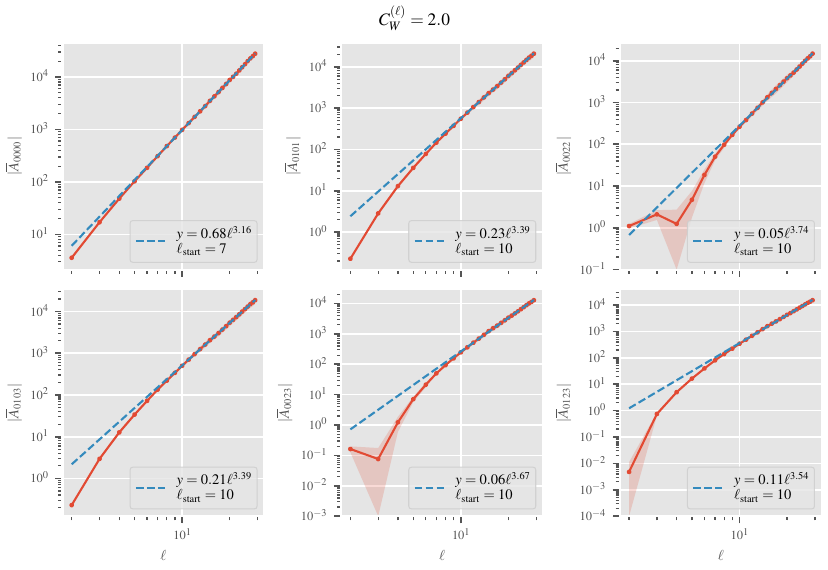}
    \end{center}
    \caption{\emph{Stability of the $A$ Tensor at Criticality.} Selected components of the Monte--Carlo estimated tensor $\overline{A}$ of a ReLU MLP as a function of layer depth $\ell$ are shown for the critical value $C_W^{(\ell)} = C_W^\mathrm{c} = 2$. The hidden layers are of size \num{200}. A regression of the asymptotic power law is shown in blue. The lowest layer included in the regression is given by $\ell_\text{start}$. $\overline{A}$ is computed from $N_\text{net} = \num{600}$ initializations and its mean and error bars are estimated from $N_\text{stats} = \num{10}$ repetitions (see text).}\label{fig:A_comparison_critical}
\end{figure}

\begin{figure}
  \begin{center}
    \includegraphics[width=0.95\textwidth]{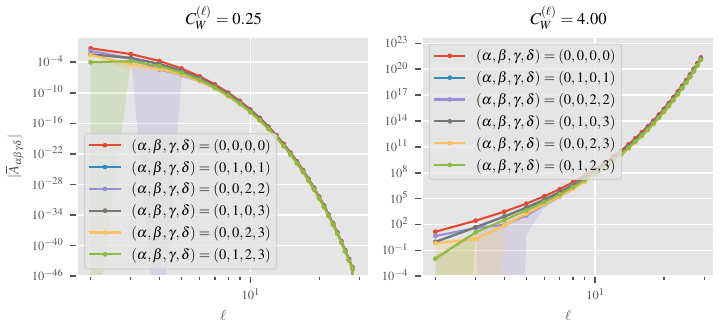}
  \end{center}
  \caption{\emph{Instability of the $A$ Tensor Away from Criticality.} Selected components of the Monte--Carlo estimated tensor $\overline{A}$ are shown for $C_W^{(\ell} < C_W^\mathrm{c}$ (left) and $C_W^{(\ell} > C_W^\mathrm{c}$ (right). $\overline{A}$ is computed from $N_\text{net} = \num{600}$ initializations and its mean and error bars are estimated from  $N_\text{stats} = \num{10}$ repetitions (see text).}\label{fig:A_comparison_non_critical}
\end{figure}

\begin{figure}
    \begin{center}
        \includegraphics[width=0.99\textwidth]{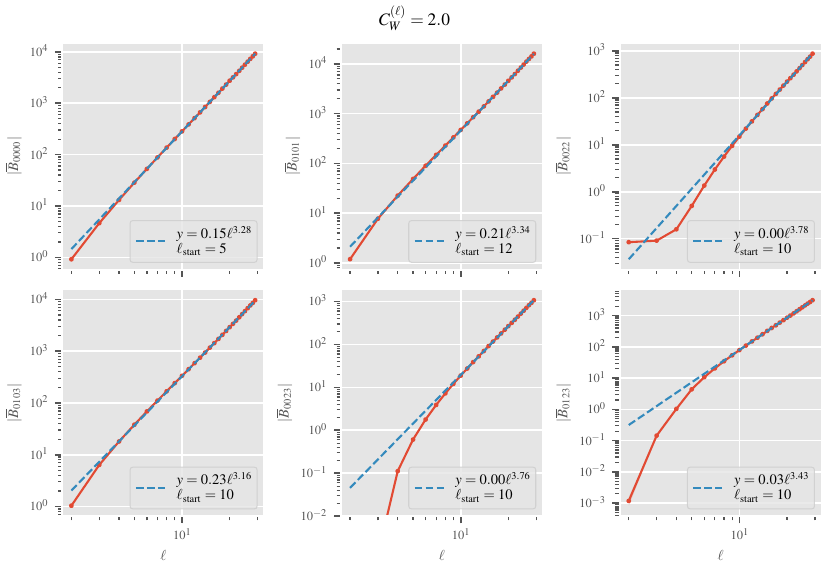}
    \end{center}
    \caption{\emph{Stability of the $B$ Tensor at Criticality.} Selected components of the Monte--Carlo estimated tensor $\overline{B}$ of a ReLU MLP as a function of layer depth $\ell$ are shown for the critical value $C_W^{(\ell)} = C_W^\mathrm{c} = 2$. The hidden layers are of size \num{200}. A regression of the asymptotic power law is shown in blue. The lowest layer included in the regression is given by $\ell_\text{start}$. $\overline{B}$ is computed from $N_\text{net} = \num{600}$ initializations and its mean and error bars are estimated from $N_\text{stats} = \num{10}$ repetitions (see text).}\label{fig:B_comparison_critical}
\end{figure}

\begin{figure}
  \begin{center}
    \includegraphics[width=0.95\textwidth]{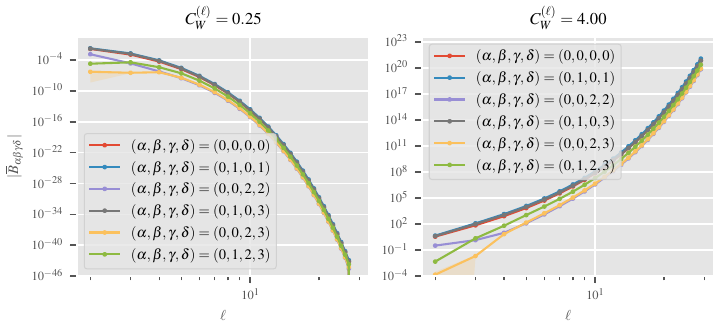}
  \end{center}
  \caption{\emph{Instability of the $B$ Tensor Away from Criticality.} Selected components of the Monte--Carlo estimated tensor $\overline{B}$ are shown for $C_W^{(\ell} < C_W^\mathrm{c}$ (left) and $C_W^{(\ell} > C_W^\mathrm{c}$ (right). $\overline{B}$ is computed from $N_\text{net} = \num{600}$ initializations and its mean and error bars are estimated from  $N_\text{stats} = \num{10}$ repetitions (see text).}\label{fig:B_comparison_non_critical}
\end{figure}

\begin{figure}
    \begin{center}
        \includegraphics[width=0.99\textwidth]{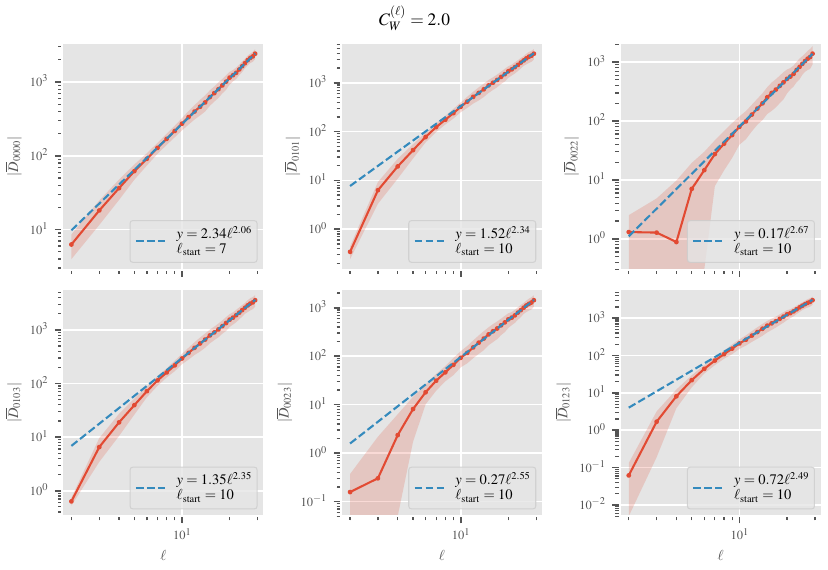}
    \end{center}
    \caption{\emph{Stability of the $D$ Tensor at Criticality.} Selected components of the Monte--Carlo estimated tensor $\overline{D}$ of a ReLU MLP as a function of layer depth $\ell$ are shown for the critical value $C_W^{(\ell)} = C_W^\mathrm{c} = 2$. The hidden layers are of size \num{200}. A regression of the asymptotic power law is shown in blue. The lowest layer included in the regression is given by $\ell_\text{start}$. $\overline{D}$ is computed from $N_\text{net} = \num{600}$ initializations and its mean and error bars are estimated from $N_\text{stats} = \num{10}$ repetitions (see text).}\label{fig:D_comparison_critical}
\end{figure}

\begin{figure}
  \begin{center}
    \includegraphics[width=0.95\textwidth]{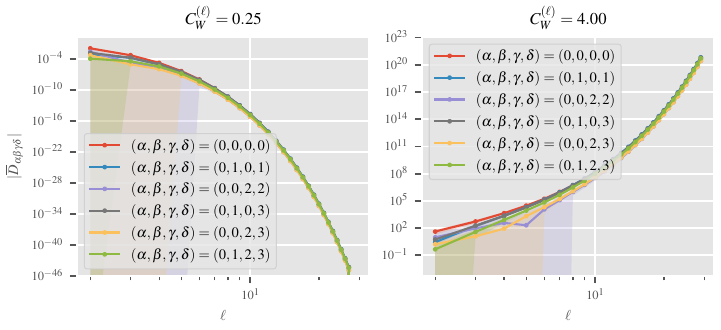}
  \end{center}
  \caption{\emph{Instability of the $D$ Tensor Away from Criticality.} Selected components of the Monte--Carlo estimated tensor $\overline{D}$ are shown for $C_W^{(\ell} < C_W^\mathrm{c}$ (left) and $C_W^{(\ell} > C_W^\mathrm{c}$ (right). $\overline{D}$ is computed from $N_\text{net} = \num{600}$ initializations and its mean and error bars are estimated from  $N_\text{stats} = \num{10}$ repetitions (see text).}\label{fig:D_comparison_non_critical}
\end{figure}

\begin{figure}
    \begin{center}
        \includegraphics[width=0.99\textwidth]{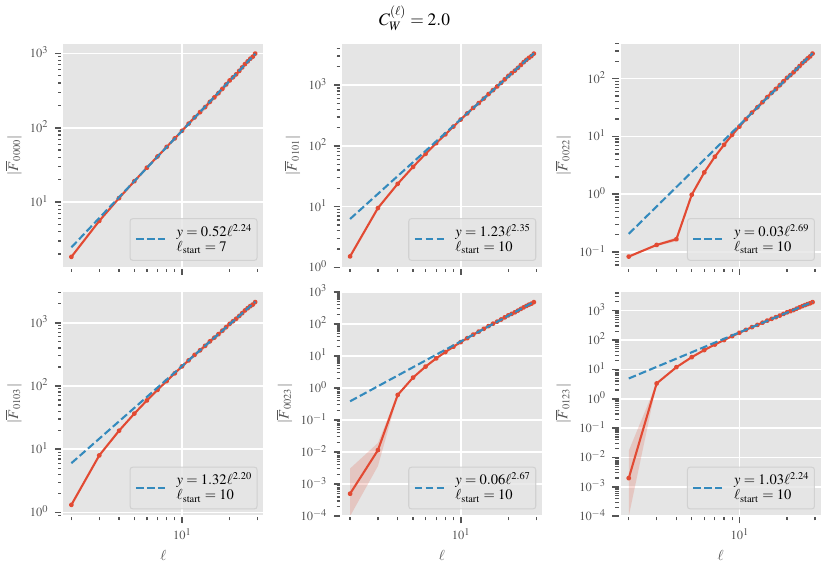}
    \end{center}
    \caption{\emph{Stability of the $F$ Tensor at Criticality.} Selected components of the Monte--Carlo estimated tensor $\overline{F}$ of a ReLU MLP as a function of layer depth $\ell$ are shown for the critical value $C_W^{(\ell)} = C_W^\mathrm{c} = 2$. The hidden layers are of size \num{200}. A regression of the asymptotic power law is shown in blue. The lowest layer included in the regression is given by $\ell_\text{start}$. $\overline{F}$ is computed from $N_\text{net} = \num{600}$ initializations and its mean and error bars are estimated from $N_\text{stats} = \num{10}$ repetitions (see text).}\label{fig:F_comparison_critical}
\end{figure}

\begin{figure}
  \begin{center}
    \includegraphics[width=0.95\textwidth]{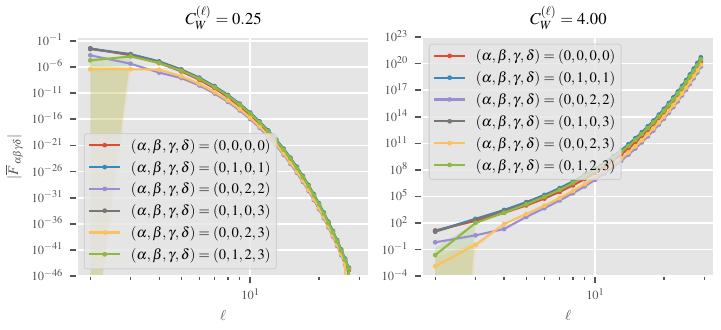}
  \end{center}
  \caption{\emph{Instability of the $F$ Tensor Away from Criticality.} Selected components of the Monte--Carlo estimated tensor $\overline{F}$ are shown for $C_W^{(\ell} < C_W^\mathrm{c}$ (left) and $C_W^{(\ell} > C_W^\mathrm{c}$ (right). $\overline{F}$ is computed from $N_\text{net} = \num{600}$ initializations and its mean and error bars are estimated from  $N_\text{stats} = \num{10}$ repetitions (see text).}\label{fig:F_comparison_non_critical}
\end{figure}

\subsection{Further Kernel-Corrections for Scale-Invariant and Scale-Dependent Activation Functions}
\label{ssub_app:kernel_corrections}

This subsection provides additional experiments to the kernel-corrections for additional activation functions.

For completeness, we first show in correspondence to Figure \ref{fig:relu_corrections} the absence of corrections to the NNGP diagonal in Figure \ref{fig:nngp_relu_corrections}.
\begin{figure}[bt]
  \centering
  \includegraphics[width=0.38\textwidth]{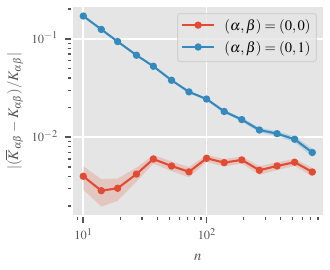}
  \vspace{-0.1cm}
  \caption{\emph{Finite-Width NNGP Corrections for ReLU.} Relative deviations of the Monte--Carlo estimated NNGP to its infinite-width counterpart as a function of hidden layer width $n=n_\ell$. A four layer ReLU MLP with $C_W=2$ is sampled over \num{5e6} initializations. Error bars of the sample mean are included for both the single and distinct input component.}\label{fig:nngp_relu_corrections}
\end{figure}

Next, we will compare the finite-width correction experiments of two scale-invariant functions, namely the identity and the Leaky ReLU, with the scale-dependent GeLU activation to show the different qualitative behavior. For all setups, we used the inputs $x_0$ and $x_1$ again, given in \eqref{eq:x0_input} \& \eqref{eq:x1_input}.

\textbf{Estimating the Finite-Width Corrections.} For a four-layer MLP without biases, we estimate the empirical NNGP and NTK according to \eqref{eq:nngp_trace_avg} and \eqref{eq:ntk_trace_avg}.
The infinite-width solutions $K$ and $\Theta$ are computed using the \texttt{neural-tangents} \cite{novak2020a} library. Error estimates are obtained by dividing the sample variance of the estimated kernels by $\sqrt{N_\text{net}}$ and further rescaling them by $1/K$ or $1/\Theta$, respectively, since we are considering the relative deviations.

\textbf{The Linear Case.} Using the identity as the activation function, we obtain the result in Figure \ref{fig:identity_corrections}. Both the single- and distinct-input component do not show any corrections with width, as expected. The variance of the weights is $C_W^{(\ell)} = 2$.

\begin{figure}
  \begin{center}
    \includegraphics[width=0.95\textwidth]{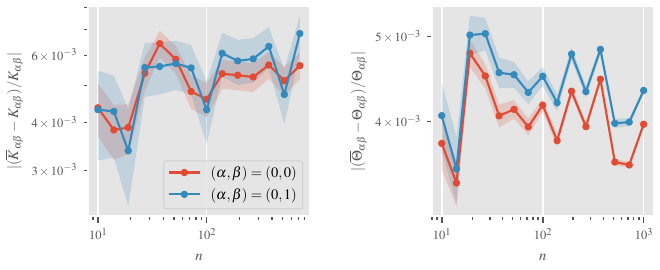}
  \end{center}
  \caption{\emph{Finite-Width Corrections for the Identity.} Relative deviations of the Monte--Carlo estimated NNGP and NTK to its infinite-width counterparts as a function of hidden layer width $n=n_\ell$. A four layer linear MLP with $C_W=2$ is sampled over $\num{5e6}$ initializations. Error bars of the sample mean are included for all lines. Both a single and distinct input component of each kernel is shown.}\label{fig:identity_corrections}
\end{figure}

\textbf{The Leaky ReLU Case.} Using
\begin{equation}
  \sigma(x) = \alpha \, \mathrm{min}(x, 0) + \mathrm{max}(x, 0) \, ,  \label{eq:leakyrelu}
\end{equation}
with $\alpha = 0.1$ as the activation function, we observe similar behavior as for the ReLU as can be seen in Figure \ref{fig:leakyrelu_corrections}. While the distinct-input component shows a convergence to the infinite-width limit, the single-input component is already exact at any finite width, as expected due to the scale-invariant nature. Again, $C_W^{(\ell)} = 2$.

\begin{figure}
  \begin{center}
    \includegraphics[width=0.95\textwidth]{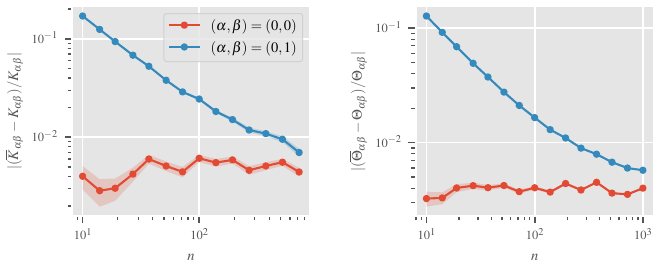}
  \end{center}
  \caption{\emph{Finite-Width Corrections for Leaky ReLU.} Relative deviations of the Monte--Carlo estimated NNGP and NTK to its infinite-width counterparts as a function of hidden layer width $n=n_\ell$. A four layer Leaky ReLU MLP with $C_W=2$ is sampled over $\num{5e6}$ initializations. Error bars of the sample mean are included for all lines. Both a single and distinct input component of each kernel is shown.}\label{fig:leakyrelu_corrections}
\end{figure}

\textbf{The GeLU Case.} To show the distinct behavior of scale-dependent activation functions, we repeat the same experiment for the GeLU function. As shown in Figure \ref{fig:gelu_corrections}, not only the distinct-input, but also the single-input component approach the infinite-width solution only with increasing width. Note that due to the scale-dependence, the scale of the input vectors affects the convergence behavior. Intuitively, large inputs emphasize the asymptotic regions of the activation function, rendering the corrections more similar to the ReLU case, whereas small inputs focus on the nearly linear region close to the origin, leading to a behavior resembling the linear activation function's. Therefore, we rescaled the inputs $x_0$ and $x_1$ by a factor of \num{0.53} to make the distinct nature of this scale-dependent function more pronounced. The variance of the weights was set to $C_W^{(\ell)} = \num{1.98305826}$ \cite{roberts2022}, in order to suppress significant change of magnitude of the preactivation values after a few layers.
\begin{figure}
  \begin{center}
    \includegraphics[width=0.95\textwidth]{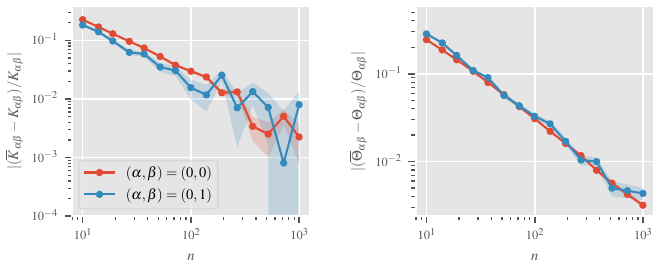}
  \end{center}
  \caption{\emph{Finite-Width Corrections for GeLU.} Relative deviations of the Monte--Carlo estimated NNGP and NTK to its infinite-width counterparts as a function of hidden layer width $n=n_\ell$. A four layer GeLU MLP with $C_W=2$ is sampled over $\num{5e5}$ initializations. Error bars of the sample mean are included for all lines. Both a single and distinct input component of each kernel is shown.}\label{fig:gelu_corrections}
\end{figure}

\subsection{Compute Resources}

All presented experiments were conducted on a cluster equipped with NVIDIA A40 GPUs with 48GB of VRAM. The stability analysis for the NNGP and $V$ could each be computed on a single GPU in less than two hours for all three values of $C_W$. Tensors involving the gradient are more expensive to compute, thus the layer width was reduced to \num{200}. On a single GPU, a single such tensor for three different values of $C_W$ was computed in less than 6 hours. Note that multiple tensors can be computed in parallel using the same ensemble.

Similarly, the finite-width corrections are fast to compute for the NNGP but more expensive for the NTK. For each activation function, it took less than three hours to obtain the results presented here.

Solving the recursions was done entirely on a laptop CPU and takes between a few minutes to less than an hour depending on the tensor at a depth not greater than 5.


}


\end{document}